\documentclass[10pt,logo,copyright]{nvidia_release_template/nvidia_report}
\usepackage[numbers,sort&compress]{natbib}

\usepackage{mathtools}
\newcommand{\mathbbm}[1]{\mathbb{#1}}
\usepackage{wasysym}
\usepackage{mathrsfs}
\usepackage[mathcal]{euscript}

\usepackage{float}
\usepackage{adjustbox}
\usepackage{subcaption}
\usepackage{pgfplots}
\pgfplotsset{compat=newest}
\usepackage[most]{tcolorbox}
\tcbuselibrary{skins}
\usepackage{wrapfig}
\usepackage{placeins}%

\usepackage{multirow}
\usepackage{makecell}
\usepackage{diagbox}
\usepackage{tabularx}
\newcolumntype{Y}{>{\centering\arraybackslash}X}
\newcolumntype{C}{>{\centering\arraybackslash}X}

\usepackage[normalem]{ulem}
\usepackage{multicol}
\usepackage{outlines}
\usepackage{etoolbox}
\usepackage{siunitx}
\usepackage{listings}

\usepackage[nameinlink]{cleveref}
\crefname{equation}{Eq.}{Eqs.}
\crefname{figure}{Fig.}{Figs.}
\crefname{section}{Sec.}{Secs.}
\crefname{appendix}{App.}{Apps.}
\crefname{table}{Tab.}{Tabs.}

\graphicspath{{./}{figs/}{figs/imgs/}{figs/exps/}{tables/}{secs/}{supp/}}

\titleformat{\paragraph}[runin]
  {\normalfont\normalsize\bfseries}
  {}
  {0pt}
  {#1}
\titlespacing*{\paragraph}{0pt}{1.5ex plus 0.5ex minus 0.2ex}{1em}

\newcommand{\sysFull}{SimFoundry\xspace}
\newcommand{\sysName}{SimFoundry\xspace}

\definecolor{researchgreen}{HTML}{2E8B57}

\newcommand{\openpi}{$\pi_{0.5}$}
\newcommand{\openpizero}{$\pi_{0}$}
\newcommand{\grootsix}{GR00T N1.6}
\newcommand{\grootseven}{GR00T N1.7}
\newcommand{\dreamzero}{DreamZero}

\newcommand{\fruitsTask}{\textbf{Serve Fruits}\xspace}
\newcommand{\cupTask}{\textbf{Cup in Bowl}\xspace}
\newcommand{\markerCupTask}{\textbf{Marker in Cup}\xspace}
\newcommand{\clearTableTask}{\textbf{Clear Table}\xspace}
\newcommand{\dishwareTask}{\textbf{Stack Dishware}\xspace}
\newcommand{\markerTask}{\textbf{Store Marker}\xspace}
\newcommand{\cabinetTask}{\textbf{Store Marker}\xspace}
\newcommand{\trashTask}{\textbf{Throw Away Trash}\xspace}
\newcommand{\potTask}{\textbf{Pot on Stove}\xspace}

\newcommand{\y}{\textcolor{researchgreen}{\ding{51}}}
\newcommand{\n}{\textcolor{red}{\ding{55}}}

\newcommand{\sceneModel}{$V_{scene}$ }
\newcommand{\imageModel}{$V_{image}$ }
\newcommand{\articulationModel}{$V_{articulation}$ }

\newcommand{\segimg}{$V_{seg}^{image}$ }
\newcommand{\segmesh}{$V_{seg}^{mesh}$ }

\newcommand{\sceneimg}{$\mathbf{I}_s\,$}
\newcommand{\scenedepth}{$\mathbf{D}_s\,$}

\title{SimFoundry: Modular and Automated Scene Generation for Policy Learning and Evaluation}
\date{2026-06-25}

\author{
\parbox{\textwidth}{
\raggedright
{\normalfont\bfseries\fontsize{9}{13}\selectfont
Nadun Ranawaka$^{1,2*}$,
Josiah Wong$^{1,3*}$,
Wei-Lin Pai$^{3}$,
Wei-Teng Chu$^{3}$,
Tianyuan Dai$^{1,4}$,
Masoud Moghani$^{1,5}$,
Hang Yin$^{3}$,
Yunfan Jiang$^{1,3}$,
Wesley Durbano$^{1*}$,
Brandon Huynh$^{1*}$,
Yu Fang$^{1}$,
Danfei Xu$^{1,2}$,
Ruohan Zhang$^{3}$,
Li Fei-Fei$^{3}$,
Linxi Fan$^{1}$,
Bowen Wen$^{1}$,
Ajay Mandlekar$^{1\dagger}$,
Yuke Zhu$^{1,4\dagger}$
}\\
{\normalfont\fontsize{8}{10}\selectfont
$^{1}$NVIDIA\quad
$^{2}$Georgia Institute of Technology\quad
$^{3}$Stanford University\quad
$^{4}$The University of Texas at Austin\quad
$^{5}$University of Toronto\\
$^{*}$Equal contribution\quad
$^{\dagger}$Equal advising
}
}
}


\begin{document}

\maketitle

\begin{center}
\includegraphics[width=0.97\textwidth]{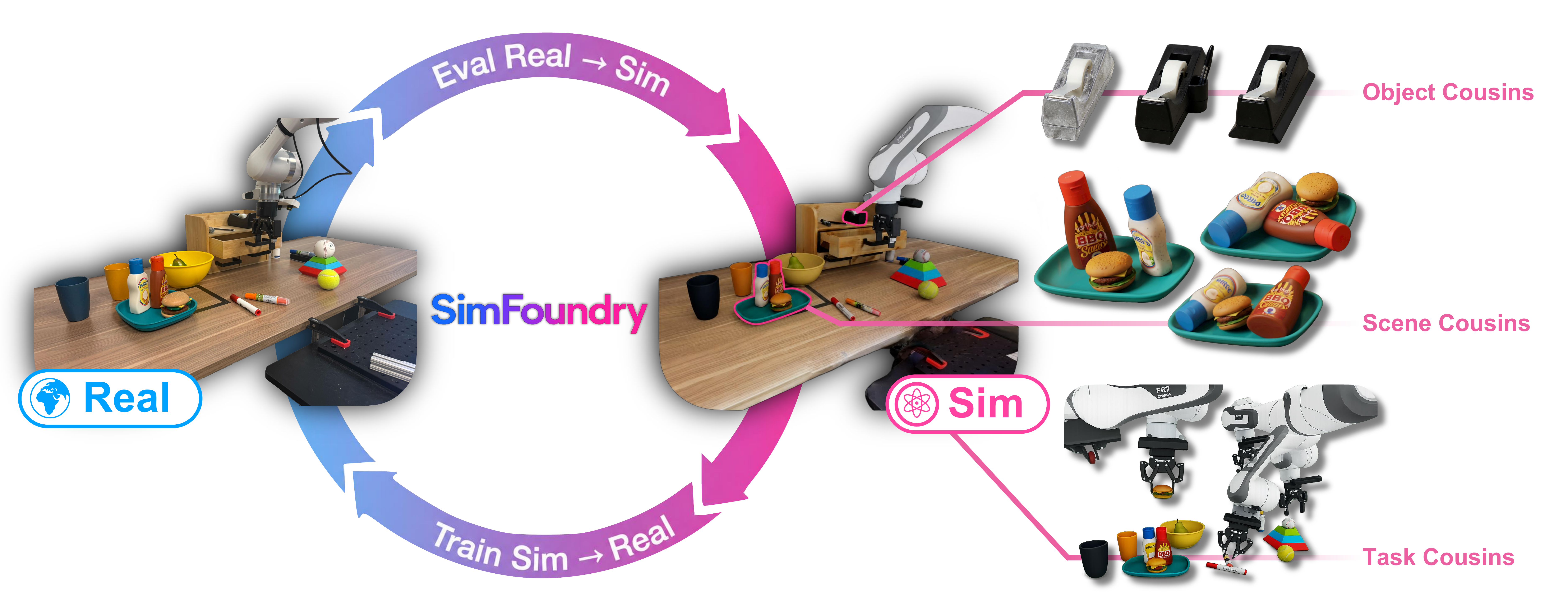}
\end{center}
{\captionsetup{hypcap=false}
\captionof{figure}{\textbf{Overview.} \sysFull{} takes a single real-world input video and
automatically reconstructs an interactive, sim-ready digital twin of the scene.
Based on the reconstructed digital twin, \sysFull{} can further generate an unlimited number of
\emph{digital cousins} --- affordance-preserving variants of the original scene,
spanning three different axes of variation, which we term \textit{object}, \textit{scene}, and \textit{task cousins}, respectively.
These generated simulation environments support both real-to-sim policy evaluation and
sim-to-real policy training, enabling policies to be benchmarked and improved at
scale before deployment in the real world.
}
\label{fig:teaser}}

\vspace{0.5em}

\begin{abstract}
Training and evaluating robot policies in the real world is costly and difficult to scale.
We introduce \sysName{}, a modular and automated system for zero-shot real-to-sim scene construction from a video.
\sysName{} generates sim-ready digital twins and supports object, scene, and task editing, enabling the automated generation of diverse digital cousins: affordance-preserving variations of reconstructed real-world scenes.
Policies trained on \sysName{} data transfer zero-shot to challenging real tasks involving multi-step manipulation, articulated object interaction, and bimanual interaction, and its digital cousins (variations of the original scene, objects, and tasks) facilitate generalization to new real-world conditions. 
Across 7 manipulation tasks and 5 policy architectures, \sysName{} simulation evaluations strongly predict real-world performance, with mean Pearson correlation 0.911 and mean maximum ranking violation 0.018. When evaluating sim-trained policies zero-shot in the real world, policies trained with object, scene, and task cousins in simulation show average task success rate improvements of 17\%, 21\%, and 40\%, respectively. Additional details at \texttt{https://research.nvidia.com/labs/gear/simfoundry/}.
\end{abstract}

\abscontent

\clearpage

\section{Introduction}

Robotic foundation models~\cite{black2024pi_0, bjorck2025gr00t} trained on large-scale robot manipulation datasets have enabled robots to perform a wide range of manipulation tasks autonomously.
However, sourcing high-quality robot manipulation data in large volumes is labor-intensive, often involving large-scale robot teleoperation efforts spanning many months or years~\cite{ebert2021bridge, brohan2022rt, khazatsky2024droid, black2024pi_0}.
Moreover, evaluating trained foundation models in a systematic and scientific manner on real-world manipulation problems of interest can be costly and require thousands of trials across different tasks to make rigorous comparisons~\cite{barreiros2025careful}.

In response to these bottlenecks, recent work has explored simulation as a scalable alternative for training and evaluating robot manipulation models. Automated data generation tools can synthesize large volumes of diverse, high-quality demonstrations with minimal human effort~\cite{dalal2023imitating, mandlekar2023mimicgen, jiang2024dexmimicgen, garrett2024skillmimicgen,li2025momagen}, and have been used to train and improve real-world agents~\cite{maddukuri2025sim, wei2025empirical, cheng2025generalizable, haldar2026point}. Recent work has also shown that simulation-based evaluations can strongly correlate with real-world results, offering a time- and cost-efficient alternative to physical benchmarking~\cite{li2024evaluating, badithela2025reliable}. However, manually constructing simulation environments remains challenging, especially when they must align with real-world scenes and tasks in visuals, geometry, and dynamics.

To address the issues of manual simulation development, real-to-sim scene construction~\cite{torne2024reconciling, dan2025x} has emerged as a paradigm for generating synthetic scenes grounded in the real world. By leveraging 3D reconstruction and generative models, users can create ``sim-ready'' environments that support physically grounded robotic interaction with minimal manual effort. While this reduces environment-authoring overhead and enables both sim-to-real transfer~\cite{dai2024automated, torne2024reconciling, zhao2025robot, dan2025x, han2025re, jiang2025gsworld} and predictive real-world benchmarking~\cite{jain2025polaris, zhang2025real, jangir2025robotarena}, few works provide a unified system that automates scene reconstruction while also performing real-to-sim policy evaluation and training policies that transfer across domains. Recent systems that primarily generate simulation-ready 3D scenes~\cite{xia2026sage, pfaff2026scenesmith, wang2025tabletopgen} often lack the physical interaction, task specification, or data-generation machinery needed to close the sim-to-real policy-learning loop. Conversely, systems designed for simulation-based policy evaluation often assume manually tuned scenes, focus on short-horizon atomic manipulation, or do not support automatically generating diverse objects, scenes, and tasks for policy training~\cite{jain2025polaris, han2025re, jiang2025gsworld, yin2026genie, kim2026molmospaces}.

To this end, we introduce \sysFull{}, a unified and modular system that turns a single real-world input video into interactive simulation environments for both policy evaluation and policy training.
\sysName{} unifies three capabilities that are often treated separately in prior work: reconstructing a sim-ready digital twin, expanding that reconstruction into diverse training environments, and using the resulting simulations to both benchmark and train robot policies.
Its modular design decomposes real-to-sim construction into interchangeable components for perception, asset generation, pose alignment, articulation, physics annotation, and data generation, allowing improved foundation models to be incorporated as they become available without redesigning the full system.
To scale beyond a single reconstructed scene, \sysName{} automatically generates \textit{digital cousins}: affordance-preserving variations that maintain the task-relevant semantics of the original scene while varying object instances, spatial layouts, and feasible manipulation tasks.
These digital cousins, spanning object-, scene-, and task-level variations, enable large-scale synthetic data generation for training robot policies that are robust to changes in object morphology and scene layout and that generalize to related manipulation tasks. \sysName{} environments also produce simulation evaluations that strongly correlate with real-world policy performance.
Finally, policies trained on \sysName{} simulation trajectories can successfully deploy in their real-world counterparts.

\noindent \textbf{Summary of Contributions:}
\newline\noindent $\bullet$ We introduce \sysName{}, a fully automated and modular real-to-sim system that generates interactive, sim-ready scenes from a single video. \sysName{} supports rigid and articulated objects, physics annotations, and automated object, scene, and task cousins, augmenting a single scene into diverse training environments. Across 12 reconstruction scenes, \sysName{} achieves zero-shot F1 scores of $0.81$--$0.92$, which can be further improved to $0.93$--$0.99$ with only $3$ minutes of per-object tuning.
\newline\noindent $\bullet$ We demonstrate \sysFull{} on manipulation tasks that exceed prior real-to-sim work in manipulation complexity, including multi-step tasks, articulated-object interaction, and bimanual coordination on both DROID and YAM robot embodiments. Across $7$ tasks and $5$ policy types, \sysName{} simulation evaluations strongly correlate with real-world performance with a mean Pearson correlation of $0.911$ and an MMRV of $0.018$, outperforming the state-of-the-art baseline~\cite{jain2025polaris} by over $0.59$ on the Pearson correlation.
\newline\noindent $\bullet$ We demonstrate that \sysName{}-generated data can train policies that transfer to the real world and generalize beyond the reconstructed twin. Policies trained on \sysName{} simulation data achieve strong, and sometimes near-perfect, real-world success rates on both YAM and DROID.
Multi-task \sysName{} data further improves generalist policies by up to $31\%$ in simulation and $18\%$ in the real world, while reaching $29\%$ success rate on held-out real tasks.

\raggedbottom
\section{Related Work}

\begin{table*}[!t]
\centering
\setlength{\tabcolsep}{2.2pt}
\setlength{\aboverulesep}{0pt}
\setlength{\belowrulesep}{0pt}
\renewcommand{\arraystretch}{1.18}
\rowcolors{2}{white}{gray!10}
 \caption{\textbf{System comparison.} \sysFull provides a unified and modular pipeline for real-to-sim scene generation that is more feature complete than prior works.}
\label{tab:comparison}

\footnotesize
\begin{adjustbox}{max width=\textwidth}
\begin{tabular}{@{}>{\raggedright\arraybackslash}p{2.15cm}*{14}{c}@{\hspace{4pt}}>{\columncolor{ForestGreen!8}}c@{}}
\toprule
& \rotatebox[origin=c]{60}{\makecell[l]{ACDC\\\cite{dai2024automated}}}
& \rotatebox[origin=c]{60}{\makecell[l]{RialTo\\\cite{torne2024reconciling}}}
& \rotatebox[origin=c]{60}{\makecell[l]{DRAWER\\\cite{xia2025drawer}}}
& \rotatebox[origin=c]{60}{\makecell[l]{RoLA\\\cite{zhao2025robot}}}
& \rotatebox[origin=c]{60}{\makecell[l]{R2R2R\\\cite{yu2025real2render2real}}}
& \rotatebox[origin=c]{60}{\makecell[l]{SIMPLER\\\cite{li2024evaluating}}}
& \rotatebox[origin=c]{60}{\makecell[l]{PolaRiS\\\cite{jain2025polaris}}}
& \rotatebox[origin=c]{60}{\makecell[l]{RobotArena-$\infty$\\\cite{jangir2025robotarena}}}
& \rotatebox[origin=c]{60}{\makecell[l]{R2S-Soft\\\cite{zhang2025real}}}
& \rotatebox[origin=c]{60}{\makecell[l]{Re$^3$Sim\\\cite{han2025re}}}
& \rotatebox[origin=c]{60}{\makecell[l]{GSWorld\\\cite{jiang2025gsworld}}}
& \rotatebox[origin=c]{60}{\makecell[l]{SAGE\\\cite{xia2026sage}}}
& \rotatebox[origin=c]{60}{\makecell[l]{MolmoSpaces\\\cite{kim2026molmospaces}}}
& \rotatebox[origin=c]{60}{\makecell[l]{GenieSim\\\cite{yin2026genie}}}
& \rotatebox[origin=c]{60}{\makecell[l]{\textbf{\sysName{}}\\\textbf{(Ours)}}} \\
\midrule

\makecell[l]{Sim-to-real\\training}           & \y & \y & \y & \y & \y & \n & \n & \n & \n & \y & \y & \n & \y & \y & \y \\
\makecell[l]{Real-to-sim\\policy eval}        & \n & \n & \n & \n & \n & \y & \y & \y & \y & \y & \y & \n & \y & \n & \y \\
\makecell[l]{Automatic scene\\construction}   & \y & \n & \y & \y & \n & \n & \n & \y & \y & \y & \y & \y & \n & \y & \y \\
\makecell[l]{Articulated\\objects}            & \y & \y & \y & \n & \y & \y & \n & \n & \n & \n & \n & \y & \y & \y & \y \\
\makecell[l]{Multi-\\embodiment}
& \n & \n & \n & \y & \n & \y & \n & \y & \n & \n & \y & \n & \y & \n & \y \\
\makecell[l]{Asset\\generation}& \n & \y & \y & \y & \y & \n & \y & \y & \y & \y & \y & \y & \n & \y & \y \\
\makecell[l]{Background\\reconstruction}      & \n & \y & \y & \y & \n & \y & \y & \y & \y & \y & \y & \n & \n & \n & \y \\
\makecell[l]{Object\\cousins}        & \y & \n & \n & \n & \n & \n & \n & \n & \n & \n & \n & \y & \y & \y & \y \\
\makecell[l]{Scene\\cousins}        & \n & \n & \n & \n & \n & \n & \n & \n & \n & \n & \n & \y & \y & \y & \y \\
\makecell[l]{Task\\cousins}        & \n & \n & \n & \n & \n & \n & \n & \n & \n & \n & \n & \n & \n & \y & \y \\

\bottomrule
\end{tabular}
\end{adjustbox}
\end{table*}

\textbf{3D Asset Generation and Alignment.}
3D asset reconstruction and generation has evolved along multiple fronts. Retrieval-based methods align CAD models from databases to single-view images~\cite{kuo2020mask2cad, kuo2021patch2cad,
gumeli2022roca, avetisyan2019scan2cad, dai2024automated, gao2024diffcad}. In parallel, generative models synthesize high-fidelity meshes from one or a few images~\cite{lai2025hunyuan3d25highfidelity3d, xiang2025trellis2, carion2025sam3, xu2024instantmesh, hong2023lrm, wu2024unique3d, tochilkin2024triposr}, and recent extensions
expand to handle articulated objects with automatically inferred movable parts~\cite{wang2019shape2motion, yan2020rpm, weng2024neural,
jiang2022ditto, liu2024singapo, chen2025freeart3d,
chen2024urdformer, yuan2025larm, li2025art, le2024articulate}.
Precise object alignment in multi-object scenes draws on vision foundation models
for depth~\cite{wen2025foundationstereo, wen2025fast, lin2025depth}, segmentation~\cite{ravi2024sam, liu2024grounding}, and 6-DoF pose and scale estimation~\cite{wen2024foundationpose, lee2025any6d, wen2023bundlesdf}. \sysName{} is inherently modular to compose these primitives, allowing newer tools to be swapped in and outputs to be refined via human interventions.

\textbf{Real-to-Sim for Simulation Environment Creation and Applications.} The advent of high-quality 3D reconstruction and generative 3D synthesis have unlocked several works that automate simulation environment construction from real-world
captures~\cite{lim2021planar, jiang2022ditto, antonova2022bayesian,
dai2024automated, torne2024reconciling, han2025re,
jiang2025gsworld, jain2025polaris, jangir2025robotarena,
zhang2025real, wang2025embodiedgen, qureshi2025splatsim, dan2025x,
pfaff2026scenesmith, xia2026sage, kim2026molmospaces,
wang2025tabletopgen, xia2025holoscene}. These systems broadly target one of two goals: closing the real-to-sim-to-real loop for agent training~\cite{dai2024automated, torne2024reconciling,
torne2024robot, xia2025drawer, zhao2025robot, gu2025igen,
chhablani2025embodiedsplat, escontrela2025gaussgym}, or providing reconstructed environments for reliable policy evaluation~\cite{li2024evaluating, jain2025polaris, zhang2025real,
jangir2025robotarena, badithela2025reliable, han2025re,
jiang2025gsworld, yin2026genie}. A separate body of work sidesteps physical simulation entirely and uses reconstructed scenes only for rendering~\cite{yu2025real2render2real}, which limits applicability to contact-rich robot tasks.
\sysName{} belongs to a select group of real--to--sim systems~\cite{han2025re, jiang2025gsworld, yin2026genie} that demonstrate both successful sim-to-real agent transfer and strong correlation between simulated and physical policy evaluations.
However, \sysName{} goes beyond these systems, in its ability to handle more diverse task characteristics (including bimanual, articulation, and multi--step manipulation) and its support for developing multiple types of cousins of a digitized real--world scene to scale up the diversity of reconstructed environments. Table~\ref{tab:comparison}
summarizes these differences and Appendix~\ref{app:related} provides a more detailed related work discussion.

\section{Preliminaries}
\begin{figure*}[t!]
    \centering
    \includegraphics[width=\textwidth]{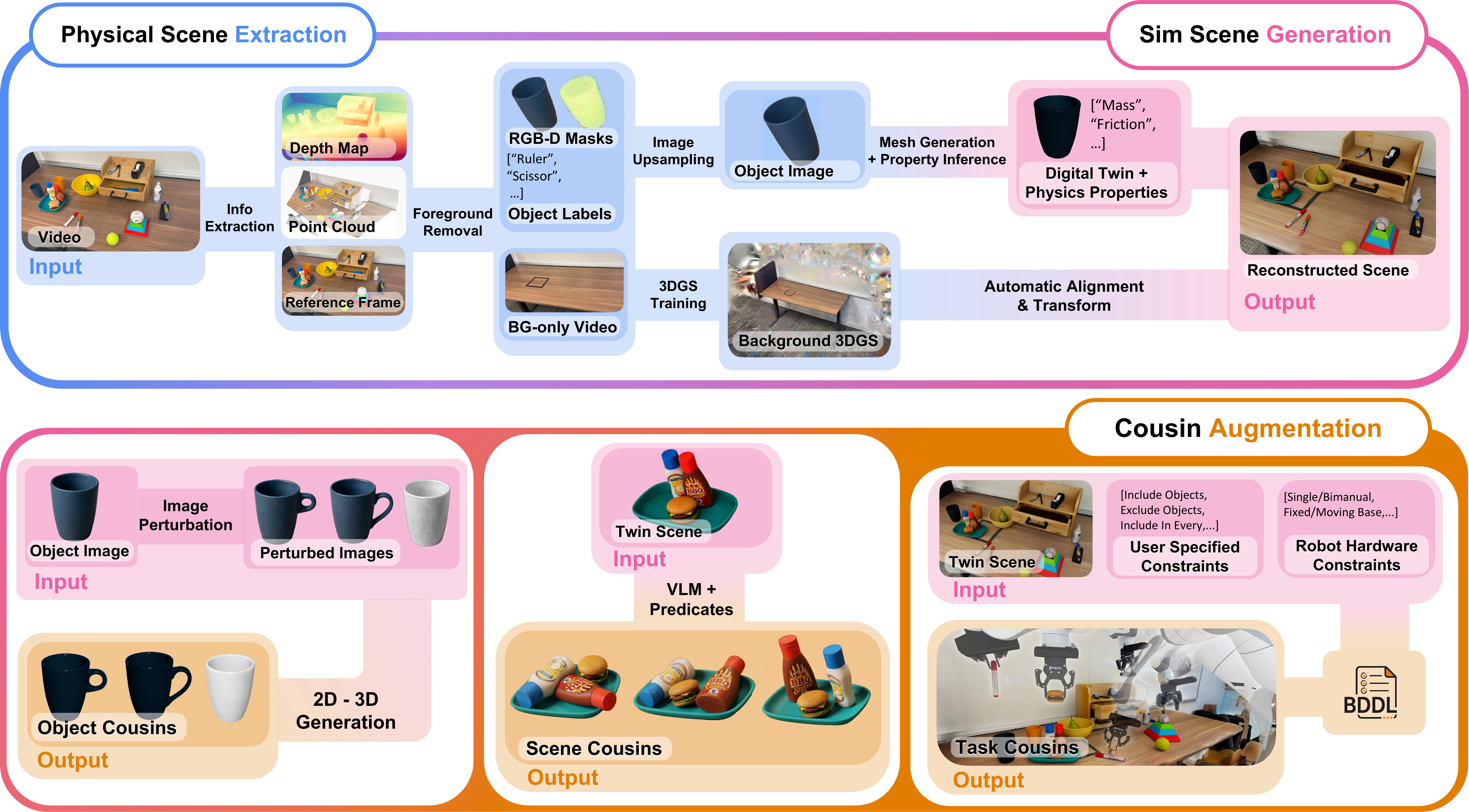}
    \caption{\textbf{Method Overview.} \sysFull extracts per-object relevant information (segmentation masks, depth, etc.), generates 3D visual meshes via 2D-to-3D generation models, and compiles the final output scene by annotating relevant physical parameters and sanity checking the overall scene configuration in a physics simulator. \sysFull additionally supports diverse simulated augmentations along these axes of variation on object, scene, and task: \textit{object cousins} can be generated by modifying input objects in their image space and re-generating corresponding 3D meshes; \textit{scene cousins} can augment the configuration of objects; and \textit{task cousins} can propose viable interactions within the scene. }
    \label{fig:pull}
\end{figure*}

\textbf{Overview.} In this work, we seek to apply \sysName{} to first reconstruct real world scenes $\mathcal{S}_{real}$ in simulation $\mathcal{S}_{sim}$ by converting an input video
into a set of object meshes $\mathcal{M}_i$, scales $\textbf{s}_i$, and poses $\textbf{p}_i$, where $i \in \{1, \ldots, N\}$, leveraging multiple foundation models $V_*$ to achieve this.
We then apply \sysName{} to downstream robotics applications.
We broadly define a policy $\pi_{\theta}$ mapping observations at the current timestep $o_t$ to actions $a_t$, $\pi_{\theta}: \mathcal{O} \rightarrow \mathcal{A}$, implemented as a neural network parameterized by $\theta$.
We focus on two applications: \textbf{real-to-sim evaluation}, where existing real-world policies are evaluated in simulation to characterize their performance, and \textbf{sim-to-real training}, where policies are trained in simulation and then deployed in the real world.

\textbf{Real and Simulation Policy Correlation.} We measure real and sim policy correlation using the Pearson Correlation Coefficient ($r$) and Mean Maximum Rank Violation (MMRV), both of which have been proposed by prior work~\cite{jain2025polaris, li2024evaluating}. Ideal correlation has $r \rightarrow 1$, which measures linear correlation between real and simulation task results, and MMRV $\rightarrow 0$, which measures the average worst rank-violation of policies as evaluated in simulation versus their actual ranks in the real world. When measuring task success, we measure end-to-end task success (a discrete 0 or 1), which is defined as the completion of all task criteria (see Appendix~\ref{app:task_rubric}).

\textbf{Real-to-Sim Reconstruction and Synthetic Simulation Data.}
We define \textit{digital twins} as being strict replicas of the geometry and object layouts of a real-world scene. In contrast, \textit{digital cousins}~\cite{dai2024automated, maddukuri2025sim} are virtual scenes that maintain the semantic and geometric affordances of a real-world scene without explicitly modeling it, and serve as a form of object instance randomization. MimicGen~\cite{jiang2024dexmimicgen} is a recent method proposed to quickly generate large amounts of synthetic trajectories by splicing together various subtask trajectories sampled from a set of source demonstrations.

\section{\sysFull: A modular, automated real-to-sim generation pipeline}

\sysFull{} generates interactive simulated scenes through three stages, as seen in \autoref{fig:pull}: \textbf{Extraction}, which infers relevant per-object information from a video; \textbf{Generation}, which creates, aligns, annotates, and stabilizes sim-ready assets; and \textbf{Augmentation}, which produces digital cousins in the form of object, scene, and task variations. Appendix~\ref{sec:vlm_details} has additional details, including the underlying foundation models used (denoted as $V_*$).

\textbf{Extraction.} We assume the input is a raw RGB video. We first convert the input into a representative RGB frame \sceneimg and estimate a corresponding depth map \scenedepth using off-the-shelf depth estimation models $V_{im2depth}$~\cite{wen2025foundationstereo,lin2025depth}. Using the camera intrinsics $\mathbf{K}$, we lift this RGB-D observation into a scene point cloud $\mathbf{P}_s$, which is used for scene alignment and object pose estimation. We then query \segimg~\cite{carion2025sam3} to extract the ground plane and align the reconstruction with the simulator world frame.

Next, we use a scene-understanding VLM \sceneModel to detect the objects in the scene and \segimg to iteratively segment out the foreground objects ${o_1,\dots,o_n}$. For each object, we extract its segmentation mask $m_i$ from \segimg along with corresponding RGB and depth pixels $(p_i^{rgb}, p_i^{depth})$. After each extraction, we remove the object from the RGB-D observation using image and depth inpainting, and repeat this process until no foreground objects remain. This stage outputs per-object RGB-D crops and masks, which are used for mesh generation and alignment. Further details on video preprocessing, inpainting, and iterative decomposition are provided in Appendix~\ref{sec:extraction_details}.

\textbf{Generation.} Given each object crop $p_i^{rgb}$, we use \imageModel to upsample and a 2D-to-3D mesh model $V_{mesh}$ to generate a visual mesh $\mathcal{M}_i$. We then estimate and refine the object pose $\mathbf{p}_i$ by aligning the mesh to the reconstructed scene using the scene RGB-D observation, object mask, and point cloud geometry, with additional refinement from another model $V_{pose}$ such as FoundationPose~\cite{wen2024foundationpose}. Objects identified as articulated, such as cabinets or drawers, are processed by a separate articulation module that detects movable parts, segments the mesh, and generates joint parameters using \articulationModel and prior articulation-generation methods~\cite{le2024articulate, qiu2025articulate}; full details are provided in Appendix~\ref{sec:articulated_object_generation}.

Finally, for each generated object, we produce collision geometry using CoACD~\cite{wei2022approximate} and assign physical properties such as mass and friction by querying \sceneModel. Once all objects are generated, aligned, and annotated, we compose the scene in PyBullet~\citep{pybullet}, resolve object penetrations to obtain a stable configuration, and export the resulting sim-ready scene to downstream robotics simulators such as IsaacLab~\cite{mittal2025isaac}. Further details are listed in Appendix~\ref{app:physical_stability}.

\textbf{Augmentation.}
Once the initial scene is reconstructed, \sysName{} expands it into a family of digital cousins: affordance-preserving simulated variants that retain its task-relevant semantics while varying three axes: object instance, scene layout, and task specification. \autoref{fig:real2sim_reconstruction} illustrates this process across diverse real-world scenes. The middle row shows that SimFoundry reconstructs the objects, layouts, and scene structure of the real-world inputs, while the bottom row demonstrates how the reconstructed twins can be expanded into plausible digital cousins. These cousins alter object geometry, appearance, spatial configuration and task specifications and for brevity, we refer to digital cousins generated by varying these axes as \textit{object cousins}, \textit{scene cousins}, and \textit{task cousins}, respectively. These terms indicate the dominant axis of variation rather than mutually exclusive classes: a digital cousin may combine object, scene, and task variations.
Further details are provided in Appendix~\ref{app:scene_variations}.

\textit{Object cousins} generate new object instances that maintain the affordances of the original object while varying geometry, topology, and appearance. For example, a reconstructed mug, drawer, or plate can be converted into multiple plausible alternatives with different shapes, handles, textures, or proportions. These object-level cousins provide instance diversity while preserving task-relevant functionality.

\textit{Scene cousins} vary the spatial arrangement of objects in the reconstructed scene using semantic spatial predicates such as \textbf{OnTop} and \textbf{RightOf}. Rather than simply perturbing object poses, these cousins produce meaningful alternative layouts, such as moving an object from beside a receptacle to inside or on top of it. We can also add controllable distractor objects from a library of sim-ready assets. These scene-level cousins introduce structured geometric diversity and help policies generalize beyond the original layout.

\textit{Task cousins} use the reconstructed scene to propose additional feasible manipulation tasks grounded in the available objects and affordances. \sysName{} converts these tasks into simulation-compatible goal specifications, enabling procedural demonstration collection without manually authoring each environment or task. This allows the same reconstructed scene to support multi-task data generation, including related tasks that share objects, goal conditions, or intermediate behaviors with the original task.

Together, these mechanisms provide controllable diversity across objects, layouts, and tasks.
In our experiments, object cousins improve robustness to unseen object instances, scene cousins improve generalization to novel layouts, and task cousins improve both zero-shot and few-shot downstream task performance.

\input{tables/scenes}

\textbf{Background Reconstruction and Alignment.}
The Extraction-Generation-Augmentation pipeline produces a physically-grounded foreground scene of per-object meshes.
To recover a photorealistic background, \sysName{} can fuse reconstructed objects with a 3D Gaussian Splat~\cite{kerbl20233d} background.
We support two pipelines to this end. The \emph{automatic} pipeline operates on the same single raw video used by the Extraction stage: it removes foreground objects via prompted video segmentation and two-pass inpainting, recovers metric depth and camera poses, trains a depth-supervised splat, and bridges it into the simulator world through a derived rigid transform,
requiring no additional capture or user input. The \emph{manual} pipeline instead needs the user to take a second video of the scene with foreground objects physically removed. It then trains a splat on this background-only video,
and aligns it to the reconstructed scene through our interactive editor. The two routes emit identical asset structures and, as we show in Appendix~\ref{app:3dgs-background-analysis}, yield comparable reconstruction quality; they trade off capture effort against compute and fidelity on texture-less surfaces and silhouettes. We detail both pipelines and their respective strengths in Appendix~\ref{sec:bg_details}. We also use mesh background reconstruction, generated by apps such as Scaniverse\footnote{https://dev.scaniverse.com/}, in our robotics experiments.

\section{Experiments}
\label{sec:experiments}

We highlight two key applications of \sysName{} --- using \sysName{} environments as a way to benchmark real-world manipulation policies (Sec.~\ref{subsec:real_to_sim_eval}) and training robot manipulation agents from generated \sysName{} environments that transfer zero-shot to the real-world (Sec.~\ref{subsec:exp_sim_to_real}).
Sec.~\ref{subsec:exp_system} contains additional experiments that analyze \sysName{} reconstruction performance. 
Our experiments are performed on two robot embodiments --- the DROID~\cite{khazatsky2024droid} platform, and a YAM workcell~\cite{i2rt2025yam}. The tasks we evaluate (shown in \autoref{fig:real2im_correlation}) span diverse characteristics, including short-horizon pick and place, bimanual coordination, and long-horizon language following (see Appendix~\ref{app:task} for details).

\subsection{Real-to-Sim Policy Evaluation}
\label{subsec:real_to_sim_eval}

\paragraph{Setup.} We aim to show that policy evaluations in \sysName{} simulation scenes can correlate strongly with real-world policy evaluation results.
We consider two sets of policies and tasks --- pre-trained generalist policies (\openpizero~\cite{black2024pi_0}, \openpi~\cite{intelligence2025pi_}, \grootsix~\cite{bjorck2025gr00t}, \grootseven, and \dreamzero~\cite{ye2026world}) that are deployed zero-shot (no finetuning) on 4 less difficult tasks, and policies that are finetuned (\openpizero, \openpi, and \grootsix) using 50 real-world demos per task and deployed on 3 more challenging tasks. We use separate task sets for each policy group to ensure meaningful (non-zero) policy evaluation results for zero-shot and few-shot deployment. Appendix~\ref{app:policy} contains details on policy training and selection, and Appendix~\ref{app:real-to-sim-evals} contains the evaluation procedure.

\begin{figure}[t]
    \centering

    \includegraphics[width=0.48\textwidth,keepaspectratio]{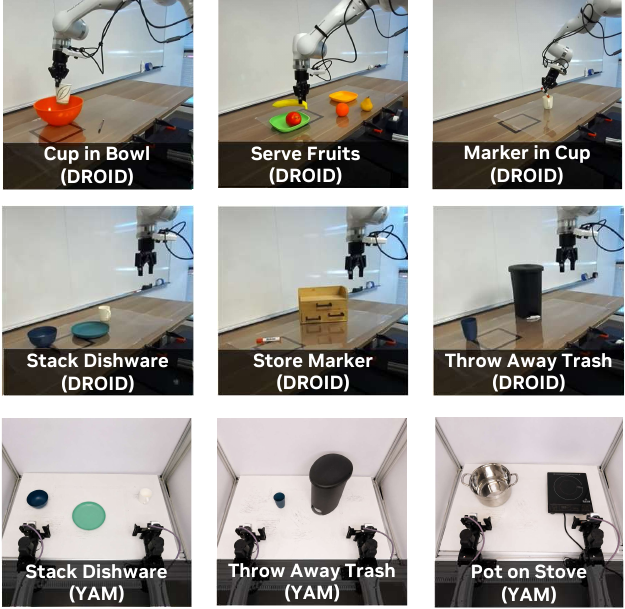}
    \hfill
    \includegraphics[width=0.48\textwidth,keepaspectratio]{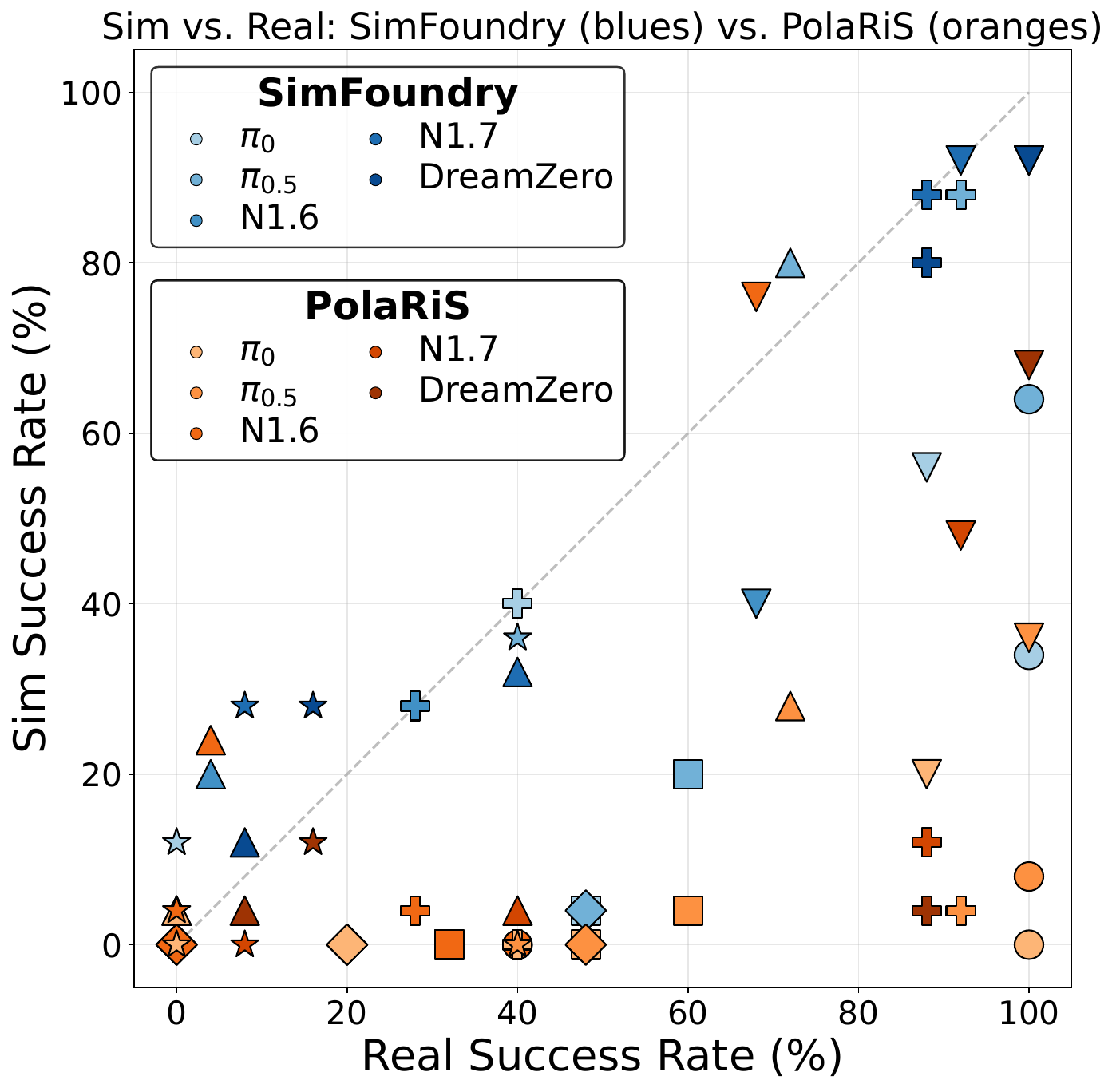}

    \caption{\textbf{Tasks and Real-to-Sim Policy Evaluation correlations.} (Left) We apply \sysName{} to a DROID setup using a single Franka arm (top two rows), and a bimanual setup with two YAM arms (bottom row). Our tasks span multiple types of manipulation, including multi-step, articulated object interaction, and bimanual coordination (\clearTableTask not shown, more details in Appendix~\ref{app:task}). (Right) Under zero-shot policy evaluation without simulator-specific finetuning, SimFoundry outperforms the state-of-the-art baseline PolaRiS~\cite{jain2025polaris} in simulation-based evaluation correlations. Each marker shape represents a different task from the left. Additional details in Appendix~\ref{app:results} and \autoref{fig:real2sim_fullsize}.}
    \label{fig:real2im_correlation}
\end{figure}

\paragraph{\sysName{} scene evaluations strongly correlate with real-world performance across diverse policies.} As shown in \autoref{fig:real2im_correlation}, \sysName{} evaluations closely match real-world results and preserve policy rankings, with a mean Pearson correlation of 0.911 and MMRV of 0.018 (\autoref{tab:real2sim-simfoundry}). \sysName{} evaluations also reveal model strengths: \grootseven \hspace{0.2mm} can outperform others on precise grasping (e.g., \markerCupTask), while $\pi_{0.5}$ shows stronger language following (e.g., \fruitsTask), offering actionable guidance for model development. These correlations hold across policy types, including two VLA families ($\pi$ and GR00T) and the world-action model DreamZero.

\paragraph{Sub-task evaluations improve correlations, especially on multi-step tasks.} We introduce a sub-task evaluation procedure that increases policy eval correlations from a mean Pearson score of 0.90 to 0.95. By evaluating at a sub-task level, users can more accurately target future data collection for model improvement by focusing on specific sub-tasks. This procedure also helps improve eval correlations for long-horizon tasks, where success can be bottlenecked by a few, more difficult sub-tasks.  Full details on the evaluation procedure and results are in Appendix~\ref{sec:subtask_eval}.

\paragraph{\sysName{} outperforms state-of-the-art simulation evaluation frameworks and makes fewer assumptions.} We compare \sysName{} to PolaRiS~\cite{jain2025polaris},  a state-of-the-art method for real-to-sim policy evaluation (full details of comparison in Appendix~\ref{sec:PoLaRis_baseline_details}). 
We use the same protocol to evaluate the same real-world policies in PolaRiS, and find that \sysName{} has a mean Pearson correlation that is over 0.59 higher than PolaRiS. PolaRiS obtains improved correlation by shallowly finetuning policies on PolaRiS simulation data. However, we do not adapt the policies to either simulation framework, and evaluate them completely zero-shot in simulation.

\subsection{Sim-to-Real Policy Training}
\label{subsec:exp_sim_to_real}

We show that \sysName{} can generate synthetic data for training policies that can be deployed in the real world. We study three settings: zero-shot sim-to-real transfer (policies trained only on \sysName{} data), sim-and-real co-training (policies trained on large-scale \sysName{} data plus limited real data), and multi-task transfer (policies trained on multi-task \sysName{} data that transfer to new real-world tasks).
We make use of \sysName{}'s ability to produce automated object, scene, and task cousins, and show that they are critical to enable real-world policy generalization. A summary of our results is given in \autoref{fig:all_cousins_exps}, while detailed results per experiment are given in Appendix~\ref{app:results}.

\begin{figure}[t]
    \centering
    \includegraphics[width=0.96\textwidth]{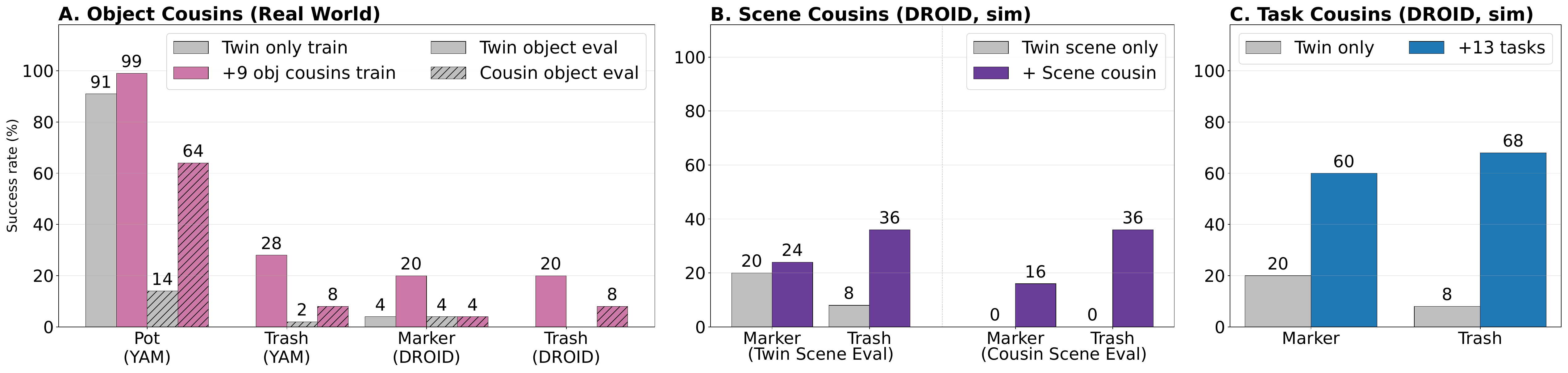}
    \caption{\textbf{SimFoundry Data Diversity Improves Policy Performance.} (A) Across multiple robot embodiments and multiple tasks, leveraging additional object cousins~\cite{dai2024automated} improves direct Sim-to-Real policy transfer on the original target scene objects and additional held-out unseen objects. (B) Scene cousins improve policy performance on the original scene and allow policy transfer to cousin scenes. (C) Adding task cousins improves performance on related downstream tasks by enabling intra-task transfer. \textbf{Note:} Pot refers to the \potTask task, Trash refers to the \trashTask task and Marker is for the \cabinetTask task.
      }
    \label{fig:all_cousins_exps}
\end{figure}

\paragraph{Policies trained on \sysName{}-generated data transfer zero-shot to the real world.}
Across both YAM and DROID, policies trained on \sysName{} data transfer effectively to real scenes, reaching $99\%$ success on \potTask with YAM and $100\%$ success on \dishwareTask with DROID (Table~\ref{tab:sim2real_yam_droid}). On YAM, this is achieved with a simple flow-matching policy trained from scratch; on DROID, we finetune the \openpi~\cite{intelligence2025pi_} DROID checkpoint on sim-generated demonstrations.

\paragraph{Co-training with sim and real data further improves performance.}
Although \sysName{} supports zero-shot transfer, adding small amounts of real data further boosts performance. On DROID, co-training improves most \openpizero{} and \openpi{} results in both sim and real (\autoref{fig:all_cousins_detailed}). For example, \openpi{} real-world success on \markerTask increases from $60\%$ to $92\%$, while \openpizero{} gains $36\%$ sim success on \trashTask. This suggests that \sysName{} reconstructions are faithful enough to complement real demonstrations during training.

\paragraph{Object cousins improve robustness to unseen objects.}
Across both embodiments, increasing object diversity improves policy performance. Adding object cousins yields a $50$-point real-world gain on held-out \potTask objects, and improves DROID performance in both sim and real with gains up to $20$ points on \trashTask. These results show that \sysName{} generates useful object-level diversity for training policies that generalize beyond the reconstructed twin. Additional details and ablations are provided in Appendix~\ref{app:results}.

\paragraph{Scene cousins improve layout generalization.}
On DROID, adding scene cousins boosts success in simulation by up to $28$ points on \trashTask in the twin scene (\autoref{fig:all_cousins_exps}B). Scene cousins also enable transfer to novel layouts, reaching $16\%$ success on \markerTask cousin scenes where the twin-only policy achieves $0\%$.

\paragraph{Multi-Task Sim-to-Real and Task Generalization.}

\begin{wraptable}{hr}{0.45\textwidth}
\vspace{-12pt}
\centering
\caption{\textbf{Multi-task policy evaluation.} Success rates for policies evaluated on seen and held-out tasks in simulation and the real world. }
\label{tab:multi_task_policy_eval}
\setlength{\tabcolsep}{3pt}
\renewcommand{\arraystretch}{1.05}
\scriptsize
\begin{tabularx}{\linewidth}{@{}>{\raggedright\arraybackslash}Xccc@{}}
\toprule
& \openpi{}-DROID
& \textbf{\openpi{}}-FT
& \openpi{}-DROID-FT \\
\midrule
Sim              & 30 & 51 & \textbf{61} \\
Sim -- held out  & 37 & \textbf{45} & 33 \\
Real             & 28 & 45 & \textbf{46} \\
Real -- held out & 26 & \textbf{29} & 26 \\
\bottomrule
\end{tabularx}
\vspace{-16pt}
\end{wraptable}

We reconstruct a cluttered scene, use \sceneModel{} to propose tasks, generate demonstrations entirely in simulation, and finetune both base \openpi{} and \openpi-DROID\@. We evaluate on 13 generated-data tasks and 7 held-out tasks in simulation and the real world. Results are presented in \autoref{tab:multi_task_policy_eval}.

\paragraph{\sysName{} can train generalist policies and task cousins improve few-shot downstream learning.}

\sysName{}-finetuned policies outperform the base DROID checkpoint by up to $31\%$ in simulation and $18\%$ in the real world, and \openpi-FT reaches $29\%$ success on held-out tasks without task-specific demonstrations (Table~\ref{tab:multi_task_policy_eval}). With the total number of demonstrations fixed, replacing some target-task data with related task-cousin demonstrations improves downstream simulation performance, especially on harder tasks: in simulation, 13 task cousins increase success on \trashTask by $60\%$ and \markerTask by $40\%$ (\autoref{fig:all_cousins_exps}C).

\subsection{System Analysis}
\label{subsec:exp_system}

We analyze \sysName{} along three axes: reconstruction fidelity, the human effort required to refine that fidelity, and the scalability of the end-to-end pipeline (see Appendix~\ref{sec:3d_reconstruction_evaluation_details} for full details).

\textbf{\sysName{} scene reconstruction fidelity outperforms state-of-the-art methods.}
Under fully automated operation, \sysName{} surpasses SAM3D across three 3D geometric metrics (Table~\ref{tab:scene_reconstruction_quantitative}). For instance, \sysName{} achieves higher F1 score (0.81--0.92) than SAM3D (0.66--0.71), lower chamfer distance and position error, showing that the pipeline recovers precise scene geometry without human input.

\textbf{\sysName{} environment generation scales well with compute and human time.} 
The pipeline reconstructs objects at an average rate of roughly 5 minutes per object across diverse real-world scenes (Table~\ref{tab:additional_real2sim_reconstruction}), and an additional 3 minutes of per-object operator tuning yields consistent gains on every metric (e.g. F1 scores rise to 0.93--0.99, as shown in Table~\ref{tab:scene_reconstruction_quantitative}), demonstrating that fidelity can be traded against effort on demand.

\flushbottom
\section{Limitations}

Our system relies heavily upon off-the-shelf foundation models. While this enables broad modularity and the ability to swap components as better models become available, it also naturally inherits the failure modes of each underlying model. Additionally, we make several assumptions that limit our current pipeline to tabletop-style layouts.  Relaxing this assumption to support multi-level or non-planar environments is a natural direction for future work. A more thorough discussion of limitations is given in Appendix~\ref{app:limitations}.

\section{Conclusion}
\sysName{} is a fully automated pipeline that natively reconstructs interactive sim-ready scenes from a single video, handles articulated object generation and scenes with clutter and occlusion, and generates object, scene, and task cousins. We find that \sysName{} can measure task success in simulation that correlates to real-world policy performance and outperforms prior work in sim-based policy evaluations. Second, \sysName{}-generated data can train policies that transfer to the real world, while object, scene, and task cousins improve robustness to unseen objects, novel layouts, and related downstream tasks.
\sysName{} enables simulation to accelerate real-world policy development by facilitating synthetic data generation for training policies and reliable simulation-based evaluation to compare policies and yield actionable insights.

\clearpage

\section{Acknowledgments}

The authors would like to thank Omkaar Buddhikot, Amitoj Sandhu, Nadia Laswi, Ramanpreet Singh, Mona Abbas, and Osiriz Durana for their help with data collection and model evaluation. We also thank Jeremy Chimienti, Danyi Chen, and Lion Park for their help with hardware support, and Scott Reed, You Liang Tan, and Fengyuan Hu for feedback and valuable discussions. Nadun Ranawaka is partially supported by the Agricultural Technology Research Program at the Georgia Institute of Technology.

\bibliographystyle{plainnat}
\bibliography{references}

\clearpage

\appendix
\captionsetup[figure]{skip=6pt}
\captionsetup[table]{skip=6pt}

\part*{Appendix}

\renewcommand\thesection{\Alph{section}}

\counterwithin{figure}{section}
\counterwithin{table}{section}

\section{Overview}
\label{app:overview}

The Appendix contains the following content.

\begin{itemize}

\item \textbf{FAQ} (Appendix~\ref{app:faq}): answers to some common questions

\item \textbf{Limitations} (Appendix~\ref{app:limitations}): more thorough list and discussion of \sysName limitations

\item \textbf{Full Related Work} (Appendix~\ref{app:related}): more thorough discussion on related work

\item \textbf{Scene Reconstruction} (Appendix~\ref{app:scene_recons}): additional details on \sysName scene reconstruction method

\item \textbf{Digital Cousins Augmentation} (Appendix~\ref{app:scene_variations}): additional details on how \sysName produces cousins at the object, scene and task level

\item \textbf{Detailed Experiment Results} (Appendix~\ref{app:results}): tables containing detailed results for our real-to-sim and sim-to-real evaluations as well as additional discussion on these results

\item \textbf{Policy Details} (Appendix~\ref{app:policy}): details on how policies are obtained for all \sysName experiments

\item \textbf{Robot Platform and Task Details} (Appendix~\ref{app:task}): details on the robot platforms, how they are modeled in simulation, and the tasks used in the experiments

\item \textbf{Real-to-Sim Policy Evaluations} (Appendix~\ref{app:real-to-sim-evals}): additional details on real-to-sim policy evaluation experiments

\item \textbf{Human Interaction} (Appendix~\ref{app:human}): details on how users can interact with \sysName to improve generation quality or control specific elements

\item \textbf{System Analysis} (Appendix~\ref{app:system}): analysis of several characteristics of \sysName, including reconstruction fidelity and environment generation throughput. This section also details the background reconstruction methods supported by \sysName{}

\end{itemize}

\clearpage
\section{FAQ}
\label{app:faq}

\begin{enumerate}

\item \textbf{Why should I use \sysName compared to alternative methods that generate simulation environments?}

There have been many impressive systems showcasing automated, diverse, and scalable real-to-sim reconstruction pipelines. We highlight the following key distinguishing details that are especially advantageous:

\begin{itemize}
    \item \textbf{Towards Feature-Complete Automation.} Our approach is \textit{fully} automated, supporting programmatic scene, articulated object, and background reconstruction, all within a single unified pipeline.
    \item \textbf{Empirically Validated for Robotics Tasks.} We show that \sysName is concretely useful for downstream robotics applications, both in real-to-sim eval and sim-to-real training settings.
\end{itemize}

See Appendix~\ref{app:related} for more discussion.

\item \textbf{How are \sysName scenes tested for physical stability?}

Our reconstructed scenes are spawned within a PyBullet physics simulator instance, and subsequently stepped until objects settle. This guarantees physical stability during subsequent initializations, though objects may drift with respect to their original fitted poses. See Appendix~\ref{app:physical_stability} for more details.

\item \textbf{How much manual effort is needed to produce \sysName scenes and use them for the robot manipulation applications shown in the paper?}

After initial manual scene scans using a smartphone camera, our \sysName environments used in our robotics experiments were generated using our fully automated pipeline and then quickly interactively tuned with a human operator, requiring a few minutes worth of iteration.

\item \textbf{What are some typical runtimes for how long it takes to generate a scene?}

On average, it takes roughly 5 minutes per object when reconstructing a real scene in simulation, which is the time cost amortized across the entire pipeline.

\item \textbf{Does \sysName support transparent objects?}

Yes, with a caveat. \sysName merely inherits the strengths and limitations of the underlying 2D to 3D mesh model used to generate simulation meshes. The default model used is Hunyuan 2.1, which does not support transparency, though we additionally support other models such as TRELLIS.2, which does support transparent objects.

\item \textbf{For the DROID experiments, how much of the policy performance can be attributed to the pretrained checkpoint?} We find that for the tasks showing sim-to-real transfer on DROID, the \openpi \hspace{0.2mm} and \openpizero \hspace{0.2mm} checkpoints pre-trained on the DROID dataset perform poorly without task-specific finetuning data. On \markerTask and \trashTask, both checkpoints get a $0\%$ success rate, while for \dishwareTask, \openpizero-DROID gets $52\%$ and \openpi-DROID gets $48\%$ success rate (improving to $100\%$ with sim-only finetuning). This illustrates that \sysName{} data is valuable for finetuning pretrained checkpoints, especially on difficult tasks.

\item \textbf{Why do you use a binary success metric instead of a normalized task reward?}
We primarily report end-to-end task success because it provides a stricter test of real-to-sim fidelity. Normalized rewards can give partial credit for completing intermediate subtasks, whereas binary success requires the policy to complete the full task under the reconstructed scene dynamics, visuals, and geometry. Empirically, we find that success-rate correlations are harder to achieve in this setting, especially for long-horizon tasks and real-world-finetuned policies, where small reconstruction errors can cause policies to go out-of-distribution and fail. Therefore, maintaining a strong success-rate correlation in this setting provides stronger evidence that \sysName{} reconstructions faithfully preserve the factors that determine real-world policy performance.

\end{enumerate}

\clearpage
\section{Limitations}
\label{app:limitations}

As all VLMs in this paper are queried through a remote third-party provider, identical inputs can yield non-deterministic outputs across runs; for instance, we occasionally observe inconsistent inpainting from the Gemini image model, producing degenerate or duplicated extracted objects.

Reconstruction fidelity is further bounded by the quality of the inferred point cloud. For monocular inputs in particular, the scale and shape of the recovered geometry may not fully match the
real-world scene, reducing the accuracy of the reconstructed output.
Our articulation results likewise depend on accurate 3D segmentation of the object mesh, which can be difficult for meshes produced by image-to-mesh models or for objects with occluded internal structure.

Our physics-stability procedure assumes that objects rest on a single flat reference surface, which restricts the pipeline to tabletop-style scene layouts. Future work could address this to extend to more complex and varied scenes.

Finally, the automatic background pipeline removes the second-capture and manual-alignment work, but at the cost of runtime: the two-pass video inpainting required to produce a dense, temporally consistent, and clean RGB-D stream for splat training takes roughly 90 minutes per scene on a single GPU\@. This overhead is largely hidden in a multi-GPU setting, where background reconstruction can run in
parallel with the Extraction, Generation, and Variation stages rather than serially after them. Also, we use mesh reconstructions of the background for some of our robotics experiments in simulation, due to issues with near-field clipping of the generated 3DGS and to reduce rendering latency.

\clearpage
\section{Full Related Work}
\label{app:related}

\subsection{Real--to--Sim for Simulation Environment Creation and Applications}
We expand on the main-text discussion and position \sysName{} against four categories of prior automated real-to-sim work.

\paragraph{Real--to--sim--to--real manipulation.}
This category digitizes physical scenes into simulation for real--world robot training~\cite{dai2024automated,
torne2024reconciling, torne2024robot, xia2025drawer, zhao2025robot,
gu2025igen}. These systems are typically demonstrated on single-step pick-and-place tasks with rigid objects, whereas \sysName{} supports a wider manipulation regime that includes
bimanual setups, articulated-object interactions, and multi-step tasks. A further distinction is that most prior pipelines produce
a static digital twin of the captured scene, which bounds environmental diversity and limits the generalization gains achievable from simulated training. \sysName{} instead generates
multiple \emph{cousins} of each reconstructed
scene---object, scene, and task cousins---which prove to be crucial to improve sim-to-real policy transfer, extending the digital-cousins formulation of \cite{dai2024automated}.

\paragraph{Real--to--sim--to--real for navigation and locomotion.}
Related approaches close the same loop for robot navigation~\cite{chhablani2025embodiedsplat} and locomotion~\cite{escontrela2025gaussgym}, but focus on tasks with substantially different physics and contact
requirements from dexterous manipulation.

\paragraph{Real--to--sim policy evaluation.}
Another body of work focuses on using reconstructed simulation environments as a way to reliably evaluate manipulation policies~\cite{li2024evaluating, jain2025polaris, zhang2025real, jangir2025robotarena, badithela2025reliable, han2025re, jiang2025gsworld, yin2026genie}, such that the results strongly correlate with corresponding real-world evaluations.
However, unlike \sysName{}, several such systems do not show that data from their simulation environments can be used to train real--world agents~\cite{li2024evaluating, jain2025polaris, zhang2025real, jangir2025robotarena, badithela2025reliable}, and to our knowledge, no systems support tasks involving general articulated objects.

\paragraph{Real--to--render.}
A separate line circumvents physical simulation entirely and uses the reconstructed scene purely as a rendering target~\cite{yu2025real2render2real}. This bypasses the cost of
physics modeling but, by the same token, makes it difficult to apply to higher-precision contact-rich tasks that depend on accurate dynamics.

\sysName{} belongs to a select group of real--to--sim systems~\cite{han2025re, jiang2025gsworld, yin2026genie} that demonstrate both successful sim-to-real agent transfer and strong correlation between simulated and physical policy evaluations.
However, \sysName{} goes beyond these systems by handling more diverse task conditions (including bimanual, articulation, and multi--step manipulation) and supporting multiple types of cousins of a reconstructed sim-ready scene to fully unlock the diversity of the simulation (see Table~\ref{tab:comparison} for a summary).

\subsection{Imitation Learning from Human Demonstrations and Synthetic Data Generation.}
Robot teleoperation~\cite{mandlekar2018roboturk, zhao2023learning} is a common approach for collecting demonstrations to train robots to perform manipulation tasks autonomously -- here, a human uses a teleoperation device (such as a smartphone or a VR controller) to guide a robot through different tasks, and the resultant robot sensor streams and actions are logged to a dataset.
Robot manipulation policies are often trained on such datasets with Behavioral Cloning (BC)~\cite{pomerleau1989alvinn, schaal1999imitation, Ijspeert2002MovementIW, mandlekar2021matters, chi2023diffusion}.
In recent years, this approach has been scaled up to collect months of data using large teams of human operators~\cite{ebert2021bridge, brohan2022rt, o2024open, khazatsky2024droid}, and has proven to be very effective for robot manipulation~\cite{Calinon2010LearningAR, black2024pi_0, khazatsky2024droid, brohan2023rt}.
However, data collection is a bottleneck, since it is time--consuming and expensive.
A recent line of work leverages synthetic data generation (SDG) in simulation~\cite{mandlekar2023mimicgen, jiang2024dexmimicgen, garrett2024skillmimicgen, wang2023robogen, dalal2023imitating} as a compelling alternative to address the need for large-scale datasets.
Recent evidence has shown that these synthetic datasets can supplement or even replace real-world datasets to reduce the burden of real-world data collection~\cite{maddukuri2025sim, wei2025empirical, bjorck2025gr00t, cheng2025generalizable, tian2025interndata, yin2026genie, haldar2026point}.
We use such tools to highlight an important application of our system -- real--to--sim--to--real policy learning. Here, we reconstruct a simulation environment (along with controlled variations) from a real-world environment, generate synthetic data in simulation, and train manipulation agents that transfer to the real-world, all with minimal human effort.

\clearpage
\section{Scene Reconstruction}
\label{app:scene_recons}
\FloatBarrier

\begin{figure*}[h!]
    \centering
    \includegraphics[width=\textwidth]{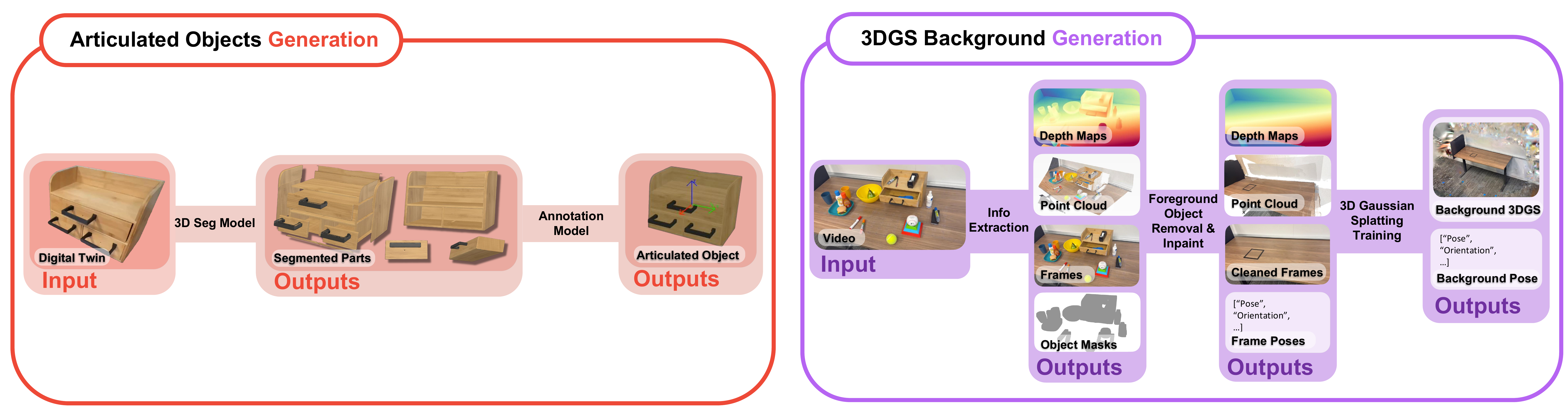}
    \caption{\textbf{Articulation and 3DGS Background Pipeline Overview.} \sysFull generates articulated objects by first decomposing a pre-existing mesh into subsequent parts, which are then annotated with relevant joint types, locations, and ranges via a VLM. \sysFull also can automatically generate a high-fidelity 3DGS background by first generating a synthetic video with removed foreground, extracting extrinsics, and training a 3DDGS to reconstruct the scene geometry.}
    \label{fig:system_articulation_3dgs}
\end{figure*}

\subsection{Extraction Details}
\label{sec:extraction_details}

The extraction stage converts a raw video into a representative RGB-D frame and converts this frame into a scene point cloud. Then, these modalities are used as inputs to iteratively segment foreground objects, removing detected objects with image and depth inpainting such that the subsequent foreground object can be detected from the residual scene. This process produces the per-object RGB-D crops and masks used by the mesh-generation and pose-alignment stages.

\paragraph{Representative Frame Selection} For video inputs, our current pipeline uses frame 0 as the default representative frame to reconstruct scenes. We ask users to start recording videos from a clear point of view that capture the whole scene; ideally from a front-facing view that minimizes occlusion.

\subsection{Foundation Model Details}
\label{sec:vlm_details}
\sysName{} is intended to be modular, and supports multiple foundation models that can be changed during execution. Below, we show the models our pipeline currently supports for each type of foundation model $V_*$:

\begin{itemize}
    \item $V_{im2depth}$: If the input is a single image or video, we utilize DepthAnything3~\cite{lin2025depth}; if the input is a stereo image pair, we utilize FoundationStereo~\cite{wen2025foundationstereo}.
    \item \segimg: We utilize SAM3~\cite{carion2025sam3}.
    \item \sceneModel: We utilize Gemini-Pro-3, though any Gemini or other general purpose VLM can be used by our pipeline.
    \item \imageModel: We utilize Gemini-Pro-3-Image-Preview, though any Gemini or other general purpose image-editing VLM can be used by our pipeline.
    \item $V_{inpaint}^{depth}$: We utilize PriorDepthAnything~\cite{wang2025depthprior}.
    \item $V_{mesh}$: We utilize either Hunyuan2.1~\cite{hunyuan3d2025hunyuan3d21} or TRELLIS.2~\cite{xiang2025trellis2}.

    \item $V_{pose}$: We utilize FoundationPose~\cite{wen2024foundationpose} to refine the 6D pose of the generated mesh with respect to the depth map.

    \item \articulationModel: We utilize Gemini-Pro-3, though any Gemini or other general purpose VLM can be used by our pipeline.
    \item \segmesh: We utilize mainly P3-SAM~\cite{p3sam}, although our pipeline also supports Segment Any Mesh~\cite{tang2024segment} and Partfield~\cite{liu2025partfield}.
    \item $V_{seg}^{video}$: We utilize SAM2~\cite{ravi2024sam2} to propagate the keyframe foreground mask produced by \segimg{} through the remainder of the video.
    \item $V_{inpaint}^{video}$: We utilize VOID~\cite{void} as a two-pass chunked inpainting model to remove foreground pixels from the masked frames and synthesize a clean static-scene RGB stream.
\end{itemize}

\subsection{Articulated Object Generation}
\label{sec:articulated_object_generation}
In this section we detail our articulated object generation pipeline, which extends prior methods such as Articulate Anymesh~\cite{qiu2025articulate} and Articulate Anything~\cite{le2024articulate}.
\paragraph{Segmentation} We first render views of the object from multiple angles, pass these into a VLM \articulationModel, and prompt \articulationModel to list the different parts of the object that can be articulated and the types of joints. For example, this would be a drawer (prismatic) for a cabinet or a door (revolute) for a microwave.
We then segment the mesh of the object with a mesh segmentation model \segmesh, such as P3-SAM~\cite{p3sam} or Segment Any Mesh~\cite{tang2024segment} which assigns an integer label to every face; we refer to each maximal group of faces sharing a label as a segment. Most existing methods only assign labels to external surfaces, but since meshes generated by TRELLIS.2~\cite{xiang2025trellis2} can have internal structures, we propagate these labels to unlabeled mesh faces using a majority-vote label propagation over the face-adjacency.
Mesh segmentation typically yields many more segments than there are articulatable parts (the semantic components named by \articulationModel in the previous step), i.e. the model is over-segmented — a single drawer may span several segments.
To recover the parts, we render the object again from multiple views with each segment shown in a distinct color, and prompt \articulationModel to assign every segment to one of the previously identified parts. Segments mapped to the same part are then merged into a single mesh, yielding one mesh per articulatable part.
This segmentation and assignment can optionally be refined by the user via a GUI.

\paragraph{Joint Generation} We adapt the actor-critic algorithm and API from Articulate Anything~\cite{le2024articulate} to generate the joint parameters.
We prompt \articulationModel to predict the joint axes and placements of each part by providing a python API which can generate URDFs.
The API allows \articulationModel to place joints relative to parts (for example, a revolute joint can be placed along the left edge of a door), which helps ground \articulationModel $\;$ and simplify its task.
\articulationModel generates code which calls this API, and the result is compiled into a URDF\@.
We then move the joints of the object according to this URDF in a simulator, and render a video of this movement.
The video is then judged by a separate critic VLM, which is asked to rate the accuracy and realism of the movement and provide feedback for improvement if necessary.
\articulationModel is prompted to improve its prediction by incorporating this feedback, and this process continues until the critic gives a score above a threshold.
\paragraph{Physical Parameters} Finally, we prompt \articulationModel to generate physical parameters such as link mass, joint friction, and damping. We provide \articulationModel with the calculated volume of each part and the entire object to better contextualize its predictions.

\subsection{Object Depenetration and Physical Stability}
\label{app:physical_stability}
Although objects' poses are estimated by foundation models, minor estimation errors can still lead to interpenetration between neighboring objects. To generate a physically plausible scene, we perform a depenetration step.

First, we generate objects' respective collision meshes using CoACD~\cite{wei2022approximate}. Then, the reconstructed scene is spawned in PyBullet, after which the physics simulation is subsequently stepped (and force-setting objects' velocities to be zero after every step to avoid potential explosions from de-penetration) until the object poses settle. This final set of poses is then cached, guaranteeing that the scene will be physically stable during subsequent initializations.

\subsection{Background Reconstruction and Alignment}
\label{sec:bg_details}

We provide two routes for reconstructing the static background as a 3D Gaussian Splat~\cite{kerbl20233d} and registering it with the \sysName{} reconstructed scene's world frame: (a) a fully automatic pipeline that reuses the single raw capture from Extraction, and (b) a manual pipeline that leverages a second foreground-free capture. The two emit identical asset structures and differ only in how the splat is obtained and aligned. We describe each in turn and conclude with a comparison of their trade-offs.

\subsubsection{Automatic Background Reconstruction and Alignment}
\label{sec:auto_bg_details}

The automatic pipeline runs end-to-end from the same raw capture used by Extraction and emits a splat rigidly aligned to the reconstructed scene's world frame. It comprises four phases: (1) frame preparation and foreground inpainting, (2) metric depth and pose recovery, (3) depth-supervised splat training, and (4) the rigid bridge into the simulator world frame.

\paragraph{Frame preparation and foreground inpainting.}
We uniformly subsample the source video to a fixed frame budget that balances dense viewpoint coverage against downstream memory. To construct a per-frame foreground mask for all foreground object, we first query \sceneModel{} on
a single keyframe to enumerate the foreground object categories, refine those prompts into pixel-accurate masks with \segimg, and propagate the merged keyframe mask temporally with a video
segmentation model $V_{seg}^{video}$. The resulting binary mask stack drives a two-pass video inpainting model $V_{inpaint}^{video}$: the first pass fills masked regions with a plausible static-scene completion, and the second re-inpaints residual hallucinations using the first pass as conditioning. We run $V_{inpaint}^{video}$ in chunks to respect GPU memory, yielding a sequence of clean RGB frames with no foreground objects.

\paragraph{Metric depth and pose recovery.}
We apply DepthAnything3~\cite{lin2025depth} to two frame streams: (a) the original frames, to recover sharp camera poses unbiased by inpainting artifacts, and (b) the inpainted frames, to recover depth consistent with the RGB stream the splat trains against. Since DepthAnything3~\cite{lin2025depth}'s forward pass is limited to a bounded number of frames, we process it in overlapping chunks; within each chunk it returns metric depth maps, per-pixel confidences, intrinsics, and world-to-camera extrinsics
in a local reference frame. We merge chunks into a single trajectory by fitting an Umeyama~\cite{umeyama} similarity between each chunk and the first on the shared camera centers in their overlap region; this recovers scale and translation to within millimeter residuals under smooth
capture motion.

The inpainted-stream and original-stream depth maps live in two independent metric worlds due to two independent DepthAnything3~\cite{lin2025depth} forward passes, and may not align well with each other. Hence, we back-project the inpainted depth
into a dense point cloud, downsample it, and apply a second Umeyama fit to align it into the original-stream world. The transformed cloud is exported as a seed PLY that initializes the splat's positions, so training begins from a geometrically plausible cloud rather than a random prior.

\paragraph{Depth-supervised splat training.}
We train a 3D Gaussian Splat in NerfStudio~\cite{nerfstudio} against the inpainted RGB frames, using the \emph{original-stream} camera poses and the seed PLY as initialization, under two training losses: a standard photometric loss between rendered and ground-truth inpainted RGB, and an $L1$ depth loss between
rendered depth and the inpainted-stream depth from DepthAnything3~\cite{lin2025depth}, masked by a per-pixel confidence threshold and weighted by a fixed coefficient. Pairing photometric and depth supervision on the \emph{same} inpainted RGB-D stream keeps the two losses mutually
consistent; the depth term suppresses floaters that purely photometric training tends to place above textureless flat surfaces to explain view-dependent shading.

To absorb residual misalignment between the original-stream poses (used for camera placement) and the inpainted-stream depth (used for geometric supervision), which arises from DepthAnything3~\cite{lin2025depth} sub-pixel pose noise and the small pose offset induced by inpainting, we enable a per-camera $\mathrm{SO}(3)\!\times\!\mathbb{R}^3$ pose optimizer that learns a small rigid perturbation per training camera. We find this to be the single most impactful design choice for splat sharpness: without it, the splat is consistently blurry regardless of frame count, resolution, or iteration budget.

\paragraph{Rigid bridge into the simulator world.}
The trained splat lives in the original-stream camera world. To register it with the reconstructed scene---expressed in a ground-plane-aligned world frame estimated by Extraction---we compose two transforms: the cam2world pose of an anchor frame from DepthAnything3~\cite{lin2025depth} on the original stream, and the cam2world of the same anchor frame from Extraction's ground-plane fit. Their
composition is a single rigid transform $M_{src\rightarrow og}$ mapping any point in the splat's world into the simulator world. The registered splat is added alongside the per-object meshes from
Generation as an additional scene asset, producing a simulation-ready scene whose static geometry and appearance faithfully reproduce the captured environment.

\subsubsection{Manual Background Alignment}
\label{sec:manual_bg_details}

When the user has physical access to the captured environment, a more stable alternative is available: record a \emph{second} video of the same scene with all foreground objects removed, train
the splat directly on this clean stream, and align the result through the \sysName{} interactive scene editor. We describe this path in three phases.

\paragraph{Clean-stream capture and pose estimation.}
The user re-films the same environment with the same camera and a similar trajectory after physically clearing the foreground objects. We process the video through the standard Nerfstudio~\cite{nerfstudio} pipeline, which uniformly extracts frames and recovers camera intrinsics and per-frame extrinsics via
COLMAP~\cite{schoenberger2016sfm} structure-from-motion. The pipeline emits the successfully registered frames together with a sparse SfM point cloud initializing the splat.

\paragraph{Splat training on the foreground-free video.}
We train a \texttt{splatfacto-big} 3D Gaussian Splat~\cite{kerbl20233d} on the registered frames via Nerfstudio's \texttt{ns-train} entrypoint, supervised only by the standard photometric loss against the captured RGB frames, and export it to a PLY via \texttt{ns-export} for downstream loading. The absence of
inpainted RGB content typically yields a noticeably sharper reconstruction than the automatic pipeline, particularly on textureless flat surfaces and along silhouettes where the automatic pipeline must synthesize plausible content.

\paragraph{Interactive alignment via the scene editor.}
Unlike the automatic pipeline---where the splat's world frame is linked to the foreground capture through shared poses and can therefore be bridged into the simulator world by a derived rigid transform---the clean-stream capture shares no common camera trajectory with the original capture, and the COLMAP
world it lives in is metrically arbitrary up to a similarity. There is thus no obvious automated way to compute its alignment with the reconstructed scene. Instead, we expose the trained splat as an
interactive prim inside the \sysName{} scene editor. The user is presented with a side-by-side rendering of the foreground scene (per-object meshes from Generation in their estimated poses) and the background splat, and applies $\mathrm{SE}(3)$ transformations (3-DoF translation, 3-DoF rotation) and an isotropic scale to the splat prim via keyboard commands. Once satisfied, the final $M_{src\rightarrow og}$ transform is serialized alongside the splat asset so subsequent launches load the manually-aligned scene
without further intervention.

\subsubsection{Comparison and Trade-offs}
\label{sec:bg_comparison}

The two pipelines emit identical asset structures and achieve comparable reconstruction quality, differing only in capture requirements and the failure modes inherited from their respective
reconstruction paths.

The \textbf{automatic pipeline} requires only the single video already captured for Extraction and no user interaction, making it the preferred choice when capture effort must be minimized, when
  only one video is available, or when the scene is impractical to clear---for example, scenes with fixed installations, public spaces, or archival footage. Because its background is bridged
  analytically into the original capture's camera frame, it inherently reproduces the exact viewpoint geometry, making it the preferred choice when high reproducibility from a specific camera
  frame is required (reflected in its consistently higher alignment scores in Table~\ref{tab:bg_comparison}). Its principal limitation is that the inpainting model must synthesize content behind removed foreground objects, which can introduce subtle hallucinations on textureless flat surfaces and along object silhouettes. Additionally, the foreground inpainting step for two-pass denoising requires around 90 minutes on a single NVIDIA RTX 3090. However, this process is independent of the main \sysName{} pipeline and can run in parallel in a multi-GPU setting.

  The \textbf{manual pipeline} trades a second capture for higher background surface fidelity: because it trains directly on a genuinely foreground-free video, it avoids inpainting artifacts
  entirely and produces sharper reconstructions on exactly the surfaces where the automatic pipeline struggles. It is preferable when reconstruction quality is the dominant concern and a second
  capture is feasible. Its costs are the additional capture effort, physical access to clear the scene, and a brief manual alignment step, since the clean stream shares no camera trajectory with
  the original capture and cannot be bridged automatically; consequently, its background cannot be registered to the original viewpoint as precisely as the automatic pipeline's, and exact
  alignment is difficult to achieve by hand.

For qualitative and quantitative comparison between two pipelines, refer to Appendix~\ref{app:3dgs-background-analysis}.

\clearpage
\raggedbottom%
\section{Digital Cousins Augmentation}
\label{app:scene_variations}

\label{sec:variation_details}

\subsection{Object Cousins Augmentation}
To systematically generate diverse yet realistic object cousins, we make use of an automated generation pipeline using a VLM and an image generation model. This approach uses the segmented object appearance to create object cousins that maintain the identity of the original object and extend its distribution in terms of shape, structure, and appearance.

The pipeline operates through the following steps:
\begin{enumerate}

\item \textbf{Object Canonicalization and Context Extraction}: For a given reconstructed object, the input of this system includes the isolated object image with a transparent background, which is generated by the reconstruction pipeline. In addition, the original scene image is also retrieved, allowing generated object variants to be compatible with their surrounding environment. The object name is also canonicalized by removing irrelevant information, including material, size, and transient states.

\item \textbf{Functional Component Decomposition}: First, the VLM is prompted to decompose the object into functional components based on grasp affordance. This process will result in a structured list of parts, such as handle, lid, body, base, etc., and this will provide a semantically meaningful basis for localized cousin generation rather than applying unconstrained variation to the object as a whole.

\item \textbf{Dimension-Specific Cousin Proposal}: For each functional component, the VLM proposes multiple candidate cousins along three predefined dimensions: \textit{geometry}, \textit{topology}, and \textit{visual appearance}. Geometry changes involve continuous shape attributes, topology changes involve the structure of the object, and visual variations involve surface-level properties such as texture or material appearance. To ensure that the model is realistic, it is specifically instructed to only create deterministic, everyday object variants and to avoid implausible or unusual modifications. To this end, the model is also not allowed to create unrealistic cousins by altering the topology of the object if it is not feasible.

\item \textbf{Scene-Aware Image Synthesis}: Each of the proposed component-level variation is then applied by the image generation model to generate a modified image of the object. The model is asked to change only the specified component, keep other components unchanged, maintain a realistic appearance, and produce the result with a semi-transparent background. The scene image is also given as input for better matching of the generated object image with the original scene image.

\item \textbf{Reasonableness Verification and Structured Output}: To filter out low-quality generations, each synthesized object is checked by the VLM for plausibility in the real world and scene consistency. Those found to be implausible or inconsistent with the scene are discarded, with a bounded fallback approach maintained for coverage over variation dimensions. The accepted results are then stored, along with relevant metadata information such as the identity of the components, variation dimension, textual descriptions, prompts, and verification results, to make them readily usable by subsequent data generation and simulation pipelines.
\end{enumerate}
The detailed prompt template used for object cousins augmentation is shown in Fig~\ref{fig:object_cousins_prompt}.

\begin{figure*}[htbp]
    \centering
    \begin{tcblisting}{
        colback=gray!5!white,
        colframe=gray!75!black,
        title=VLM Prompts for Object Cousins Generation,
        arc=2mm,
        boxrule=0.5pt,
        boxsep=0.5mm,
        left=1mm,
        right=1mm,
        top=0.75mm,
        bottom=0.75mm,
        listing only,
        listing options={
            basicstyle=\scriptsize\ttfamily,
            breaklines=true,
            breakatwhitespace=false,
            columns=fullflexible,
            keepspaces=true,
            showstringspaces=false
        }
    }
Prompt A: Functional Component Decomposition

With the given object image, decompose the object into functional components based on grasp affordance, list the component name only in number list, if there is a main body, treat its edge or bottom as a same component inside main body. Make sure to only generate the list of component names, don't generate any other additional text.

Prompt B: Topology Feasibility Check

For the {component} of the object shown, is it reasonable to make variation on topology such that the modified object still looks realistic and is reasonable enough that we would see this object in our daily task? Answer with simply yes or no. If it is likely not to be normal, make it a no.

Prompt C: Component-wise Variation Proposal

For the component '{component}', list {num_geometry} ways to generate variation through Geometry dimension, {num_topology} ways through Topological dimension, and {num_visual} ways through Visual dimension:

1. Geometry dimension
2. Topological dimension
3. Visual dimension

Make sure to be determined in the description, prevent using words like or, such as, like, or similar to in the description.
Make sure to still maintain the object to have a realistic and reasonable look like something you would see in the daily life.
For example, a red banana and a banana with a hole in it is a bad variation, because we rarely see those in daily life.
But a banana with green skin is a good variation since the unripe banana is usually green.

Format your response as:
Component: {component}
Geometry:
1. [variation description]
...
Topology:
1. [variation description]
...
Visual:
1. [variation description]
...

If topology variation is judged infeasible, the prompt instead requests only Geometry and Visual variations.

Prompt D: Scene-Aware Object Editing

Modify the following component of the object: '{component}'.
Make sure to generate a realistic modified version, make sure it looks normal and reasonable like something you would usually see in the daily life.
Do not generating variations that are weird or unusual, such as a banana with a hole in it or a banana with a red skin.
Apply this {dimension} variation: {variation_description}.
Keep all other components unchanged.
Generate the image with transparent background.
Make sure the modified object style fits the scene context.

Prompt E: Real-World and Scene Reasonableness Verification

You will see a scene image and a candidate object image. Judge whether the candidate object is reasonable in the real world and also reasonable in the given scene. Be strict and practical: reject unusual, implausible, or scene-inconsistent objects.

Return format:
RealWorldReasonable: <yes/no>; RealWorldReason: <short reason>
SceneReasonable: <yes/no>; SceneReason: <short reason>
    \end{tcblisting}
    \caption{\textbf{Object Cousins Prompt Templates.} Prompt templates used in the object cousins generation pipeline include functional decomposition, component-wise variation proposal, scene-aware object editing, and real world/scene reasonableness verification. At runtime, the variables in curly brackets are replaced with dynamic values per object and/or component.}
    \label{fig:object_cousins_prompt}
\end{figure*}

\subsection{Scene Cousins Augmentation}
Starting from the canonical spatial arrangement of objects in the generated scene, we apply randomization to vary their relative placements \textit{semantically}.
For instance, if the reconstructed scene has a spoon placed to the right of a plate, we vary the spoon's placement to be placed on top of, or to the left of, the plate. Additionally, we select and place distractor objects that are feasible for a given scene. This augmentation has the following steps:

\begin{enumerate}
    \item \textbf{Predicate Sampling}: for a given scene and task, an anchor object is selected. This anchor object can be selected automatically by a VLM based on the task, or by the user. Then, for each of the other objects, a spatial predicate (or multiple predicates) is selected from the following list: [\textbf{LeftOf, RightOf, InFrontOf, Behind, OnTopOf, Inside}]. The selected predicates specify the semantic placement of the object relative to the anchor object. The list of possible predicates for a given task and object is specified by the user and depends on the task. The object is then instantiated based on the predicates. \textbf{Note:} multiple predicates can be combined, i.e., an object can be instantiated both \textbf{LeftOf} and \textbf{InFrontOf} the anchor.
    \item \textbf{Distractor Objects}: next, a number of distractor objects up to a specified limit can be added to the scene. The distractor objects are selected from the BEHAVIOR~\cite{li2023behavior} dataset of objects and can be filtered based on the following attributes: \textit{mass}, \textit{volume}, \textit{density}, and \textit{object category}. The selected distractor objects are placed in the scene so that they do not collide with existing objects or with each other to maintain physics stability.
\end{enumerate}

\subsection{Task Cousins Augmentation}
\label{task_cousins_appendix}
To systematically generate diverse and executable manipulation tasks for reconstructed scene, we employ an automated task proposal pipeline driven by a VLM\@. This methodology leverages both visual context and structured scene metadata to define realistic tabletop tasks tailored to the specific configuration of each scene.

The pipeline operates through the following steps:
\begin{enumerate}

\item \textbf{Scene Context Extraction}: For a given scene, the system captures a 2D image for the reconstructed scene in simulation and a list of available interactable objects from the scene's state representation.
\item \textbf{Constraint Formulation}: To ensure physical realism and executability, the system incorporates specific robot constraints (e.g., maximum gripper length, single / bimanual arm) and optional object-level constraints (e.g., mandatory inclusion or exclusion of specific items across tasks).
\item \textbf{VLM Prompting}: The VLM is provided with the scene image, the filtered object list, the physical constraints, and a predefined set of allowable object predicate states (e.g., OnTop, Inside, Under) in a simulator such as OmniGibson~\cite{li2023behavior} or Isaac Lab~\cite{mittal2025isaac}. The VLM acts as a robotics expert and proposes a specified number of distinct tasks. Crucially, the VLM is instructed to ensure that each proposed task requires a meaningful state change from the scene's initial configuration.
\item \textbf{Configuration Generation}: The VLM outputs structured task definitions, including semantic group mappings and goal conditions formulated as logical predicates. These outputs are parsed and automatically compiled into standardized files that can be utilized directly during data generation as outlined in Appendix~\ref{sec:datagen}.
\end{enumerate}
This automated approach enables the rapid, scalable generation of varied task distributions that are directly compatible with the reconstructed scene by \sysName{}, facilitating extensive data collection and policy evaluation without the bottleneck of manual task engineering.

To ensure reproducibility, we provide the exact prompt template used to query the VLM in Figure~\ref{fig:vlm_prompt}. Variables enclosed in curly braces (e.g., \texttt{\{num\_tasks\}}) are dynamically populated based on the scene configuration and user constraints at runtime.

\begin{figure*}[htbp]
    \centering
    \begin{tcblisting}{
        colback=gray!5!white,
        colframe=gray!75!black,
        title=VLM Prompt for Task Proposal,
        arc=2mm,
        boxrule=0.5pt,
        boxsep=0.5mm,
        left=1mm,
        right=1mm,
        top=0.75mm,
        bottom=0.75mm,
        listing only,
        listing options={
            basicstyle=\scriptsize\ttfamily,
            breaklines=true,
            breakatwhitespace=false,
            columns=fullflexible,
            keepspaces=true,
            showstringspaces=false
        }
    }
You are a professional robotic expert in sim-to-real, robot learning, and VLAs.

Look at the provided image of a tabletop scene in a simulation (OmniGibson). Using ONLY the objects listed below, propose exactly {num_tasks} distinct tasks that a {robot_type} arm can perform in this scene, ensure that each task goal requires a change from the current configuration of the scene, for example, do not propose tasks to put a pear in bowl if it is already in the bowl. Each task should have clear goal conditions expressible with object states.

Important:
{robot_constraint}
{object_constraints}

Scene object list (id -> category, name; use these names as group identifiers in predicates):
{object_list}

Output format: for each of the {num_tasks} tasks, output a YAML block that can be merged into a task config. Use this exact structure for each task.

For each task provide:
1. task_name: short snake_case name (e.g. stack_cup_on_plate)
2. semantic_group_mapping: map each logical group name you use (e.g. cup, plate) to a list containing exactly one scene object "name" from the list above (e.g. cup: [blue_cup_syvtml_19], plate: [teal_plate_qdudop_16]).
3. goal_predicates_all: list of predicates that must ALL be true for success. Each predicate has: state, state_kwargs (null if not needed), value (true/false), group, other_group (only for binary states like OnTop, Touching).
4. goal_predicates_any: list of predicates where ANY being true yields success (optional; use null if not needed).

Allowed states (from OmniGibson object_states): {states}

For binary relations use group and other_group. For unary states (Open, ToggledOn, etc.) use group and set other_group to null or omit.

Example predicate format:
  goal_predicates_all:
    - state: OnTop
      state_kwargs: null
      value: true
      group: cup
      other_group: plate
  goal_predicates_any: null

Output exactly {num_tasks} tasks. Separate each task with "---" and a task number (e.g. "--- Task 2"). Each task block must be valid YAML with keys: task_name, semantic_group_mapping, goal_predicates_all, goal_predicates_any.
    \end{tcblisting}
    \caption{\textbf{Task Cousins Prompt Template.} The exact prompt template used to query VLM for proposing tabletop manipulation tasks. Bracketed variables are populated dynamically per scene.}
    \label{fig:vlm_prompt}
\end{figure*}

\subsubsection{Task Cousins Example}
\label{sec:task_variation_example}
\begin{figure}[H]%
    \centering

    \begin{subfigure}{0.48\textwidth}
        \centering
        \includegraphics[width=\linewidth]{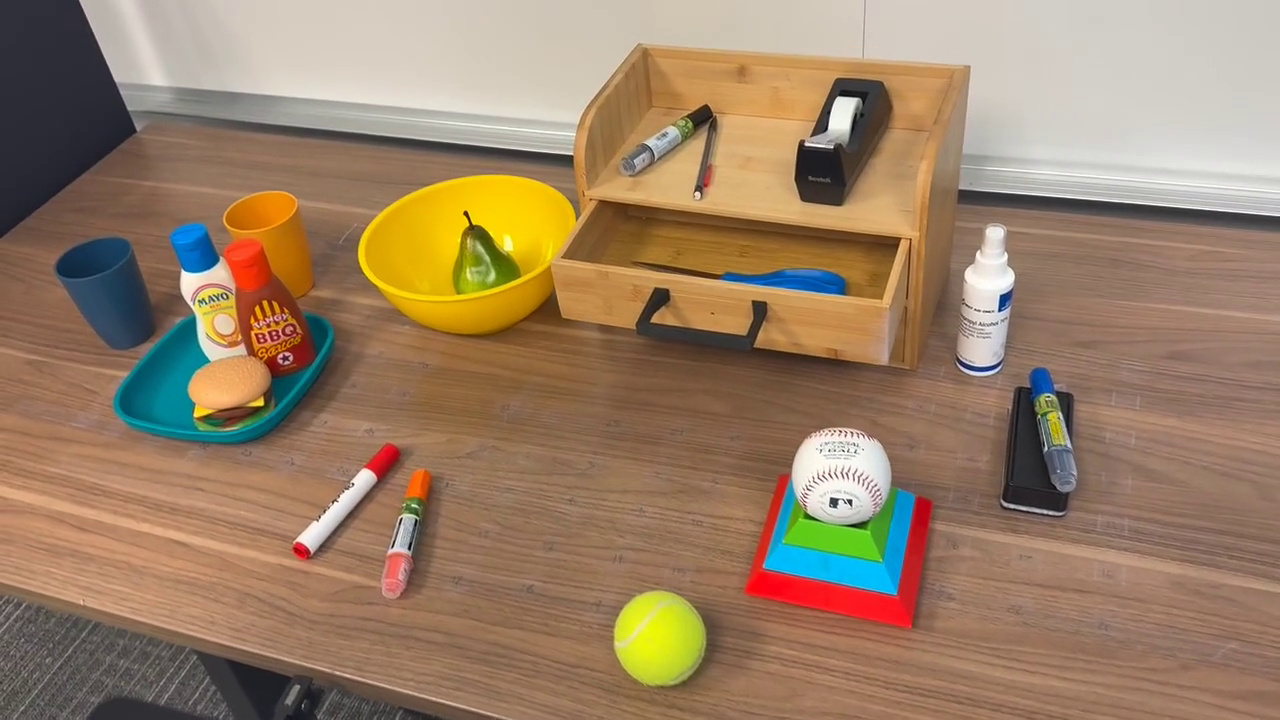}
        \caption{Real Scene}
        \label{fig:setup_a}
    \end{subfigure}\hfill%
    \begin{subfigure}{0.48\textwidth}
        \centering
        \includegraphics[width=\linewidth]{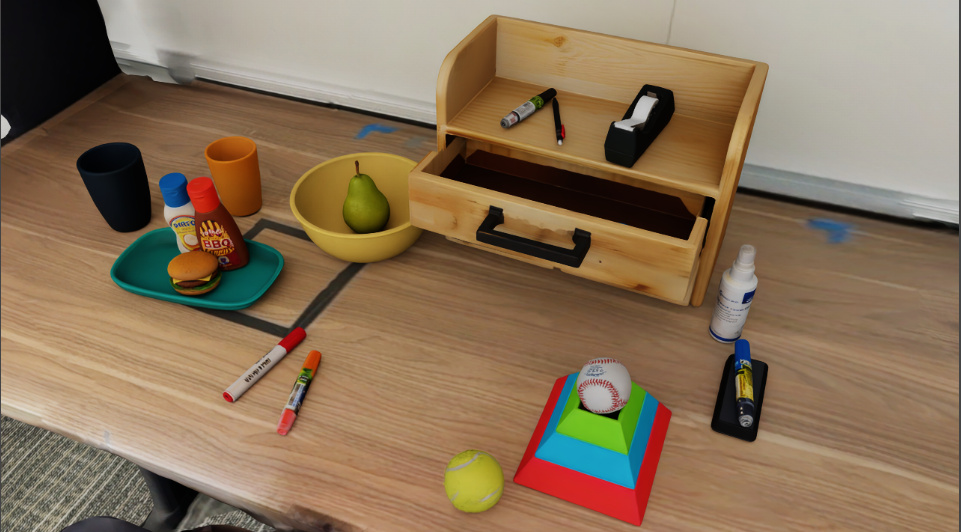}
        \caption{\sysName{} Output}
        \label{fig:setup_b}
    \end{subfigure}

    \caption{\textbf{Task Cousins Generation Example.} Real and Sim Scene used in task cousins generation example}
    \label{task_cousins_example}
\end{figure}

First, we record the cluttered scene video and run \sysName{} to generate the reconstructed scene, shown side by side in Fig~\ref{task_cousins_example}. We then run the task cousins augmentation pipeline described in Section~\ref{task_cousins_appendix} to generate 13 proposed tasks in this scene. We then collect 10 demos for each proposed task via human teleoperation and run MimicGen~\cite{mandlekar2023mimicgen} to generate 100 demos for each task. Finally, we use all subsequent demos to finetune \openpi~and rollout a single multi-task policy in the sim and real setup.
The simulation and real-world evaluation results are shown in \autoref{tab:multi_task_policy_eval}. This example demonstrates how \sysName{}'s sim-ready scenes could unlock a large variety of tasks and offer a method to scale up multi-task policy training.

\newcommand{\robotcardwithphototwo}[4]{%
\begin{tcolorbox}[
    enhanced,
    colback=gray!5,
    colframe=blue!50!black,
    title=Task: #1,
    fonttitle=\bfseries,
    coltitle=white,
    boxrule=0.5pt,
    sharp corners=south,
    top=1mm,
    bottom=1mm,
    left=1mm,
    right=1mm,
    sidebyside,
    sidebyside align=top,
    lefthand ratio=0.5,
    before skip=10pt,
    after skip=10pt
]
\small
\textbf{Instruction:} #2 \\[1ex]
\textbf{Goal:} #3
\tcblower
\centering
\includegraphics[width=\linewidth, height=2.5cm, keepaspectratio]{#4}
\end{tcolorbox}
}

\robotcardwithphototwo{Place Baseball in Bowl}
    {Pick up the baseball, then put it in the bowl.}
    {Baseball Inside of Bowl}
    {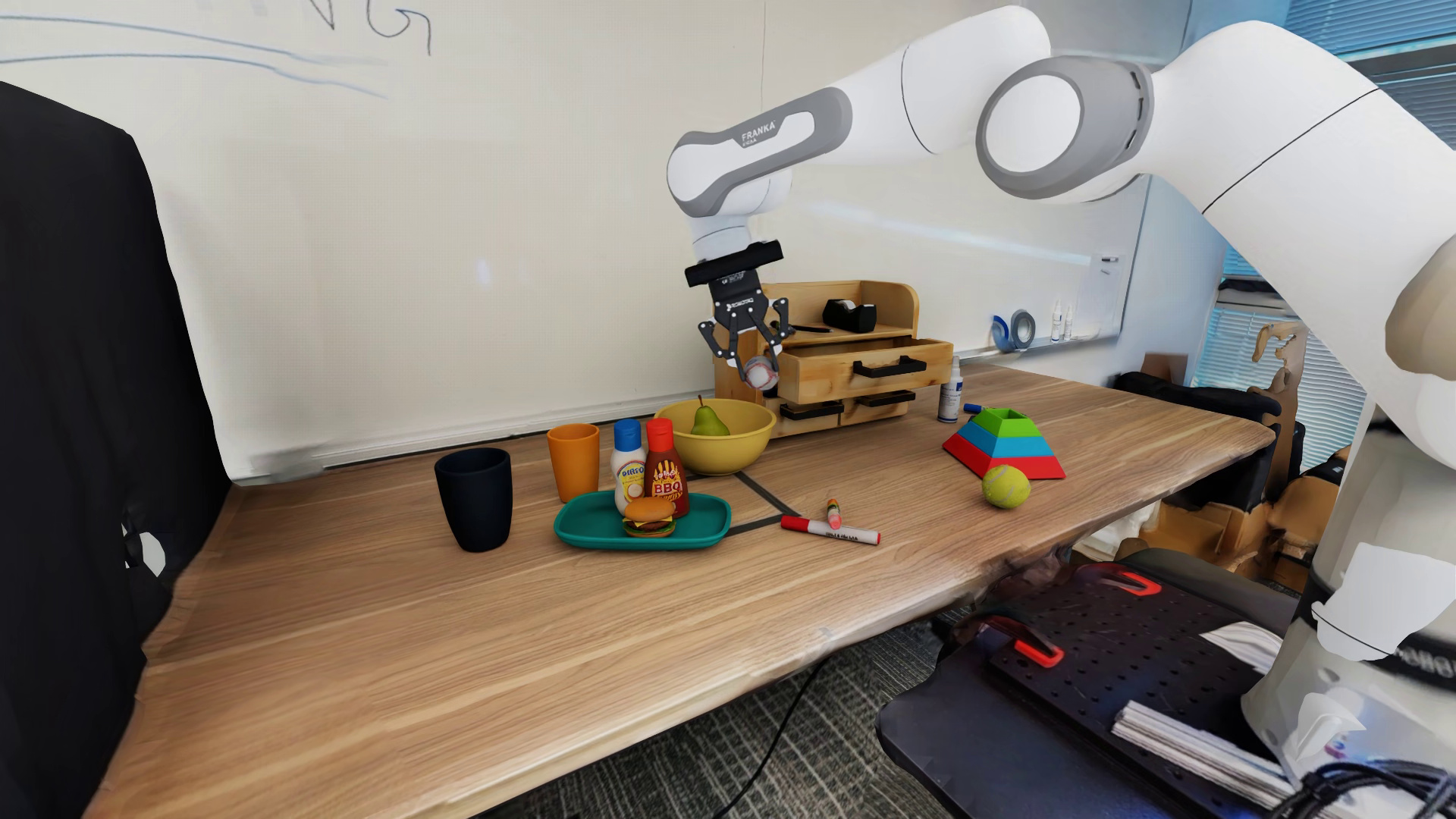}

\robotcardwithphototwo{Place Black Eraser on Organizer}
    {Pick up the black eraser, then put it on the organizer.}
    {Black Eraser OnTop of Organizer}
    {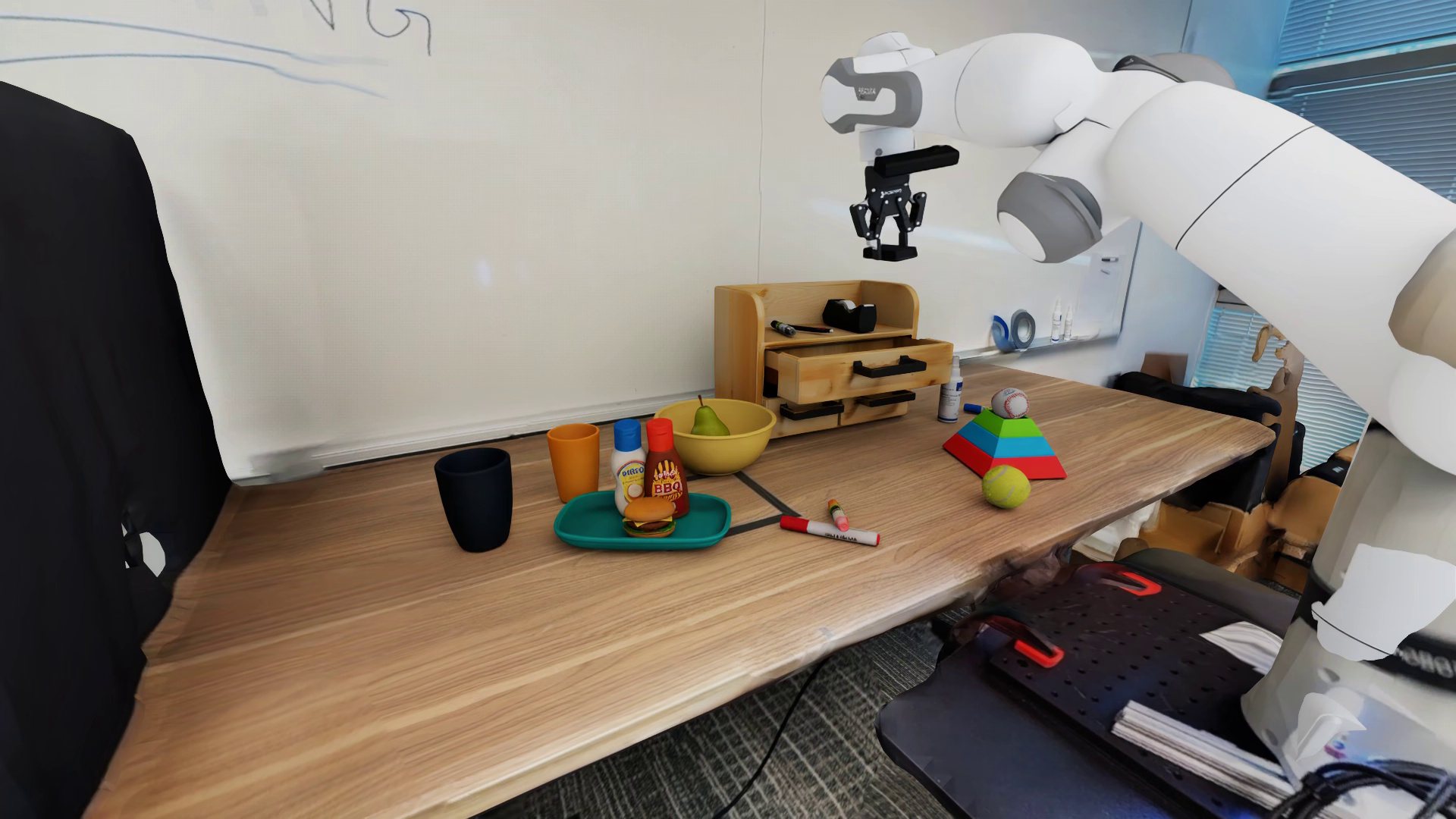}

\robotcardwithphototwo{Place Burger in Bowl}
    {Pick up the burger, then place the burger in the bowl.}
    {Burger Inside of Bowl}
    {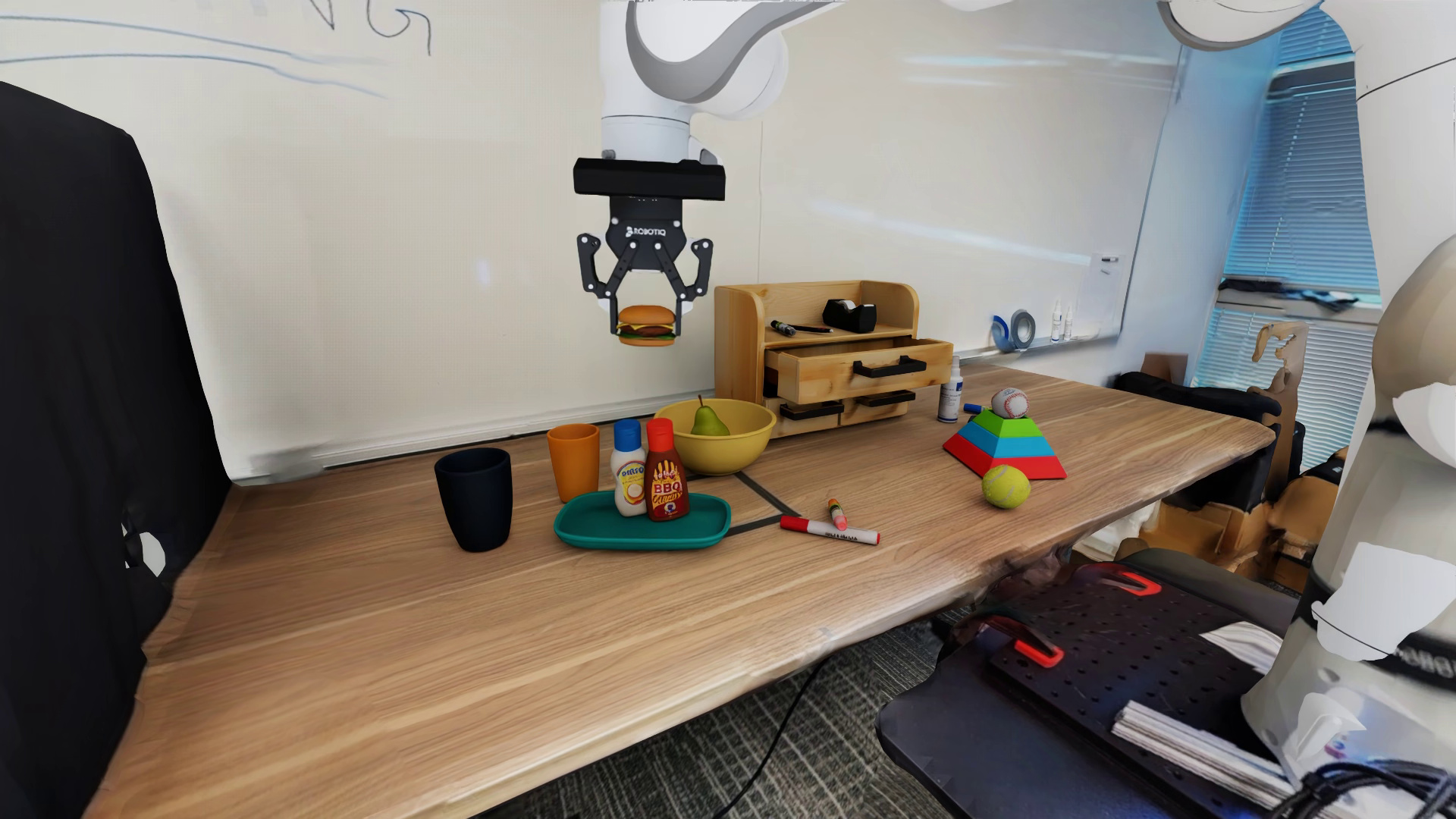}

\robotcardwithphototwo{Place Orange Marker on Organizer}
    {Pick up the orange marker, then put the orange marker on the organizer.}
    {Orange Marker OnTop of Organizer}
    {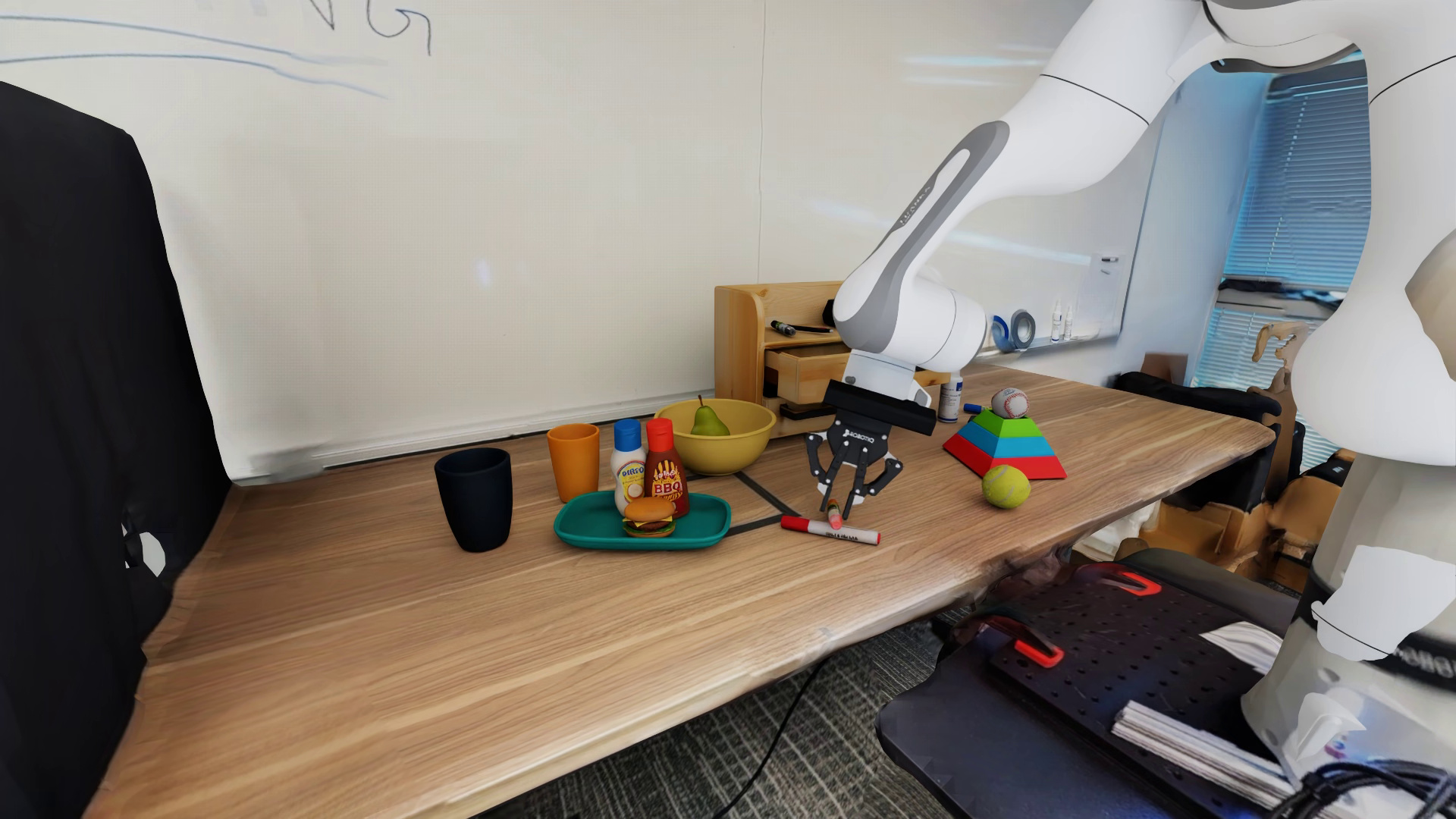}

\robotcardwithphototwo{Place Red Bottle in Bowl}
    {Pick up the red bottle, then put the bottle in the bowl.}
    {Red Bottle Inside of Bowl}
    {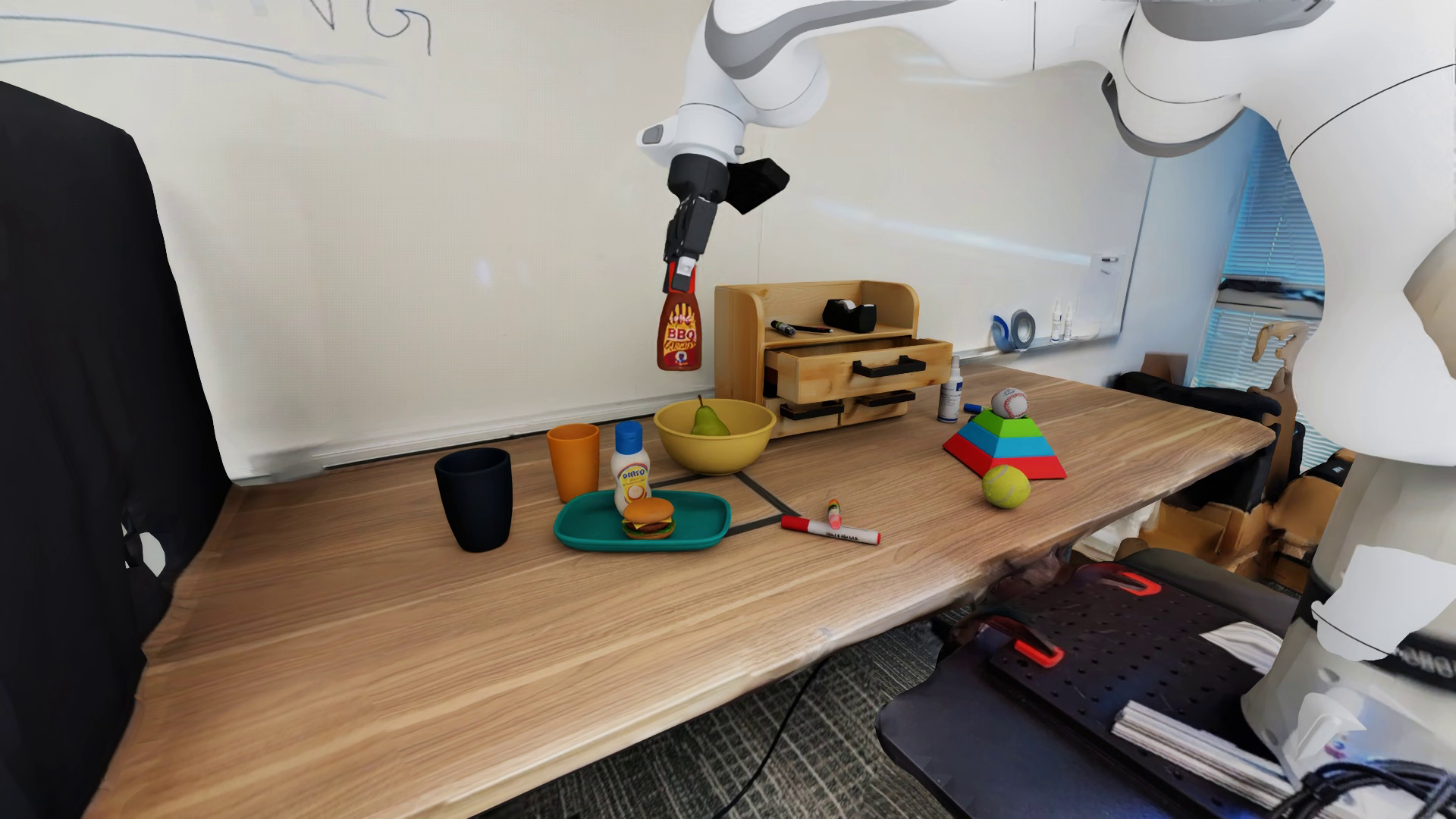}

\robotcardwithphototwo{Place Red Marker in Cup}
    {Pick up the red marker, then put the red marker in the cup.}
    {Red Marker Inside of Blue Cup}
    {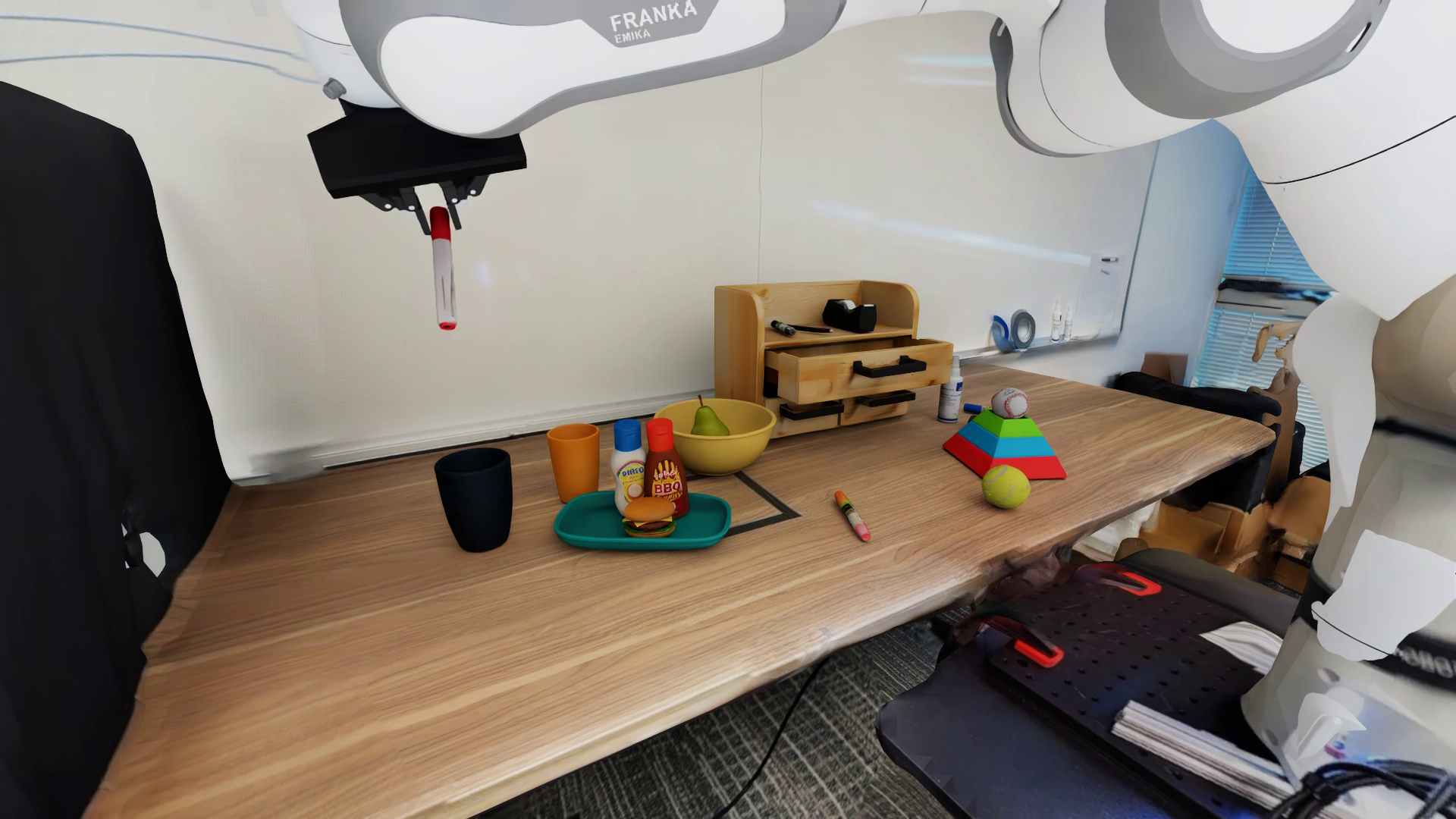}

\robotcardwithphototwo{Place Tennis Ball in Bowl}
    {Pick up the tennis ball, then put the tennis ball in the bowl.}
    {Tennis Ball Inside of Bowl}
    {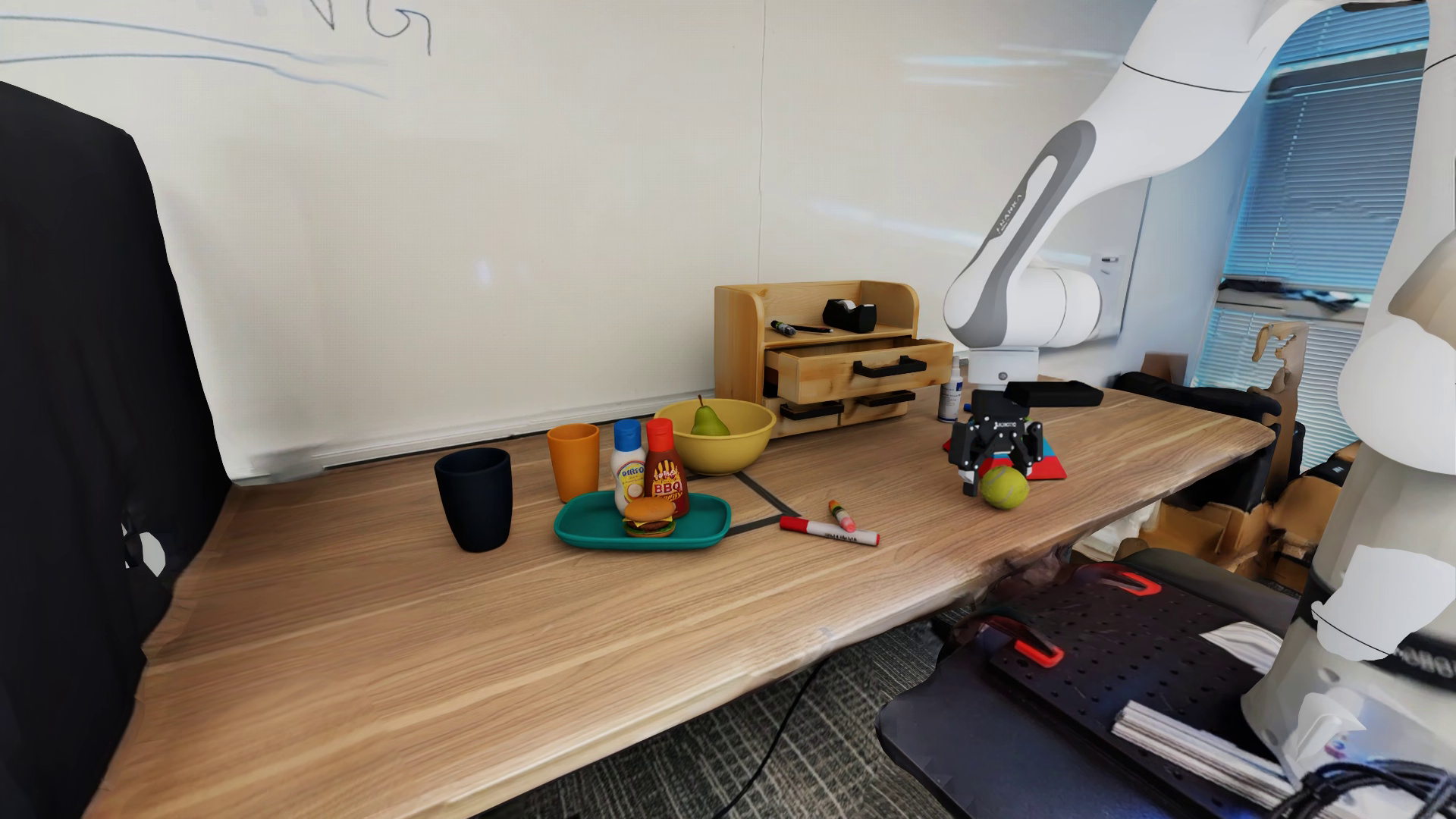}

\robotcardwithphototwo{Place Tennis Ball on Organizer}
    {Pick up the tennis ball, then put the tennis ball on the organizer.}
    {Tennis Ball OnTop of Organizer}
    {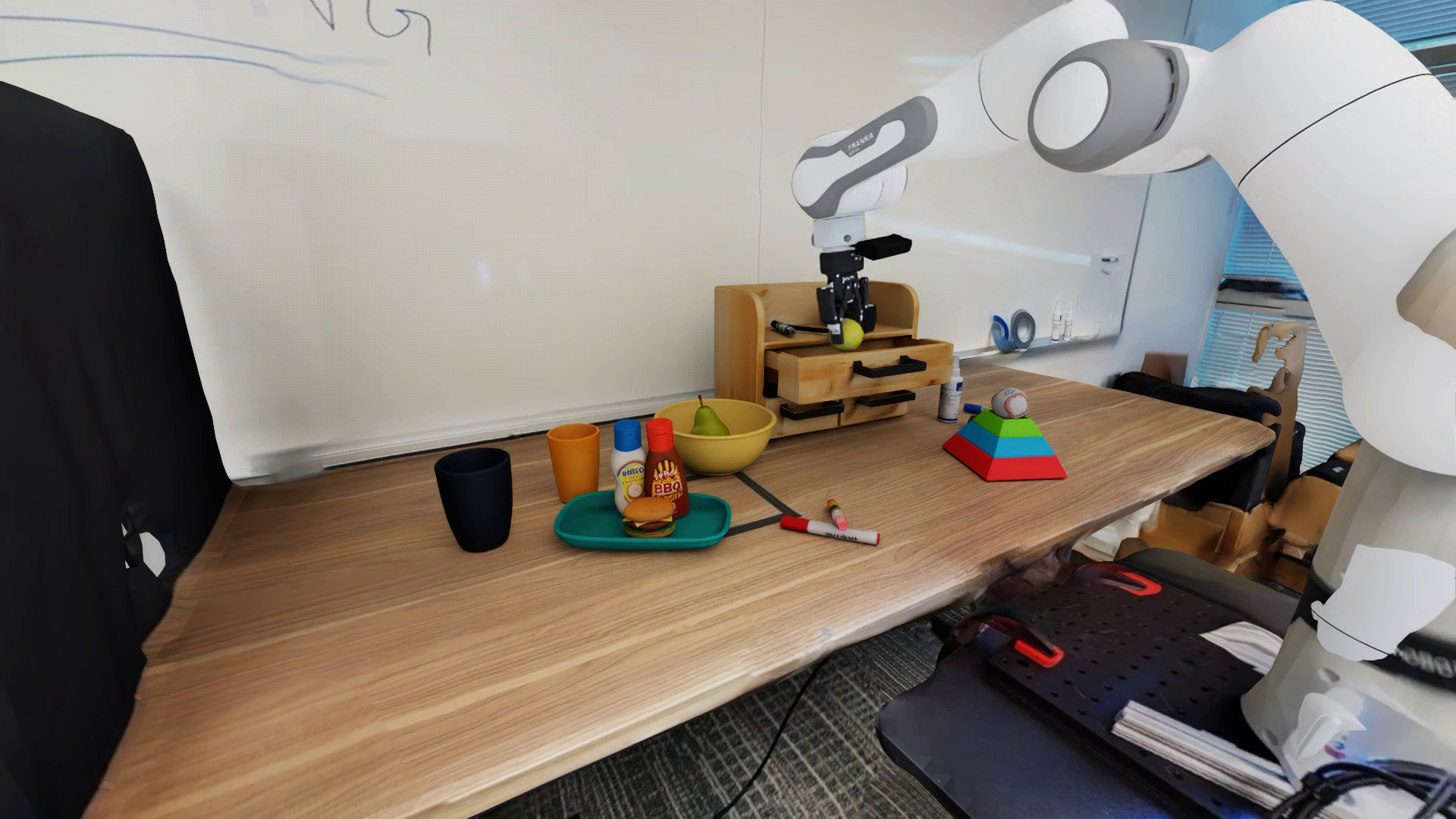}

\robotcardwithphototwo{Place Black Marker on Plate}
    {Pick up the black marker, then put the black marker on the plate.}
    {Black Marker OnTop of Plate}
    {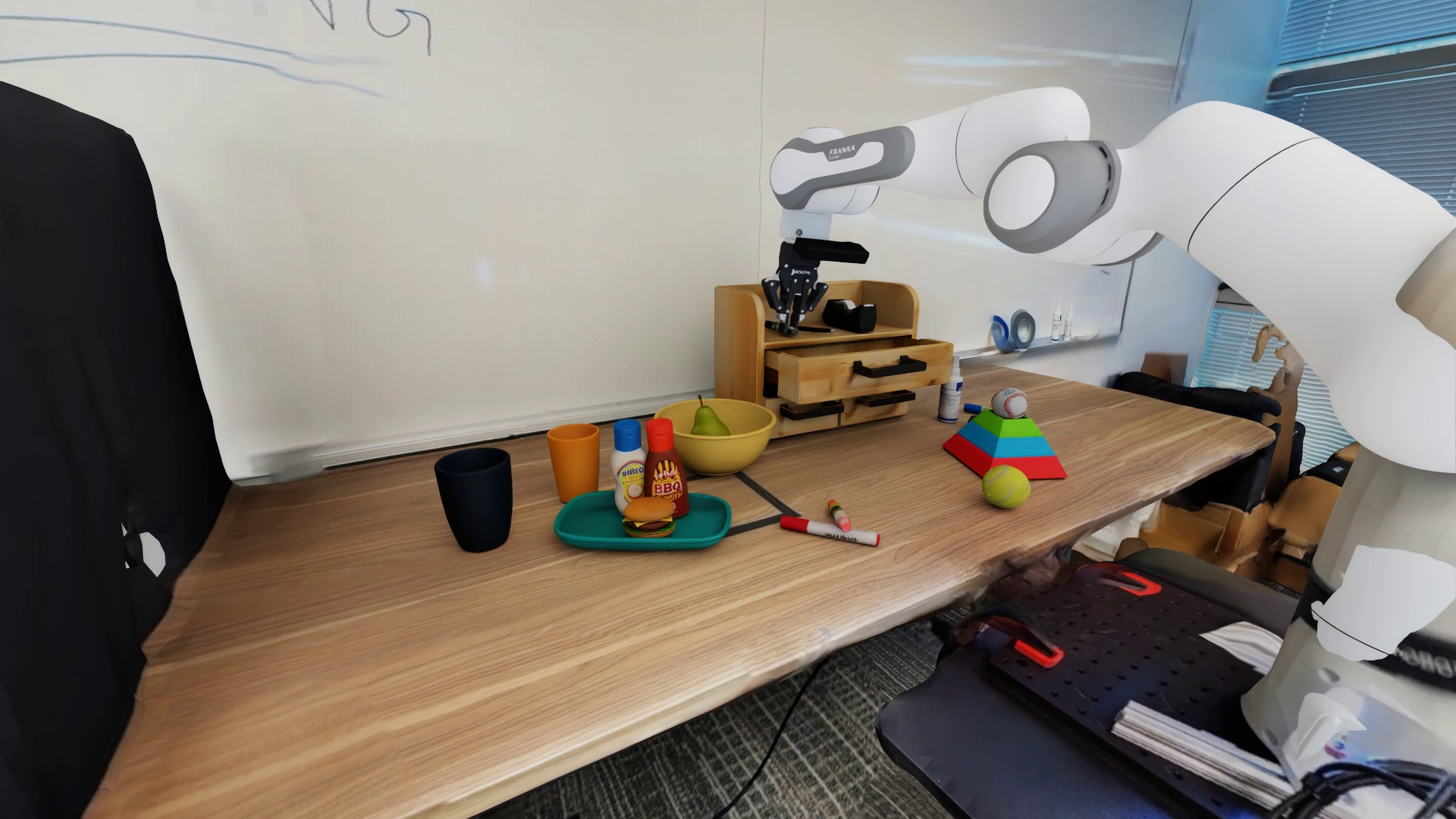}

\robotcardwithphototwo{Place Black Eraser in Bowl}
    {Pick up the black eraser, then put the black eraser in the bowl.}
    {Black Eraser OnTop of Bowl}
    {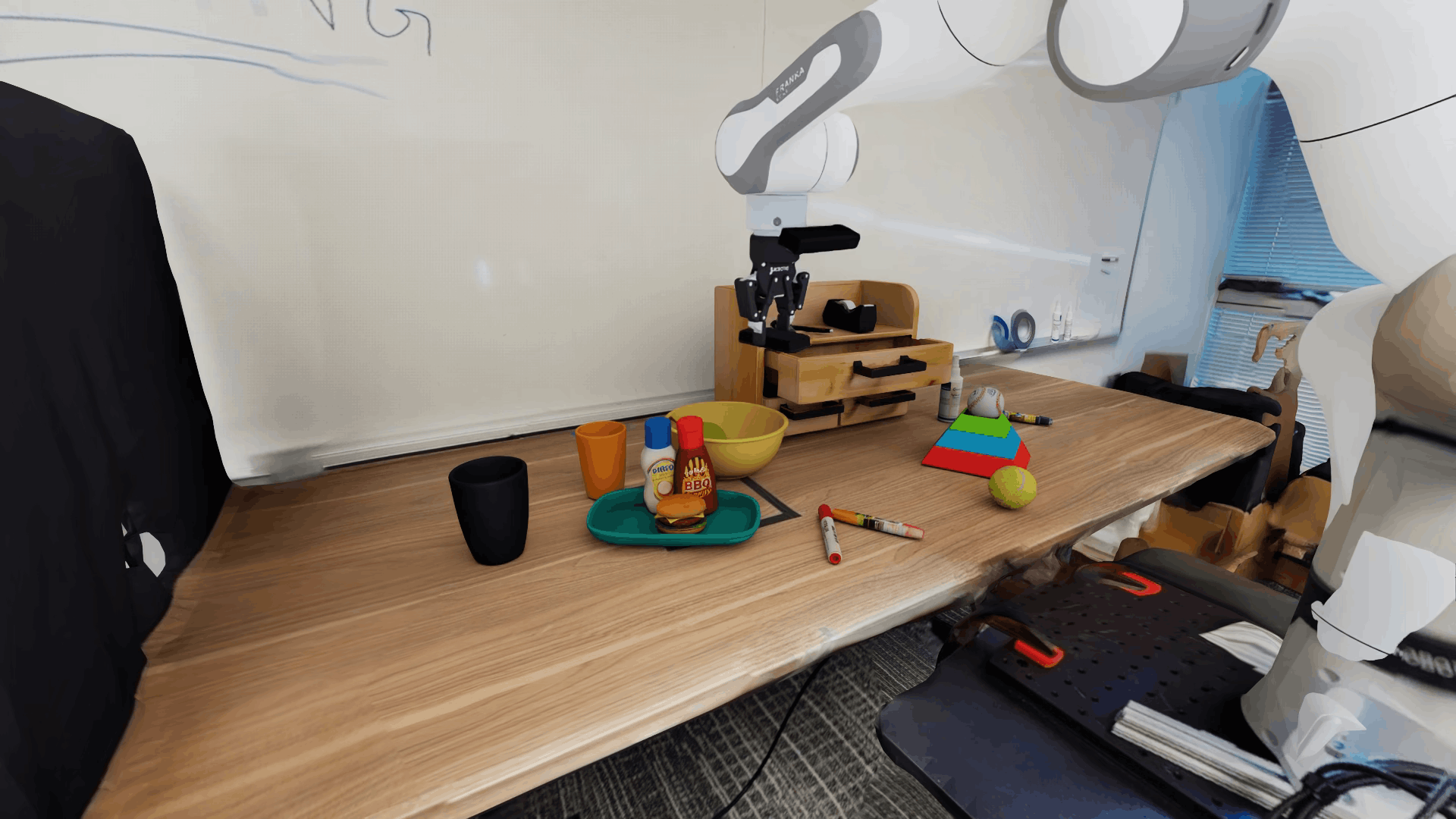}

\robotcardwithphototwo{Place Orange Cup in Bowl}
    {Pick up the orange cup, then put the orange cup in the bowl.}
    {Orange Cup OnTop of Bowl}
    {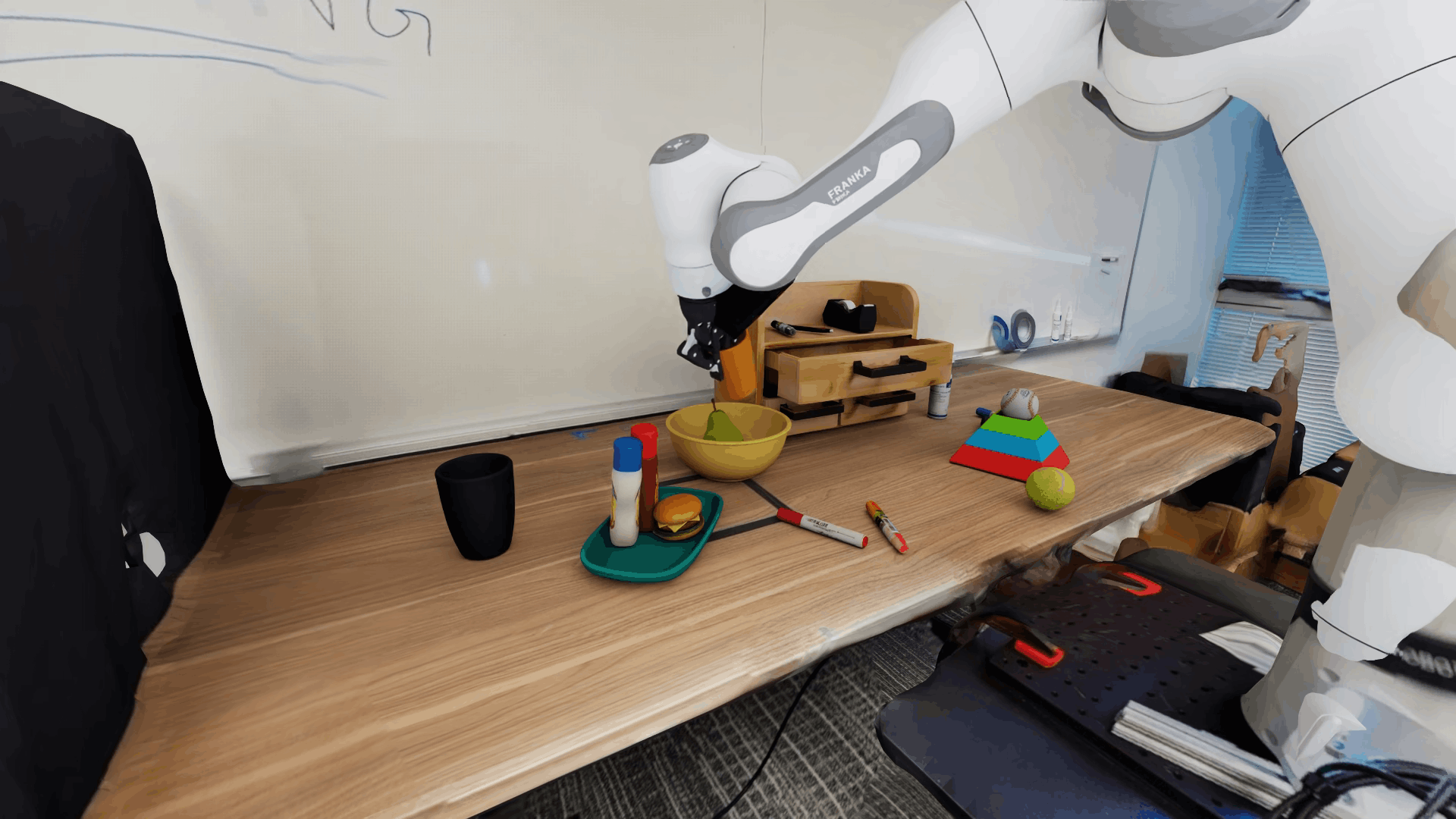}

\robotcardwithphototwo{Place Orange Cup on Organizer}
    {Pick up the orange cup, then put the orange cup on the organizer.}
    {Orange Cup OnTop of Organizer}
    {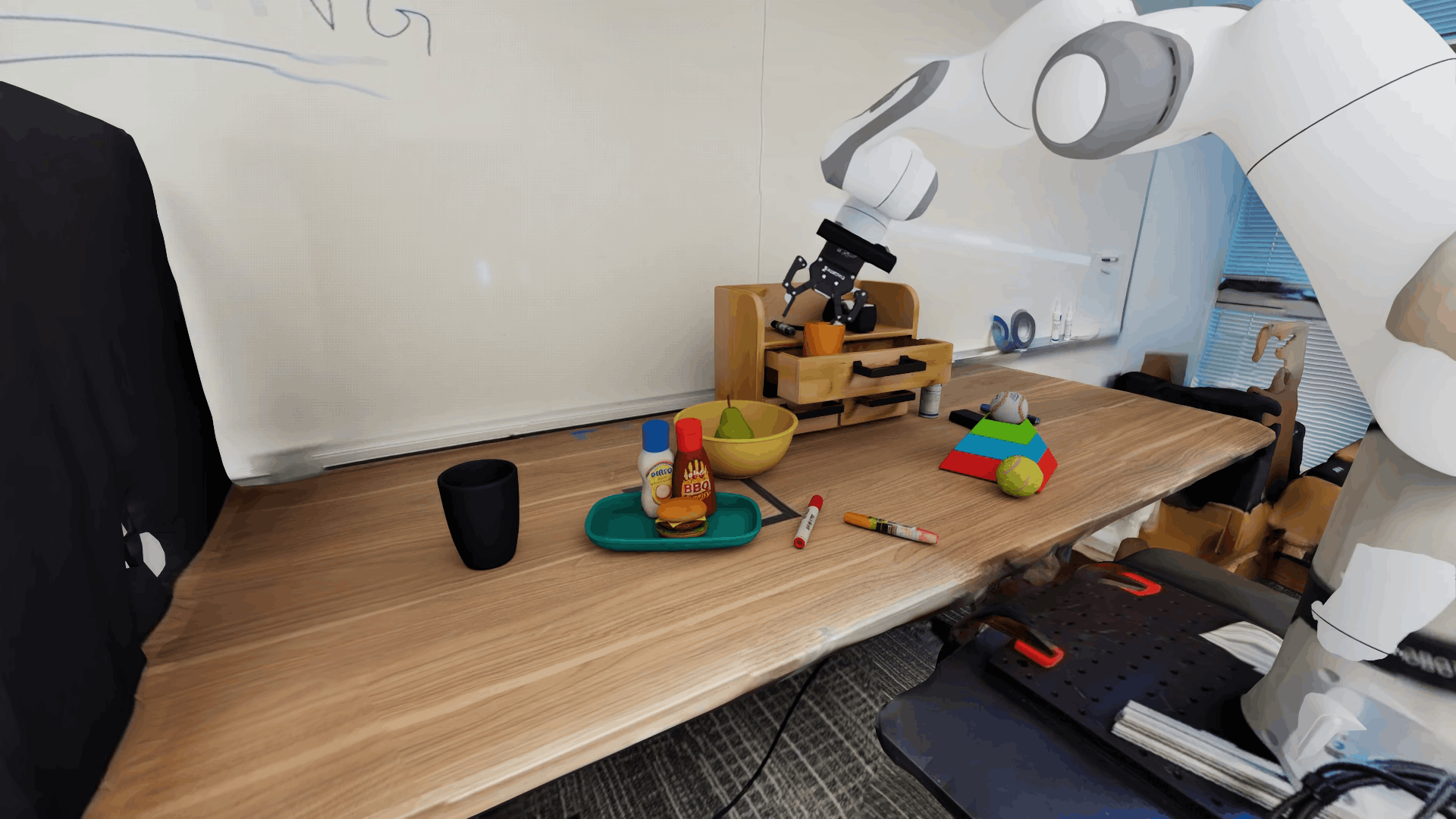}

\robotcardwithphototwo{Place Red Marker on Organizer}
    {Pick up the red marker, then put the red marker on the organizer.}
    {Red Marker OnTop of Organizer}
    {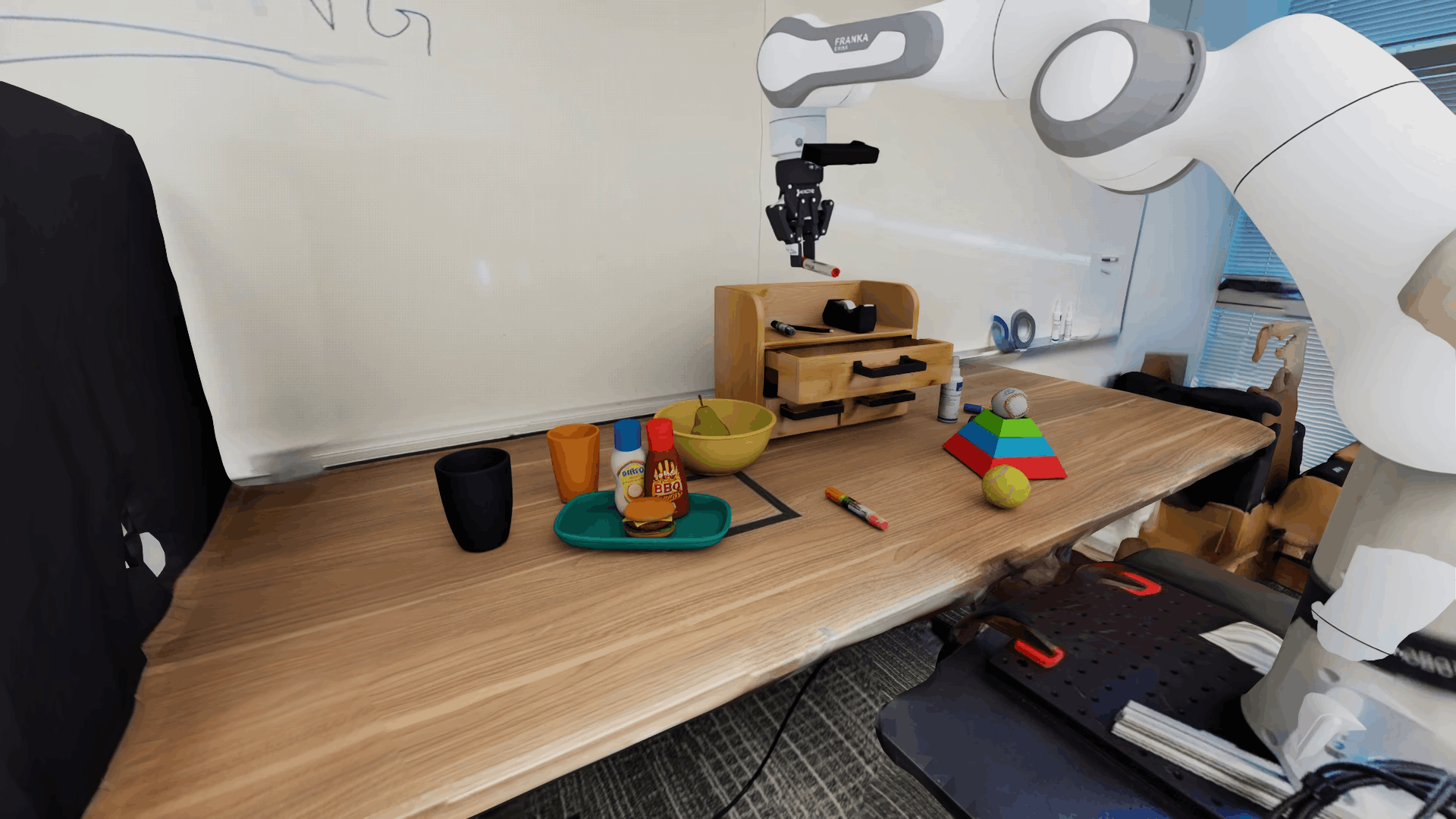}

\clearpage
\flushbottom%
\section{Detailed Experiment Results}
\label{app:results}

In this section, we present the detailed numbers for experiments in Section~\ref{sec:experiments}.

\subsection{Detailed Results for Real-to-Sim Policy Evaluation}

We present the detailed numbers for policy evaluation in \sysName{} in \autoref{tab:real2sim-simfoundry} and the results in PolaRiS~\cite{jain2025polaris} in \autoref{tab:real2sim-polaris}. In general, we find that the success rates in \sysName{} align much more closely with those in the real world, with most policies performing poorly in PolaRiS, particularly for tasks with real-world finetuning.
\begin{table*}[ht]
  \centering
  \caption{\textbf{Real-world vs.\ simulation success rates (\%) for policies evaluated in \sysName{}.} The rightmost columns report per-task Pearson $r$ (higher is better) and MMRV (lower is better) computed between the real-world and simulation success rates across all policies evaluated on that task. Cells marked ``--'' indicate the policy was not evaluated on that task. }
  \label{tab:real2sim-simfoundry}
  \scriptsize
  \begin{tabular}{lcccccccccccc}
    \toprule
    & \multicolumn{2}{c}{$\pi_0$} & \multicolumn{2}{c}{$\pi_{0.5}$} & \multicolumn{2}{c}{\grootsix} & \multicolumn{2}{c}{\grootseven} & \multicolumn{2}{c}{DreamZero} & \multicolumn{2}{c}{Real$\leftrightarrow$Sim agreement} \\
    \cmidrule(lr){2-3} \cmidrule(lr){4-5} \cmidrule(lr){6-7} \cmidrule(lr){8-9} \cmidrule(lr){10-11} \cmidrule(lr){12-13}
    Task & Real & Sim & Real & Sim & Real & Sim & Real & Sim & Real & Sim & Pearson $r\uparrow$ & MMRV $\downarrow$ \\
    \midrule
    Stack Dishware & 100 & 34 & 100 & 64 & 40 & 0 & -- & -- & -- & -- & 0.883 & 0.000 \\
    Store Marker & 48 & 4 & 60 & 20 & 32 & 0 & -- & -- & -- & -- & 0.915 & 0.000 \\
    Throw Away Trash & 20 & 0 & 48 & 4 & 0 & 0 & -- & -- & -- & -- & 0.910 & 0.067 \\
    Serve Fruits & 0 & 4 & 72 & 80 & 4 & 20 & 40 & 32 & 8 & 12 & 0.960 & 0.016 \\
    Cup in Bowl & 88 & 56 & 100 & 92 & 68 & 40 & 92 & 92 & 100 & 92 & 0.907 & 0.016 \\
    Marker in Cup & 40 & 40 & 92 & 88 & 28 & 28 & 88 & 88 & 88 & 80 & 0.995 & 0.008 \\
    Clear Table & 0 & 12 & 40 & 36 & 0 & 0 & 8 & 28 & 16 & 28 & 0.810 & 0.016 \\
    \bottomrule
  \end{tabular}
\end{table*}

\begin{table*}[h]
  \centering
  \caption{\textbf{Real-world vs.\ simulation success rates (\%) in PolaRiS.}}
  \label{tab:real2sim-polaris}
  \scriptsize
  \begin{tabular}{lcccccccccccc}
    \toprule
    & \multicolumn{2}{c}{$\pi_0$} & \multicolumn{2}{c}{$\pi_{0.5}$} & \multicolumn{2}{c}{\grootsix} & \multicolumn{2}{c}{\grootseven} & \multicolumn{2}{c}{DreamZero} & \multicolumn{2}{c}{Real$\leftrightarrow$Sim agreement} \\
    \cmidrule(lr){2-3} \cmidrule(lr){4-5} \cmidrule(lr){6-7} \cmidrule(lr){8-9} \cmidrule(lr){10-11} \cmidrule(lr){12-13}
    Task & Real & Sim & Real & Sim & Real & Sim & Real & Sim & Real & Sim & Pearson $r\uparrow$ & MMRV $\downarrow$ \\
    \midrule
    Stack Dishware & 100 & 0 & 100 & 8 & 40 & 0 & -- & -- & -- & -- & 0.500 & 0.200 \\
    Store Marker & 48 & 0 & 60 & 4 & 32 & 0 & -- & -- & -- & -- & 0.822 & 0.053 \\
    Throw Away Trash & 20 & 0 & 48 & 0 & 0 & 0 & -- & -- & -- & -- & -- & 0.253 \\
    Serve Fruits & 0 & 4 & 72 & 28 & 4 & 24 & 40 & 4 & 8 & 4 & 0.480 & 0.288 \\
    Cup in Bowl & 88 & 20 & 100 & 36 & 68 & 76 & 92 & 48 & 100 & 68 & -0.396 & 0.280 \\
    Marker in Cup & 40 & 0 & 92 & 4 & 28 & 4 & 88 & 12 & 88 & 4 & 0.512 & 0.176 \\
    Clear Table & 0 & 0 & 40 & 0 & 0 & 4 & 8 & 0 & 16 & 12 & -0.037 & 0.352 \\
    \bottomrule
  \end{tabular}
\end{table*}

\begin{figure}[t]
    \centering
    \includegraphics[width=0.8\textwidth]{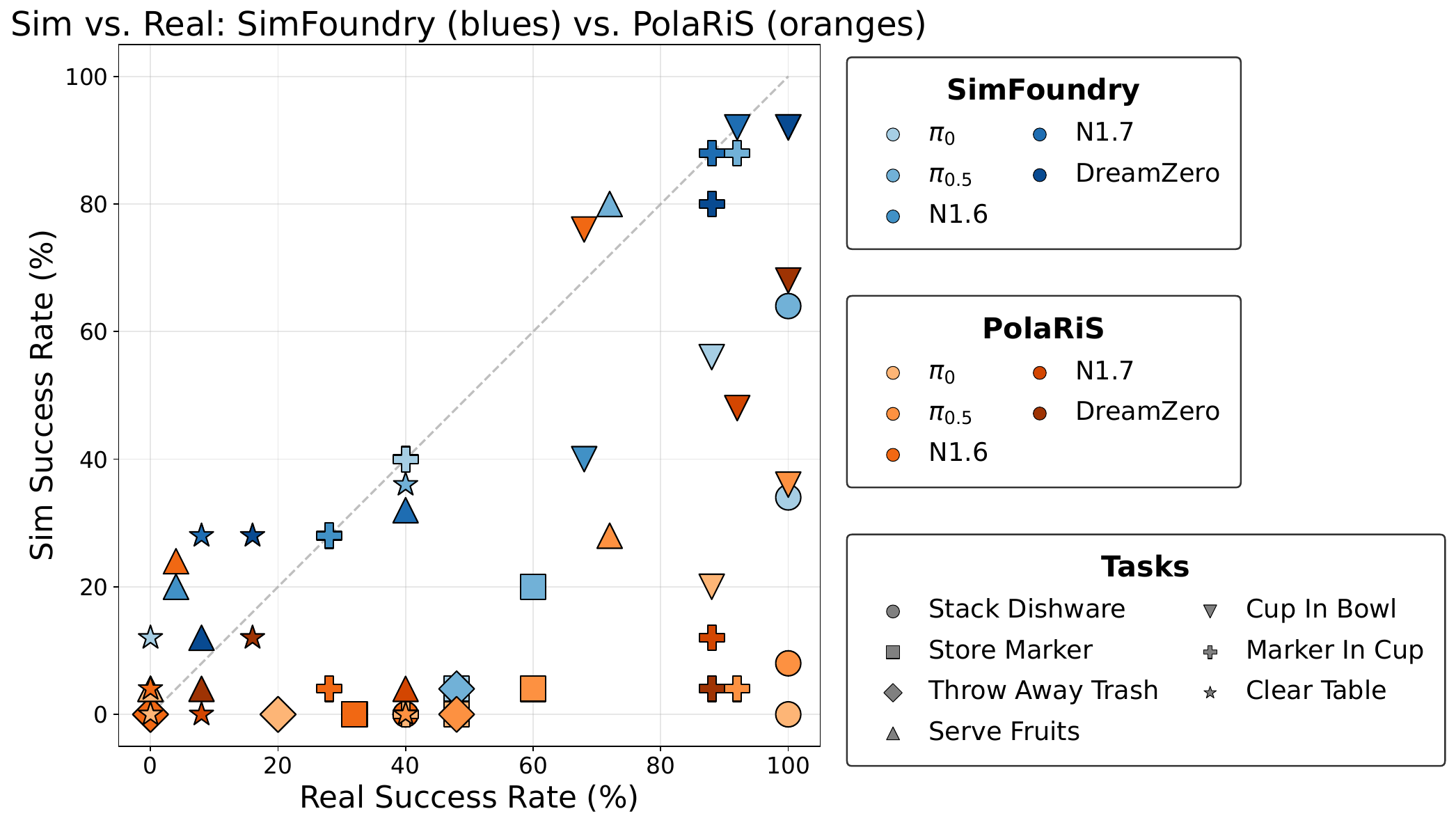}
    \caption{\textbf{Enlarged Real-to-Sim policy evaluation correlations with task
  labels.}  This figure expands the right panel of \autoref{fig:real2im_correlation}. Each point compares real-world and simulated
  success rates for a policy-task pair. The dashed diagonal indicates perfect agreement between simulated and real-world success rates. Blue points correspond to \sysName{} evaluations and orange points
  to PolaRiS evaluations. \sysName{} points lie closer to the diagonal across tasks, matching the higher Pearson correlations and lower MMRV values reported in \autoref{tab:real2sim-simfoundry} and \autoref{tab:real2sim-polaris}.}
    \label{fig:real2sim_fullsize}
\end{figure}

\subsubsection{Sub-Task Evaluations improve Real-to-Sim Correlations}
\label{sec:subtask_eval}

\paragraph{Sub-task Evaluation Protocol.}
We introduce a sub-task evaluation protocol that takes advantage of the ability to reset to arbitrary states in simulation. Starting from states where some of the initial sub-tasks are already completed, we evaluate policies on the remaining sub-tasks, allowing a more thorough policy assessment on sub-tasks that occur later in long-horizon tasks. For example, for \markerTask, we start with the cabinet drawer already open and evaluate whether the policy can complete the remaining sub-tasks. The policy success rate with initial subtasks completed in sim is compared with the full end-to-end task success rate in the real world which we found improved evaluation correlations between sim and real.

\paragraph{Sub-task evals improve correlations and provide insights into failure modes.} Evaluating from states with completed sub-tasks can improve correlations for long-horizon fine-tuned tasks and expose failure modes, providing actionable insights for policy improvement. As seen in \autoref{tab:real2sim-simfoundry-sub-tasks}, the mean pearson correlation improves from 0.902 to 0.951 with sub-task evaluations on the fine-tuned tasks. For example, in the \markerTask task, once the drawer is opened, \openpi \hspace{0.2mm} can almost always complete the rest of the task, in both sim and real.

\begin{table*}[ht]
  \centering
  \caption{\textbf{Real-world vs.\ simulation success rates (\%) correlations improve when evaluating on sub-tasks.}}
  \label{tab:real2sim-simfoundry-sub-tasks}
  \begin{tabular}{lcccccccc}
    \toprule
    & \multicolumn{2}{c}{$\pi_0$} & \multicolumn{2}{c}{$\pi_{0.5}$} & \multicolumn{2}{c}{\grootsix} & \multicolumn{2}{c}{Real$\leftrightarrow$Sim agreement} \\
    \cmidrule(lr){2-3} \cmidrule(lr){4-5} \cmidrule(lr){6-7} \cmidrule(lr){8-9}
    Task & Real & Sim & Real & Sim & Real & Sim & Pearson $r\uparrow$ & MMRV $\downarrow$ \\
    \midrule
    Stack Dishware & 100 & 64 & 100 & 80 & 40 & 24 & 0.961 & 0.000 \\
    Store Marker & 48 & 52 & 60 & 76 & 32 & 36 & 0.981 & 0.000 \\
    Throw Away Trash & 20 & 0 & 48 & 8 & 0 & 0 & 0.910 & 0.067 \\
    \bottomrule
  \end{tabular}
\end{table*}

\subsection{Detailed Results for Sim-to-Real Experiments}
The per-task success rates for the object cousin experiments are presented in \autoref{tab:sim2real_yam_droid}.
The models are trained with either -  \textbf{Twin}, i.e., only on the digital twin object, or \textbf{+ 9 cousins}, i.e., data generated with the twin plus 9 cousin objects.
The different eval settings are:
\begin{enumerate}
    \item Sim Twin: evaluation on the reconstructed twin object(s).
    \item Sim Cousins: evaluation on a held-out digital cousin(s) of the reconstructed twin. This is an object for which data is not collected in either sim or the real world.
    \item Real Twin: evaluation on the real-world object(s) that were reconstructed in simulation.
    \item Real Cousin: evaluation on held-out real-world object(s).
\end{enumerate}

\begin{figure*}
    \centering
    \includegraphics[width=0.97\textwidth]{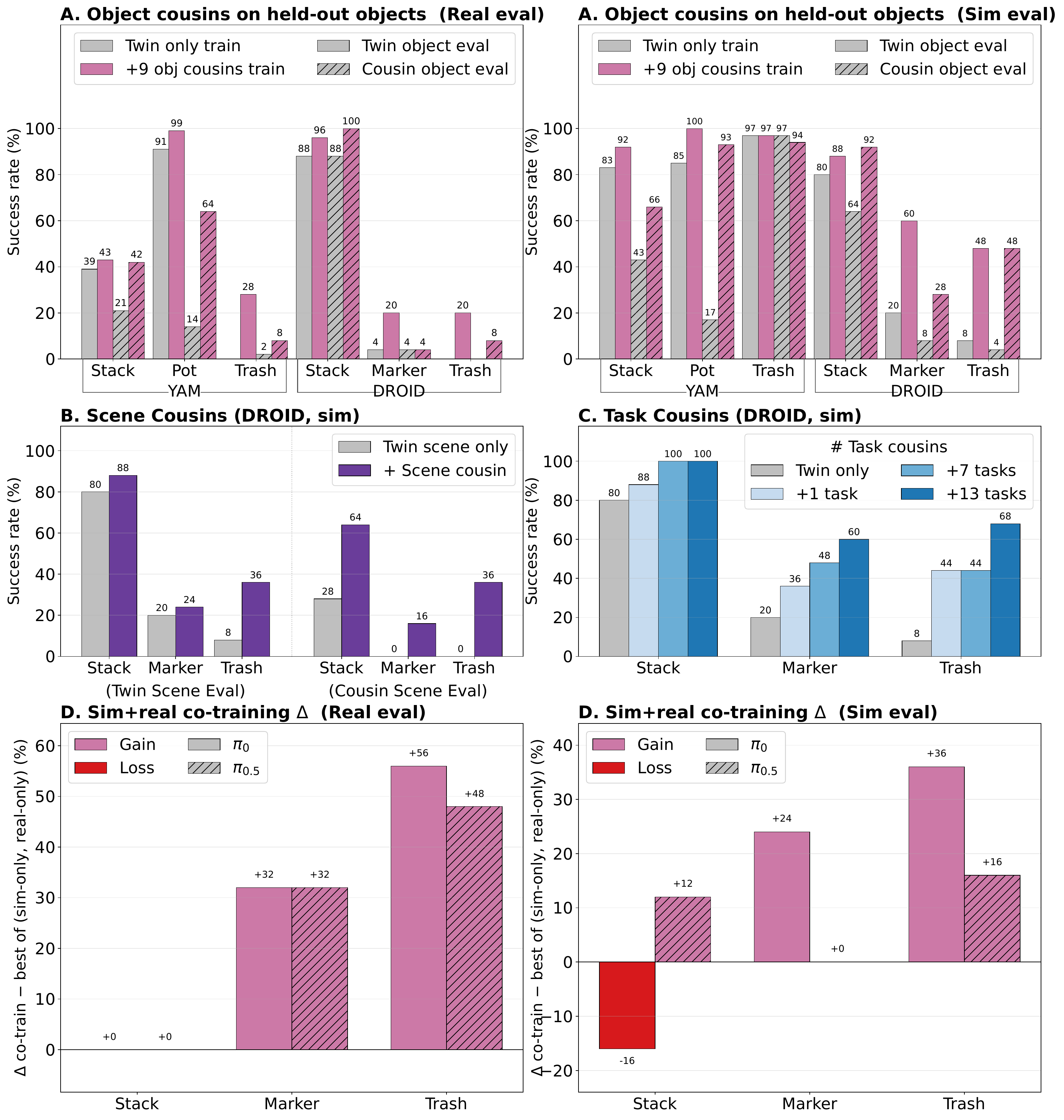}
    \caption{\textbf{\sysName{} data diversity along different axes scales data generation and policy performance.}
    We ablate how different sources of \sysName{}-generated data improve policy learning and generalization.
    \textbf{(A)} Object cousins improve robustness across DROID and YAM by training policies on affordance-preserving object variants, yielding an average zero-shot sim-to-real success improvement of $17\%$ and up to a $50\%$ real-world gain on held-out \potTask{} objects.
    \textbf{(B)} Scene cousins improve layout generalization by training on semantically modified object arrangements, producing an average improvement of $21\%$, including a $28\%$ gain on \trashTask{} in the twin scene and $16\%$ success on held-out \markerTask{} cousin layouts where the twin-only policy achieves $0\%$.
    \textbf{(C)} Task cousins improve downstream task learning by adding demonstrations from related tasks while keeping the total number of demonstrations fixed; 13 task cousins improve \trashTask{} by $60\%$ and \markerTask{} by $40\%$ in simulation.
    \textbf{(D)} Sim-and-real co-training further improves performance by combining scalable \sysName{} demonstrations with limited real data, increasing \openpi{} real-world \markerTask{} success from $60\%$ to $92\%$ and improving \openpizero{} simulated \trashTask{} success by $36\%$.
    Together, these results show that \sysName{} provides complementary forms of structured data diversity across objects, layouts, tasks, and real/sim data mixtures.}
    \label{fig:all_cousins_detailed}
\end{figure*}

\paragraph{Ablation of Object, Scene and Task Cousins.}
Figure~\ref{fig:all_cousins_detailed} summarizes how different forms of \sysName{}-generated data diversity affect policy performance. Across object, scene, and task cousins, we find that structured variation consistently improves policy robustness beyond training on the reconstructed twin alone. Object cousins improve instance-level generalization by exposing policies to affordance-preserving changes in object geometry, appearance, and topology, yielding an average task success improvement of $17\%$ during zero-shot sim-to-real transfer.  Scene cousins target a complementary form of generalization by varying semantic object relations and introducing layout diversity; these improve task success by an average of ${\sim}13\%$ on the twin scene and ${\sim}29\%$ on cousin scenes. Task cousins provide behavioral and action diversity by adding related demonstrations that share objects, predicates, or intermediate behaviors with the target task; this yields the largest average improvement of $40\%$.

In addition to these cousin-based augmentations, Figure~\ref{fig:all_cousins_detailed} also shows that \sysName{} data can effectively complement limited real-world demonstrations through co-training. While zero-shot policies trained only on \sysName{} data already transfer to the real world, adding real data further improves performance in most DROID settings.

\begin{table*}[h]
\centering
\setlength{\tabcolsep}{4pt}
\scriptsize

\begin{minipage}{0.48\textwidth}
\centering
\textbf{YAM}

\begin{tabular}{llcc}
\toprule
& & \makecell{Twin} & \makecell{$+$\\9 Cousins} \\
\midrule
\multirow{4}{*}{\makecell{Stack\\Dishware}}
  & Sim Twin     & 83 & 92 \\
  & Sim Cousins  & 43 & 66 \\
  & Real Twin    & 39 & 43 \\
  & Real Cousins & 21 & 42 \\
\midrule
\multirow{4}{*}{\makecell{Pot On\\Stove}}
  & Sim Twin     & 85 & 100 \\
  & Sim Cousins  & 17 & 93 \\
  & Real Twin    & 91 & 99 \\
  & Real Cousins & 14   & 64   \\
\midrule
\multirow{4}{*}{\makecell{Throw Away\\Trash}}
  & Sim Twin     & 97 & 97 \\
  & Sim Cousins  & 97 & 94 \\
  & Real Twin    & 0 & 28 \\
  & Real Cousins & 2 & 8 \\
\bottomrule
\end{tabular}
\end{minipage}
\hfill
\begin{minipage}{0.48\textwidth}
\centering
\textbf{DROID}

\begin{tabular}{llcc}
\toprule
& & \makecell{Twin} & \makecell{$+$\\9 Cousins} \\
\midrule
\multirow{4}{*}{\makecell{Stack \\ Dishware}}
  & Sim Twin     & 80 & 88 \\
  & Sim Cousins  & 64 & 92 \\
  & Real Twin    & 88 & 96 \\
  & Real Cousins & 88 & 100 \\
\midrule
\multirow{4}{*}{\makecell{Store \\ Marker}}
  & Sim Twin     & 20 & 60 \\
  & Sim Cousins  & 8 & 28 \\
  & Real Twin    & 4 & 20 \\
  & Real Cousins & 4 & 4 \\
\midrule
\multirow{4}{*}{\makecell{Throw Away \\ Trash}}
  & Sim Twin     & 8 & 48 \\
  & Sim Cousins  & 4 & 48 \\
  & Real Twin    & 0 & 20 \\
  & Real Cousins & 0 & 8 \\
\bottomrule
\end{tabular}
\end{minipage}
\caption{\textbf{Policy Robustness Using Object Cousins.} Across multiple robot embodiments and multiple tasks, leveraging additional object cousins~\cite{dai2024automated} improves direct sim2real policy transfer on the original target scene objects and additional held-out unseen objects.}
\label{tab:sim2real_yam_droid}

\end{table*}

\begin{table}[t]
\centering
\caption{\textbf{Boosting Performance with Scene Cousins (DROID, simulation)} Success rates shown below. }
\label{tab:scene_cousins}
\setlength{\tabcolsep}{6pt}
\scriptsize
\begin{tabular}{lcc}
\toprule
 & \textbf{twin only} & \textbf{+ scene cousin} \\
\midrule
\dishwareTask  & 80 & \textbf{88} \\
\dishwareTask - cousin  & 28 & \textbf{64} \\
\markerTask & 20 & \textbf{24} \\
\markerTask - cousin & 0 & \textbf{16} \\
\trashTask  & 8  & \textbf{36}  \\
\trashTask - cousin  & 0  & \textbf{36}  \\

\bottomrule
\end{tabular}
\end{table}

\begin{table}[t]
\centering
\caption{\textbf{Boosting Performance with Task Cousins.}  Adding additional tasks and cousins with the same or similar objects increases performance for the downstream task.}
\label{tab:task_cousins}
\setlength{\tabcolsep}{6pt}
\scriptsize
\begin{tabular}{lcccc}
\toprule
 & \textbf{twin only} & \textbf{+1 task} & \textbf{+7 tasks} & \textbf{+13 tasks} \\
\midrule
\dishwareTask  & 80 & 88 & \textbf{100} & \textbf{100} \\
\markerTask & 20 & 36 & 48  & \textbf{60} \\
\trashTask  & 8  & 44 & 44  & \textbf{68} \\

\bottomrule
\end{tabular}
\end{table}

\begin{table}[t]
\centering
\caption{\textbf{Co-training with \sysName{} generated data.} We compare success rates of models trained with simulation-only data (-\textbf{S}) to those trained with real-world demos (-\textbf{R}) as well as combinations of both simulation and real data (-\textbf{co-train}). Each model type is evaluated in both the real-world scene (-Real) and the \sysName{} reconstruction (-Sim). }
\label{tab:sim_and_real_cotraining}
\setlength{\tabcolsep}{6pt}
\scriptsize
\begin{tabular}{lcccccc}
\toprule
 & \textbf{\openpizero-S} & \textbf{\openpizero-R} & \textbf{\openpizero-co-train} & \textbf{\openpi-S} & \textbf{\openpi-R} & \textbf{\openpi-co-train} \\
\midrule
\dishwareTask - Sim   & 92 & 34 & 76  & 88 & 64 & 100 \\
\dishwareTask - Real  & 96 & 100 & 100 & 96 & 100  & 100 \\
\markerTask - Sim  & 16 & 4 & 40  & 60 & 20 & 60  \\
\markerTask - Real & 4 & 48 & 80  & 20 & 60 & 92  \\
\trashTask - Sim   & 0 & 0 & 36  & 48 & 4 & 60  \\
\trashTask - Real  & 0 & 20 & 76  & 20 & 48 & 96  \\

\bottomrule
\end{tabular}
\end{table}

\subsection{Detailed Object Cousin Ablation}

\label{sec:sim2real_zeroshot_cousin_ablation}
To better understand the importance of object cousins, we run an additional ablation on our set of bimanual tasks, testing sim2real zero-shot performance over training datasets that include only the reconstructed digital twin scene objects, or additionally including 1, 3, or 9 object cousins as part of the training mix. All runs used a fixed 1000 demonstration bandwidth and were split evenly among each scene objects' instance. Results are aggregated over 25 eval trials. To further guarantee reproducibility, we randomly sample each pose initialization and deterministically align them in both sim and real, as done in our real-to-sim policy evaluation setup (see \autoref{app:real-to-sim-evals} for more details).

Our results are shown in Table~\ref{tab:sim2real_yam_cousin_ablation}. We find in general that increasing the number of object cousins tends to improve zero-shot sim2real policy transfer, both on the original twin scene objects and held-out unseen scene objects, highlighting the potential for increasing object cousins to reliably improve policy robustness.
\begin{table}[H]
\centering
\caption{\textbf{Sim-to-real policy training results across object cousins.}}
\label{tab:sim2real_yam_cousin_ablation}
\setlength{\tabcolsep}{4pt}
\scriptsize
\textbf{YAM}\par
\vspace{2pt}
\begin{tabular}{llcccc}
\toprule
& & \makecell{Twin} & \makecell{$+$\\1 Cousin} & \makecell{$+$\\3 Cousins} & \makecell{$+$\\9 Cousins} \\
\midrule
\multirow{4}{*}{\makecell{Stack\\Dishware}}
  & Sim Twin     & 83 & 89 & 100 & 92 \\
  & Sim Cousins  & 43 & 44 & 65 & 66 \\
  & Real Twin    & 39 & 41 & 37 & 43 \\
  & Real Cousins & 21 & 32 & 27 & 42 \\
\midrule
\multirow{4}{*}{\makecell{Pot On\\Stove}}
  & Sim Twin     & 85 & 100 & 93 & 100 \\
  & Sim Cousins  & 17 & 27 & 35 & 93 \\
  & Real Twin    & 91 & 100 & 94 & 99 \\
  & Real Cousins & 14 & 38 & 16 & 64 \\
\midrule
\multirow{4}{*}{\makecell{Throw Away\\Trash}}
  & Sim Twin     & 97 & 98 & 98 & 97 \\
  & Sim Cousins  & 97 & 100 & 97 & 94 \\
  & Real Twin    & 0 & 9 & 45 & 28 \\
  & Real Cousins & 2 & 17 & 14 & 8 \\
\bottomrule
\end{tabular}
\end{table}

\clearpage
\section{Policy Details}
\label{app:policy}

\subsection{Data Generation Details}
\label{sec:datagen}
\sysName leverages a combination of human-collected demonstrations and automated data augmentation to generate synthetic datasets useful for training robot learning policies.

For a given task, we first collect a small number ($\sim 10 - 15$) of demonstrations via human operator-controlled JoyLo~\cite{jiang2025behavior} systems. Then, we augment those demonstrations using MimicGen~\cite{mandlekar2023mimicgen}, both increasing the trajectory diversity (via demonstration count) as well as visual diversity by applying domain randomization: material randomization, camera pose randomization, and (specifically in the DROID setup) table height randomization. The resulting datasets are used to train robot learning policies that can be deployed zero-shot in the real world.

\subsection{Policy Training Details}
\label{sec:policy_training}

For the DROID sim-to-real experiments, we finetune the DROID-pretrained joint-position versions of \openpizero \hspace{0.2mm} and \openpi. Each policy is trained with a batch size of 256, a learning rate of $1e-5$, and for 10k gradient steps. In simulation, the policies are evaluated every 1k steps, and the best-performing checkpoint is evaluated in the real world. We then report the best-performing checkpoint for each model.

For the YAM-based tasks, we trained flow-matching policies. Observations included joint state proprioception, top-down fixed camera RGB images, and per-wrist camera RGB images. The policy action space consisted of N-DOF joint position commands and a normalized 1-DOF continuous gripper open / close command. We train for 40k steps and similar to DROID, run sim evals periodically, selecting the best-performing checkpoints to evaluate zero-shot in the real-world.

\paragraph{Real-to-Sim Policy Details and Selection.} For the real-to-sim experiments, we once again finetune the DROID-pretrained joint-position checkpoints of \openpizero \hspace{0.2mm}, \openpi \hspace{0.2mm} and \grootsix~\cite{bjorck2025gr00t} for the following tasks - \dishwareTask, \markerTask, \trashTask. As previously, policies are evaluated every 1k steps and the best-performing checkpoint is evaluated in the real world. For the simpler tasks - \cupTask, \markerCupTask, \fruitsTask, and \clearTableTask, the following additional policies are also deployed zero-shot without any finetuning: \grootseven and \dreamzero~\cite{ye2026world} in addition to the pretrained checkpoints of the first three. All models output joint positions and gripper positions as actions.

\clearpage
\section{Robot Platform and Task Details}
\label{app:task}

\subsection{Robot Embodiments.}
\label{app:robot_embodiments}
We focus on two robot embodiments -- the DROID~\cite{khazatsky2024droid} platform, and a YAM workcell~\cite{i2rt2025yam}. The DROID platform consists of a single Franka Panda robot arm, left and right external ZED-2 cameras, and a wrist-mounted ZED-Mini camera, with the robot and external cameras mounted to a portable standing desk. For data collection, we use an Oculus VR headset to teleoperate the robot. In both simulation and the real world, a joint-position controller is used during policy rollout, with the gains tuned higher in simulation to minimize the tracking error.

The YAM workcell consists of a bimanual manipulator, a cage, a wrist-mounted RealSense D405 camera per arm, and an external top-down view camera. We use JoyLo~\cite{jiang2025behavior} to teleoperate the YAM arms, and a joint-position controller is used during both data collection and policy evaluation.

\subsection{Task Rubric}
\label{app:task_rubric}
In this sub-section, we provide the scoring rubric for each task, along with the language instruction provided to the VLAs. All sub-tasks need to be completed for a task to be successful. For our experiments, we mainly use binary success where the whole task is completed successfully, except for the sub-task evaluation protocol described in Appendix~\ref{sec:subtask_eval}.

\newcommand{\robotcard}[3]{%
    \begin{tcolorbox}[
        enhanced,
        colback=gray!5,
        colframe=blue!50!black,
        title=Task: #1,
        fonttitle=\bfseries,
        coltitle=white,
        attach title to upper,
        after title={\smallskip\par},
        boxrule=0.5pt,
        sharp corners=south,%
        top=2mm, bottom=2mm, left=2mm, right=2mm,
        before skip=10pt, after skip=10pt
    ]
    \small
    \textbf{Instruction:} #2 \\
    \textbf{Rubric:} #3
    \end{tcolorbox}
}

\newcommand{\robotcardwithphoto}[4]{%
    \begin{tcolorbox}[
        enhanced,
        colback=gray!5,
        colframe=blue!50!black,
        title=Task: #1,
        fonttitle=\bfseries,
        coltitle=white,
        boxrule=0.5pt,
        sharp corners=south,
        top=1mm, bottom=1mm, left=1mm, right=1mm,
        sidebyside,
        sidebyside align=top,
        lefthand ratio=0.65,%
        before skip=10pt, after skip=10pt
    ]
    \small
    \textbf{Instruction:} #2 \\
    \textbf{Rubric:} #3
    \tcblower
    \centering
    \includegraphics[width=\linewidth, height=2.5cm, keepaspectratio]{#4}
    \end{tcolorbox}
}

\robotcardwithphoto{Cup in Bowl}
    {Pick up the cup, and put the cup inside the bowl.}
    {1. Touch cup  2. Pick cup  3. Place in bowl}
    {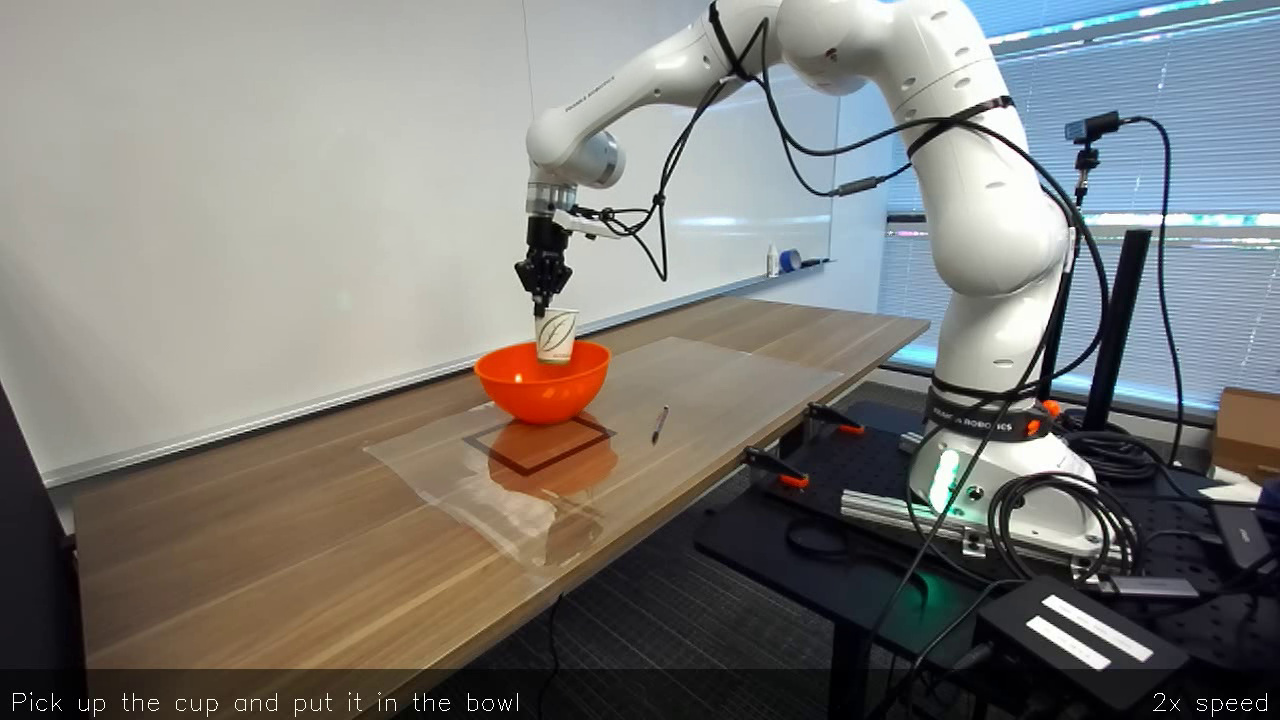}

\robotcardwithphoto{Marker in Cup}
    {Pick up the marker, and put the marker in the cup.}
    {1. Touch marker  2. Pick up marker  3. Place marker in cup}
    {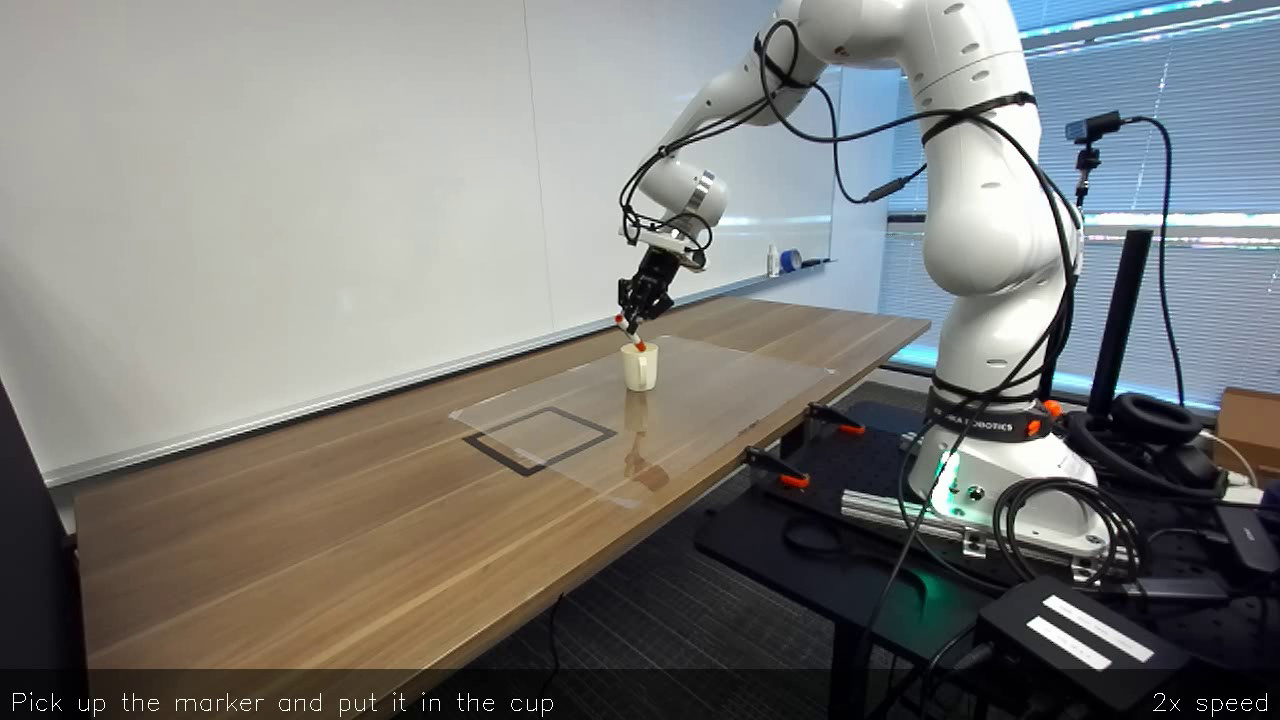}

\robotcardwithphoto{\fruitsTask}
    {Put both the banana and the apple on the green plate.}
    {1. Pick up banana  2. Pick up apple  3. Place banana on green plate  4. Place apple on green plate}
    {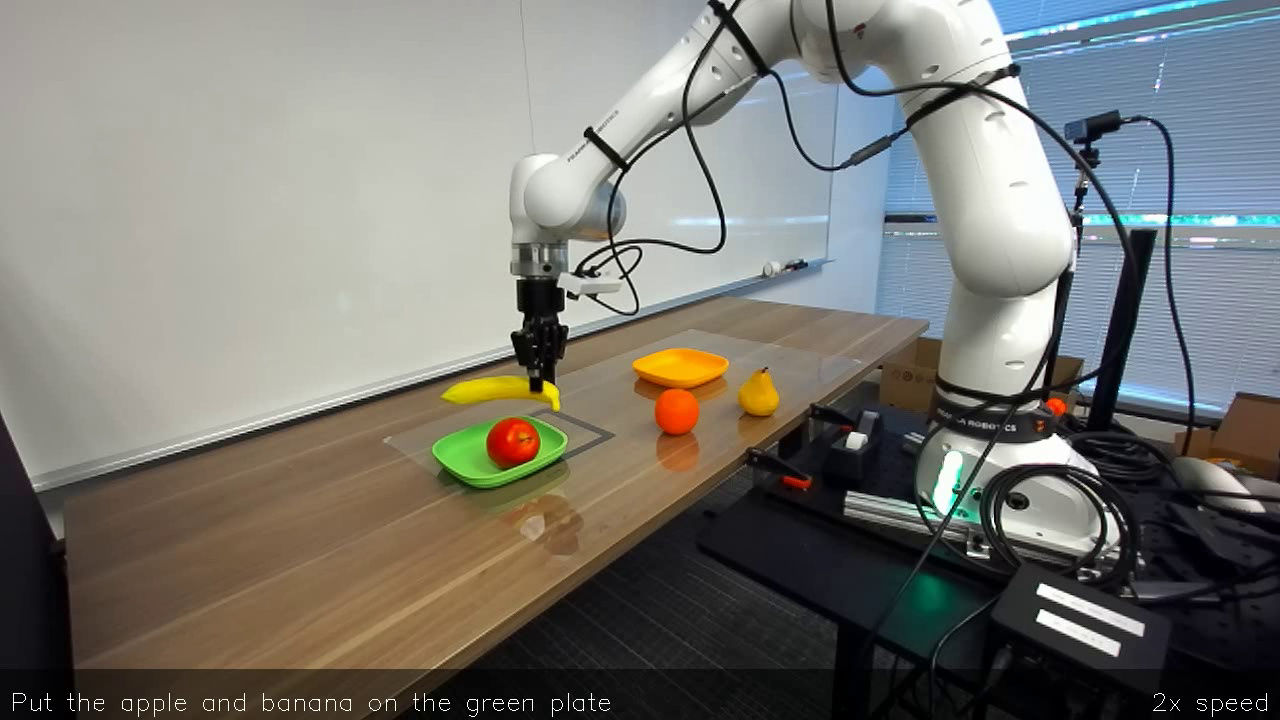}

\robotcardwithphoto{\clearTableTask}
    {Put all objects in the brown box}
    {1. Place cube in box 2. Place grapes in box 3. Place cup in box 4. Place marker in box 5. Place baseball in box }
    {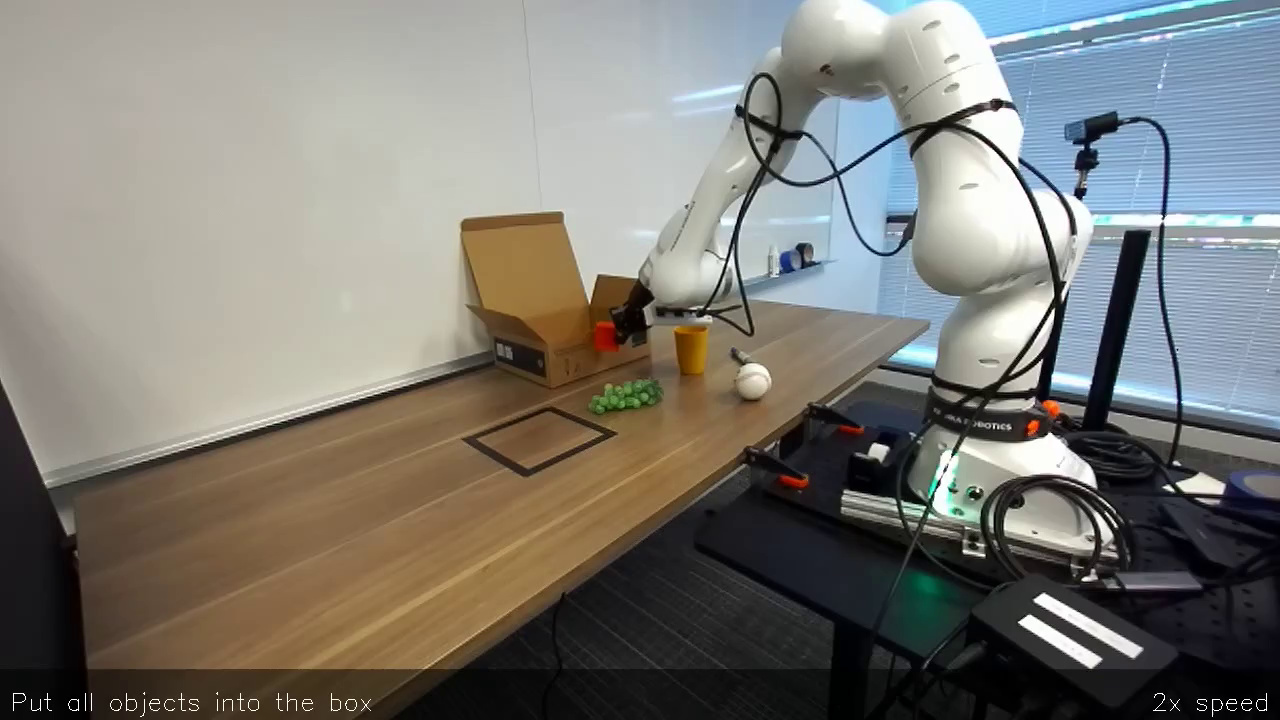}

\robotcardwithphoto{\dishwareTask}
    {Put the bowl on the plate, then put the cup in the bowl.}
    {1. Pick up bowl  2. Place bowl on plate  3. Pick up cup  4. Place cup in bowl}
    {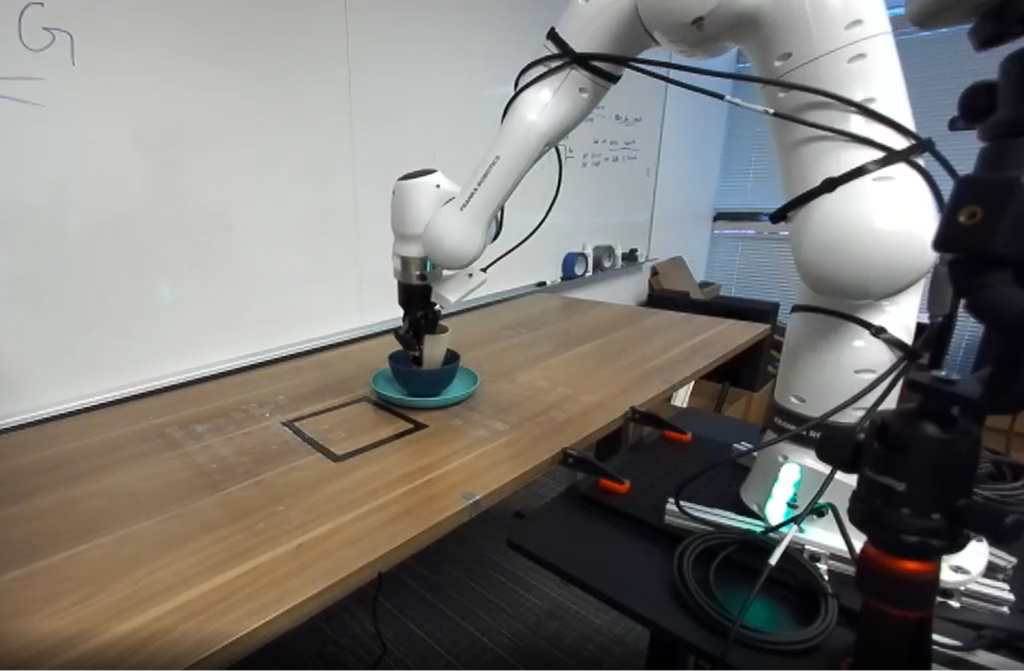}

\robotcardwithphoto{\markerTask}
    {Open drawer, pick up marker, place marker in drawer, then close drawer.}
    {1. Open cabinet drawer  2. Pick up marker  3. Place marker in drawer  4. Close drawer}
    {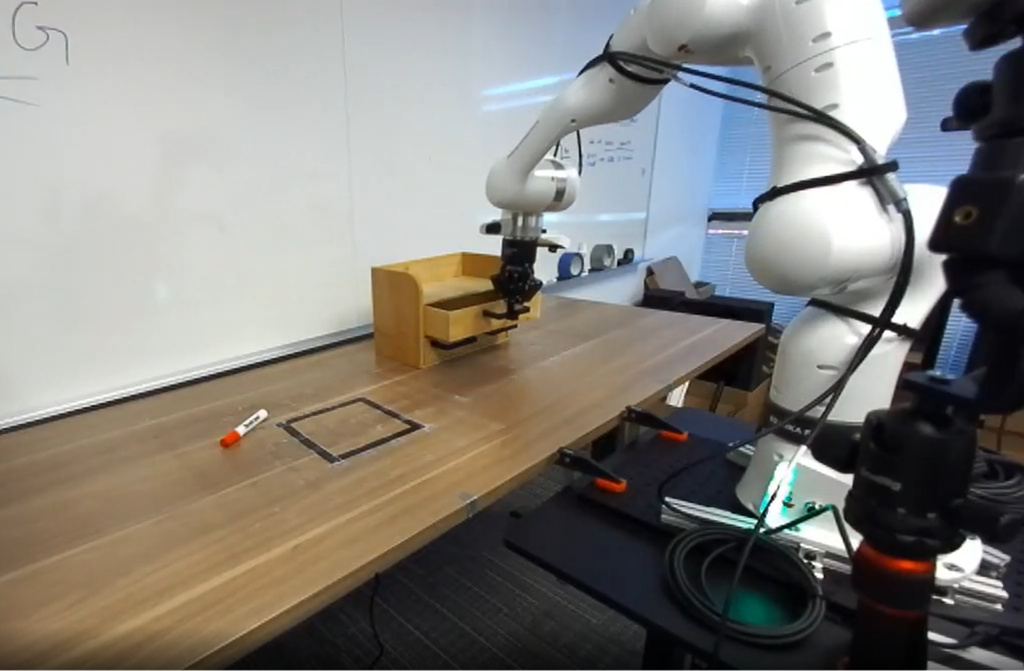}

\robotcardwithphoto{\trashTask}
    {Open the trash can, pick up the cup, put the cup into the trash can, and then close the trash can.}
    {1. Open trash can  2. Pick up cup  3. Place cup inside trash can  4. Close trash can}
    {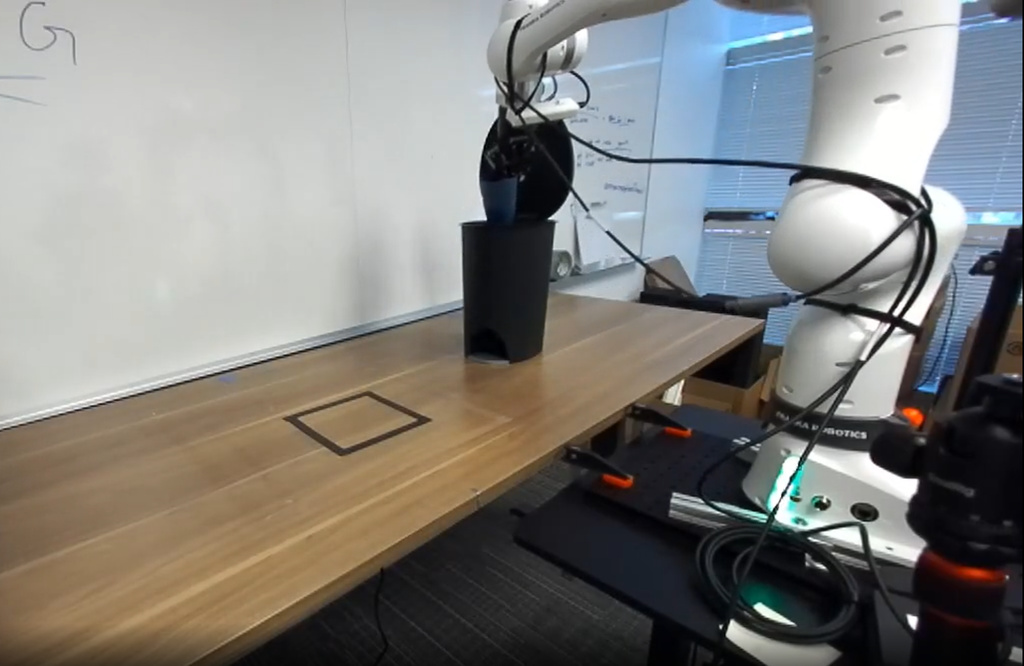}

\robotcardwithphoto{\potTask}
    {Put the pot on the stove}
    {1. Left arm grasps pot handle  2. Right arm grasps pot handle  3. Pot is lifted  4. Pot is placed on Stove}
    {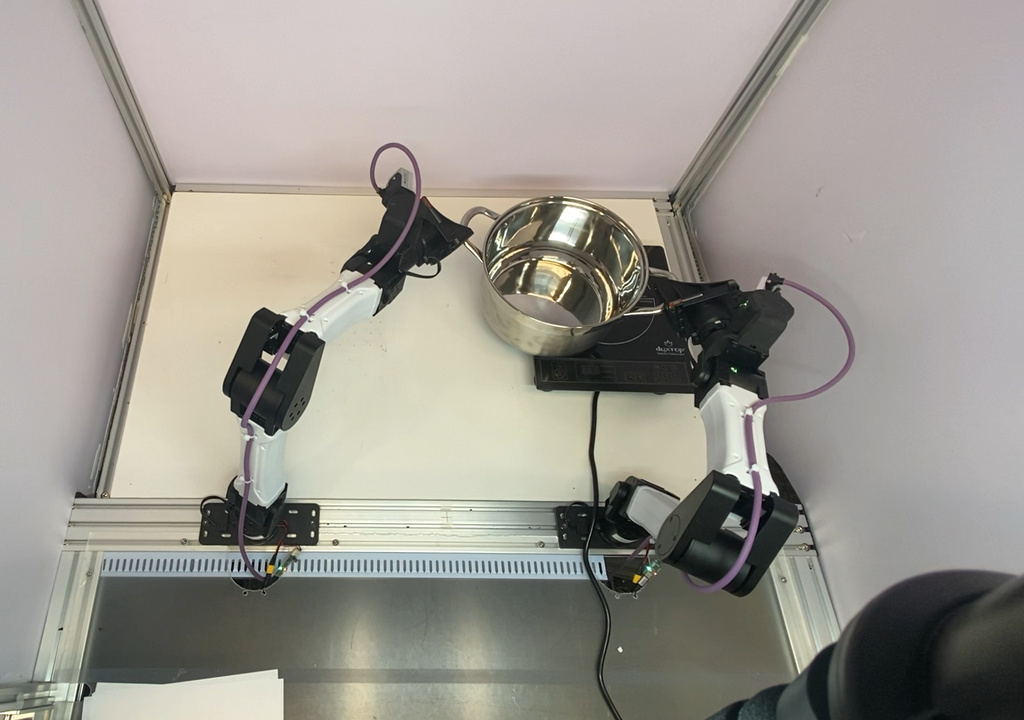}

\clearpage
\section{Real-to-Sim Policy Evaluations}
\label{app:real-to-sim-evals}

\subsection{Evaluation Protocol}
\begin{figure}[H]%
    \centering
    \includegraphics[width=0.95\textwidth]{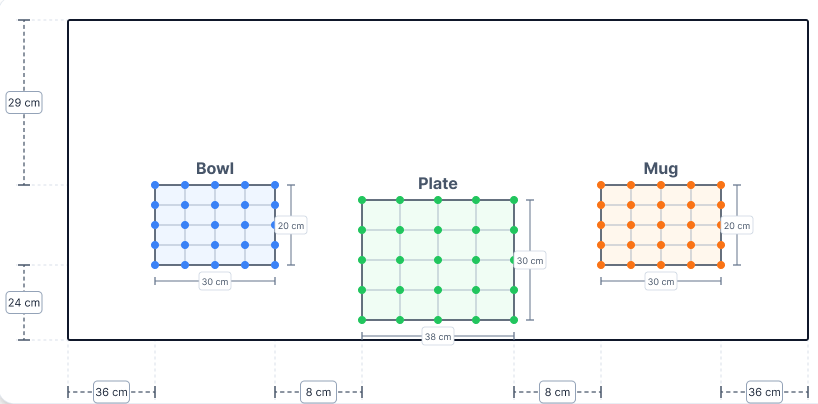}

    \caption{\textbf{Stack Dishware Evaluation Grid.} Task grid diagram detailing 25 initial starting positions of each object in \textbf{Stack Dishware}. The outer box represents the tabletop itself. Not to scale.}
    \label{fig:stackdishware_grid}
\end{figure}

For real-to-sim evaluations, we use a standardized protocol to maintain fairness and minimize variance between runs. For each task, we run 25 rollouts and each of the objects has a defined spatial reset distribution. The spatial distribution for each object is uniformly divided into a 5-by-5 grid, yielding 25 positions per object. For each rollout, one of the 25 positions is sampled independently per object, without replacement, and the center of the object is placed at this position. We also sample a rotation for each object per position, and these positions are held fixed across all checkpoints for a specific task.
\autoref{fig:stackdishware_grid} is an example of a task grid, detailing 25 possible starting positions for each object, each one represented with a dot.

The ranges of the positions are matched between the \sysName{} and real-world scenes but the exact positions may not always correspond. This was done intentionally to get a more distributional correspondence between simulation and the real world and to prevent overfitting takeaways and correlations to proprioceptive robot states.

\subsection{Metrics}
 Task success is our main metric, and for each evaluation a policy earns either a 0 or a 1 depending on if it fully accomplished the task it is being evaluated on. From task success we calculate the following metrics.

\paragraph{Real-to-Sim evaluation metrics.}
To quantify how well simulation-based evaluations predict real-world policy performance, we follow prior work~\cite{jain2025polaris,li2024evaluating} and report two complementary metrics: Pearson correlation and Mean Maximum Rank Violation (MMRV). Let $\Pi={\pi_1,\ldots,\pi_N}$ denote the set of evaluated policies. For each policy $\pi_i$, let $x_i \in [0,1]$ denote its real-world score and $y_i \in [0,1]$ denote its corresponding simulation score, computed using either task success rate or normalized reward. We collect these scores into vectors $\mathbf{x}=(x_1,\ldots,x_N)$ and $\mathbf{y}=(y_1,\ldots,y_N)$.

The Pearson correlation coefficient measures whether simulation preserves linear trends in real-world performance:
\begin{equation}
\rho(\mathbf{x}, \mathbf{y}) =
\frac{
\sum_{i=1}^{N} (x_i - \bar{x})(y_i - \bar{y})
}{
\sqrt{\sum_{i=1}^{N} (x_i - \bar{x})^2}
\sqrt{\sum_{i=1}^{N} (y_i - \bar{y})^2}
},
\end{equation}
where $\bar{x}=\frac{1}{N}\sum_i x_i$ and $\bar{y}=\frac{1}{N}\sum_i y_i$. Larger values indicate stronger agreement between simulated and real-world performance, with $\rho=1$ corresponding to perfect positive linear correlation.

Pearson correlation captures score-level agreement, but it does not directly measure whether simulation preserves policy rankings. We therefore also compute Mean Maximum Rank Violation (MMRV), which measures the average magnitude of the largest real-world performance gap involved in a simulation-induced ranking error. For each policy $\pi_i$, we identify policies $\pi_j$ whose ordering relative to $\pi_i$ differs between simulation and the real world, and take the largest real-world score difference among such inversions:
\begin{equation}
\mathrm{MMRV}(\mathbf{x}, \mathbf{y})
=
\frac{1}{N}
\sum_{i=1}^{N}
\max_{j \in \{1,\ldots,N\}}
\left[
|x_i - x_j| \cdot
\mathbbm{1}
\left(
\mathbbm{1}[y_i < y_j]
\neq
\mathbbm{1}[x_i < x_j]
\right)
\right].
\end{equation}
Here, $\mathbbm{1}(\cdot)$ is the indicator function. Intuitively, MMRV penalizes cases where simulation ranks two policies differently from the real world, with larger penalties assigned when the mis-ranked policies differ substantially in real-world performance.

\subsection{PolaRiS Real-to-Sim Experiment Details}
\label{sec:PoLaRis_baseline_details}
PolaRiS \cite{jain2025polaris} acts as a state-of-the-art baseline for evaluating real-world policies in simulation by providing a browser-based environment composer for reconstructing digital scenes that can then be brought up in Isaac Sim and evaluated using generalist policies such as \openpi.
These scenes are constructed around the DROID setup which acts as an anchor for the user to set asset initial positions and variations to create PolaRiS-ready environments for export.
We reconstructed our experiment scenes using PolaRiS's custom environment creation pipeline and tested to see how the simulated evaluation success rates correlated to real-world success rates, showing \sysName{} maintains significantly higher correlation as seen in Figure~\ref{fig:real2im_correlation}.

\subsubsection{PolaRiS Custom Environment Creation}
It is important to note that PolaRiS only provides the environment composer for scene reconstruction, meaning users must utilize recommended external software to obtain background and object digital reconstructions.
For our PolaRiS experiments, we reconstructed our DROID environment by first obtaining a video scan of the background and running it through COLMAP to obtain a sparse-reconstruction dataset of the scene. 2DGS was then used to obtain a corresponding splat and mesh of the environment, which were imported into the PolaRiS environment composer.
\sysName{} object assets were used in conjunction with the 2DGS output to recreate full task environments for some of our real-to-sim DROID tasks.

\begin{figure}[H]%
    \centering

    \begin{subfigure}{0.48\textwidth}
        \centering
        \includegraphics[width=\linewidth]{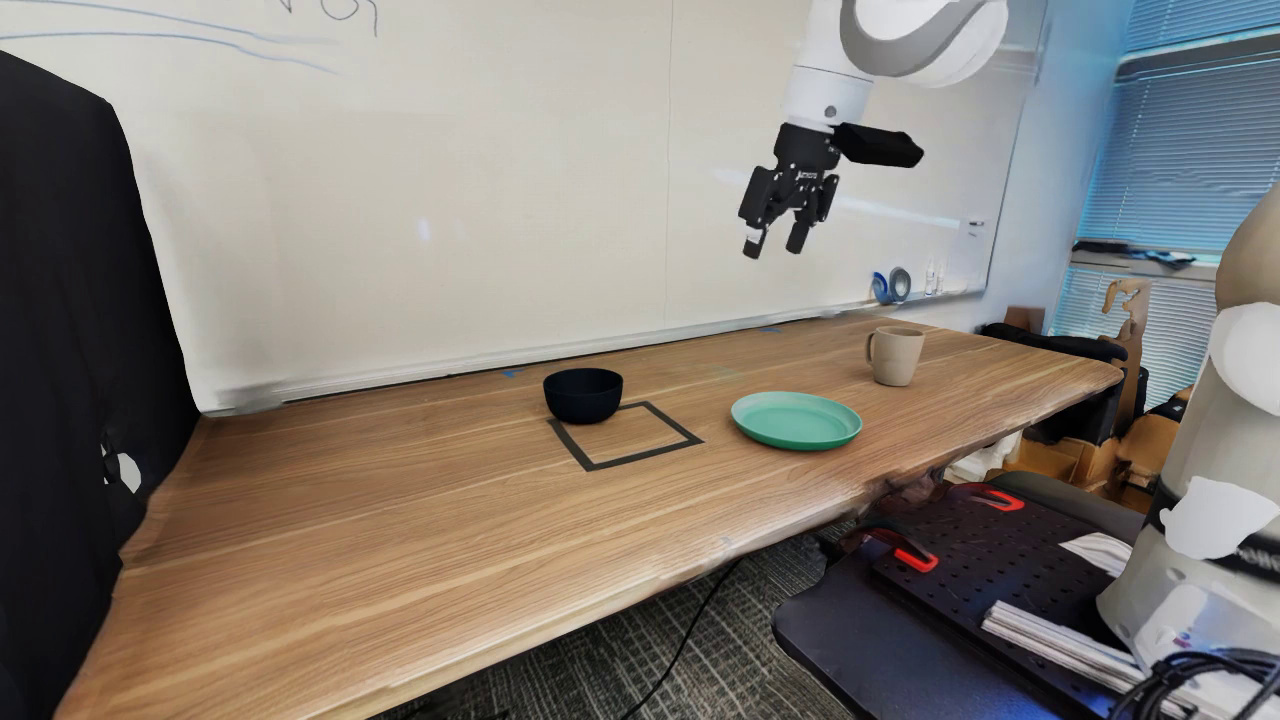}
        \caption{\sysName{} scene}
        \label{fig:sf_polaris_setup_a}
    \end{subfigure}\hfill%
    \begin{subfigure}{0.48\textwidth}
        \centering
        \includegraphics[width=\linewidth]{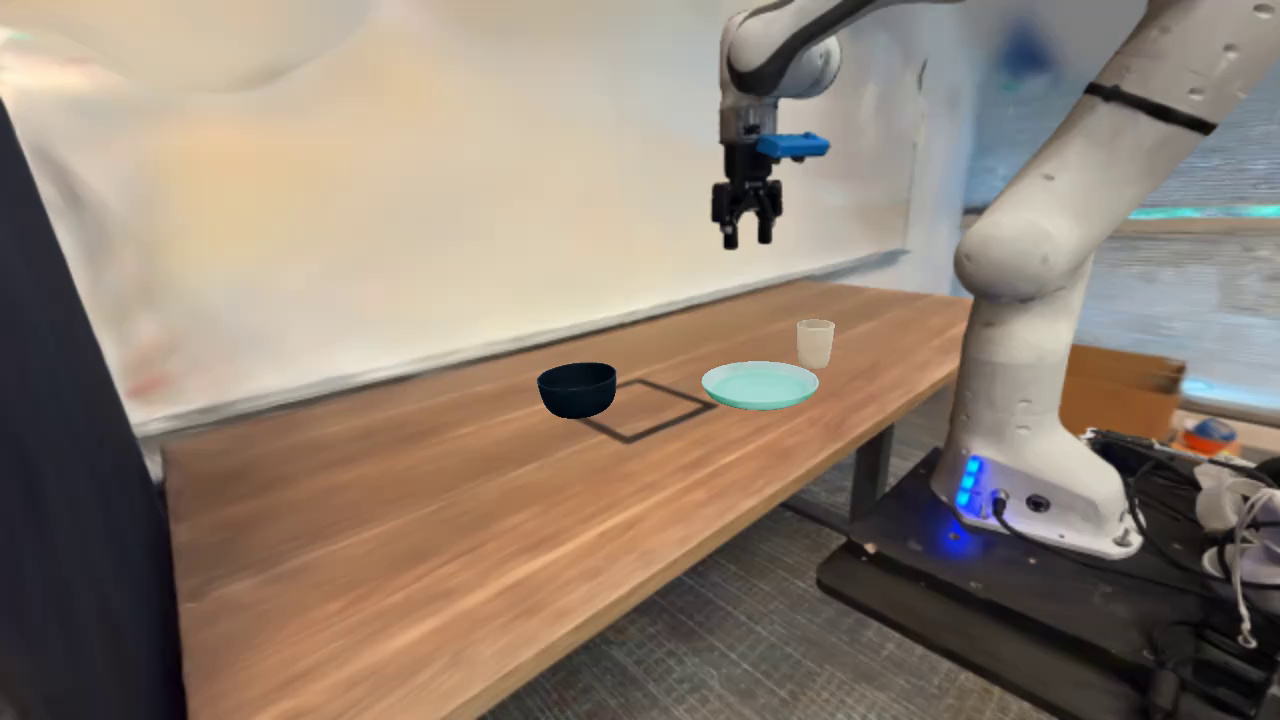}
        \caption{PolaRiS scene}
        \label{fig:sf_polaris_setup_b}
    \end{subfigure}

    \caption{\textbf{\sysName{} and PoLaRiS Scene Comparison.} \sysName{} scene compared with a PolaRiS scene for the task \textbf{Stack Dishware}}
    \label{SF_vs_polaris_scene}
\end{figure}

We place these assets in the PolaRiS scenes to match initial conditions used for real-world evaluations using the environment composer.
The resulting reconstruction is visible in Figure~\ref{SF_vs_polaris_scene}.

\subsubsection{PolaRiS Modification}
Further modifications were required to ensure that every asset gets loaded into the simulator with its textures properly and in the defined initial positions.
An invisible collider plane was added inline with our tabletop mesh to act as a flat surface that objects can rest on.
This was especially necessary for tasks with objects that can roll away, such as \markerCupTask where the marker asset was subject to the slightly uneven mesh obtained during environment digital reconstruction.
To obtain the same external viewpoint for policy observations in simulation as seen in Figure~\ref{SF_vs_polaris_scene}, we manually set the camera at the orientation matching real-world and \sysName{} evaluations.

\subsubsection{Real-to-Sim Policy Evaluations in PolaRiS}
Generalist policies were evaluated in PolaRiS using the same task initializations, policy checkpoints, and rollout protocol as in our real-world and \sysName{} evaluations.
We evaluated \openpizero, \openpi, \grootsix, \grootseven, and DreamZero in PolaRiS and the finetuned checkpoints were also evaluated on their respective tasks.
\subsubsection{PolaRiS Results Analysis}
Overall, PolaRiS yielded a low correlation to real-world policy evaluation across all tasks evaluated in our custom scene. As seen in Figure~\ref{fig:real2im_correlation} policies evaluated in PolaRiS consistently underperformed compared to their real-world success rates.
PolaRiS provides an \openpi{} policy cotrained on $10\%$ PolaRiS simulation data and $90\%$ DROID data at 1000 steps. Evaluating this policy in our custom scenes yielded higher success rates than \openpi, consistent with the PolaRiS finding that shallow simulation finetuning is crucial for improving evaluation correlation in PolaRiS\cite{jain2025polaris}. Since our simulation-based evaluation protocols focus on the zero-shot evaluation of existing policies, we exclude this adapted checkpoint from the correlation comparison in Figure~\ref{fig:real2im_correlation}.


\clearpage
\section{Human Interaction}
\label{app:human}

\subsection{Human Intervention Details}
\label{sec:human_intervention_details}

In addition to being fully automated, we provide a unified GUI with accessible touchpoints that allow human operators to easily tune our pipeline's intermediate outputs. For example, during the scene decomposition process, a human operator can intervene and enforce specific constraints on individual objects being extracted, and can quickly tweak the generated pose and scale of meshes generated during the generation process.

\subsection{Interactive Pose Refinement}
\label{sec:interactive_pose}

The Extraction stage estimates a per-object similarity transform---3-DoF translation, 3-DoF rotation, and an isotropic scale that registers each generated canonical mesh to the metric scene reconstructed pointclouds.

\paragraph{Initialization from the automatic estimate.}
The interactive tool starts with the transform emitted by the automatic Pose Matching stage, so the user always begins from the best machine
estimate rather than from scratch; in the common case where the
automatic pose is already correct, no edits are needed. Objects are processed one at a time. For each object we
load its canonical mesh and reconstruct the scene point cloud
from the stage's estimated depth and camera intrinsics
$K$ (the same per-object depth and inpainted-RGB frames used by Pose
Matching), so that the editing context is identical to the geometry
the automatic estimate was fit against.

\paragraph{Overlay visualization.}

We overlay the scene pointclouds with target object mesh, so that the user can inspect the overlay visually and use the pointcloud as reference. Because a dense point cloud can occlude the mesh and obscure these cues, the tool provides two viewing aids: the ability to toggle the point cloud visibility, and the ability to dynamically modify its density, enabling the user trade-off scene context with an unobstructed view of the mesh without altering the
underlying estimate.

\paragraph{Manual Adjustment}
The user adjusts objects' 6D poses and scales through keyboard commands. The editor also enables adjustable translation and rotation step sizes so the user can dynamically transition from coarse to fine-grained alignment within a single session. Once satisfied, the user saves the pose, after which the tool serializes the final
transform consumed by the downstream pipeline so that the manually-refined pose is loaded in place of the automatic estimate on all subsequent launches with no
further intervention.

\begin{figure}[t]
    \centering
    \setlength{\tabcolsep}{1pt}

    \begin{tabularx}{\columnwidth}{>{\scriptsize\bfseries}m{2cm} *{3}{>{\centering\arraybackslash}X}}
        \toprule

        Step & \scriptsize\textbf{1} & \scriptsize\textbf{2} & \scriptsize\textbf{3} \\

        \midrule

        Object Adjust &
        \scriptsize Baseball &
        \scriptsize Box &
        \scriptsize Mug \\
        \addlinespace[2pt]

        Full Scene Pointcloud &
        \includegraphics[width=\linewidth]{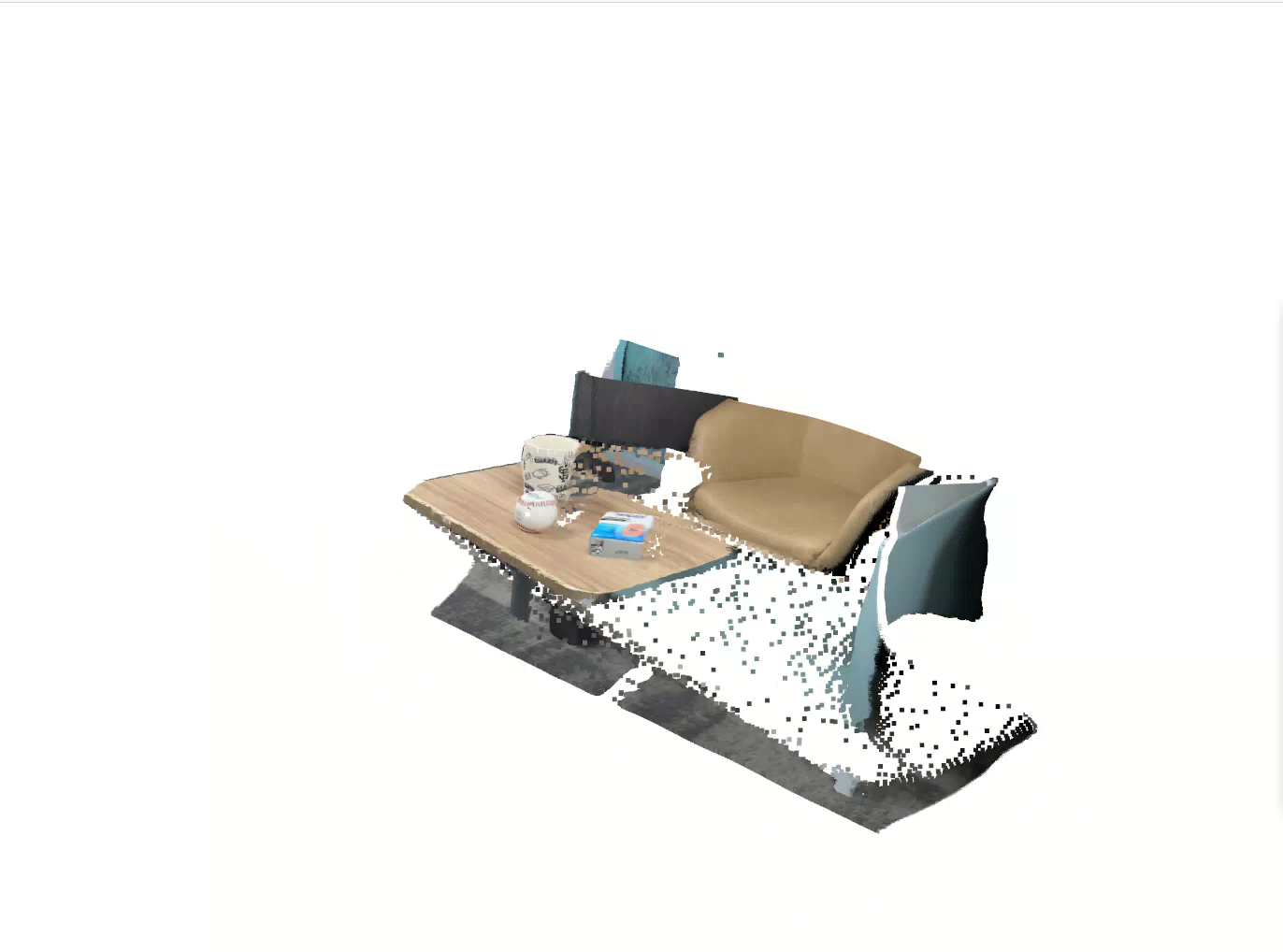} &
        \includegraphics[width=\linewidth]{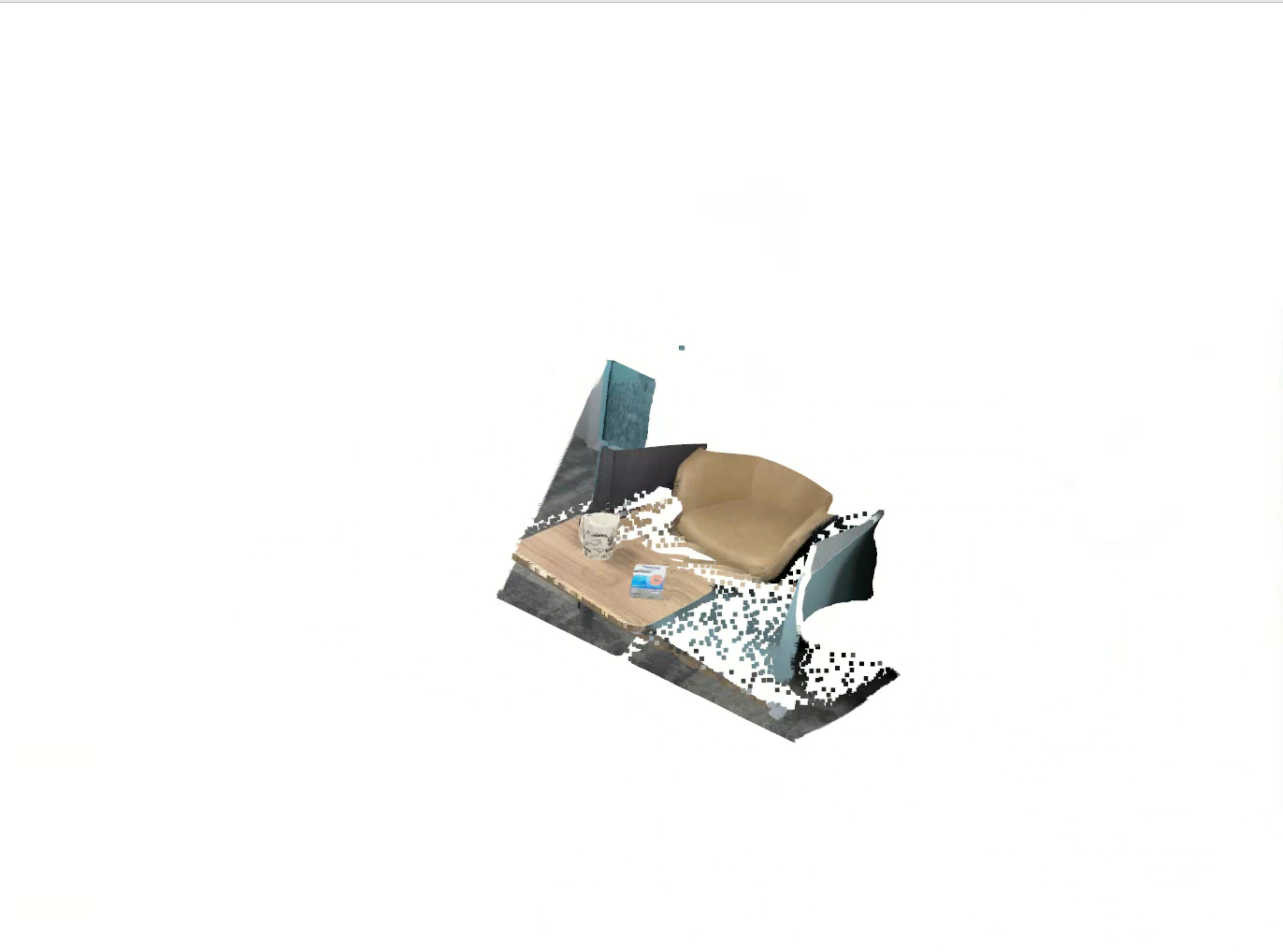} &
        \includegraphics[width=\linewidth]{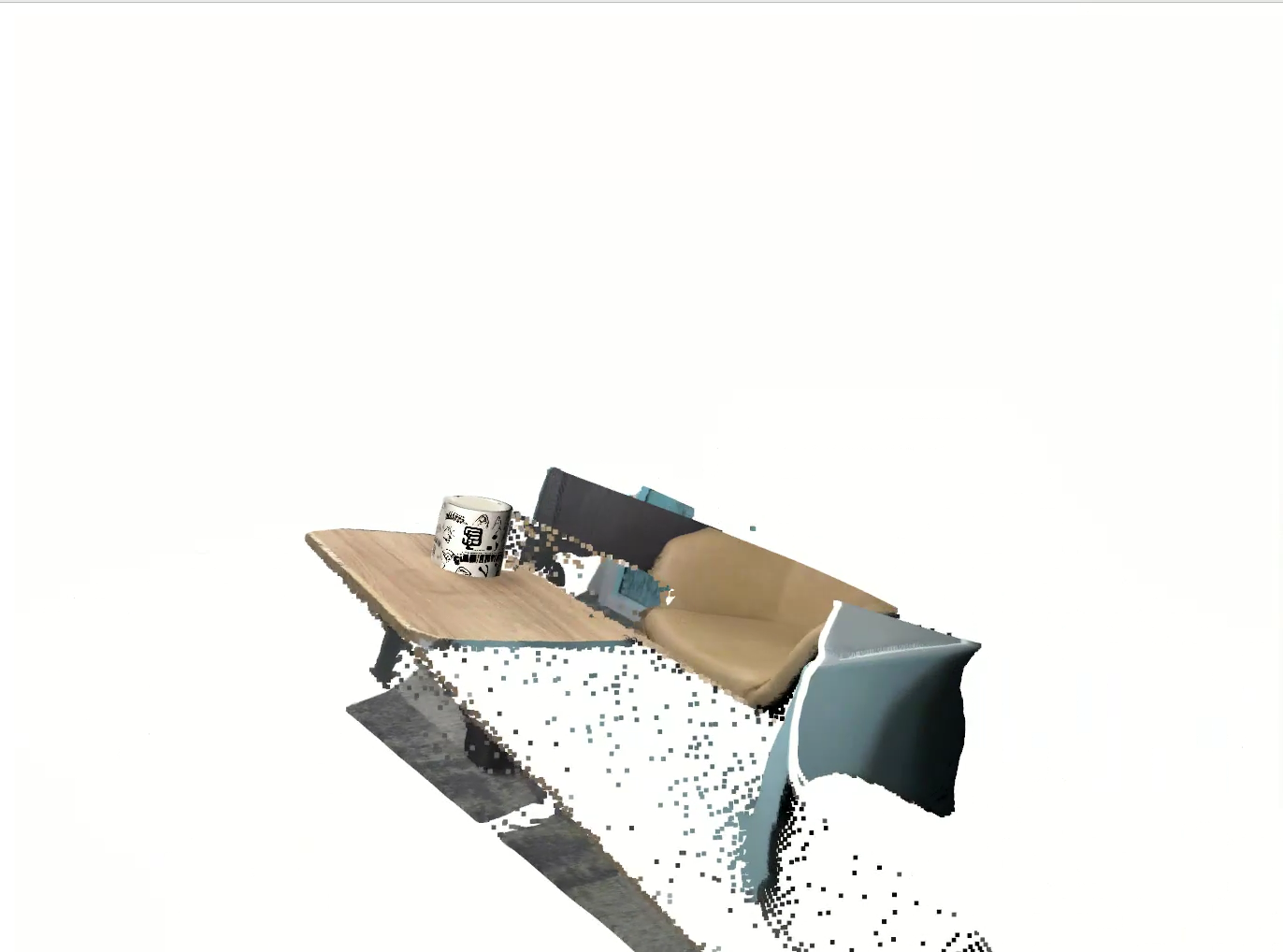} \\
        \addlinespace[2pt]

        Object Mesh \& PC Overlay &
        \includegraphics[width=\linewidth]{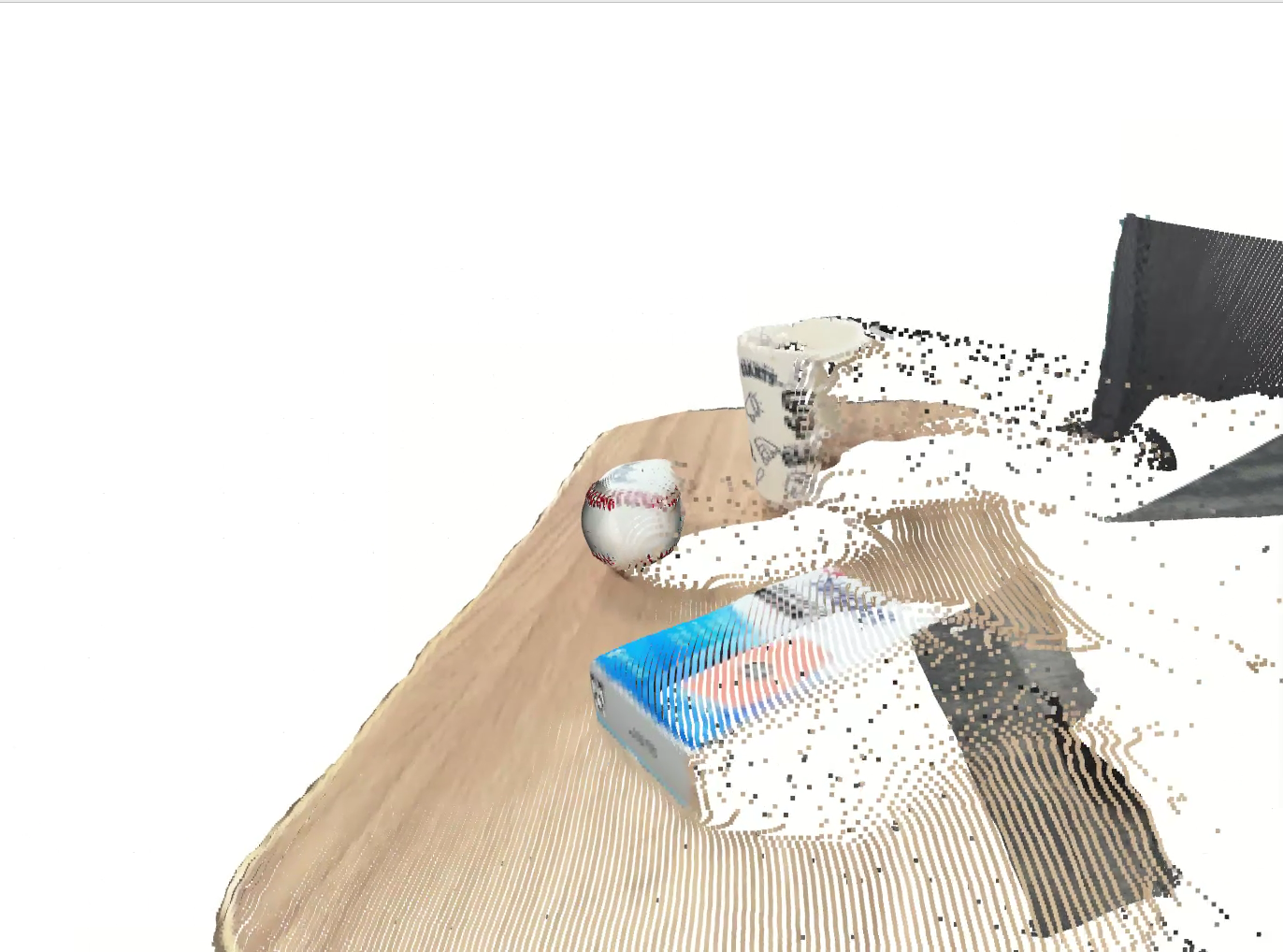} &
        \includegraphics[width=\linewidth]{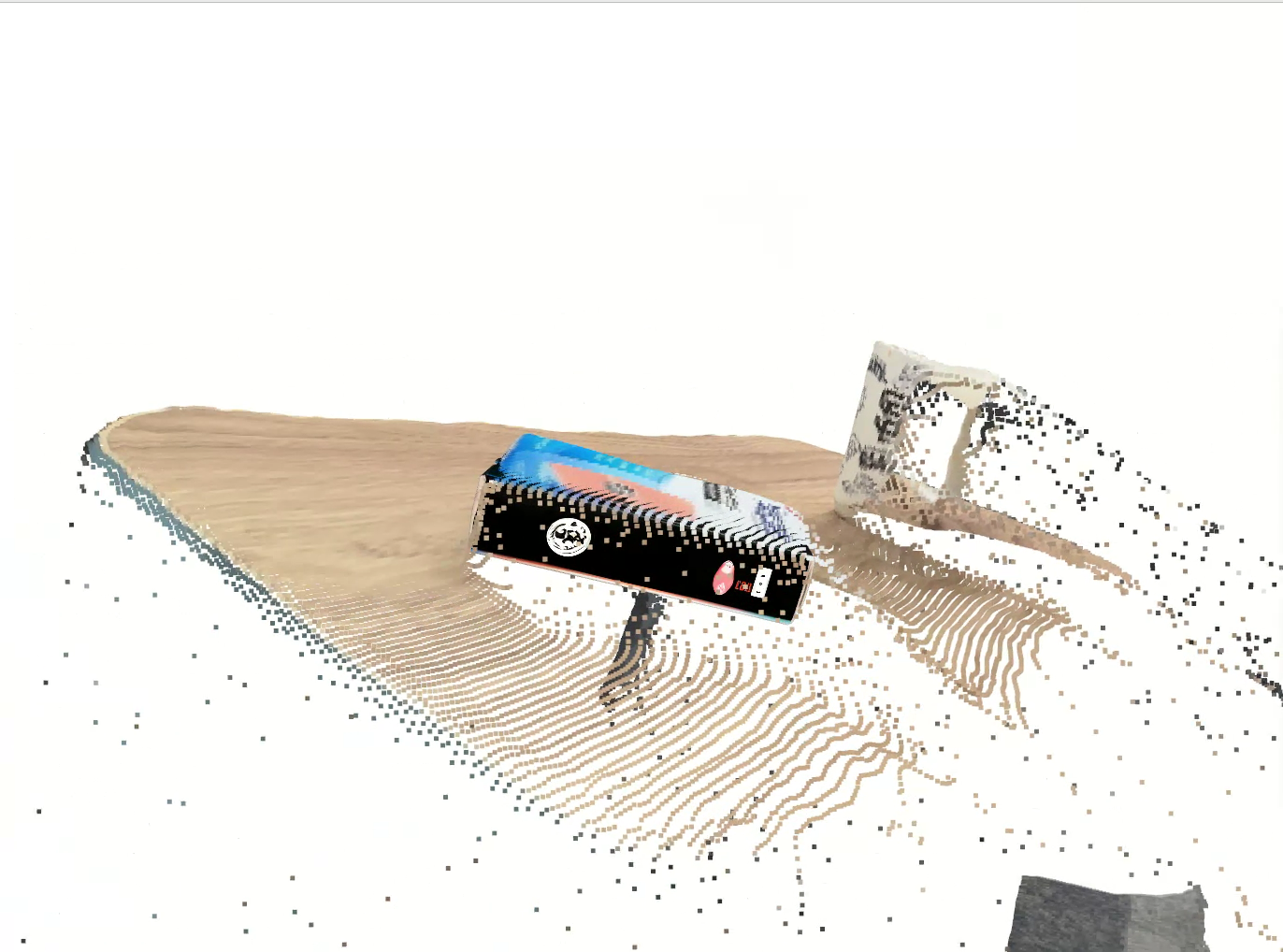} &
        \includegraphics[width=\linewidth]{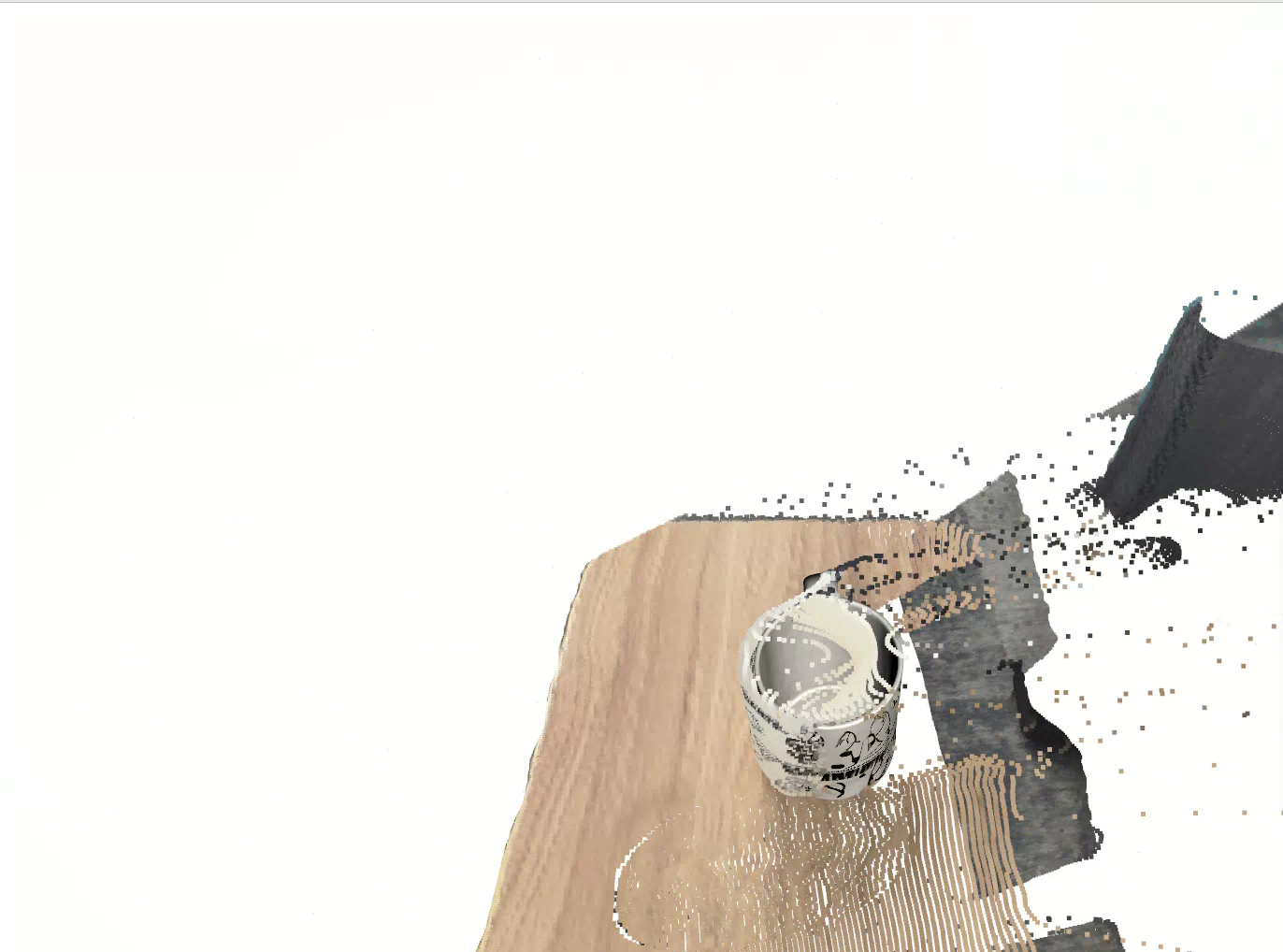} \\

        \bottomrule
    \end{tabularx}
    \caption{\textbf{Interactive Scene Editor Procedure.} The interactive scene editor launches the adjusted object mesh and the (inpainted) dense pointcloud for the scene, allowing the user to adjust object poses to align to the scene pointcloud. This process continues until all object poses have been adjusted. Note that in every step the GUI loads an inpainted pointcloud that erases previous objects to support occluded object pose tuning.}
    \label{fig:interactive_scene_edit}
\end{figure}

\paragraph{Iterative refinement across sessions.}
Saved poses are written to a fresh output so that no edit overwrites the automatic estimate or a previous manual pass. The tool can therefore be re-entered to resume from either the automatic output or any earlier interactive pass, refining the same scene over multiple sittings while retaining every intermediate version for comparison or rollback. In practice this makes the tool optional: the automatic pipeline runs unattended and suffices for the bulk of objects, while the small number of poses that matter most for the scene or a quantitative evaluation can be improved with a few minutes of guided manipulation per object. See Figure~\ref{fig:interactive_scene_edit} for an actual example of running interactive pose refinement.

\paragraph{Use Cases.} We primarily use iterative pose refinement for more accurately localizing and aligning the poses of the objects for the Real-to-Sim experiments (Section~\ref{subsec:real_to_sim_eval}). For Sim-to-Real data generation, iterative refinement has less utility, as long as the ranges of spatial initializations for the objects in simulation are large enough to cover the real-world test scenarios.

\clearpage
\section{System Analysis}
\label{app:system}
\FloatBarrier

\subsection{3D Reconstruction Evaluation Details}
\label{sec:3d_reconstruction_evaluation_details}

We describe (a) how quasi-ground truth scene reconstructions are obtained, (b) how SAM3D~\cite{sam3dteam2025sam3d3dfyimages} outputs are
placed in a common frame so that they can be compared with
\sysName{} outputs, and (c) quantitative and qualitative reconstruction results across the full set of 12 scenes.

We categorize 12 table-top scenes from difficulty low to high based on the object occlusion level using objects from the YCB dataset~\cite{calli2015ycb}, as shown in Table~\ref{tab:reconstruct_scene_difficulty}.
\begin{table}[t]
    \centering
    \setlength{\tabcolsep}{1pt}

    \begin{tabular}{>{\scriptsize\bfseries}m{1.2cm} c c c c}
        \toprule

        Difficulty & \multicolumn{4}{c}{\scriptsize\textbf{Input Scenes}} \\

        \midrule

        Easy  &
        \includegraphics[width=0.18\linewidth]{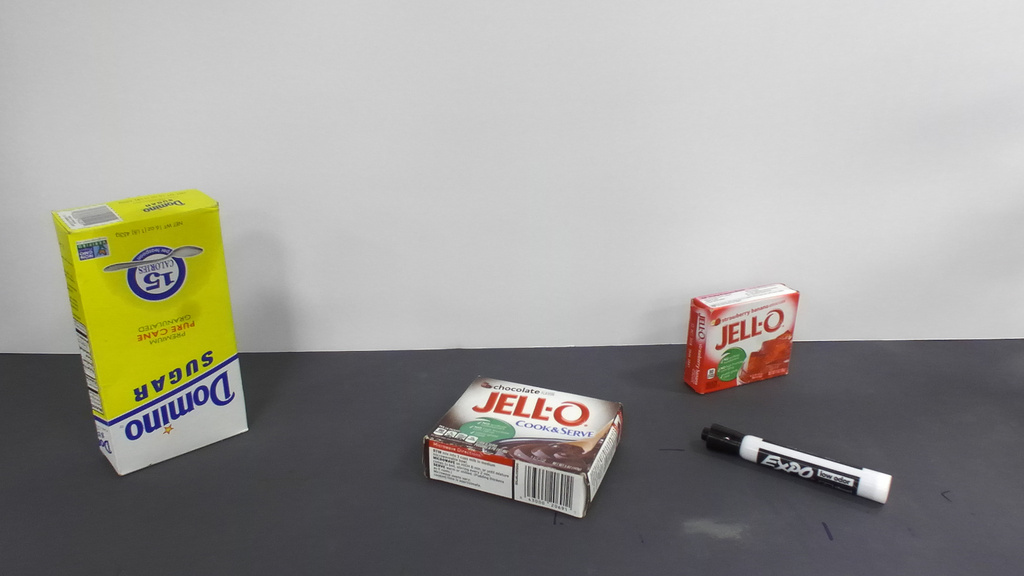} &
        \includegraphics[width=0.18\linewidth]{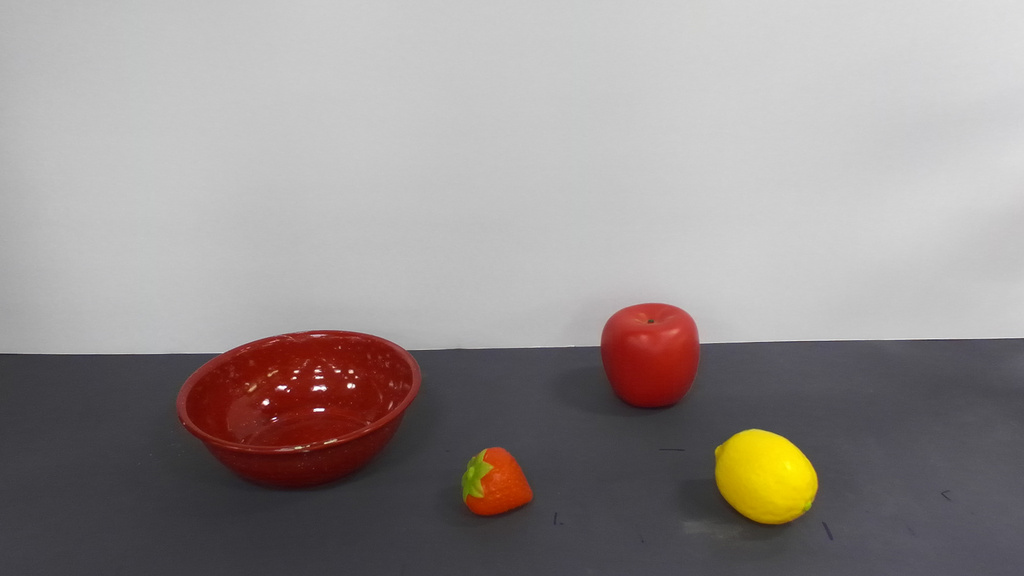} &
        \includegraphics[width=0.18\linewidth]{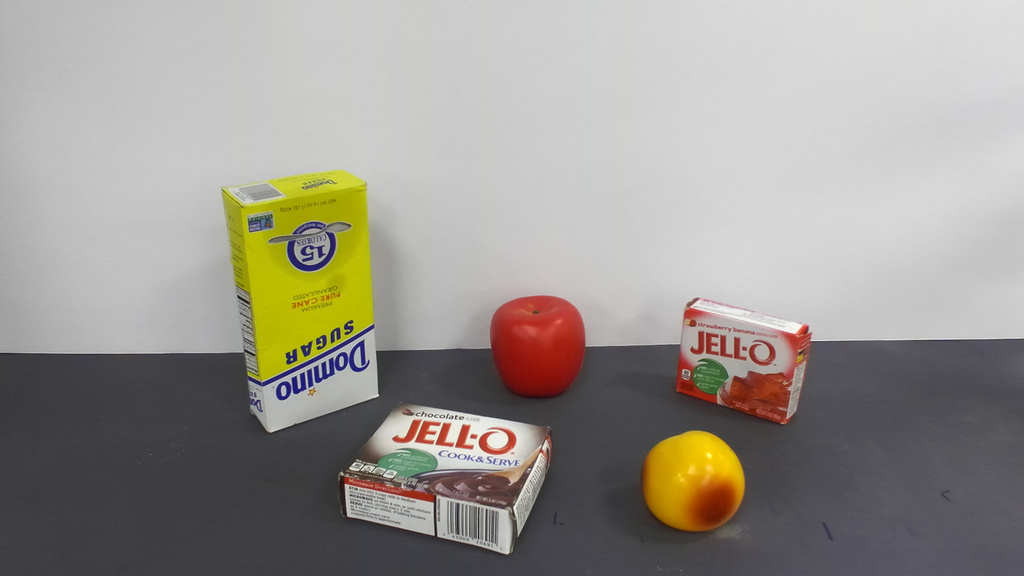} &
        \includegraphics[width=0.18\linewidth]{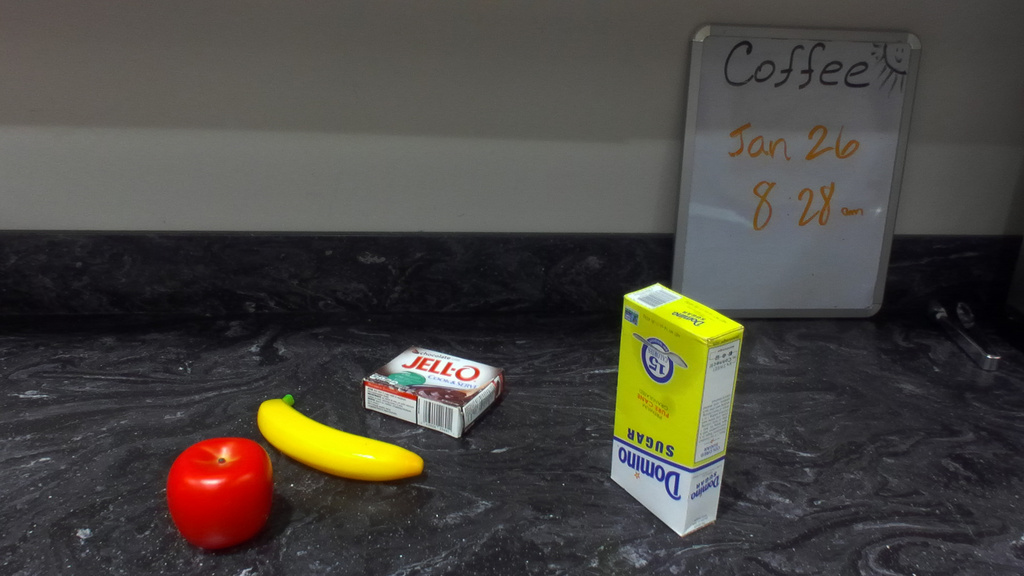} \\
        \addlinespace[2pt]

        Med &
        \includegraphics[width=0.18\linewidth]{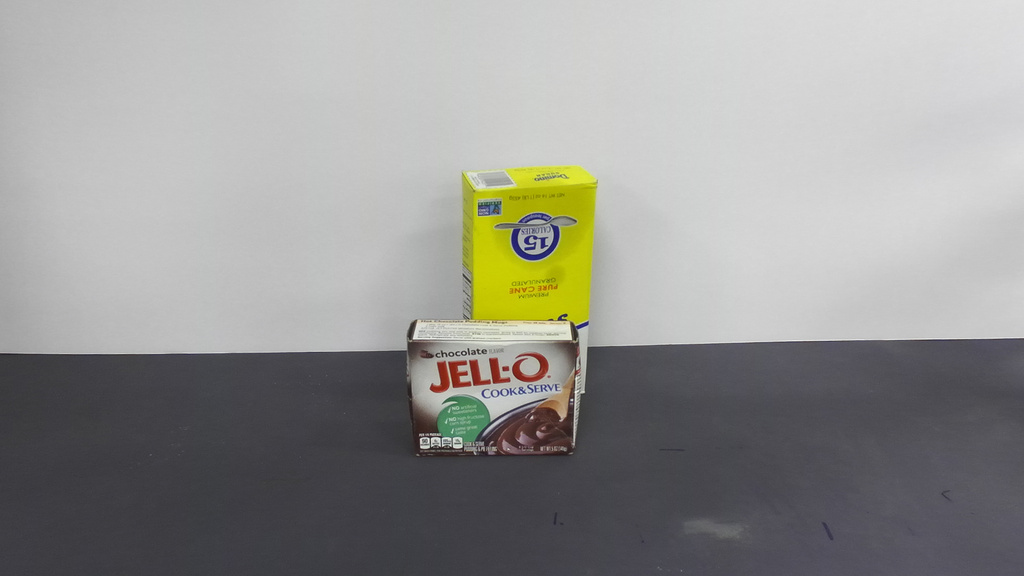} &
        \includegraphics[width=0.18\linewidth]{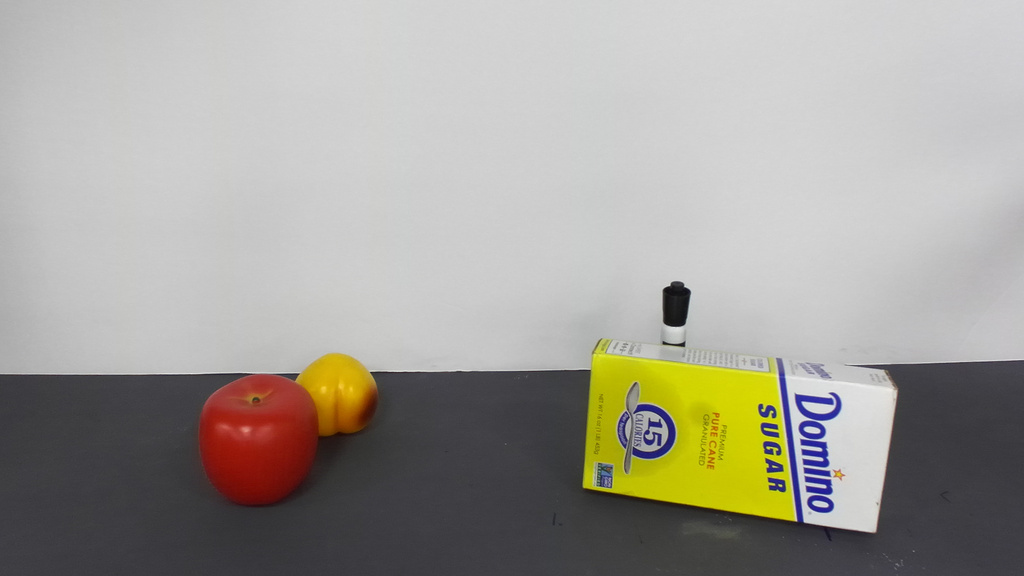} &
        \includegraphics[width=0.18\linewidth]{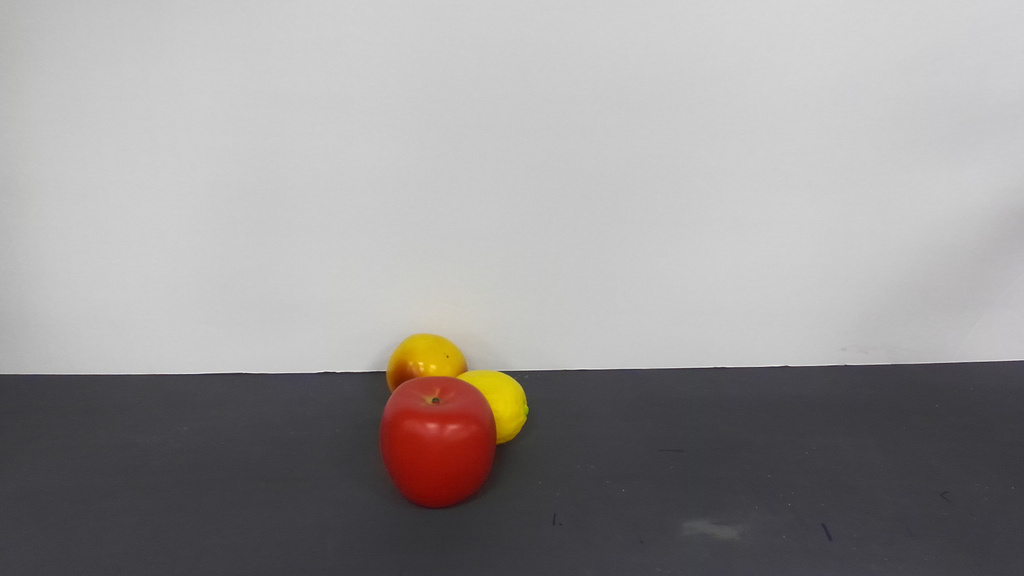} &
        \includegraphics[width=0.18\linewidth]{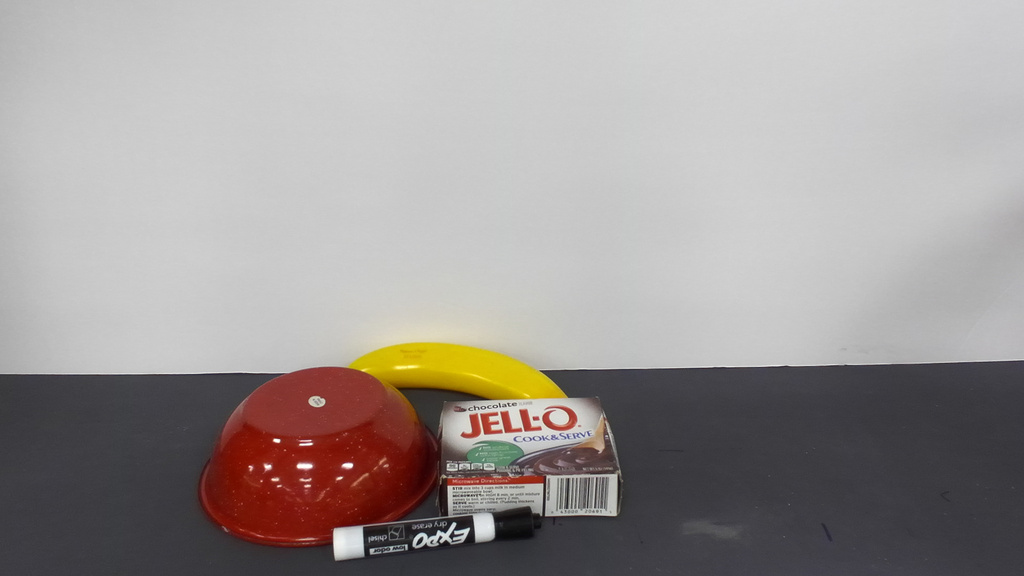} \\
        \addlinespace[2pt]

        Hard &
        \includegraphics[width=0.18\linewidth]{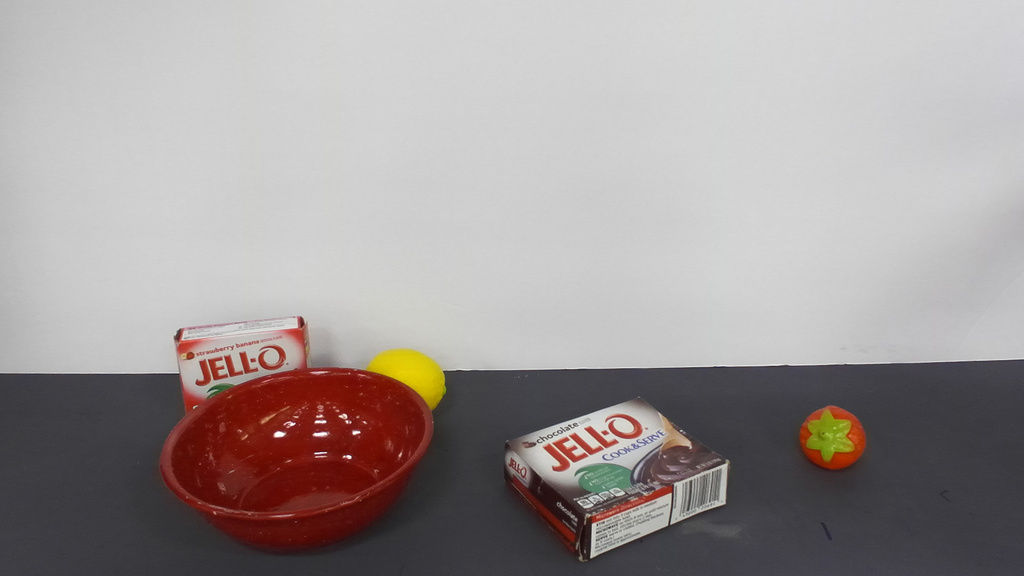} &
        \includegraphics[width=0.18\linewidth]{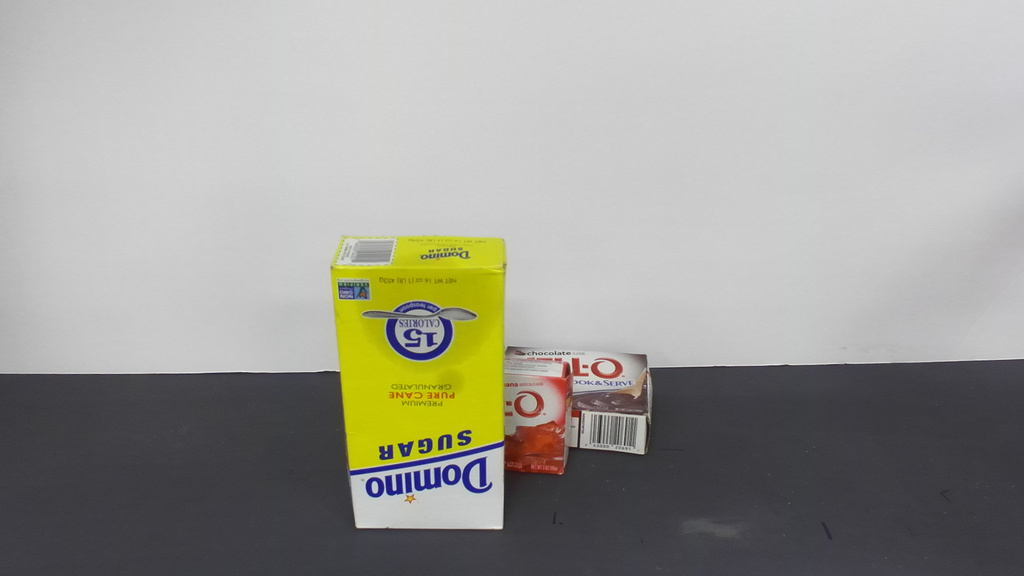} &
        \includegraphics[width=0.18\linewidth]{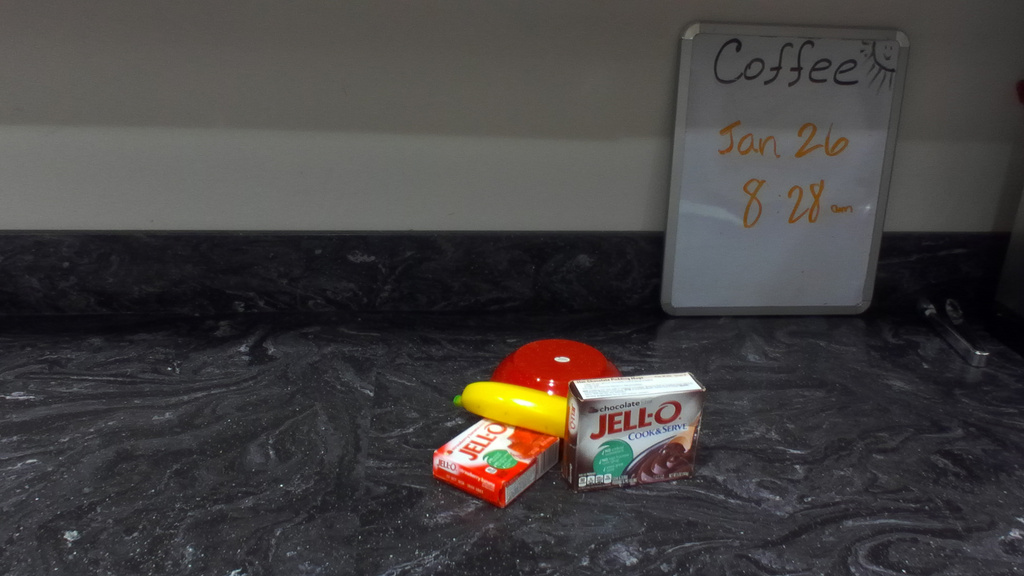} &
        \includegraphics[width=0.18\linewidth]{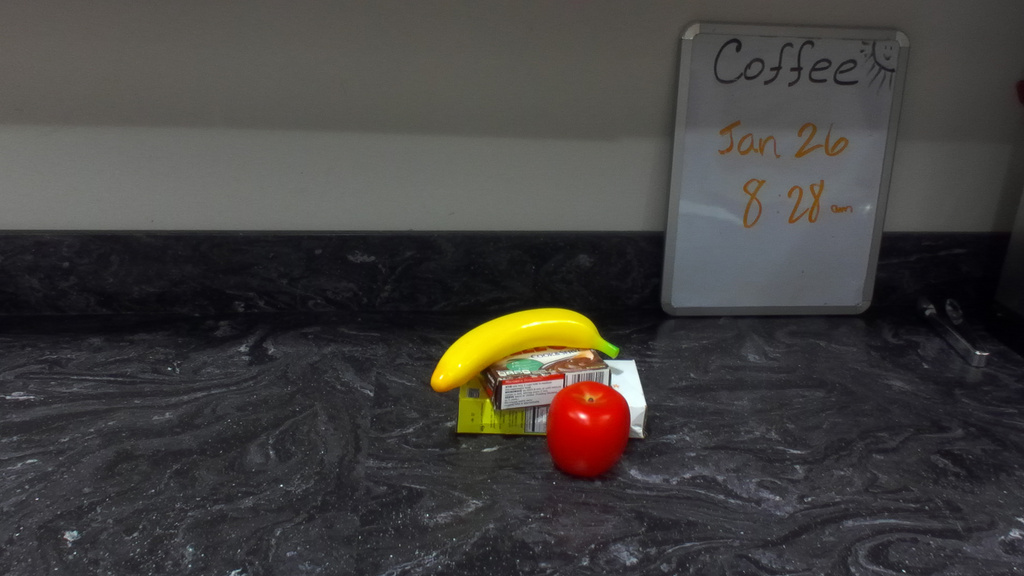} \\

        \bottomrule
    \end{tabular}
    \caption{\textbf{Reconstructed Scenes.} Categorized by Easy (No Occlusion), Mid (Slight Occlusion), and Hard (Strong Occlusion).}
    \label{tab:reconstruct_scene_difficulty}
\end{table}

\subsubsection{Quasi-Ground-Truth Scene Reconstruction}

Evaluating reconstruction fidelity in cluttered scenes requires per-object ground truth poses, which are difficult to recover directly under heavy occlusion. We therefore stage each benchmark scene incrementally and record a quasi-ground truth pose for each object while it is still fully visible. Concretely, we place the object furthest in the background first and infer its 6-DoF pose
with FoundationPose~\cite{wen2024foundationpose} and the known ground truth CAD mesh of the object from an unoccluded view, then repeat the procedure for each subsequent object until the full scene is staged. The recorded poses serve as the ground truth for all subsequent metrics. At evaluation time, only the final fully-staged (and therefore occluded) scene
image is provided to either reconstruction method.

\subsubsection{SAM3D Reconstruction Pipeline}

SAM3D produces per-object meshes and relative object poses in its own coordinate convention and with an arbitrary global scale, so a direct comparison with \sysName{} requires placing both outputs in
a shared metric world frame. We use SAM3D solely for object geometry and object relative poses, and resolve the remaining frame and scale ambiguities through three steps:

\begin{enumerate}
    \item \textbf{Convert coordinate conventions}: We map SAM3D mesh coordinates through the required axis-system changes (GLB convention $\to$ SAM3D internal $\to$ OpenCV camera convention) so the object points are expressed in the same camera convention used by our pipeline.

    \item \textbf{Place objects in a common world frame}: We apply the same world transform used by \sysName{} so SAM3D objects are moved into the identical metric world frame (same origin/z-axis convention).

    \item \textbf{Resolve SAM3D global scale ambiguity}: Because SAM3D reconstruction scale is arbitrary, we estimate one scene-level scale by matching SAM3D’s merged scene point cloud with ground-truth point cloud by calculating the smallest chamfer distance. Specifically, SAM3D's output point clouds are scaled around the camera origin, to stay physically consistent with camera-frame scaling.
\end{enumerate}

After these steps SAM3D outputs live in the same metric world as \sysName{} outputs, so any residual differences in the reported metrics are attributable to reconstruction quality rather than to frame or scale mismatch.

\subsubsection{Quantitative Reconstruction Results}

With output from \sysName{} and SAM3D, we measure reconstruction fidelity via three 3D geometric metrics: Chamfer Distance, F1-Score (with threshold 0.01 meters) and Object Bounding Box Position Error against quasi-ground-truth, the quantitative result is shown in Table~\ref{tab:scene_reconstruction_quantitative}. We find that \sysName{} outperforms SAM3D in zero-shot, and additionally further improves its geometric reconstruction fidelity with a few minutes of interactive human iteration.
\begin{table}[htbp]
\centering
\caption{\textbf{Scene Reconstruction Quantitative Results.} Entries show Average $\pm$ Standard Deviation.}
\label{tab:scene_reconstruction_quantitative}
\resizebox{\columnwidth}{!}{%
\begin{tabular}{llccc}
\toprule
\textbf{Difficulty} & \textbf{Metric} & \textbf{SAM3D Zero Shot} & \textbf{\sysName{} Zero Shot} & \textbf{\sysName{} Tuned (3min/Obj)} \\
\midrule
\multirow{3}{*}{\textbf{Easy}}
 & Chamfer Dist (m) $\downarrow$ & $0.0081 \pm 0.0024$ & $0.0042 \pm 0.0013$ & $\mathbf{0.0026 \pm 0.00026}$ \\
 & F1 Score $\uparrow$ & $0.71 \pm 0.15$ & $0.92 \pm 0.071$ & $\mathbf{0.99 \pm 0.0069}$ \\
 & Pos Error (m) $\downarrow$ & $0.016 \pm 0.0058$ & $0.0060 \pm 0.0019$ & $\mathbf{0.0041 \pm 0.00037}$ \\
\midrule
\multirow{3}{*}{\textbf{Medium}}
 & Chamfer Dist (m) $\downarrow$ & $0.0087 \pm 0.0028$ & $0.0047 \pm 0.0012$ & $\mathbf{0.0033 \pm 0.00068}$ \\
 & F1 Score $\uparrow$ & $0.66 \pm 0.18$ & $0.87 \pm 0.089$ & $\mathbf{0.97 \pm 0.026}$ \\
 & Pos Error (m) $\downarrow$ & $0.018 \pm 0.0067$ & $0.0076 \pm 0.0038$ & $\mathbf{0.0057 \pm 0.0030}$ \\
\midrule
\multirow{3}{*}{\textbf{Hard}}
 & Chamfer Dist (m) $\downarrow$ & $0.0088 \pm 0.0022$ & $0.0091 \pm 0.0076$ & $\mathbf{0.0039 \pm 0.0013}$ \\
 & F1 Score $\uparrow$ & $0.68 \pm 0.14$ & $0.81 \pm 0.071$ & $\mathbf{0.93 \pm 0.049}$ \\
 & Pos Error (m) $\downarrow$ & $0.022 \pm 0.010$ & $0.018 \pm 0.018$ & $\mathbf{0.0073 \pm 0.0022}$ \\
\bottomrule
\end{tabular}%
}
\end{table}

\subsubsection{Qualitative Reconstruction Results}

 We provide the corresponding visual reconstructions results for all 12 benchmark scenes in Table~\ref{tab:scene_reconstruction_qualitative} below, covering the full range of clutter and occlusion levels used in the evaluation.

\begin{table*}[t]
\centering
\small
\begin{tabular*}{\textwidth}{@{\extracolsep{\fill}} l l c c c c @{}}
\toprule
\makecell{Scene \\ Name} &
\makecell{Difficulty} &
\makecell{Real Scene} &
\makecell{SAM3D \\ Zero Shot} &
\makecell{\sysName \\ Zero Shot} &
\makecell{\sysName \\ Tuned {\scriptsize (3min/obj)}} \\
\midrule

Desk 1 & Easy &
\includegraphics[width=0.165\textwidth]{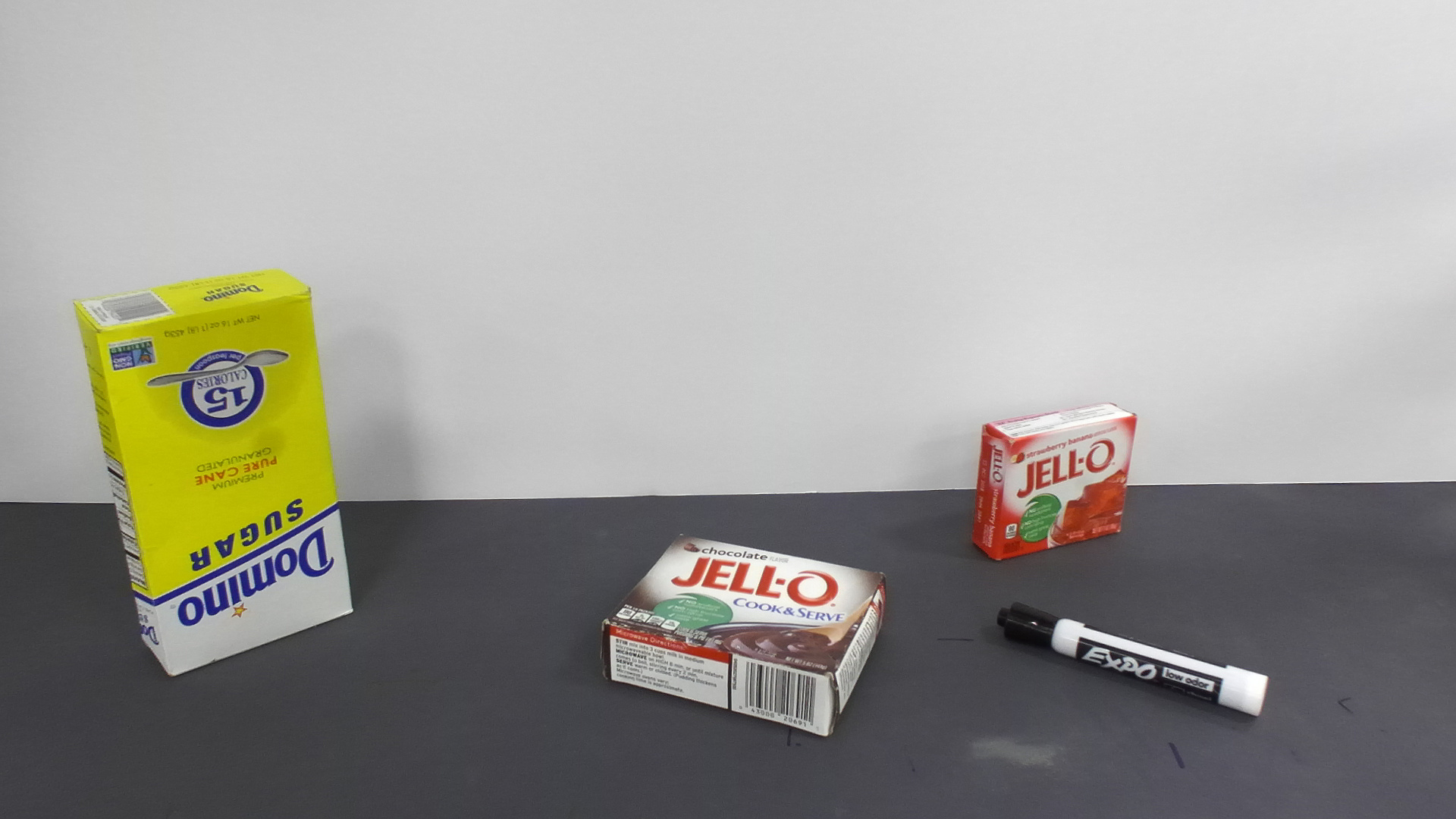} &
\includegraphics[width=0.165\textwidth]{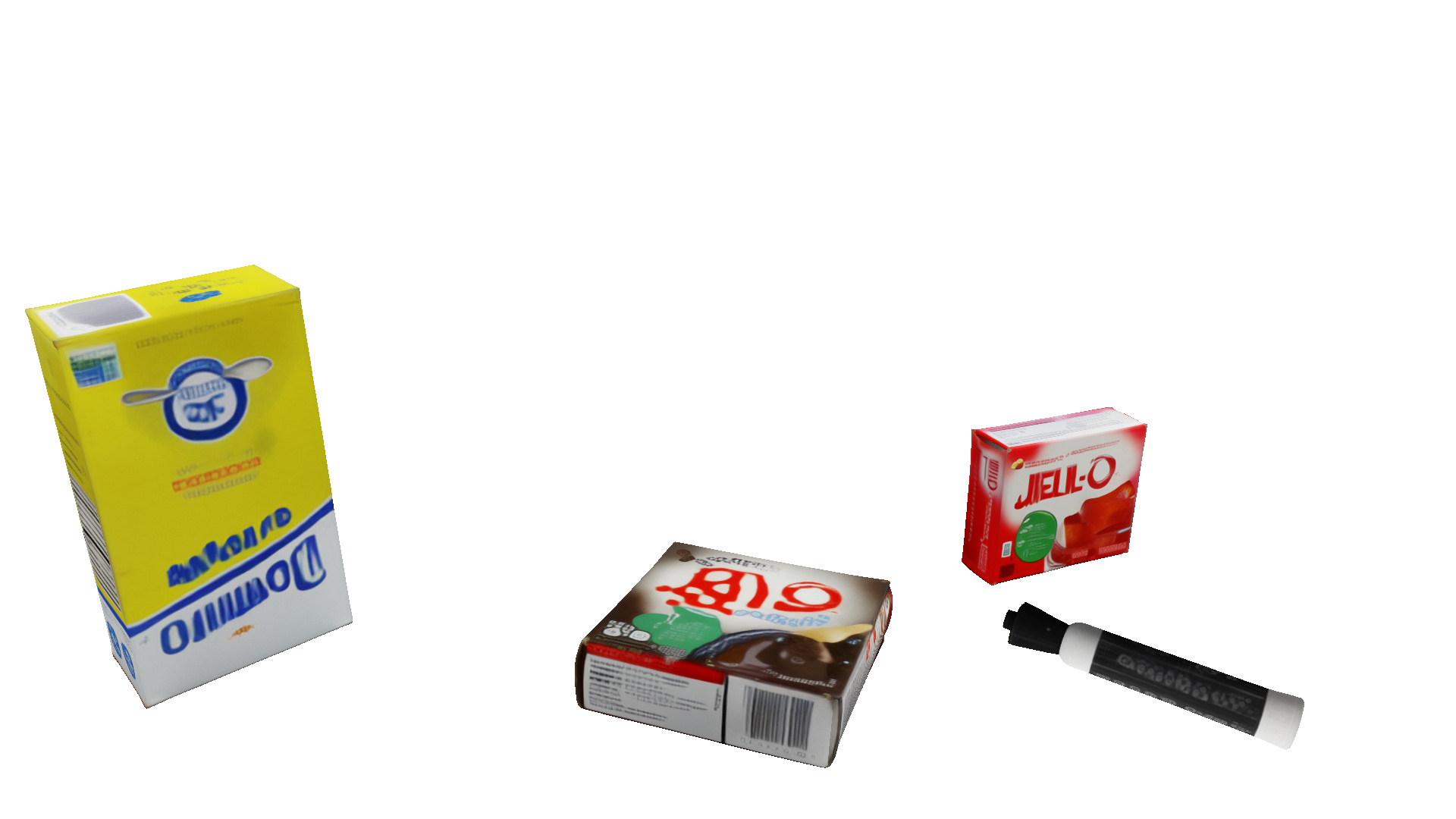} &
\includegraphics[width=0.165\textwidth]{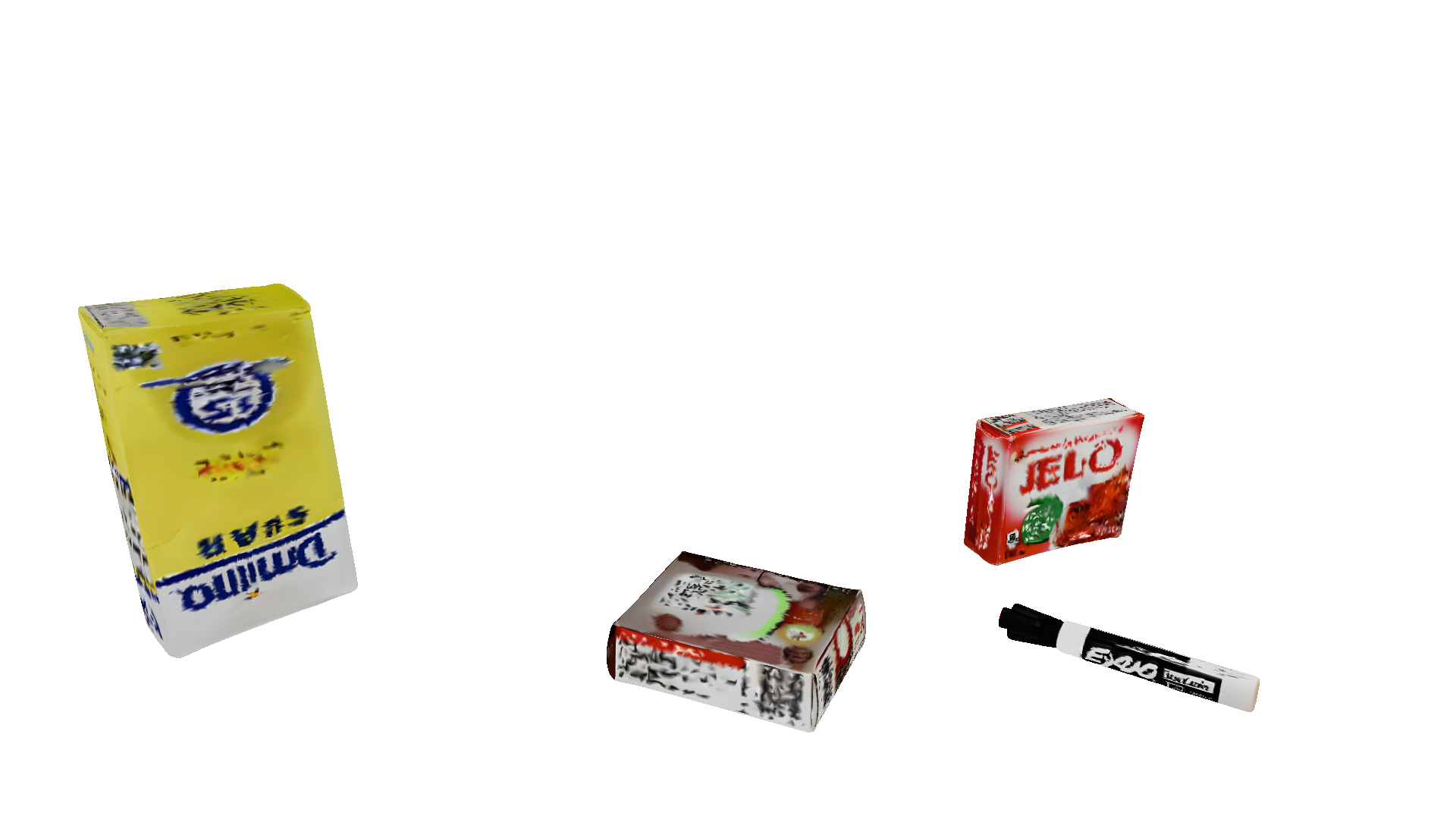} &
\includegraphics[width=0.165\textwidth]{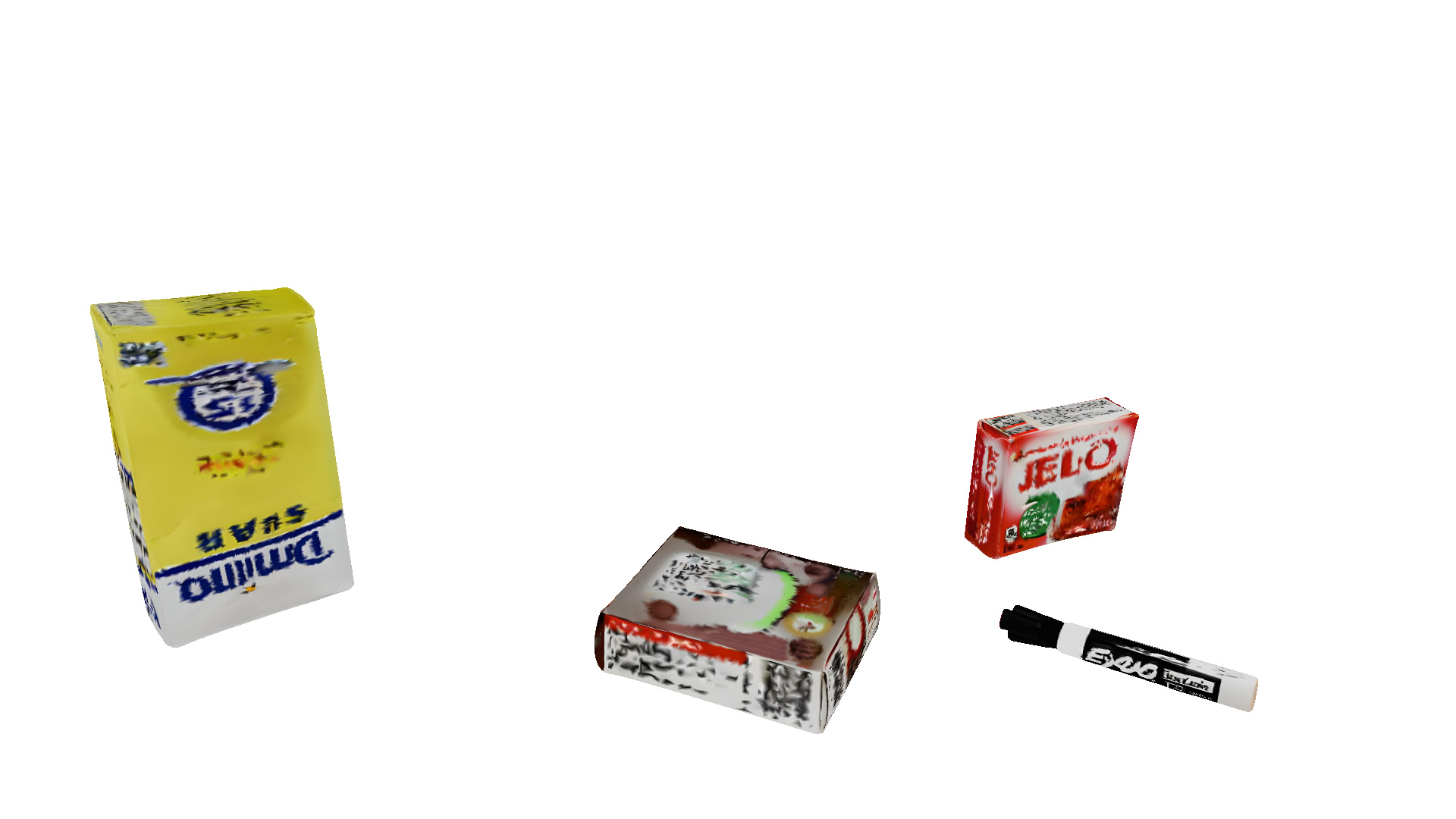} \\

Desk 2 & Easy &
\includegraphics[width=0.165\textwidth]{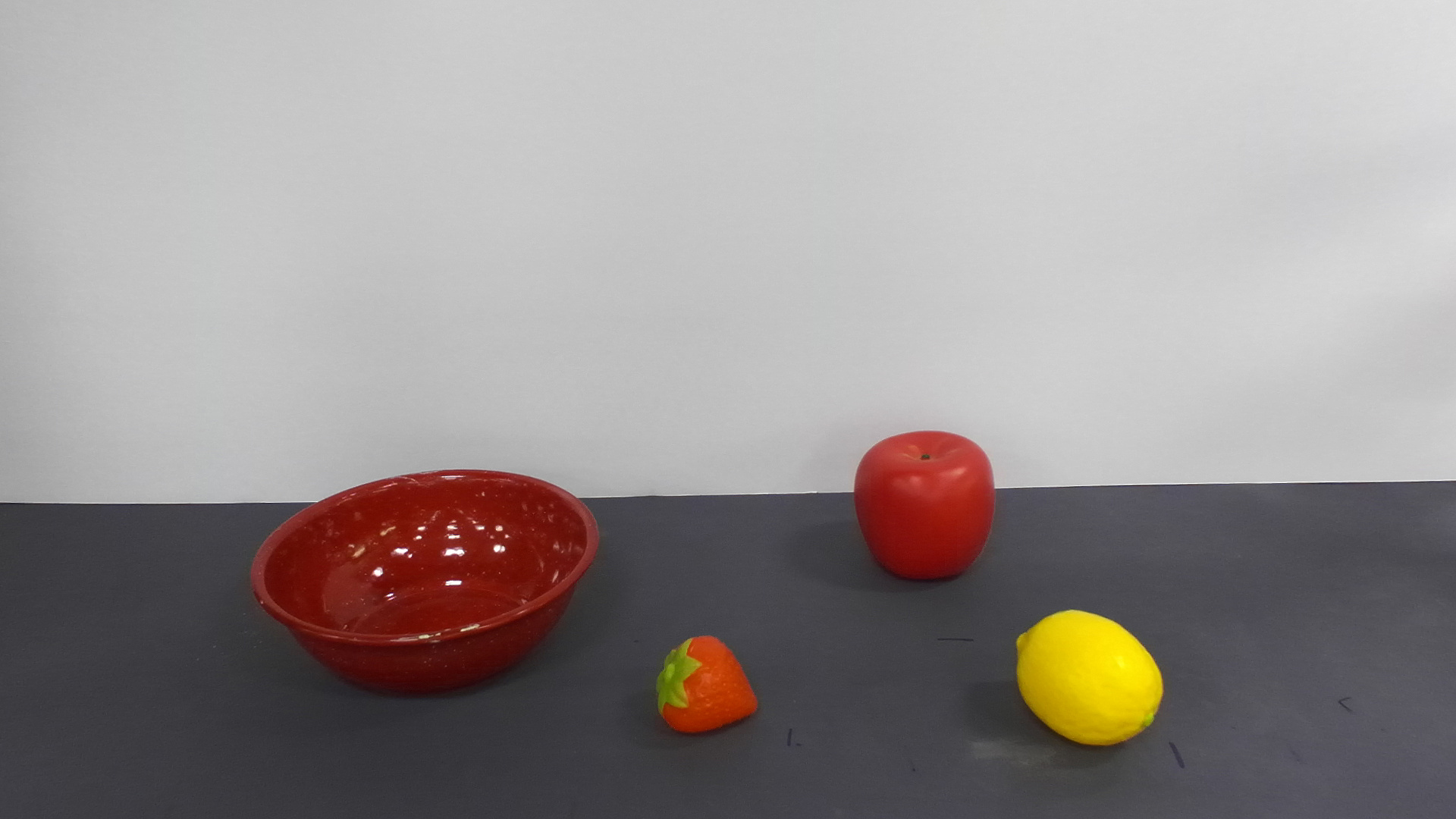} &
\includegraphics[width=0.165\textwidth]{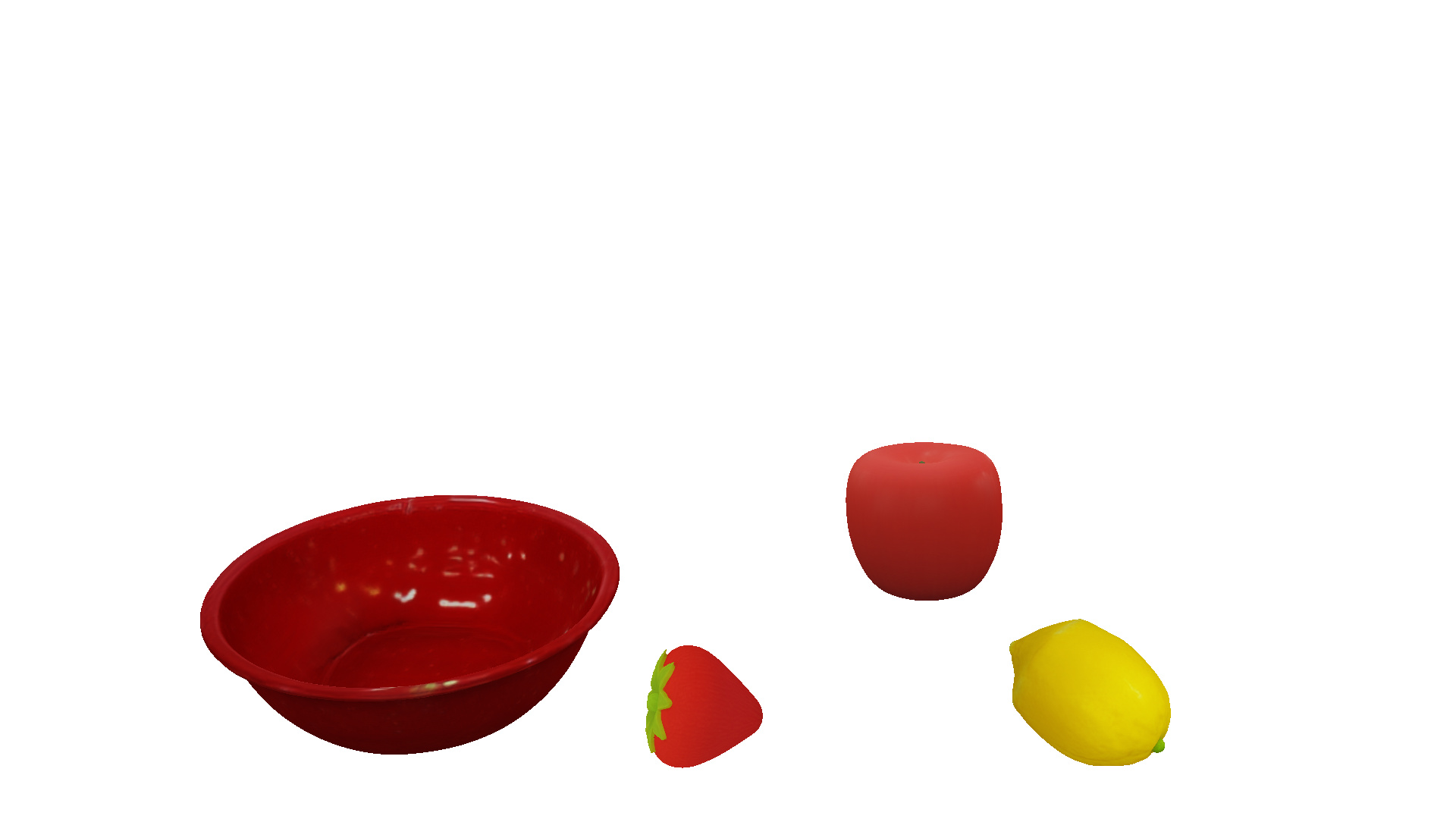} &
\includegraphics[width=0.165\textwidth]{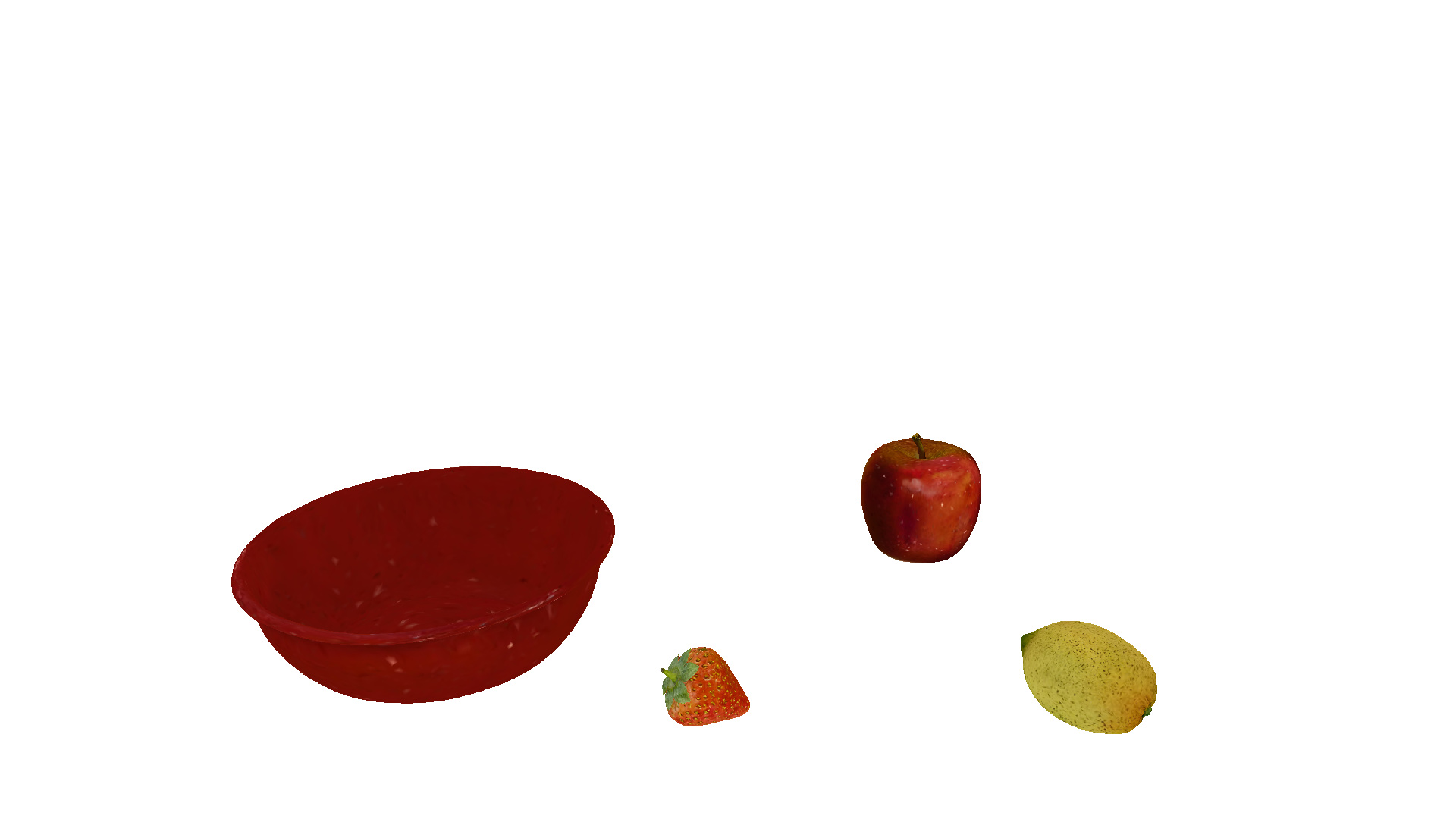} &
\includegraphics[width=0.165\textwidth]{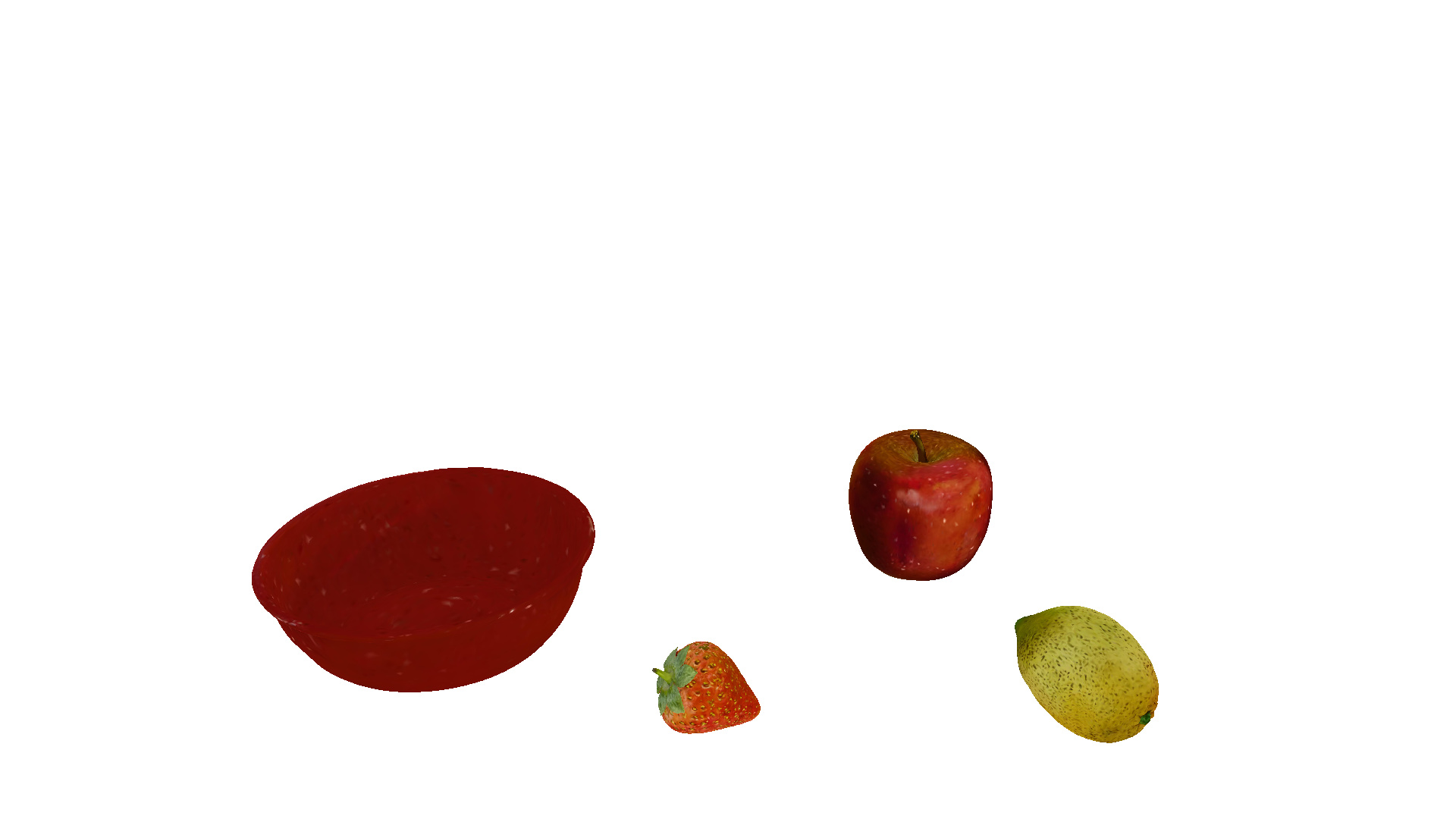} \\

Desk 3 & Easy &
\includegraphics[width=0.165\textwidth]{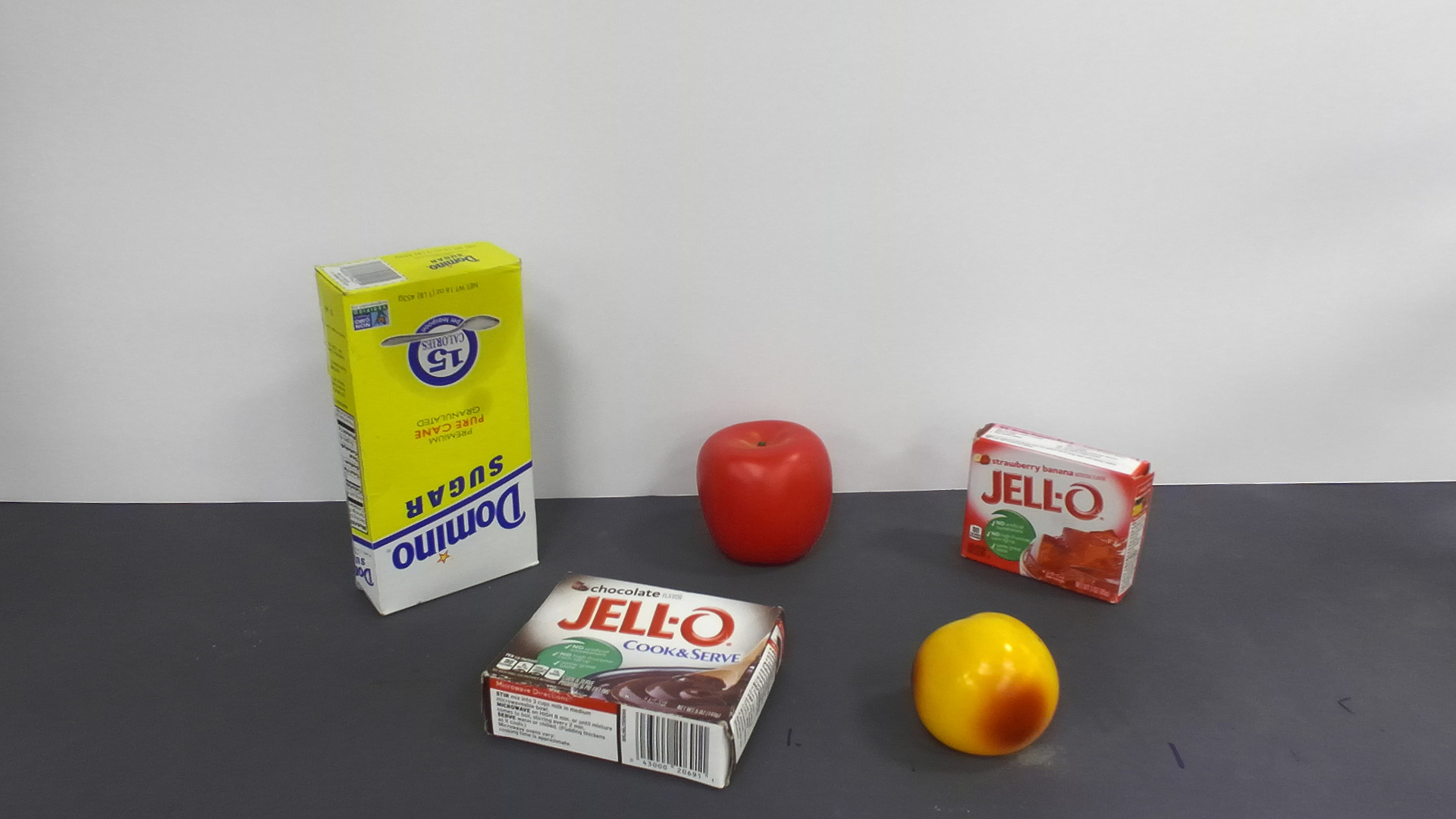} &
\includegraphics[width=0.165\textwidth]{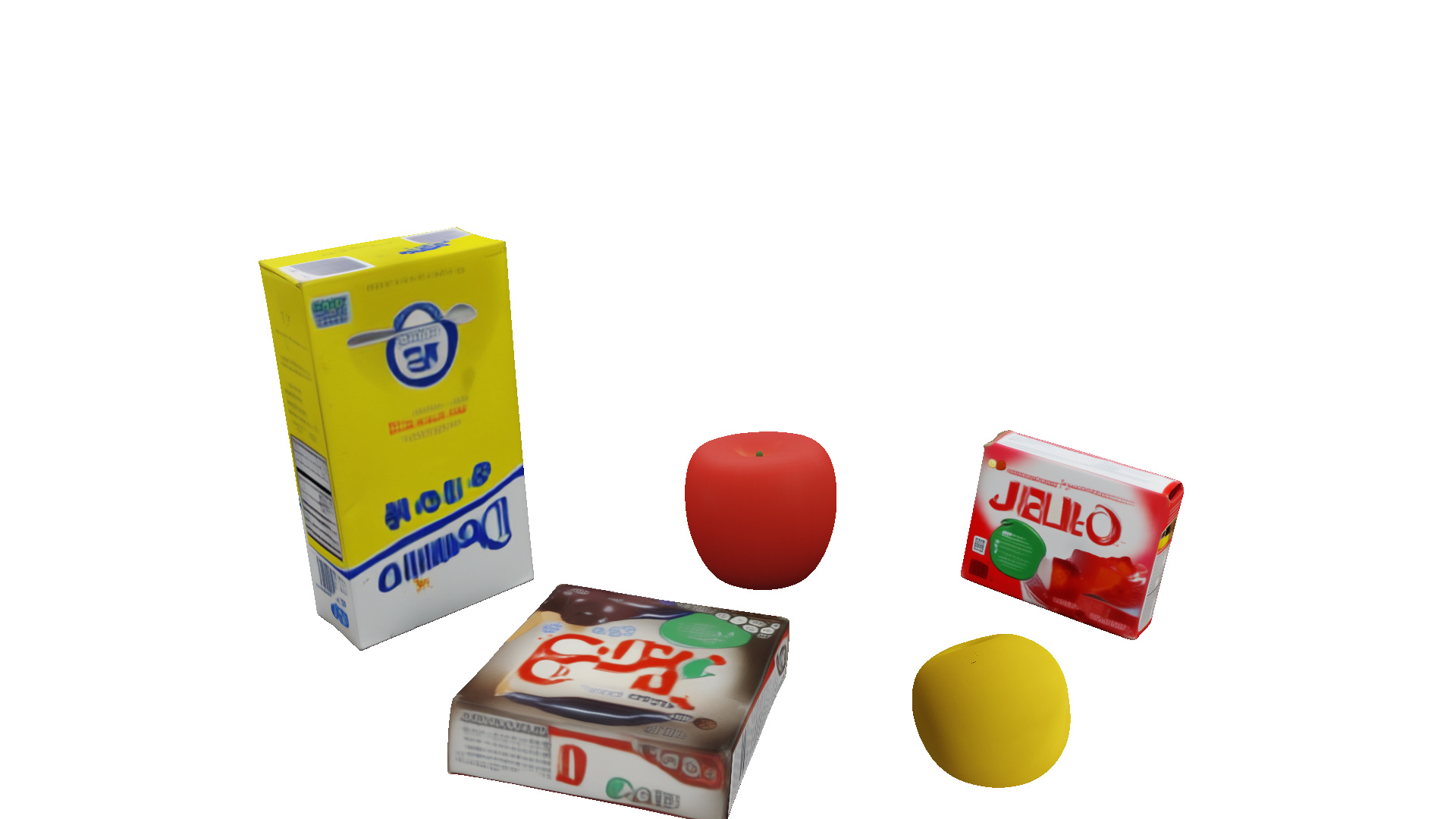} &
\includegraphics[width=0.165\textwidth]{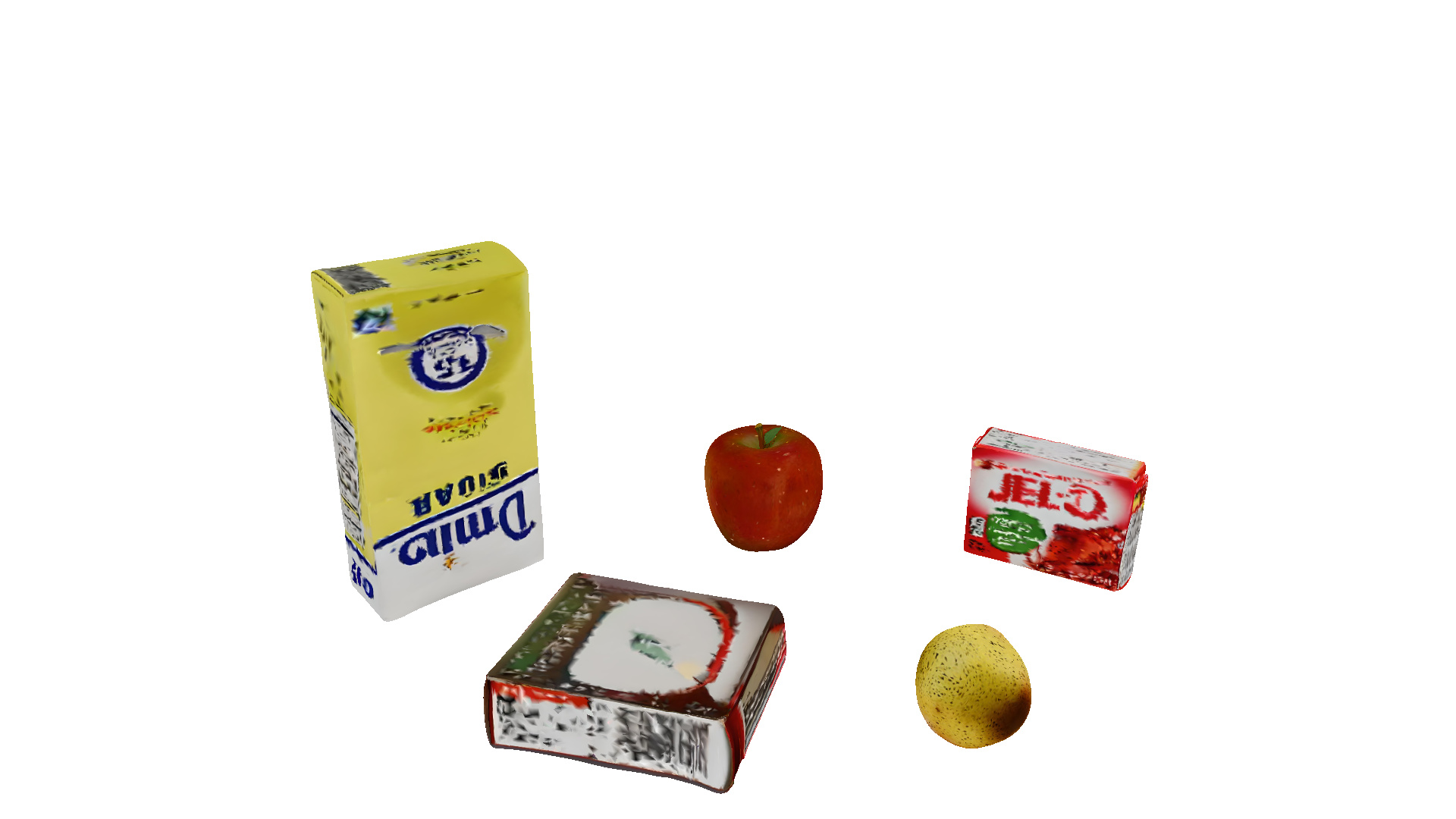} &
\includegraphics[width=0.165\textwidth]{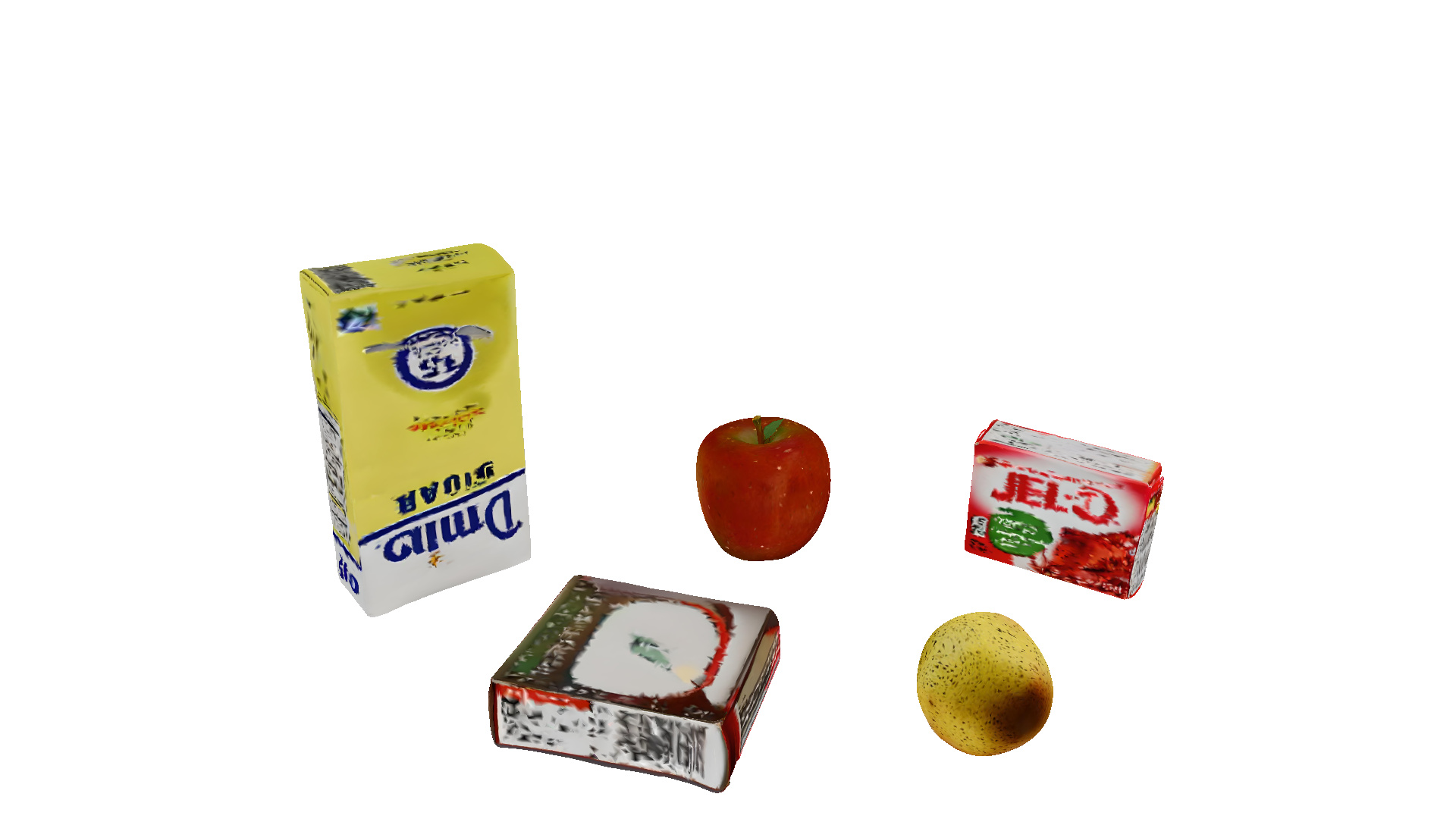} \\

Kitchen 1 & Easy &
\includegraphics[width=0.165\textwidth]{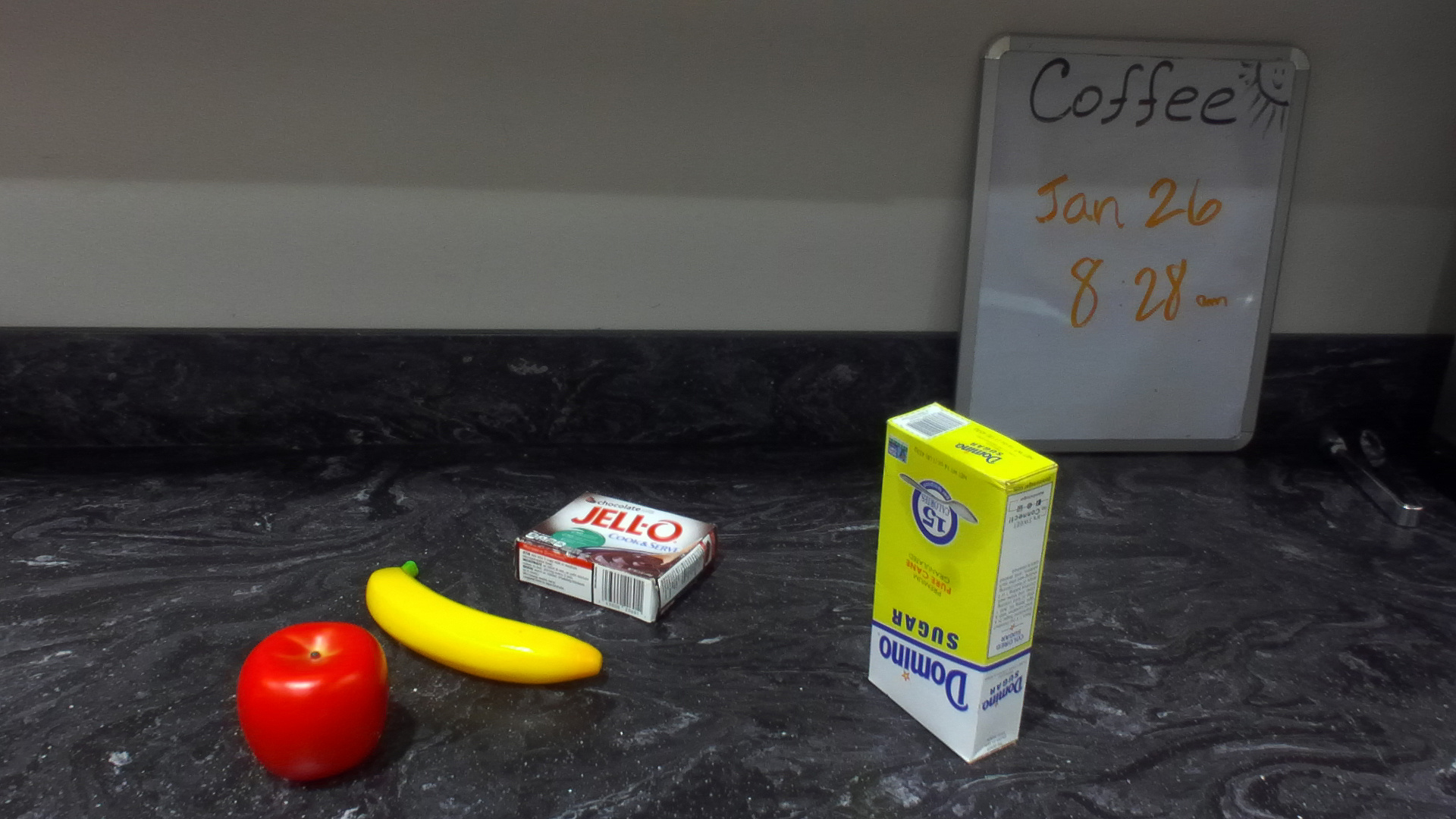} &
\includegraphics[width=0.165\textwidth]{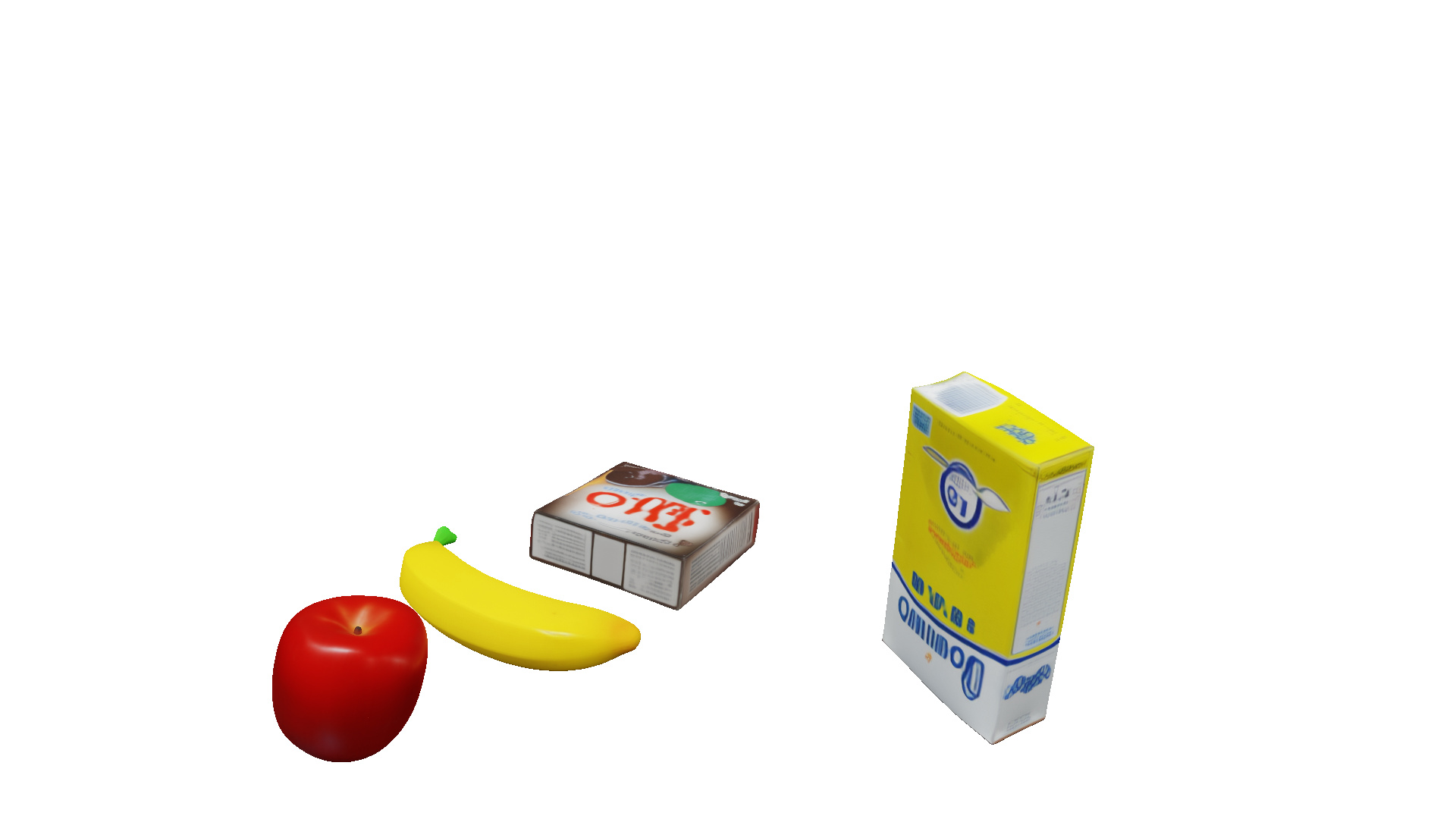} &
\includegraphics[width=0.165\textwidth]{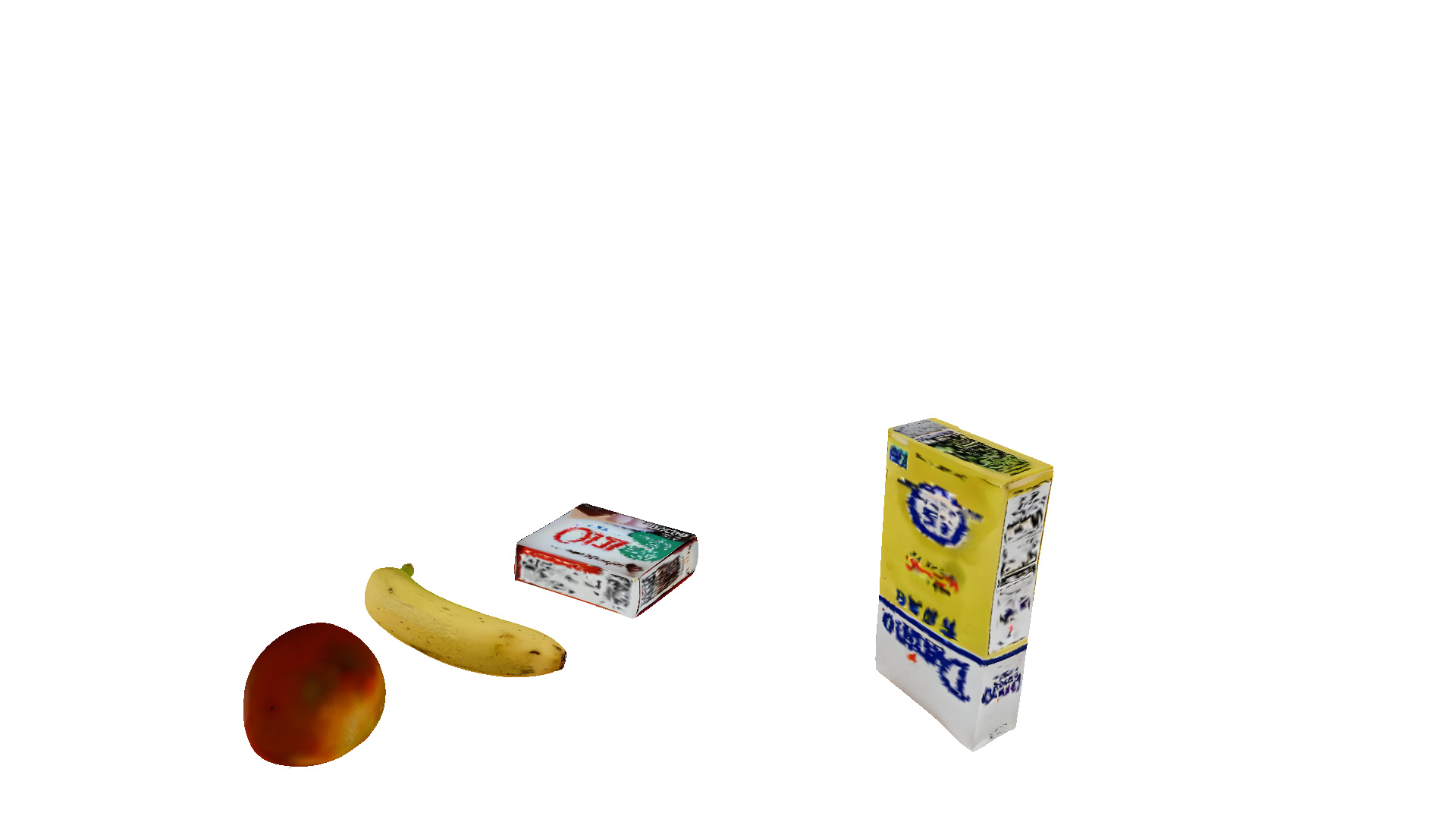} &
\includegraphics[width=0.165\textwidth]{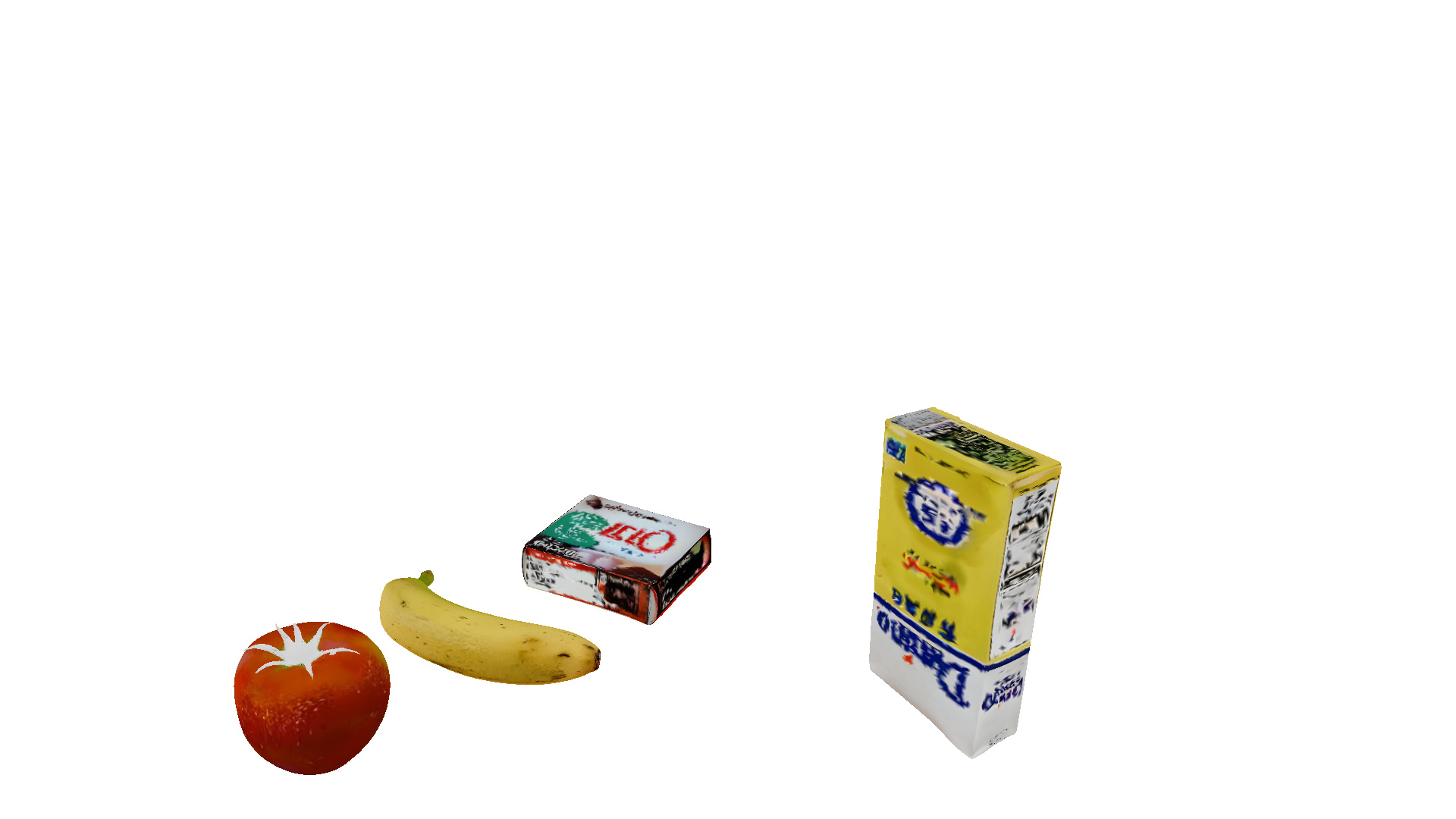} \\

\midrule

Desk 4 & Medium &
\includegraphics[width=0.165\textwidth]{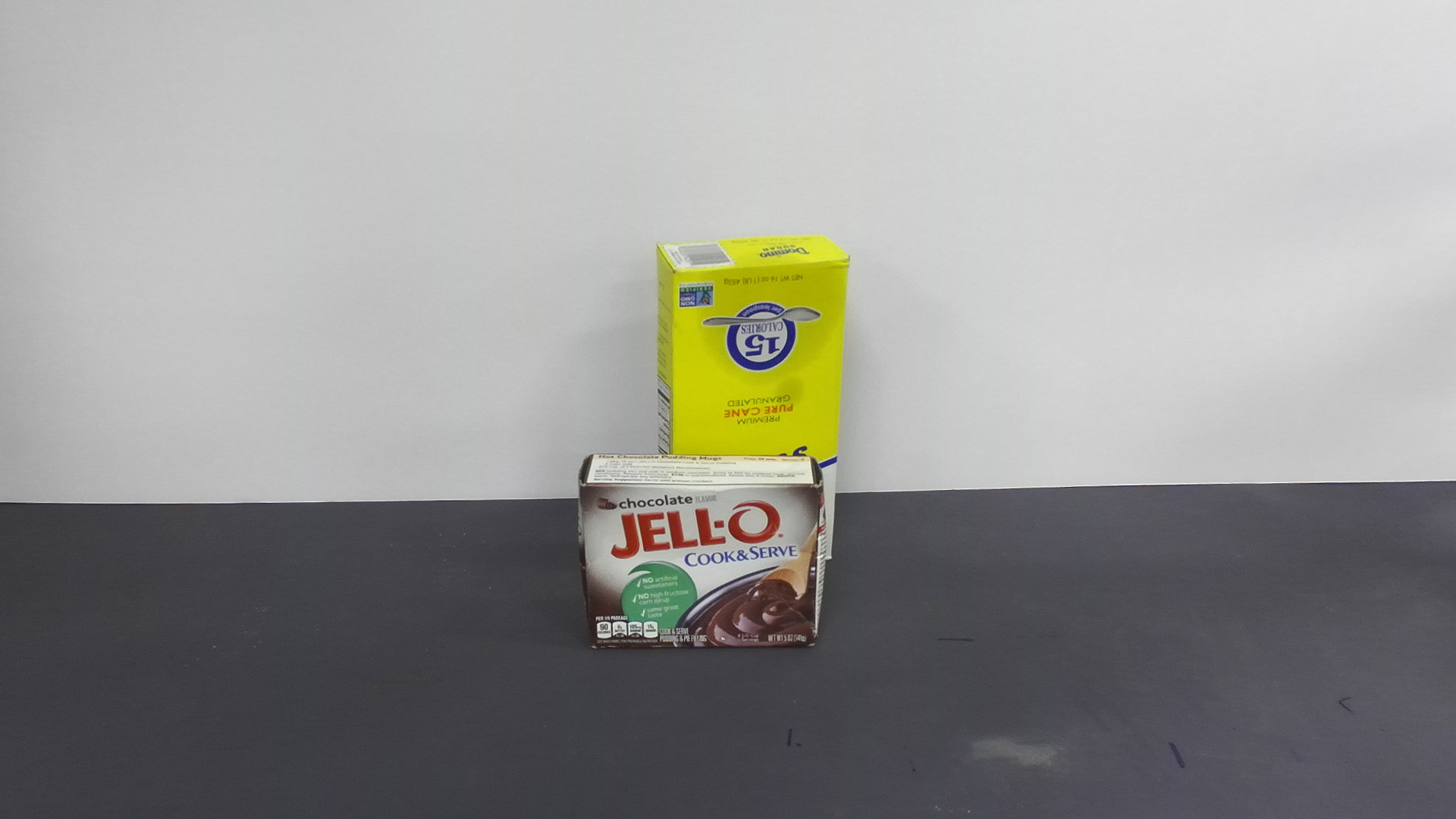} &
\includegraphics[width=0.165\textwidth]{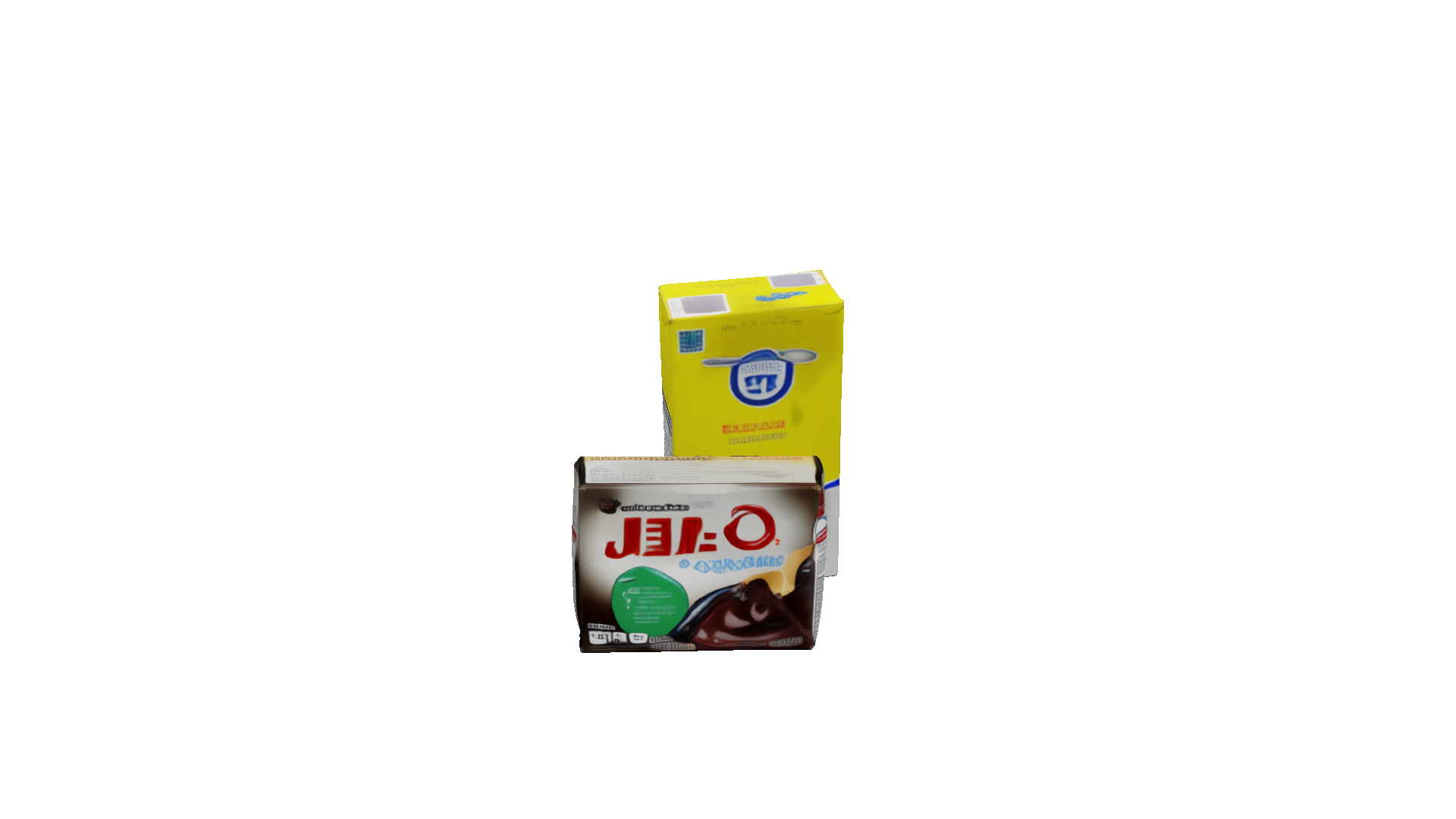} &
\includegraphics[width=0.165\textwidth]{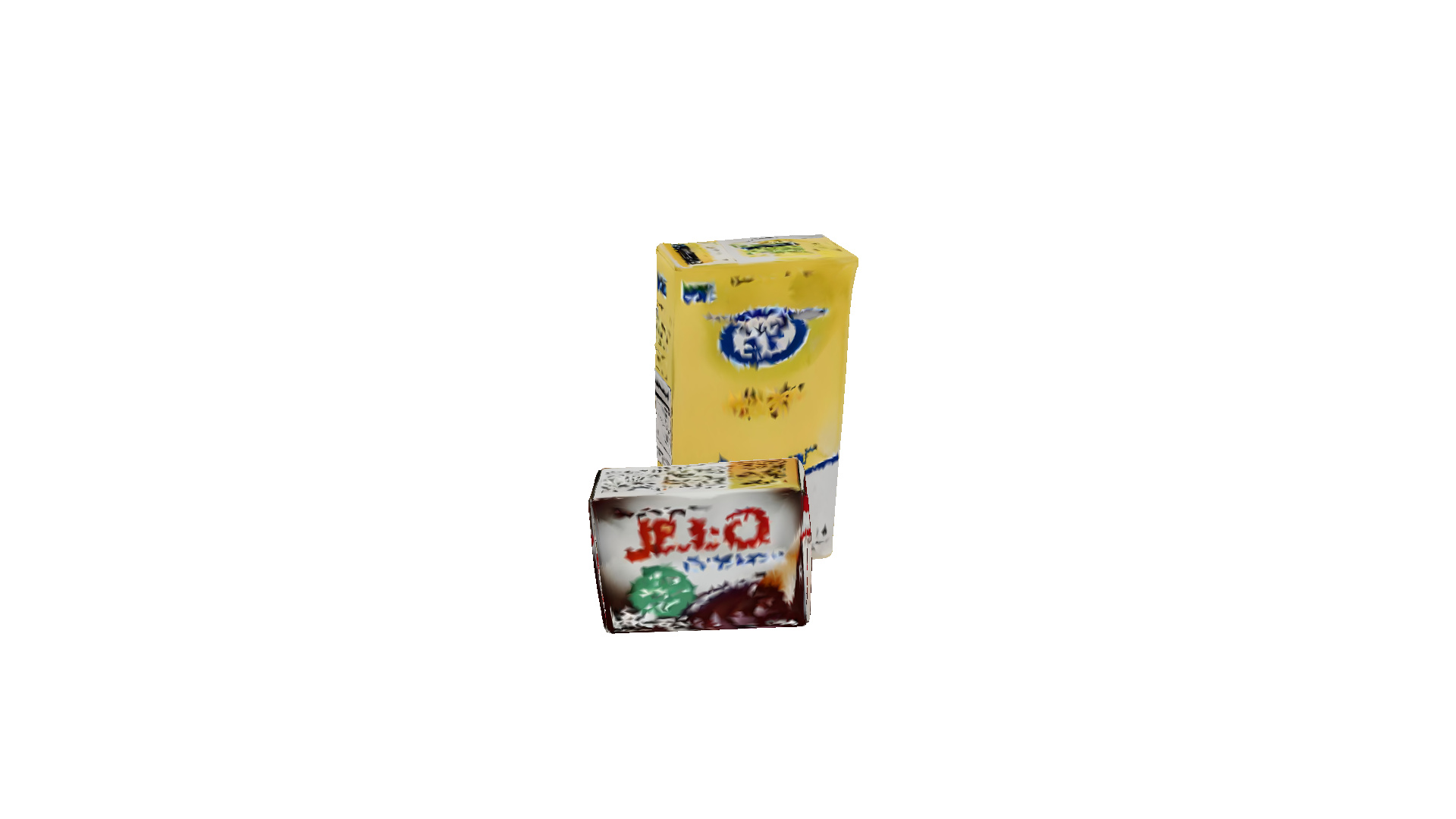} &
\includegraphics[width=0.165\textwidth]{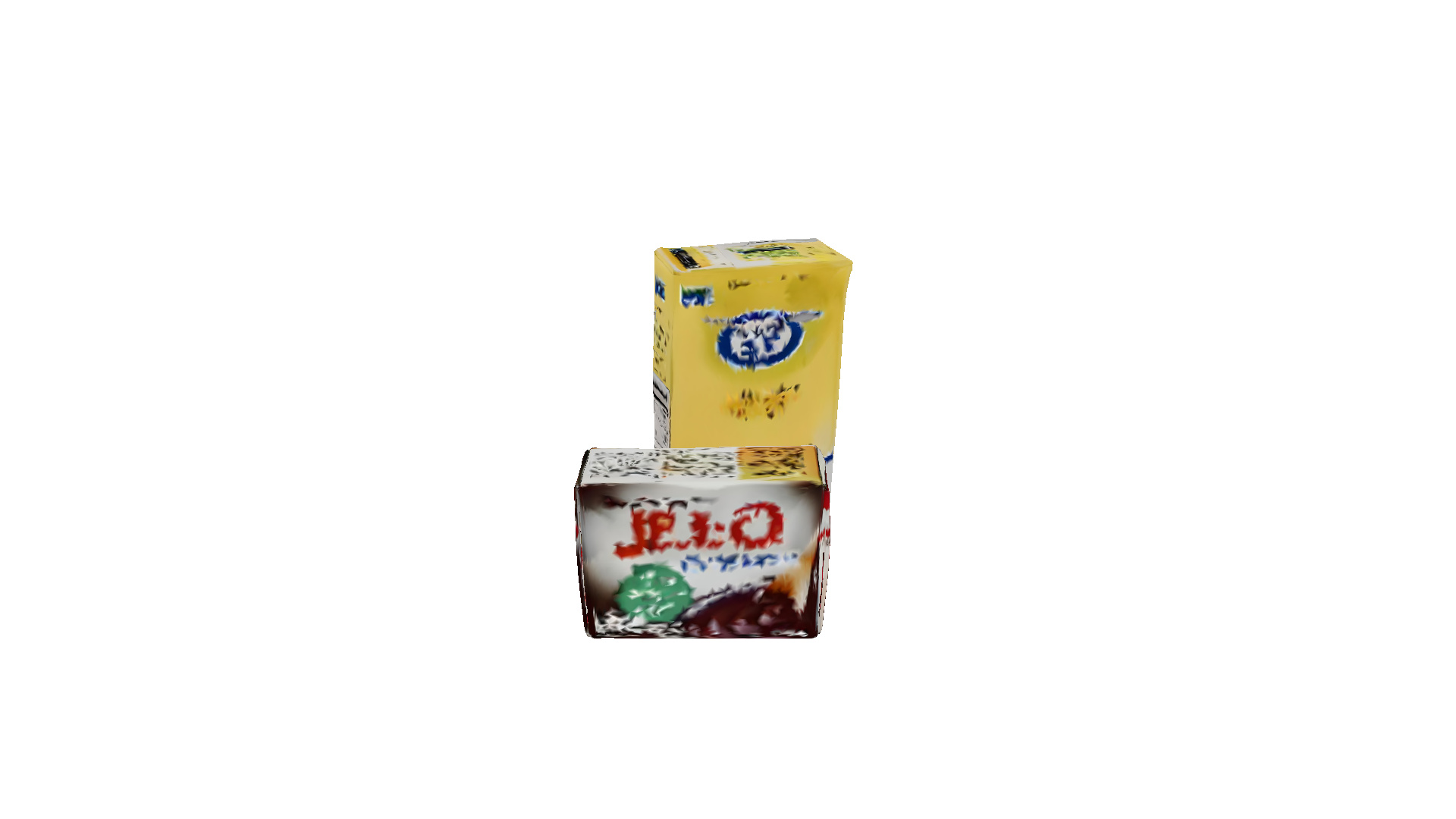} \\

Desk 6 & Medium &
\includegraphics[width=0.165\textwidth]{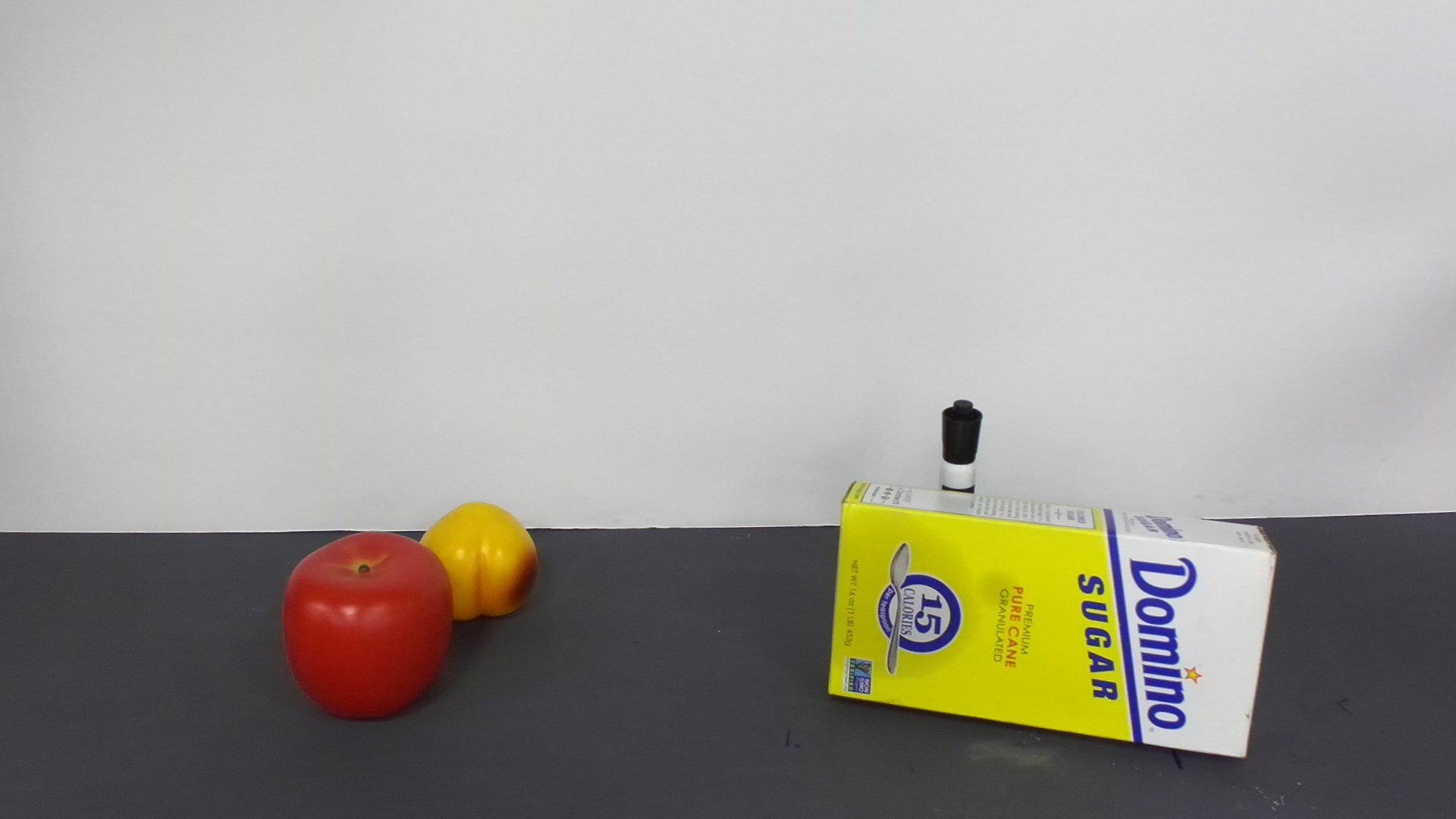} &
\includegraphics[width=0.165\textwidth]{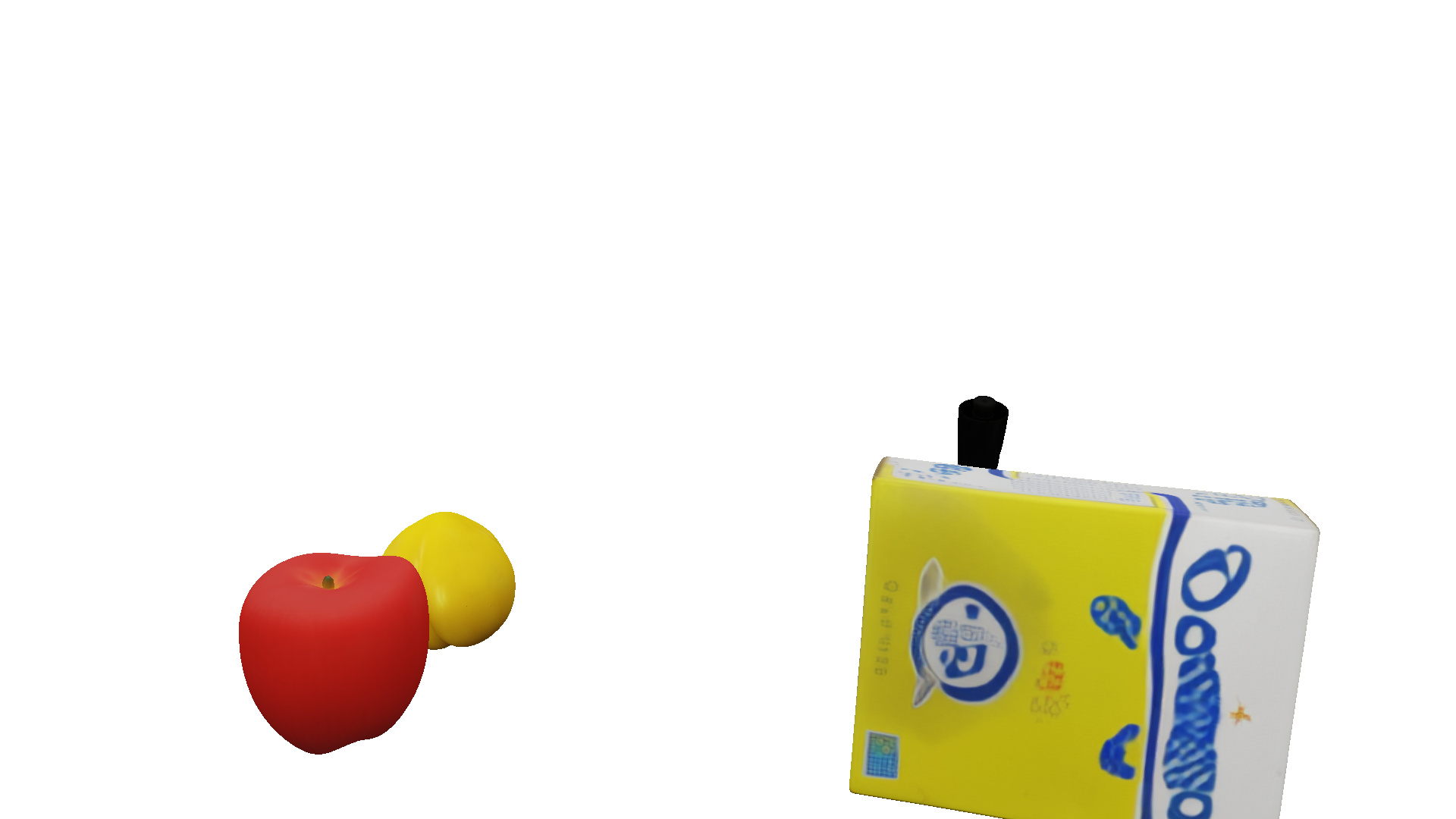} &
\includegraphics[width=0.165\textwidth]{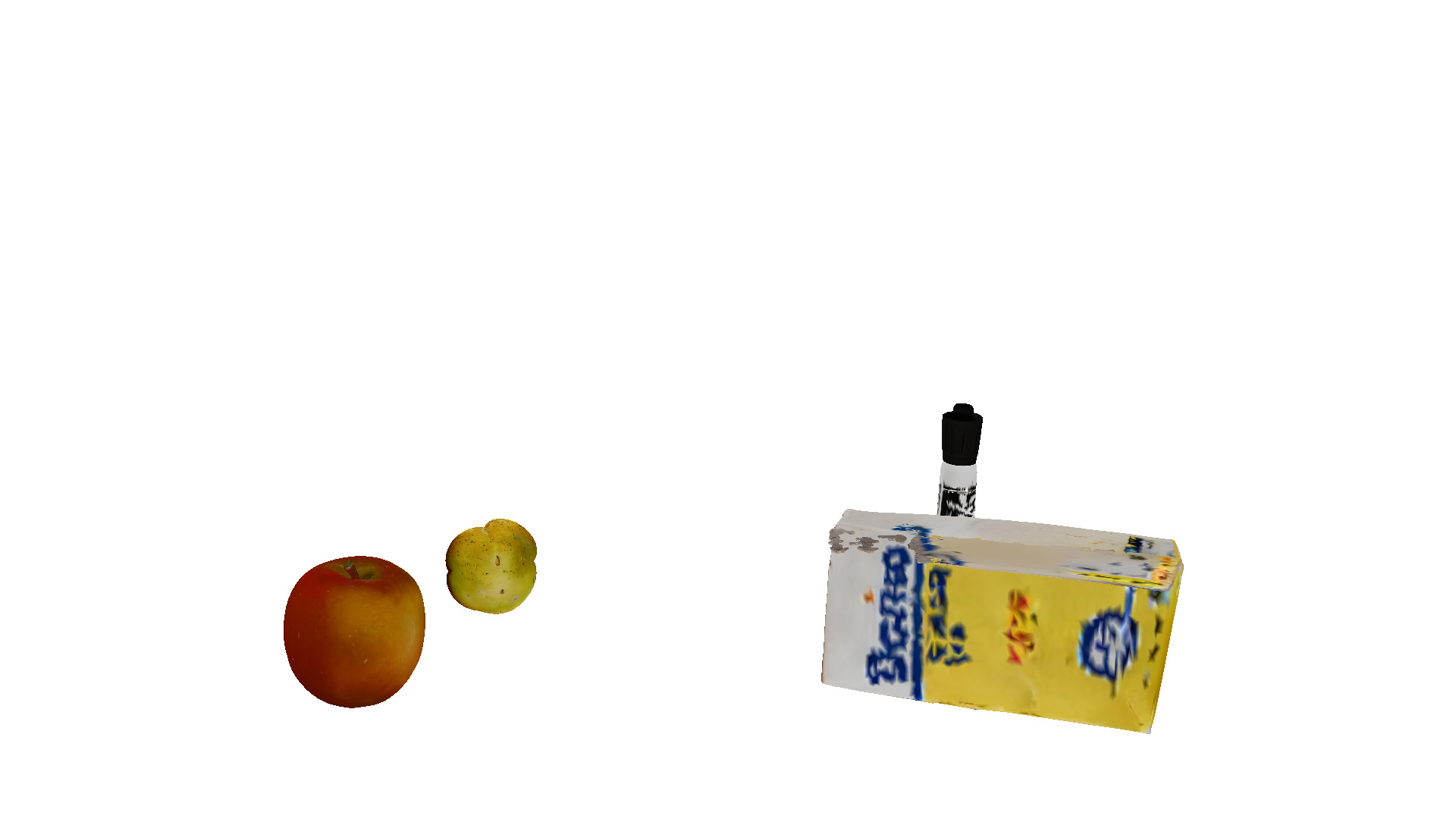} &
\includegraphics[width=0.165\textwidth]{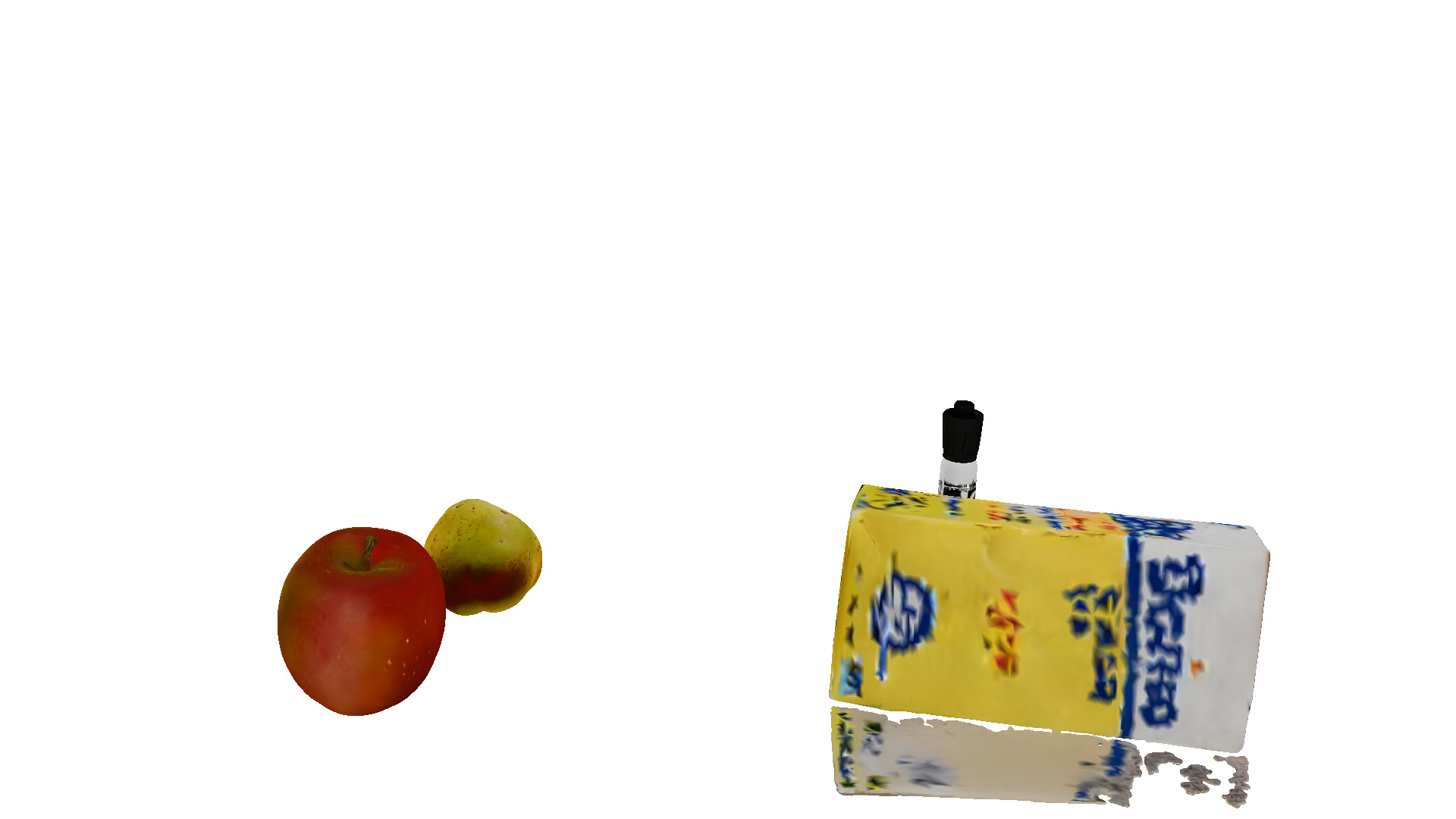} \\

Desk 8 & Medium &
\includegraphics[width=0.165\textwidth]{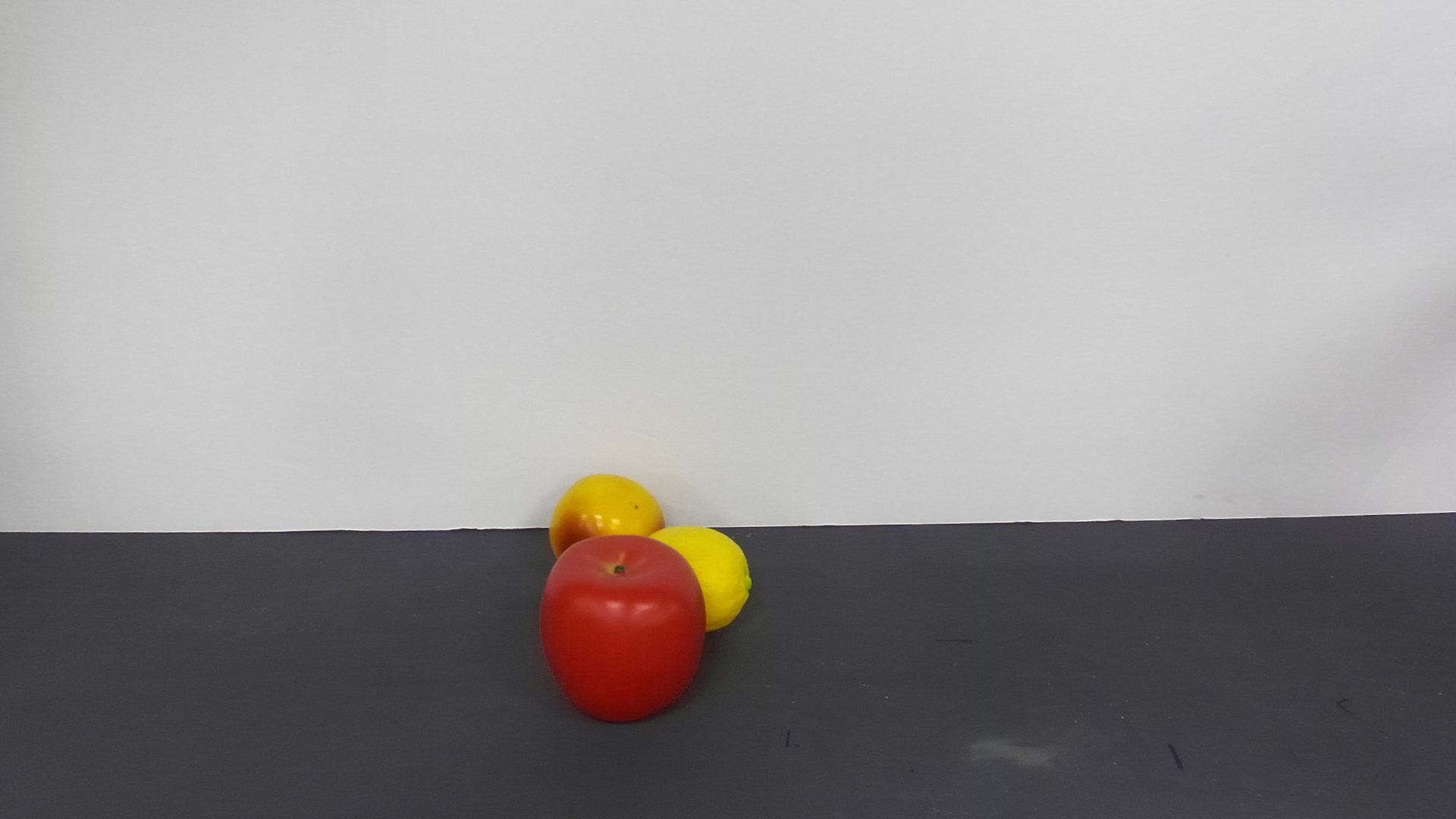} &
\includegraphics[width=0.165\textwidth]{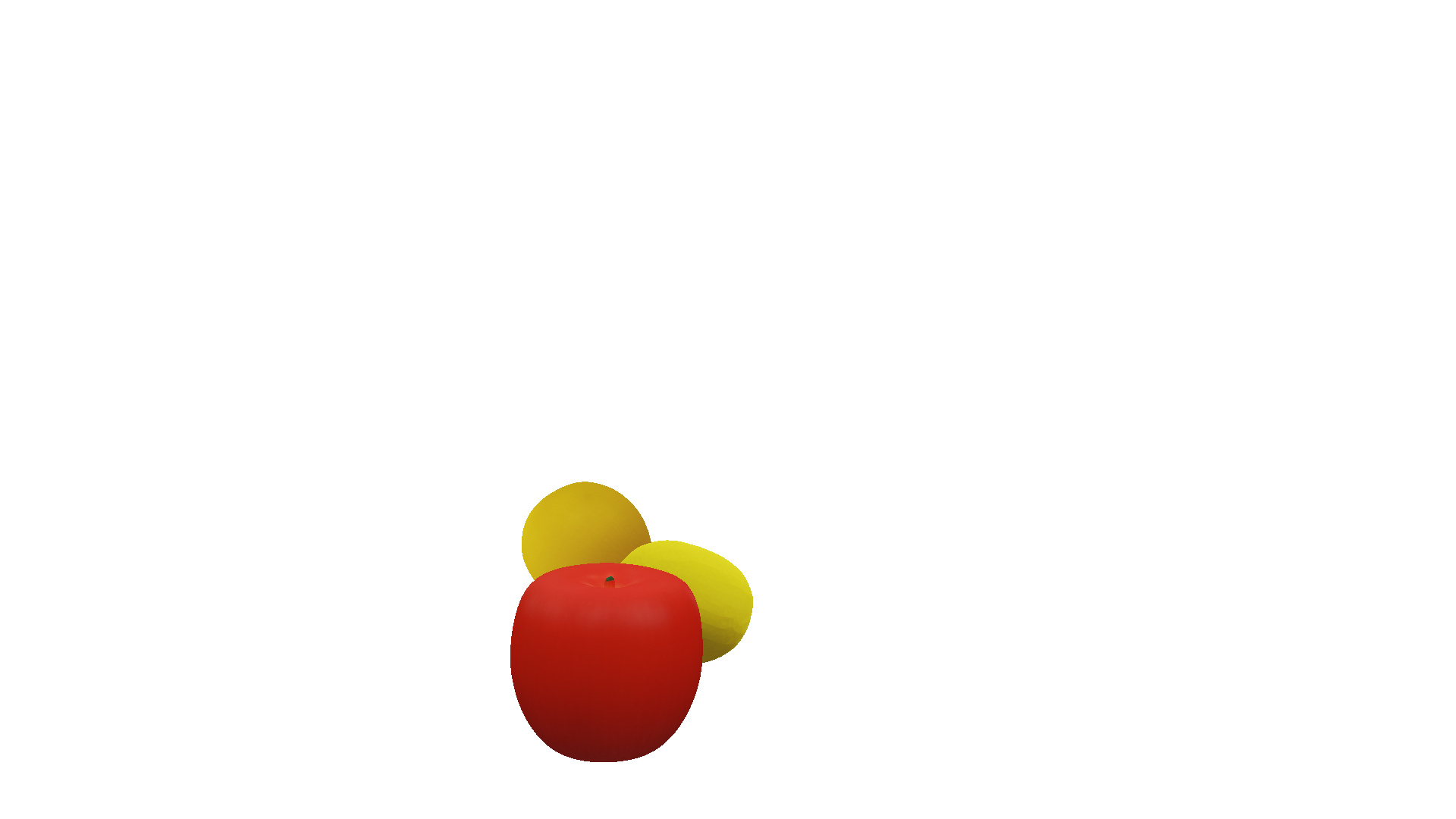} &
\includegraphics[width=0.165\textwidth]{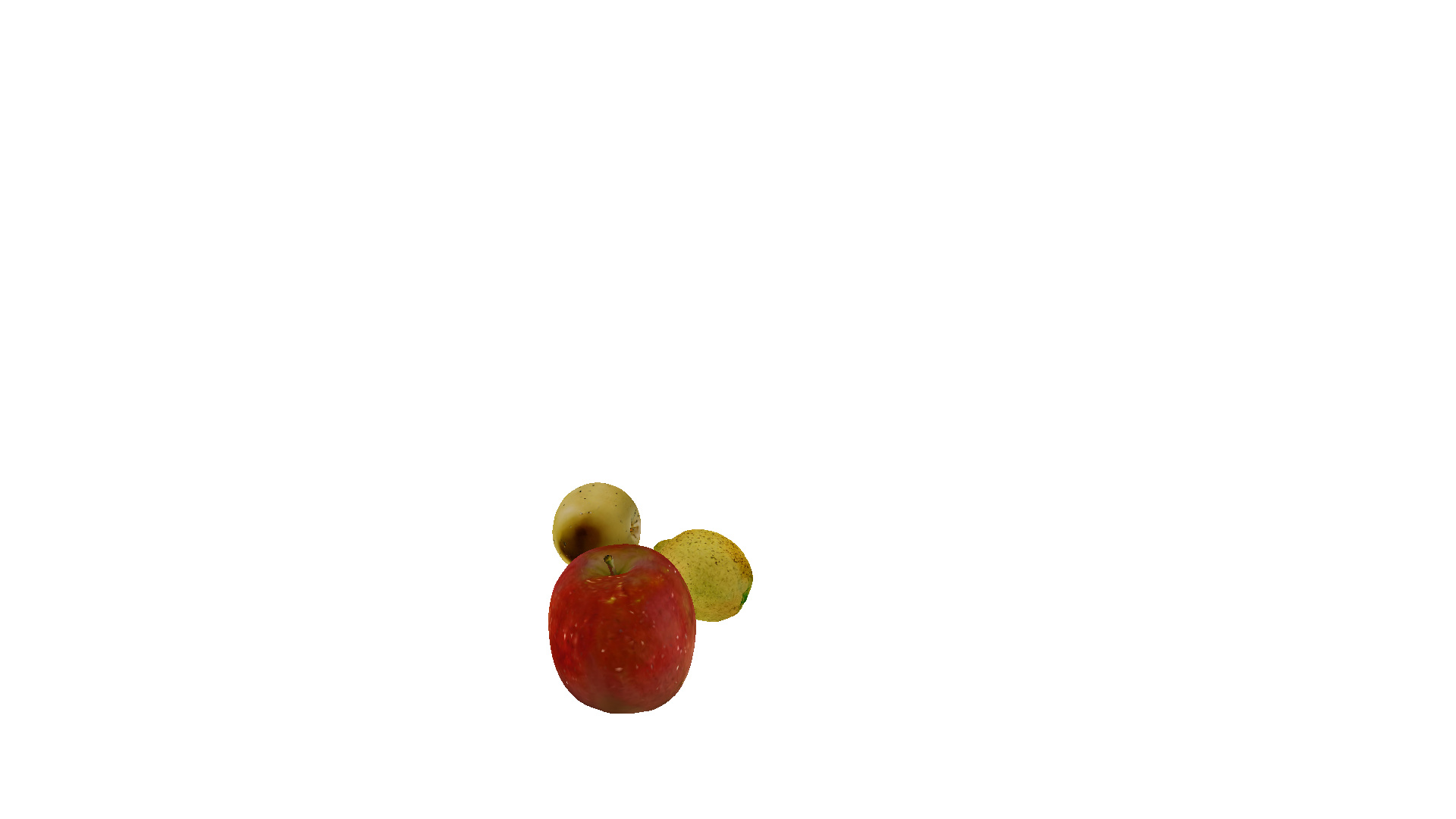} &
\includegraphics[width=0.165\textwidth]{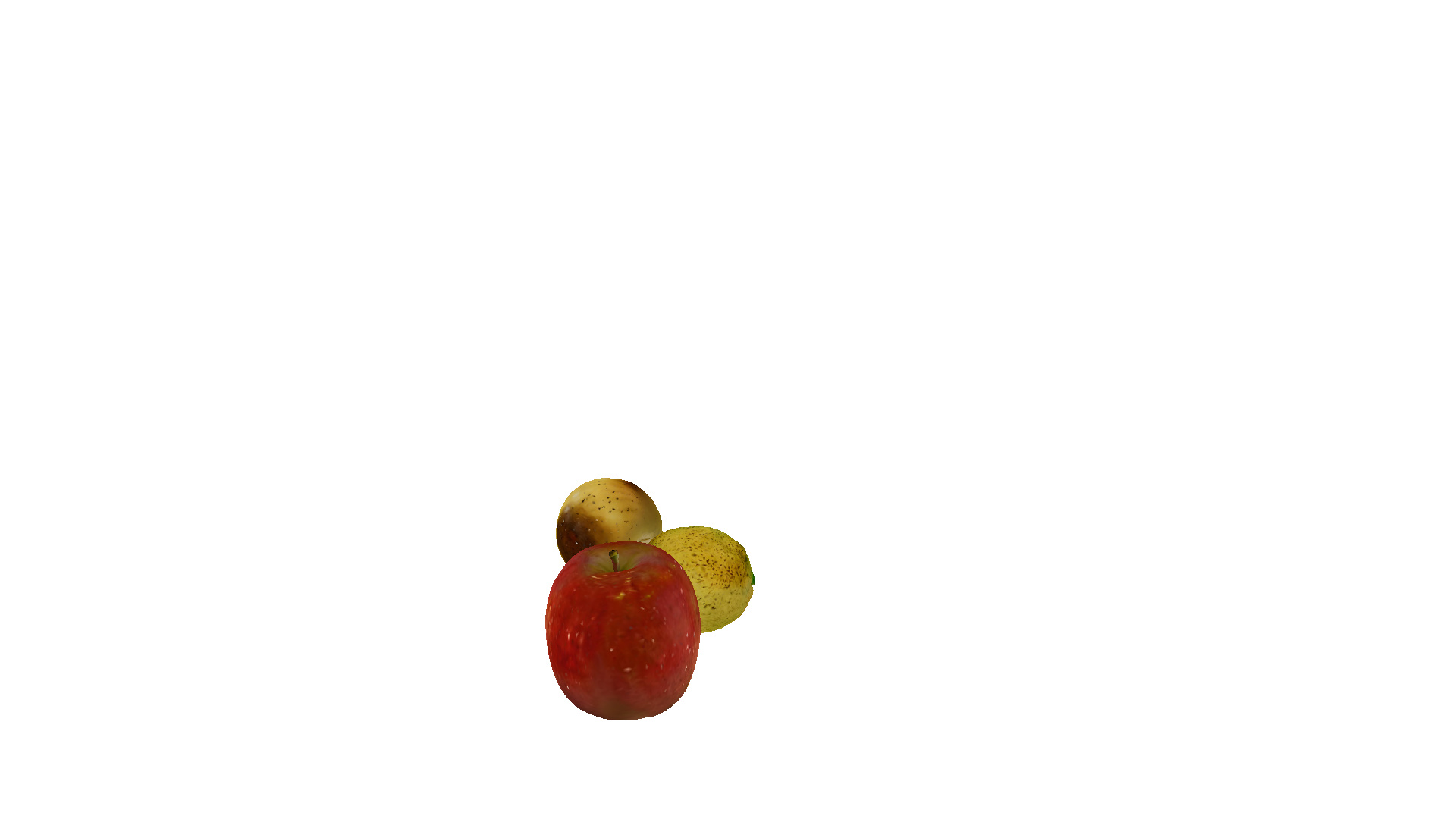} \\

Desk 9 & Medium &
\includegraphics[width=0.165\textwidth]{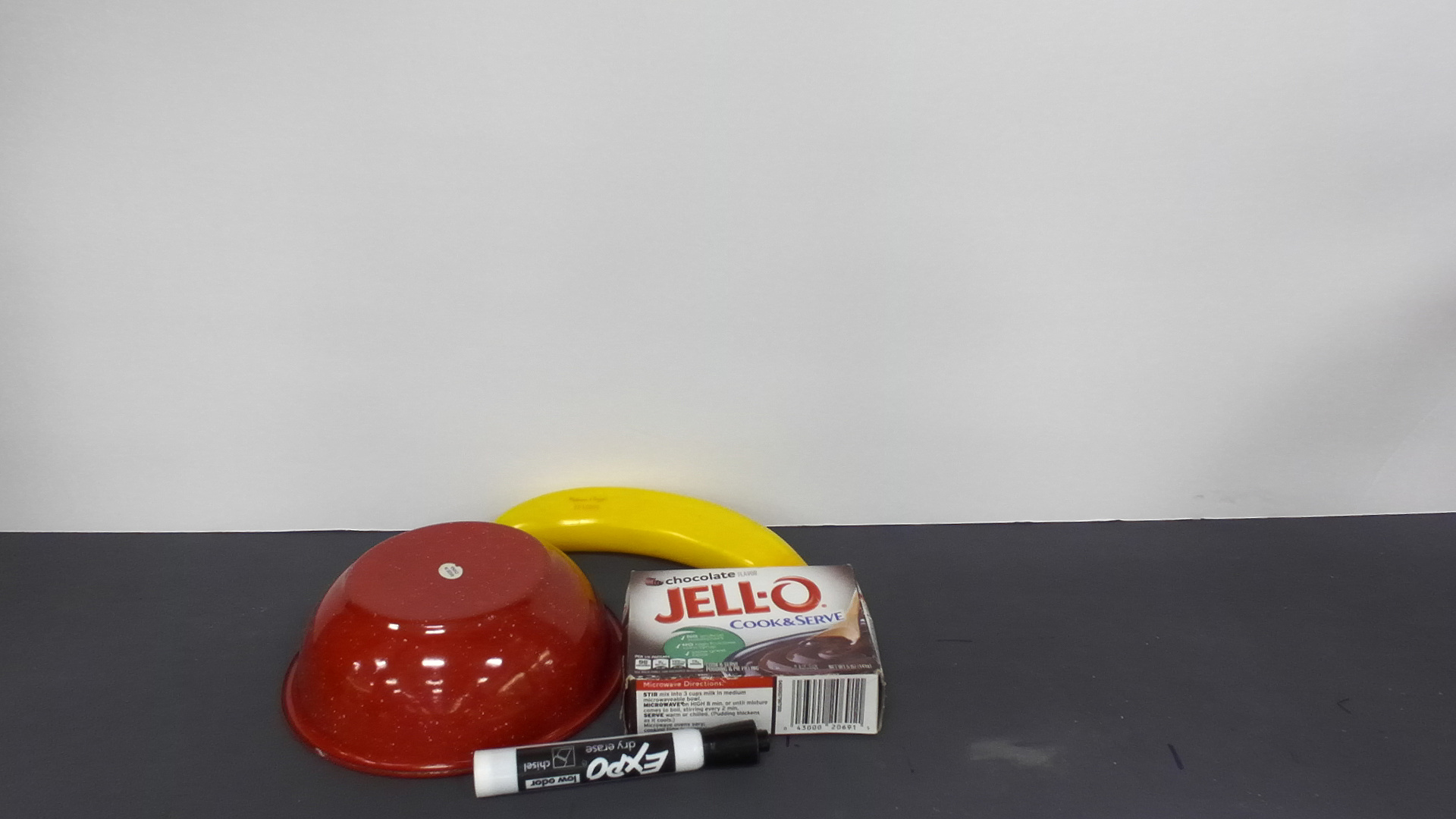} &
\includegraphics[width=0.165\textwidth]{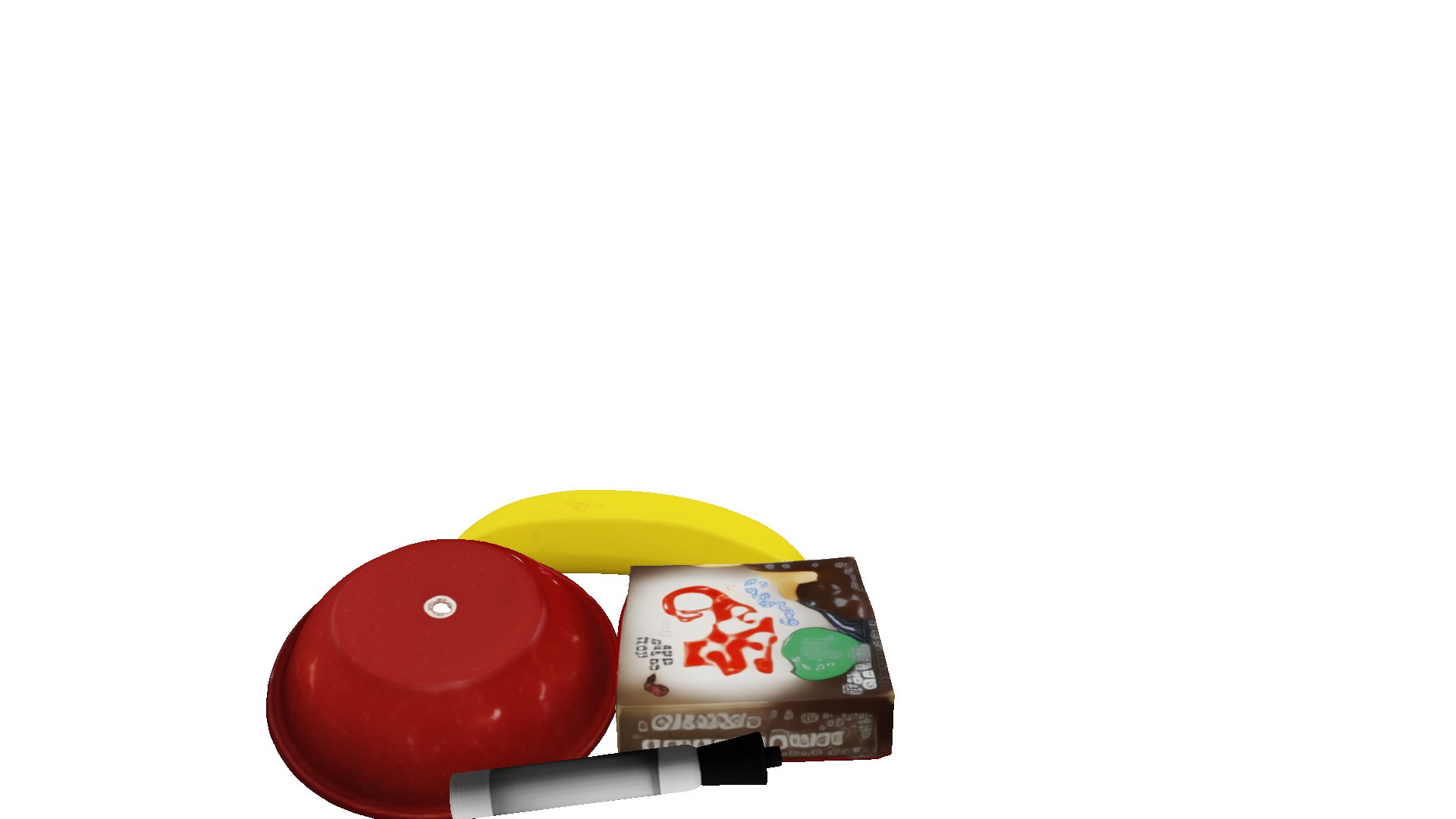} &
\includegraphics[width=0.165\textwidth]{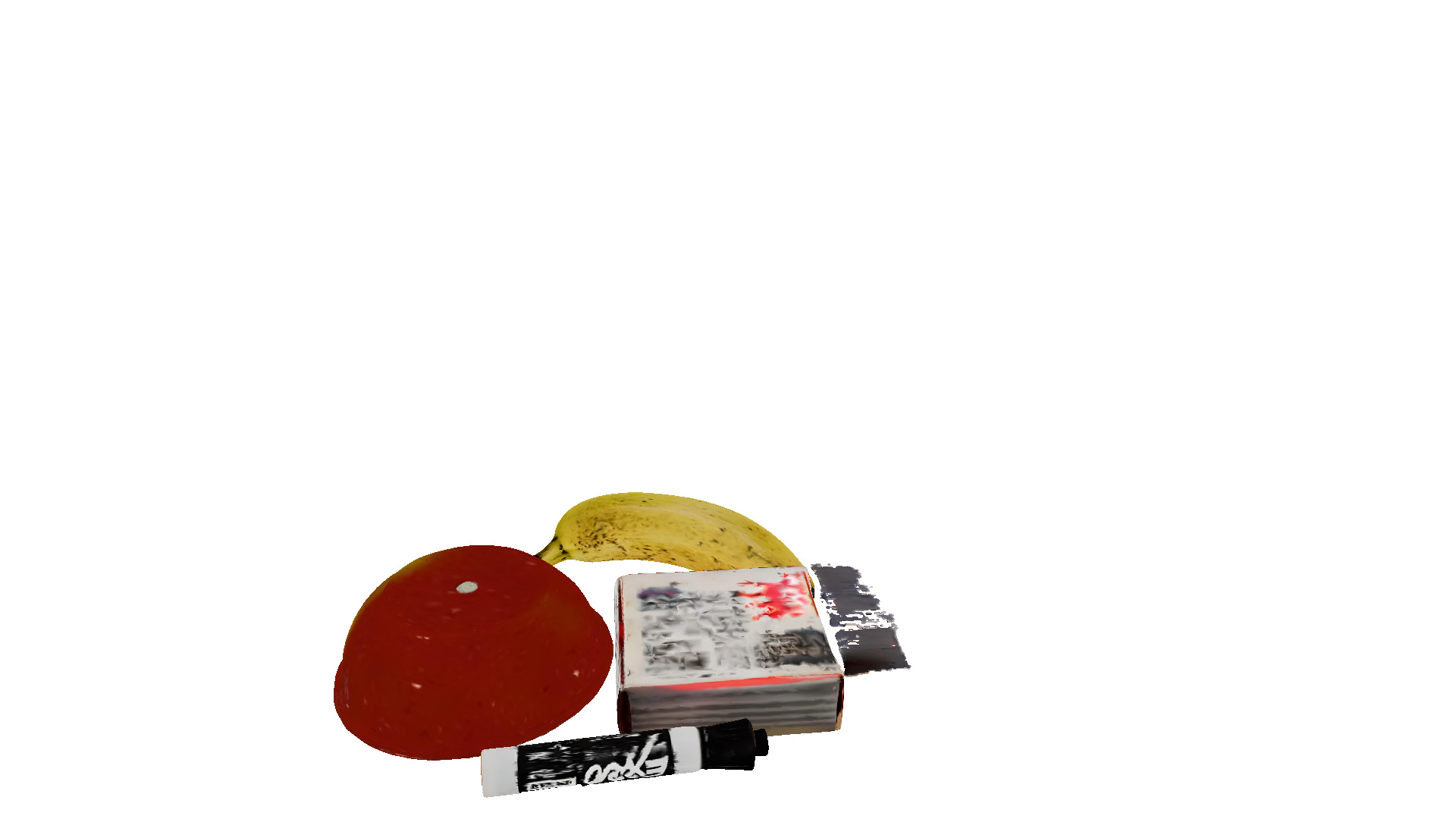} &
\includegraphics[width=0.165\textwidth]{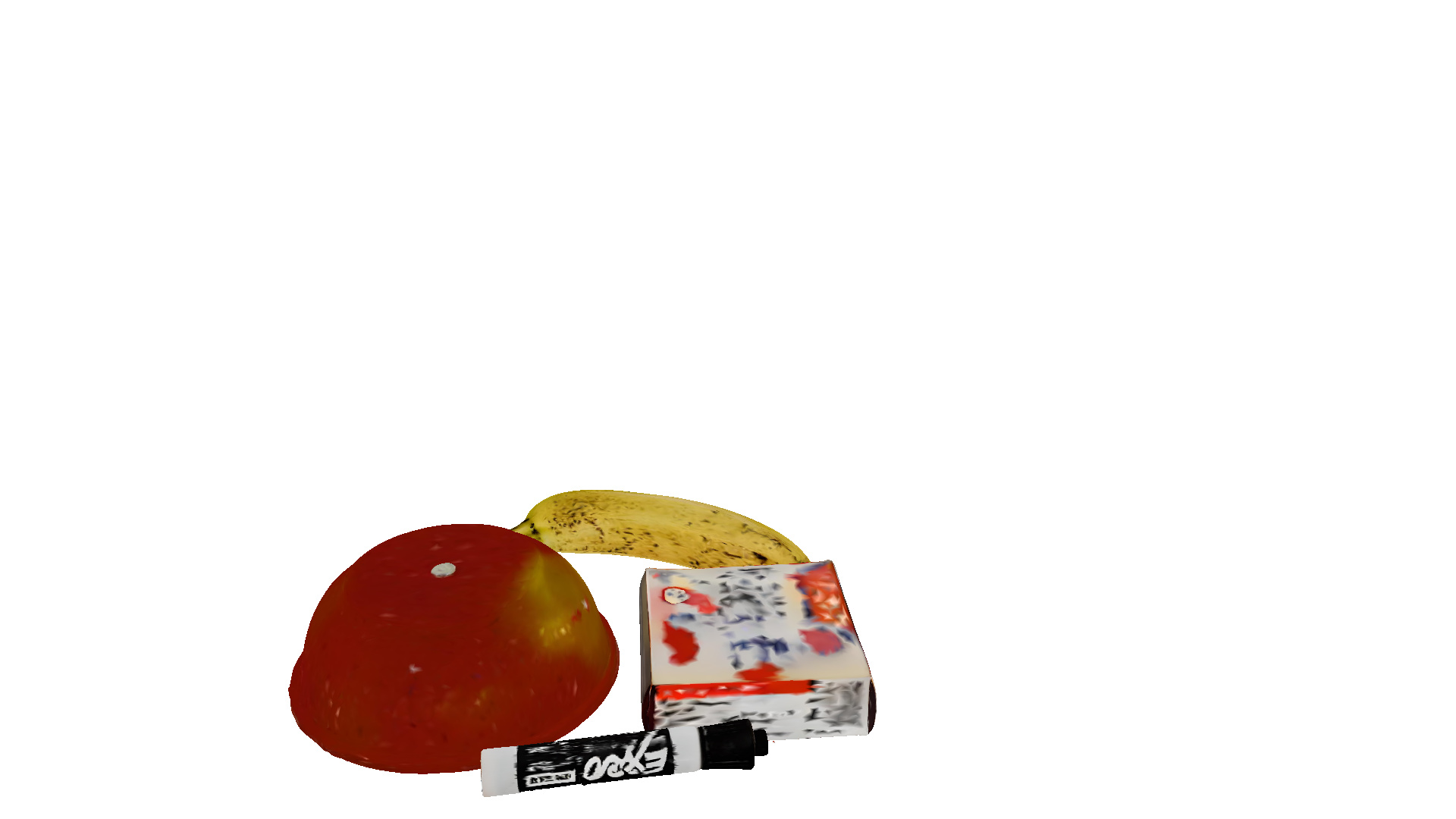} \\

\midrule

Desk 5 & Hard &
\includegraphics[width=0.165\textwidth]{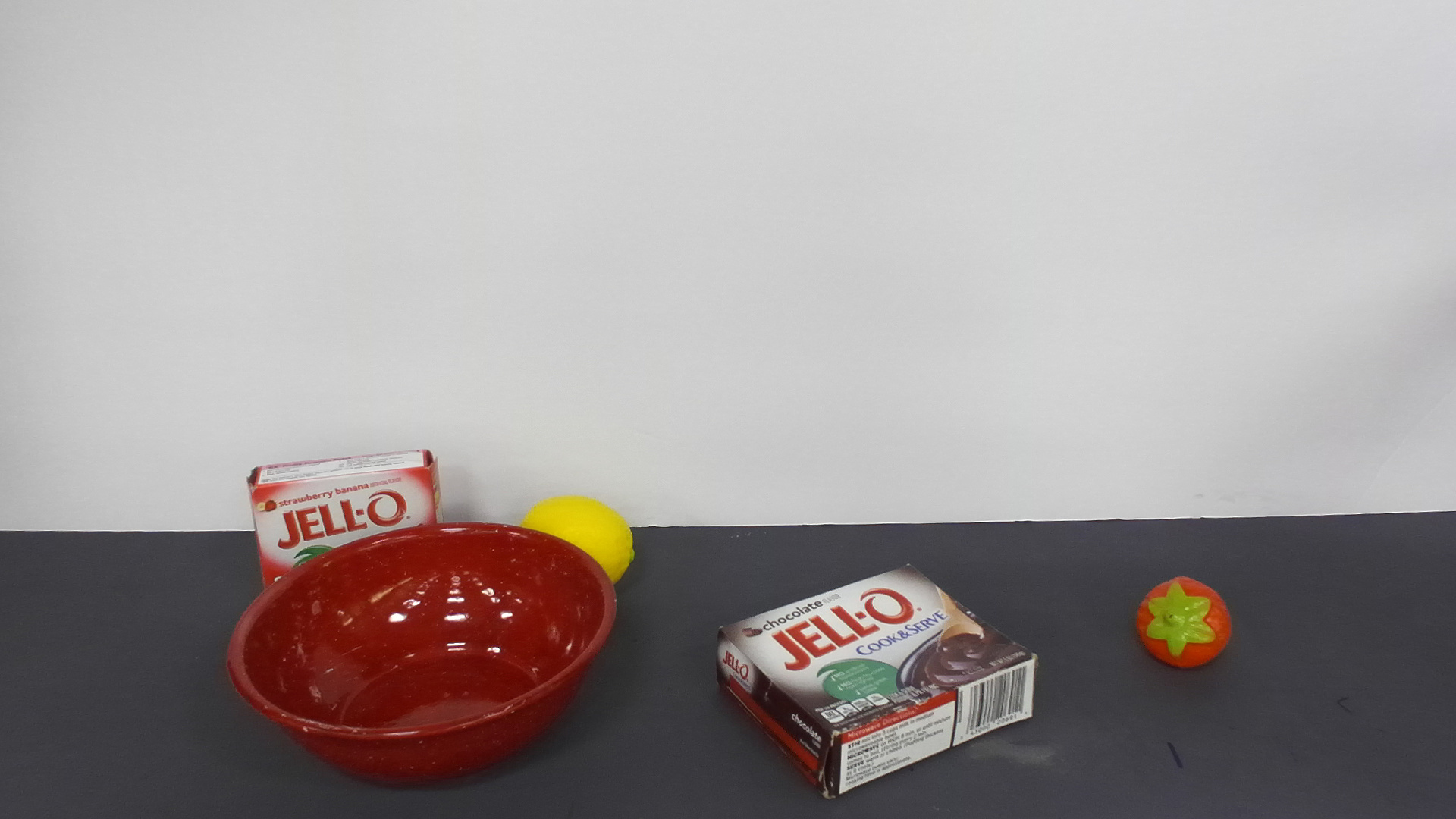} &
\includegraphics[width=0.165\textwidth]{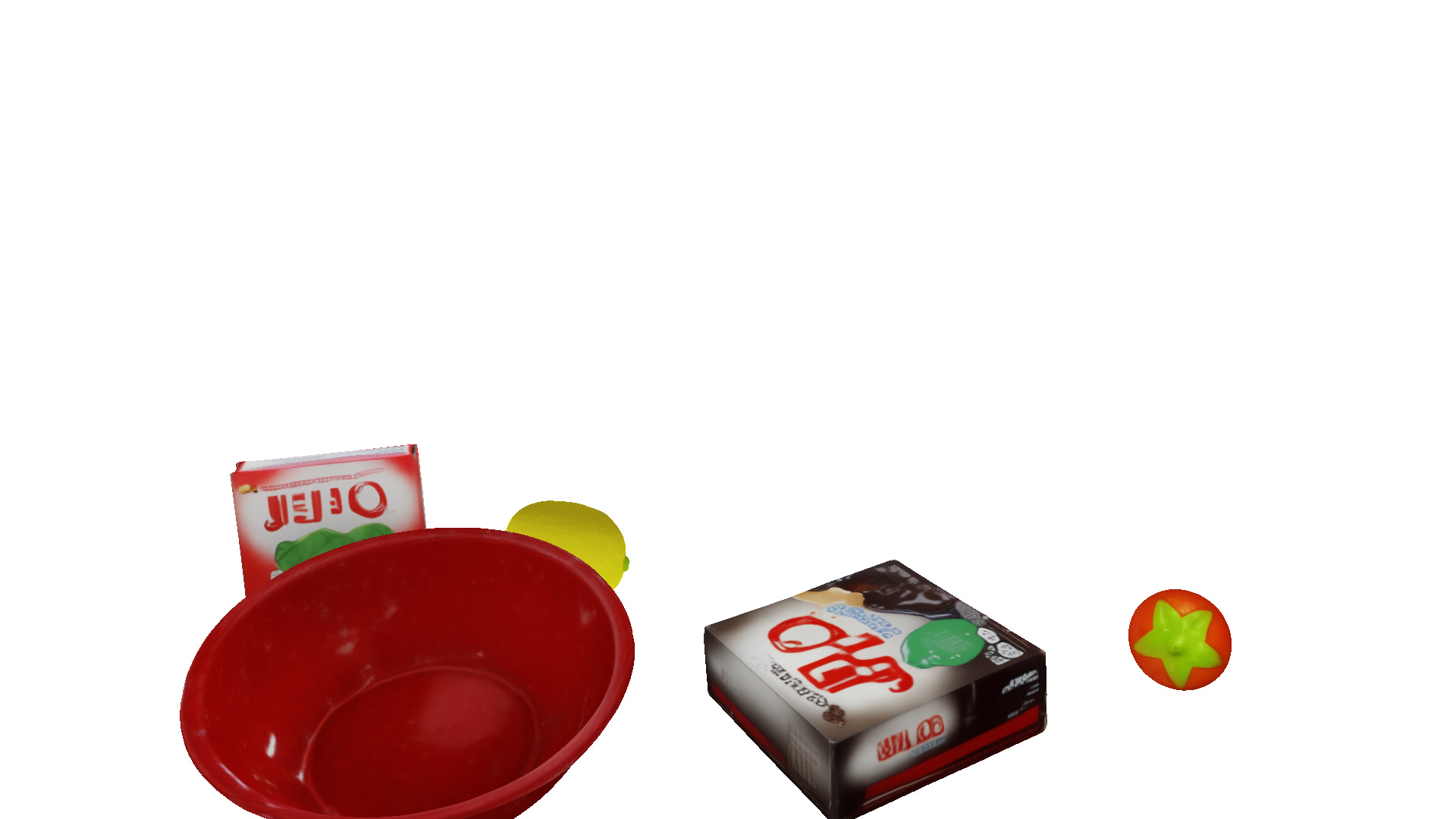} &
\includegraphics[width=0.165\textwidth]{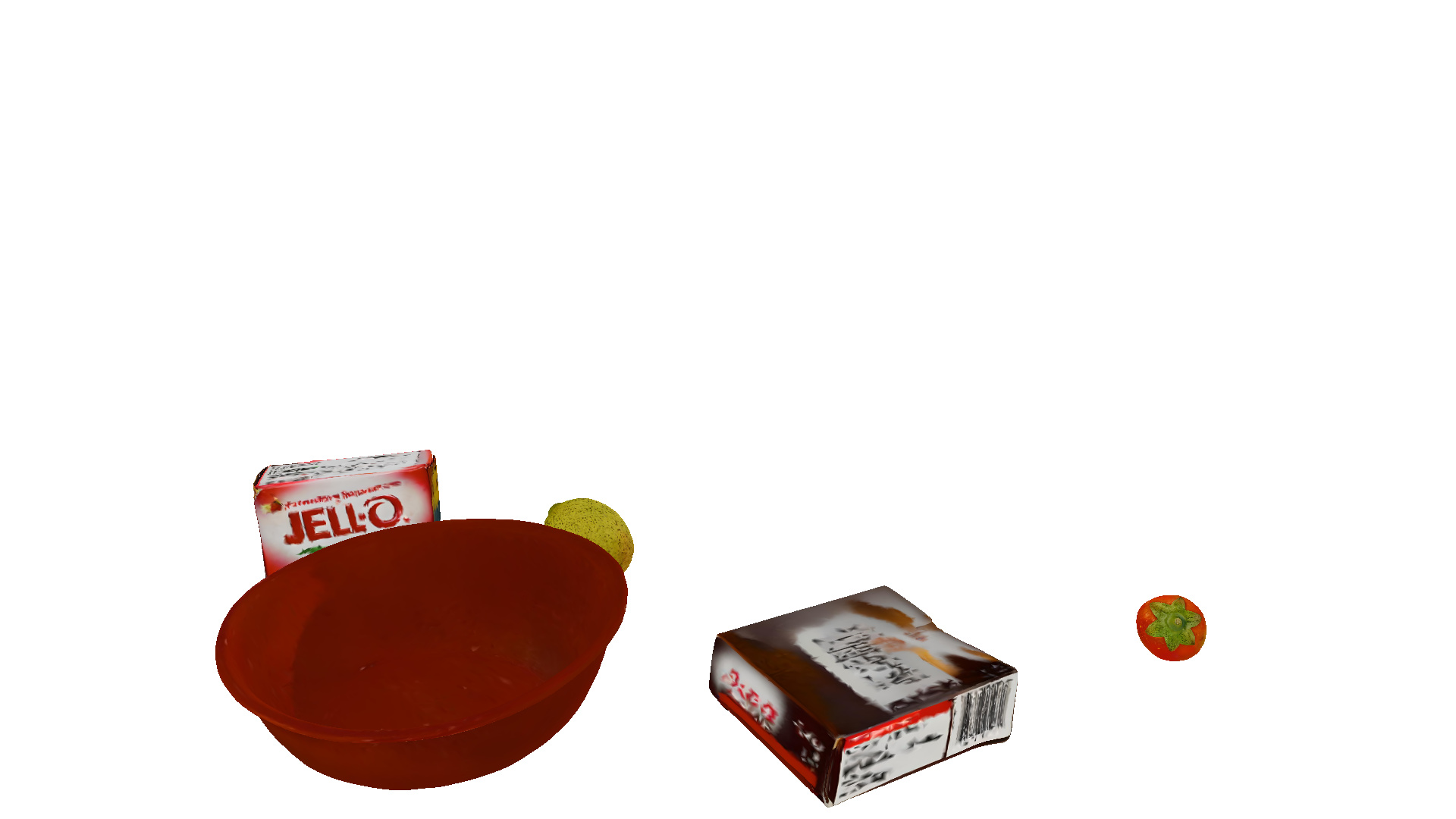} &
\includegraphics[width=0.165\textwidth]{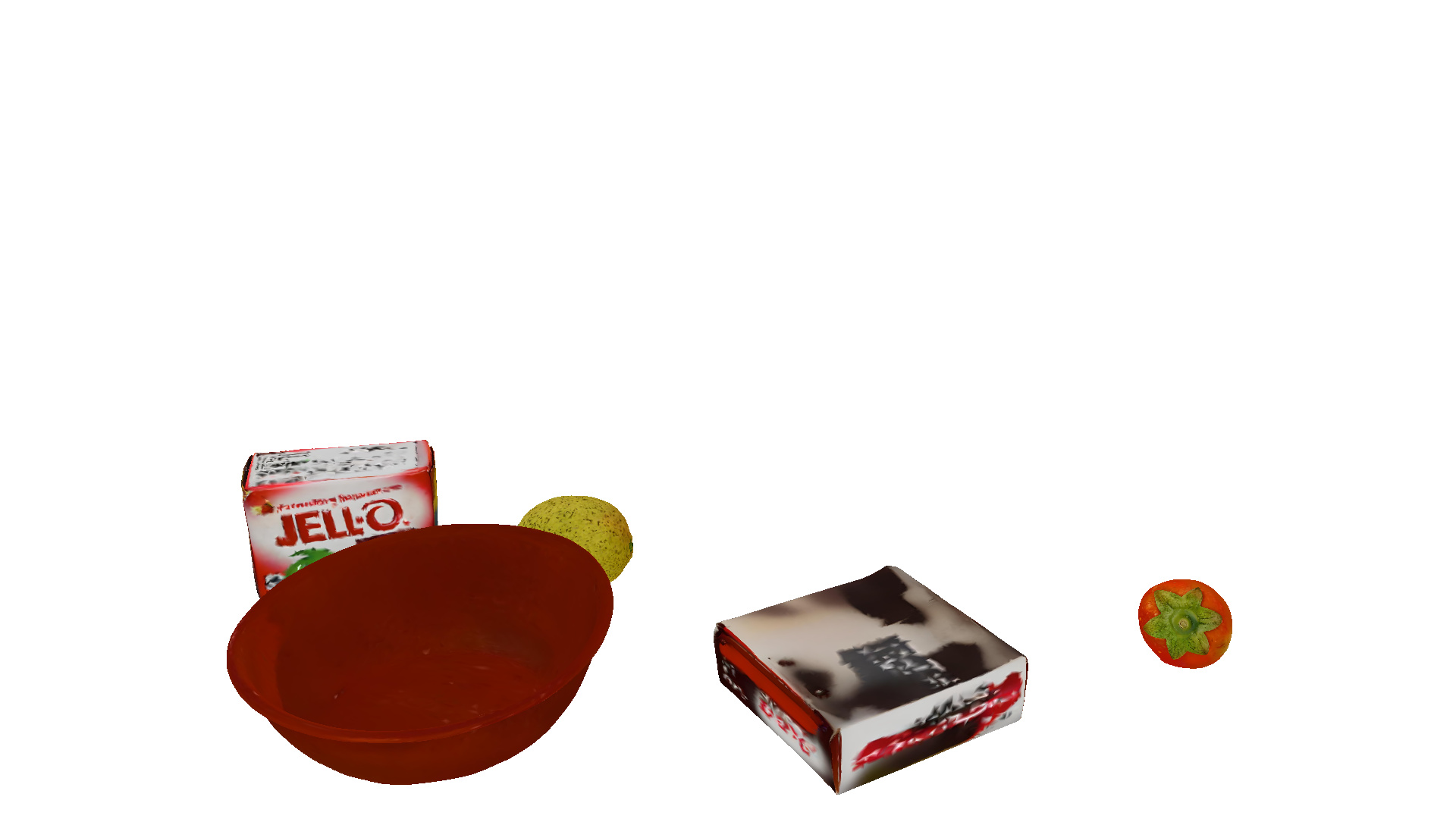} \\

Desk 7 & Hard &
\includegraphics[width=0.165\textwidth]{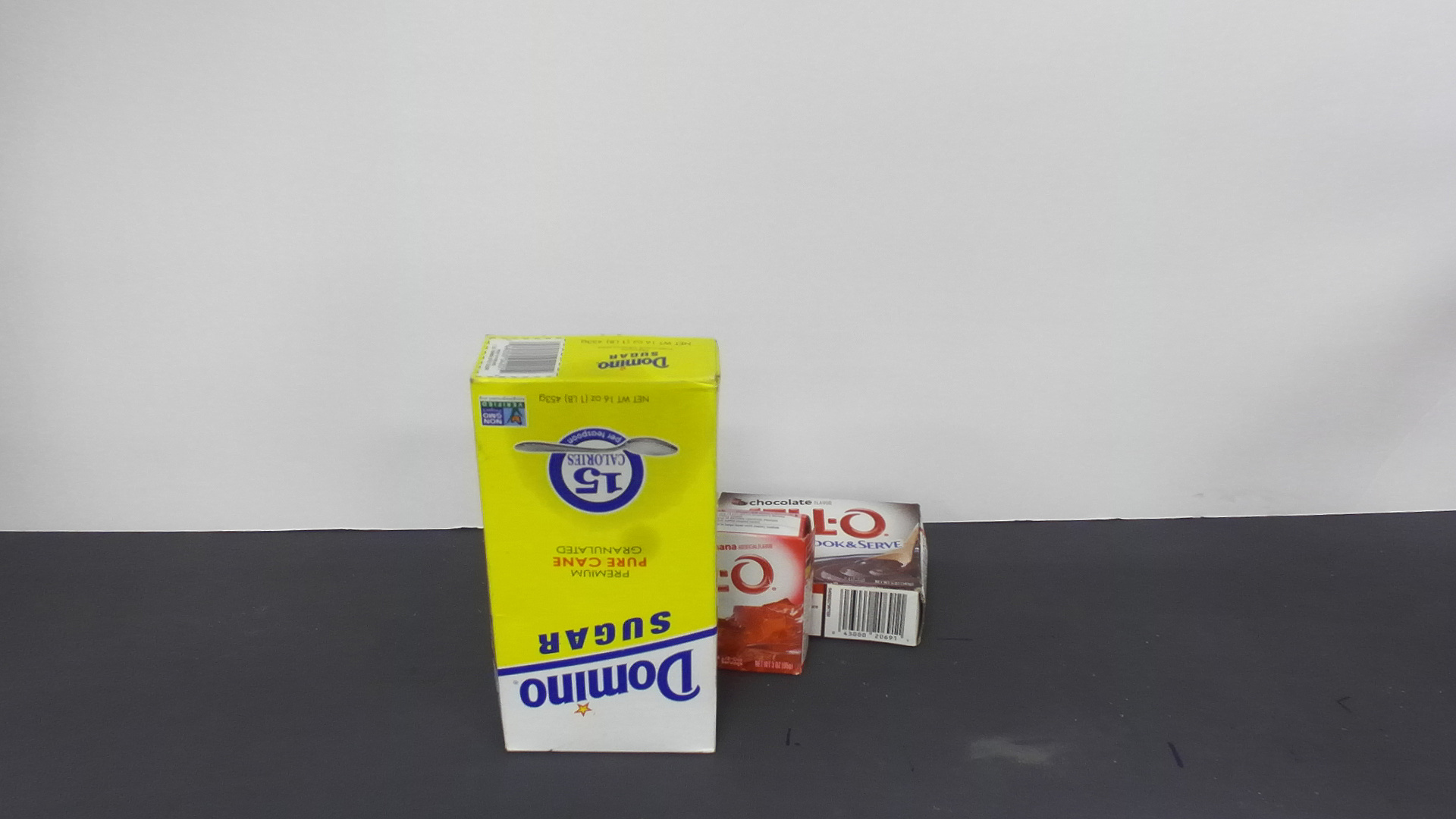} &
\includegraphics[width=0.165\textwidth]{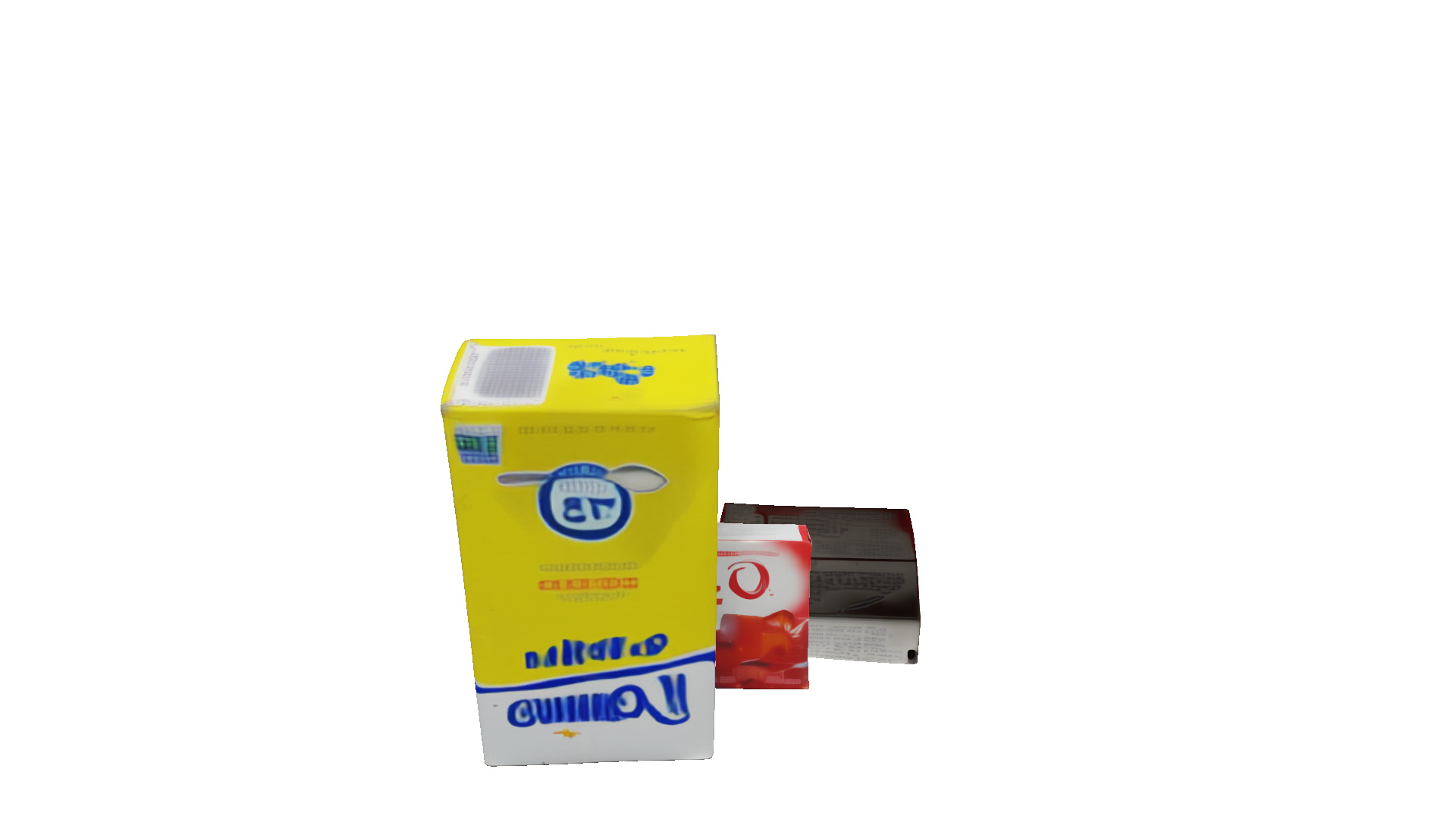} &
\includegraphics[width=0.165\textwidth]{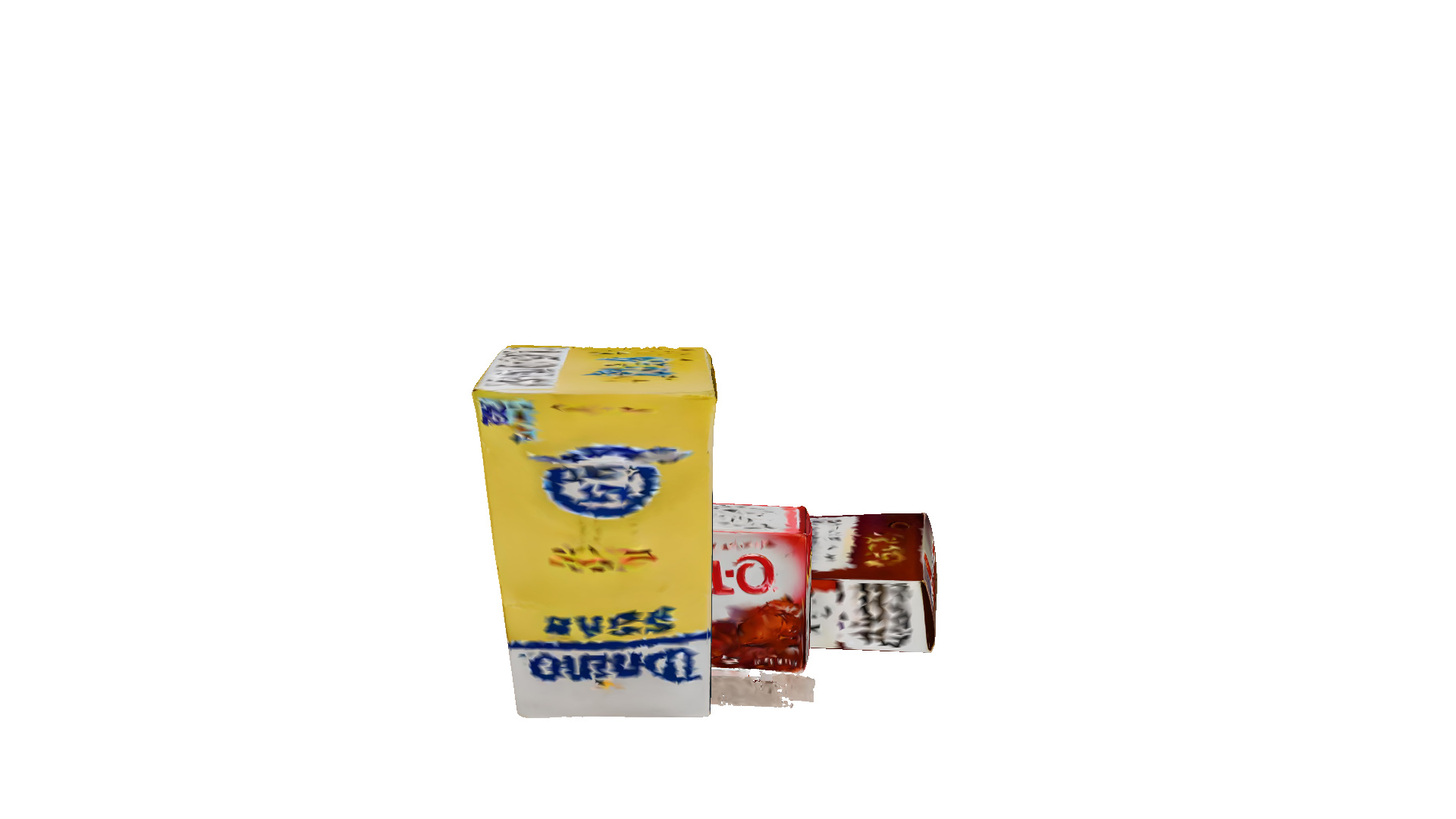} &
\includegraphics[width=0.165\textwidth]{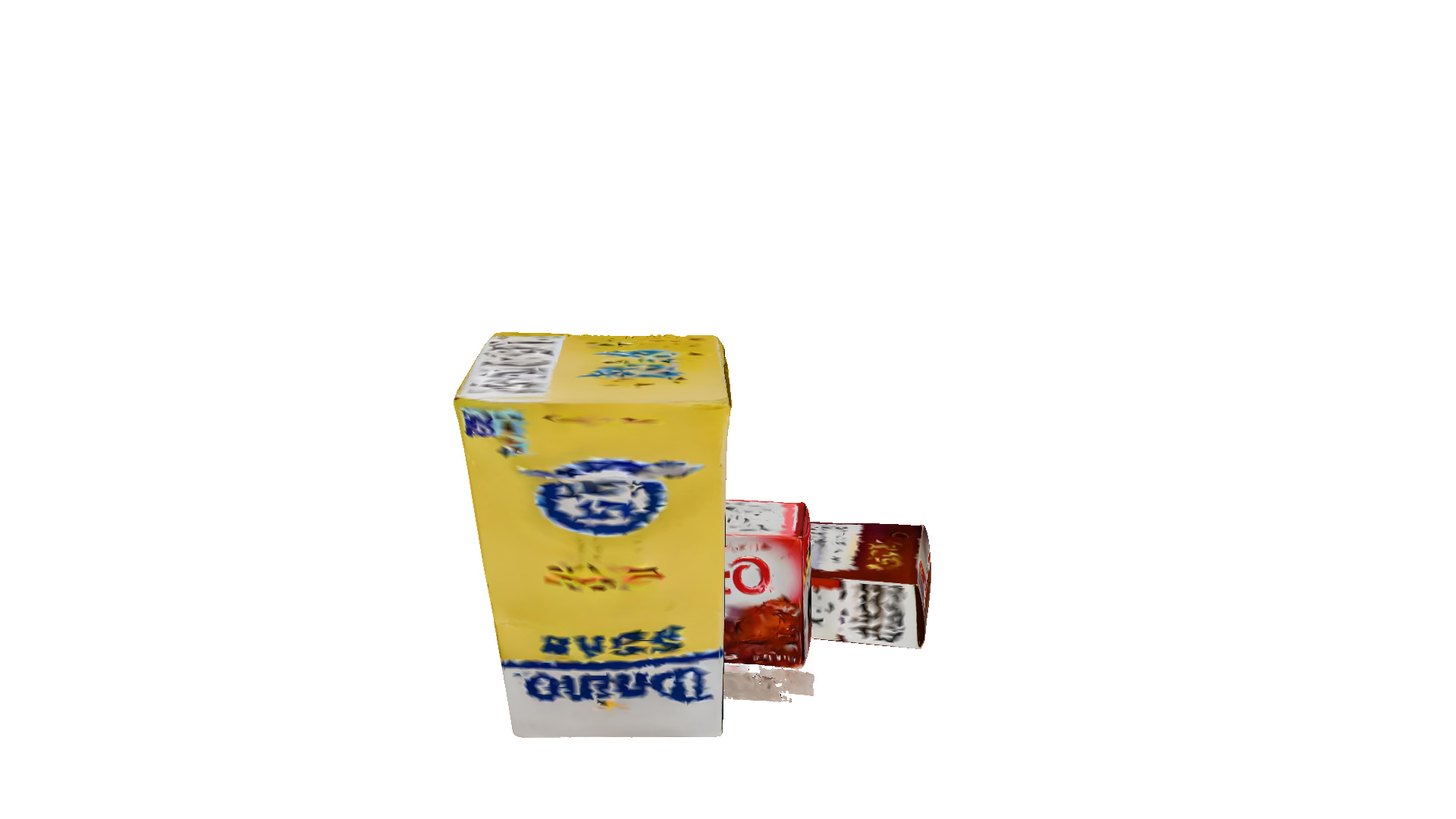} \\

Kitchen 2 & Hard &
\includegraphics[width=0.165\textwidth]{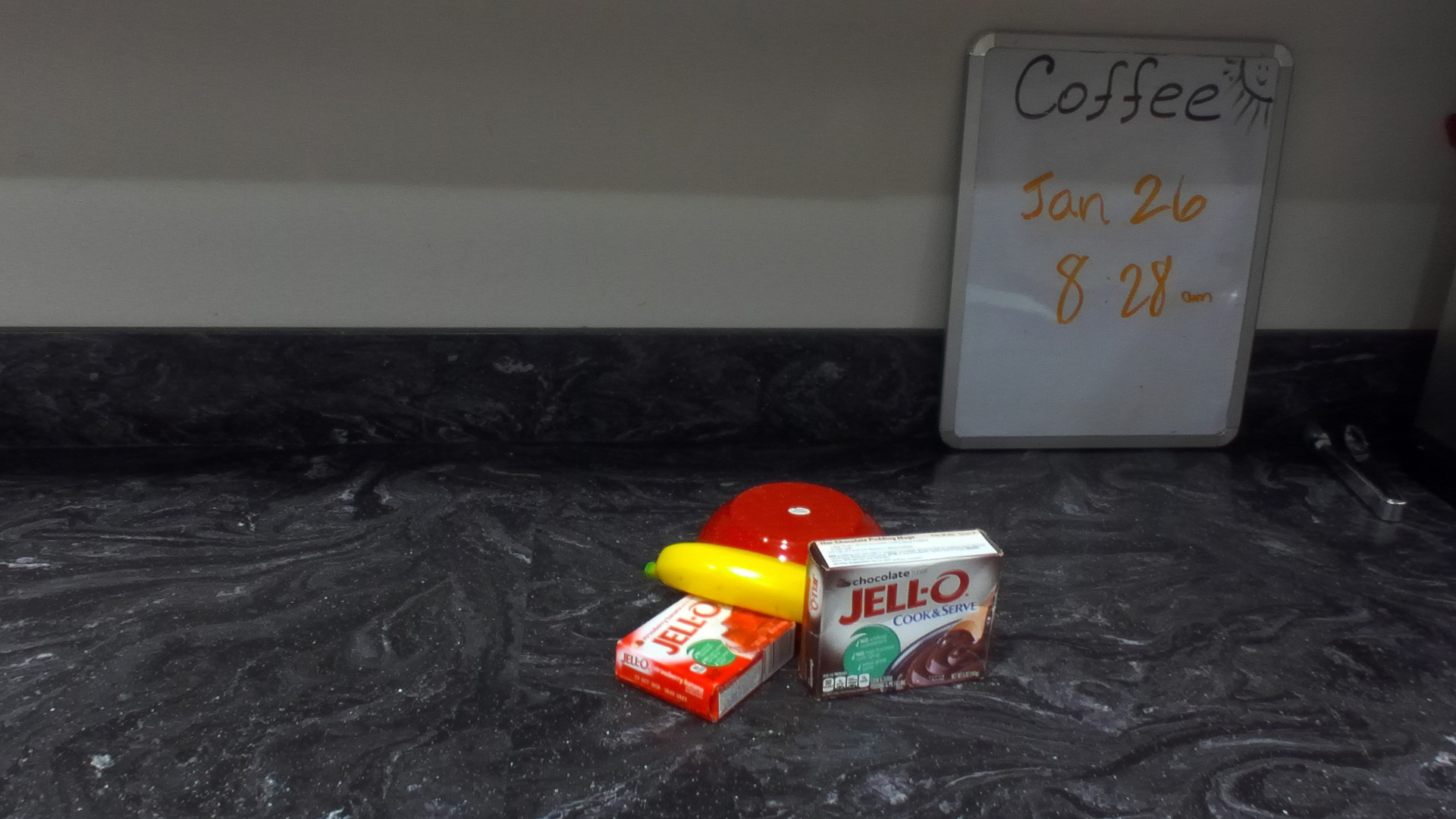} &
\includegraphics[width=0.165\textwidth]{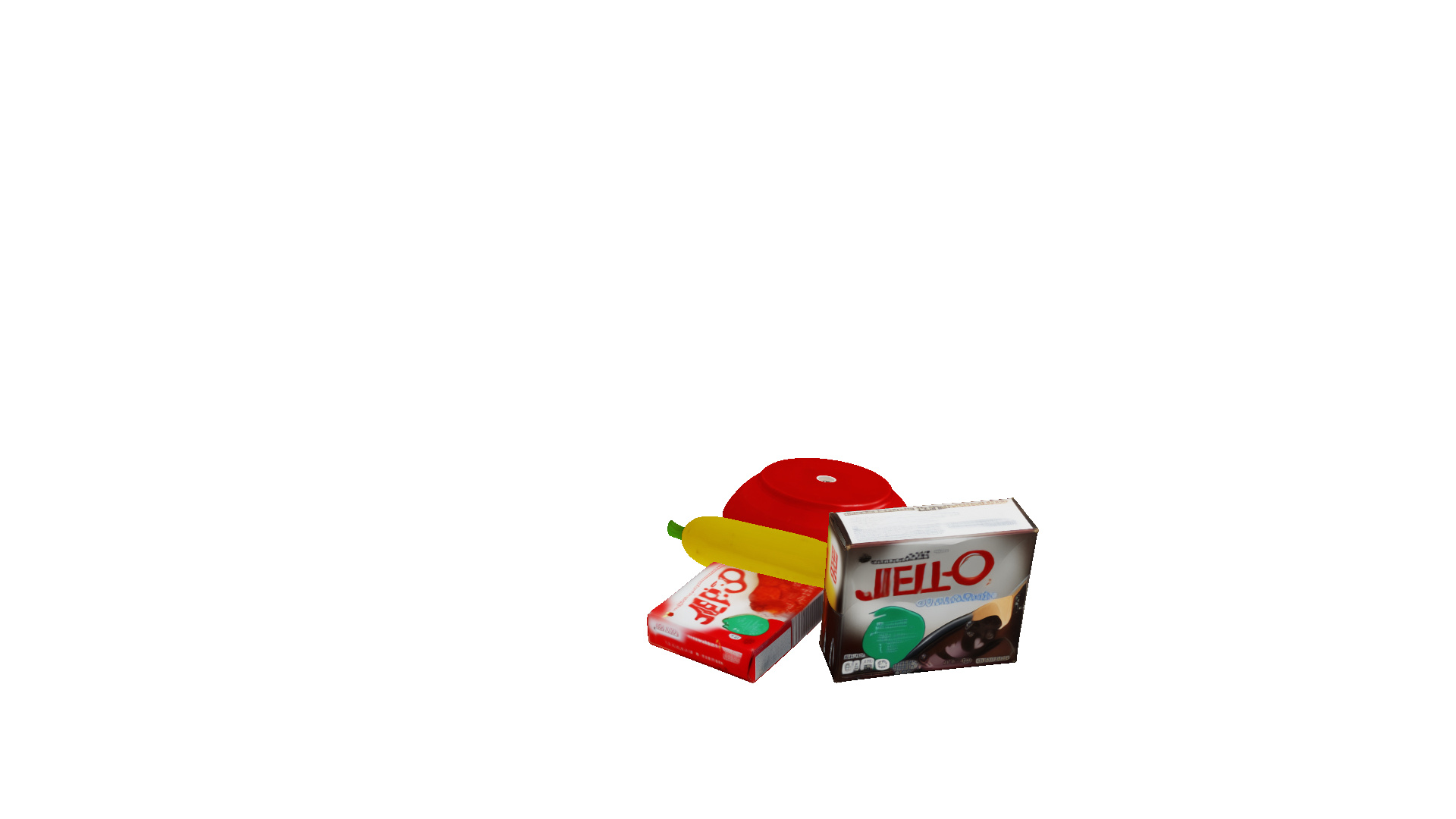} &
\includegraphics[width=0.165\textwidth]{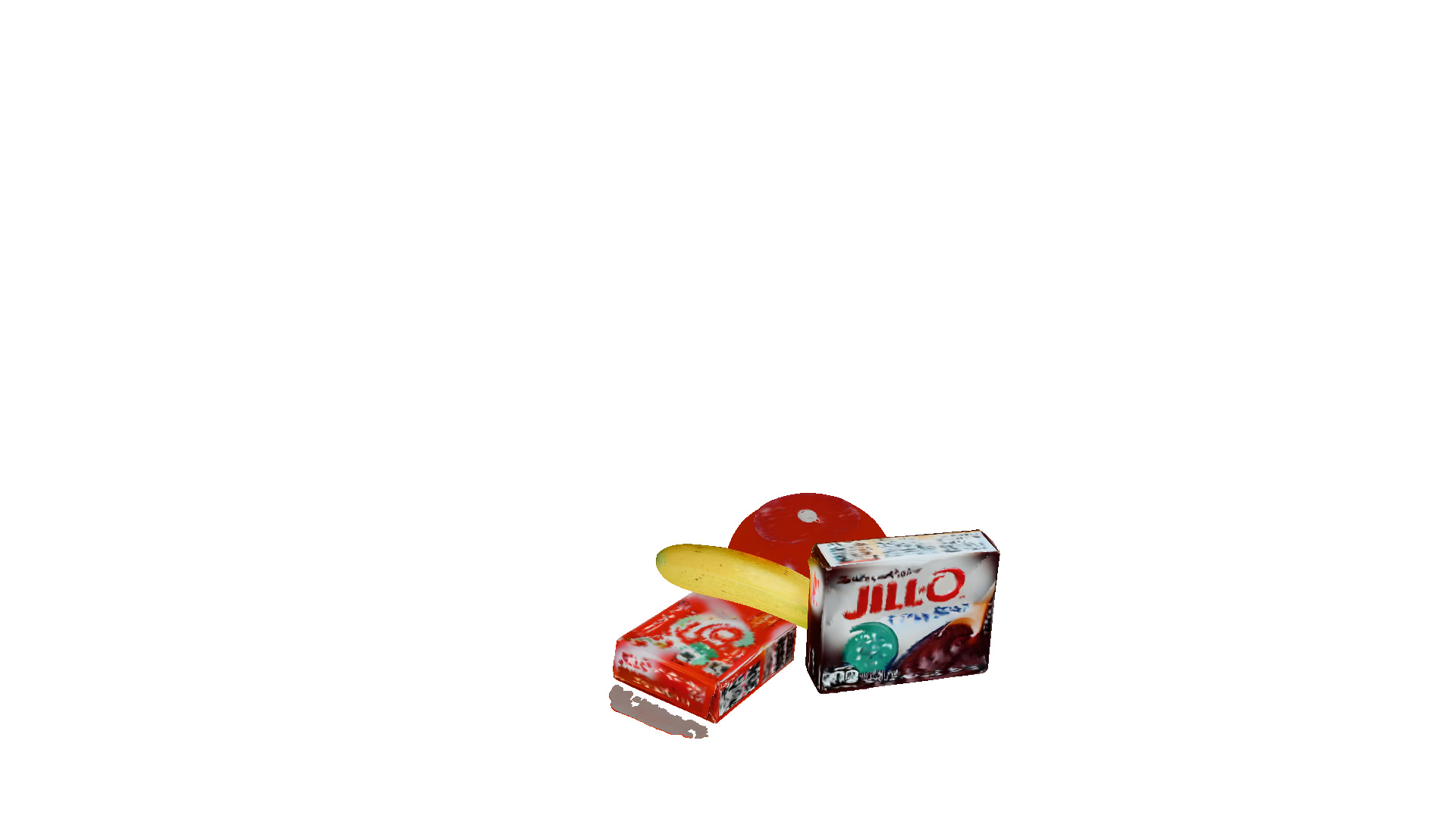} &
\includegraphics[width=0.165\textwidth]{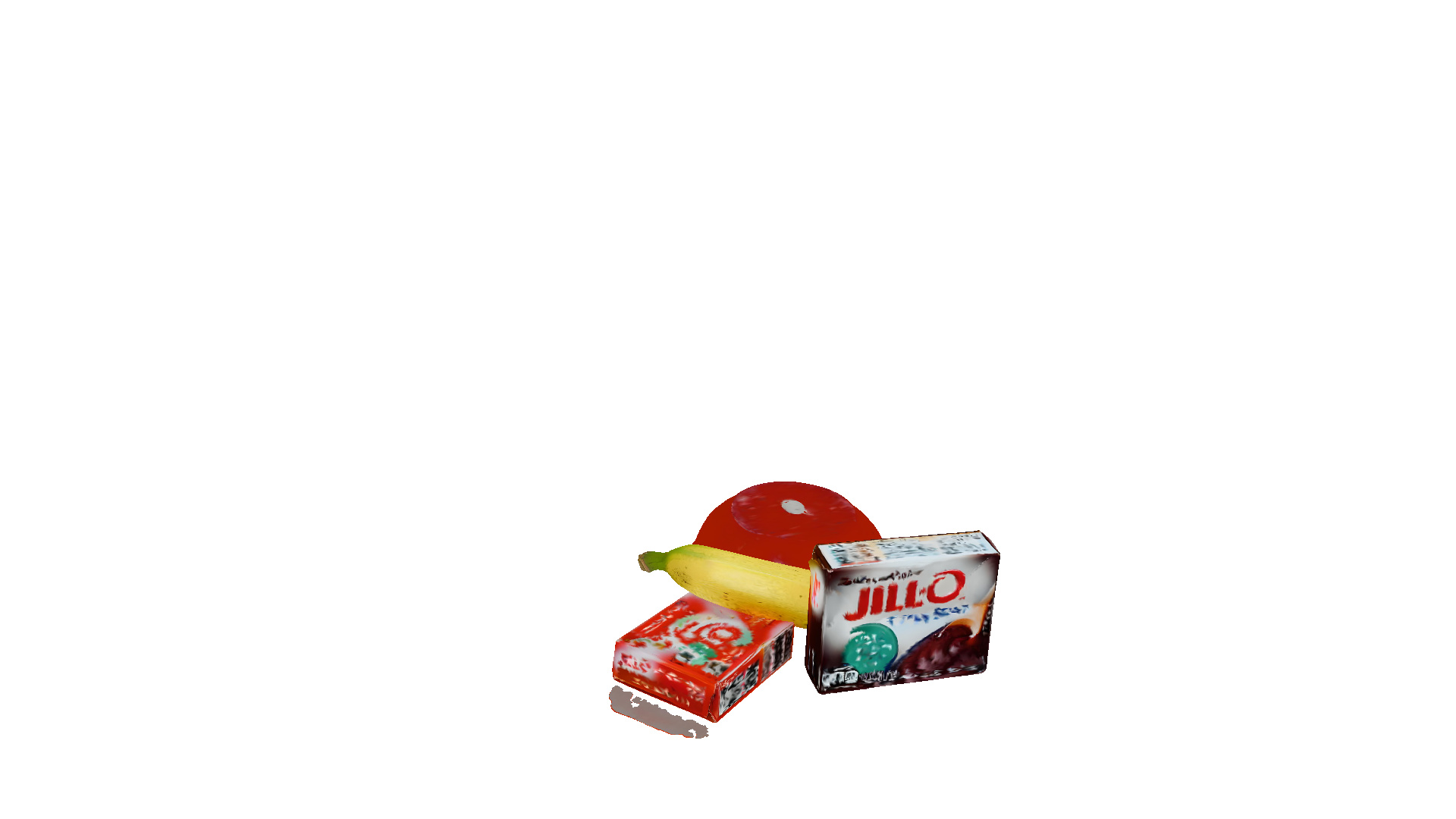} \\

Kitchen 3 & Hard &
\includegraphics[width=0.165\textwidth]{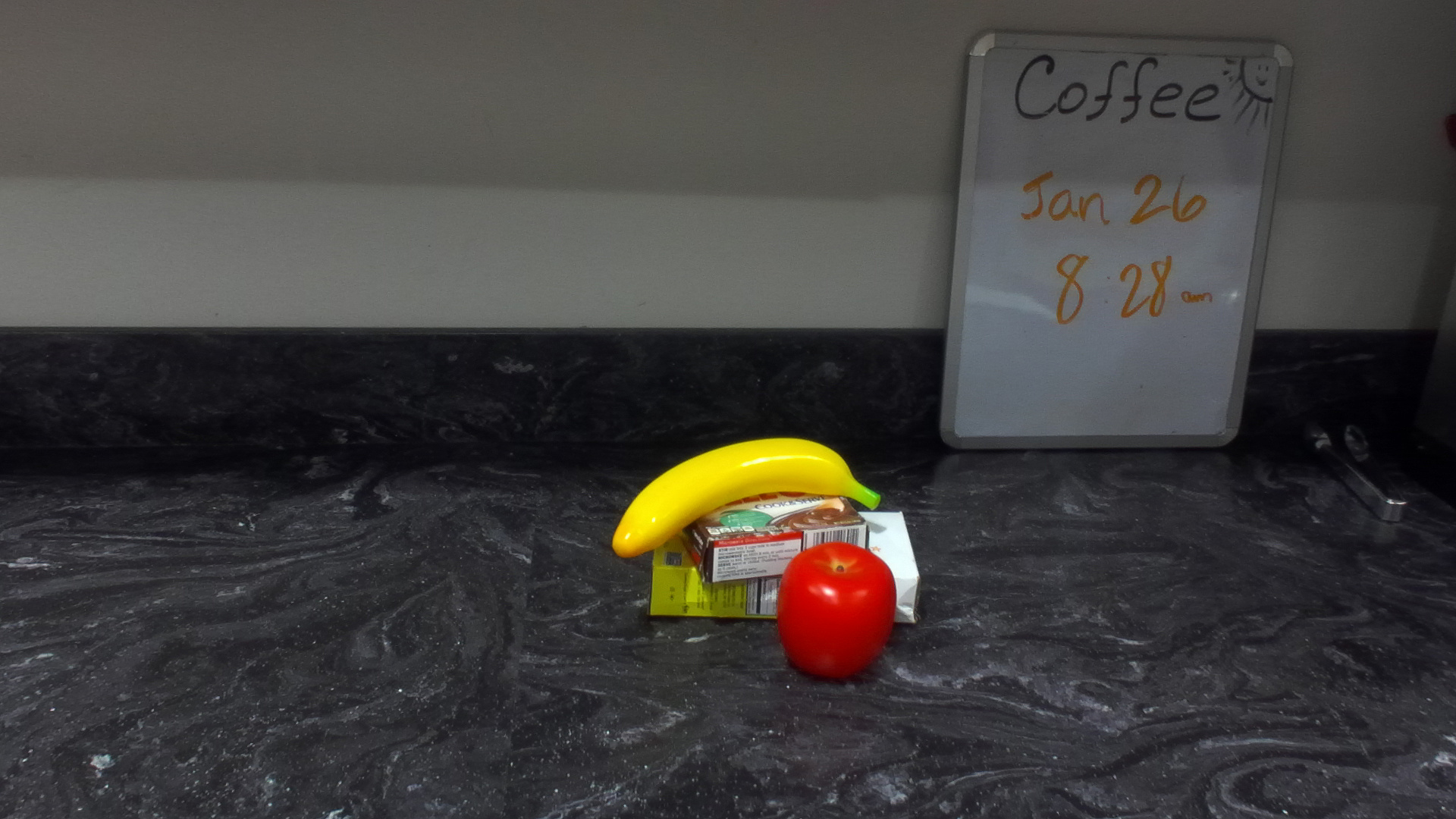} &
\includegraphics[width=0.165\textwidth]{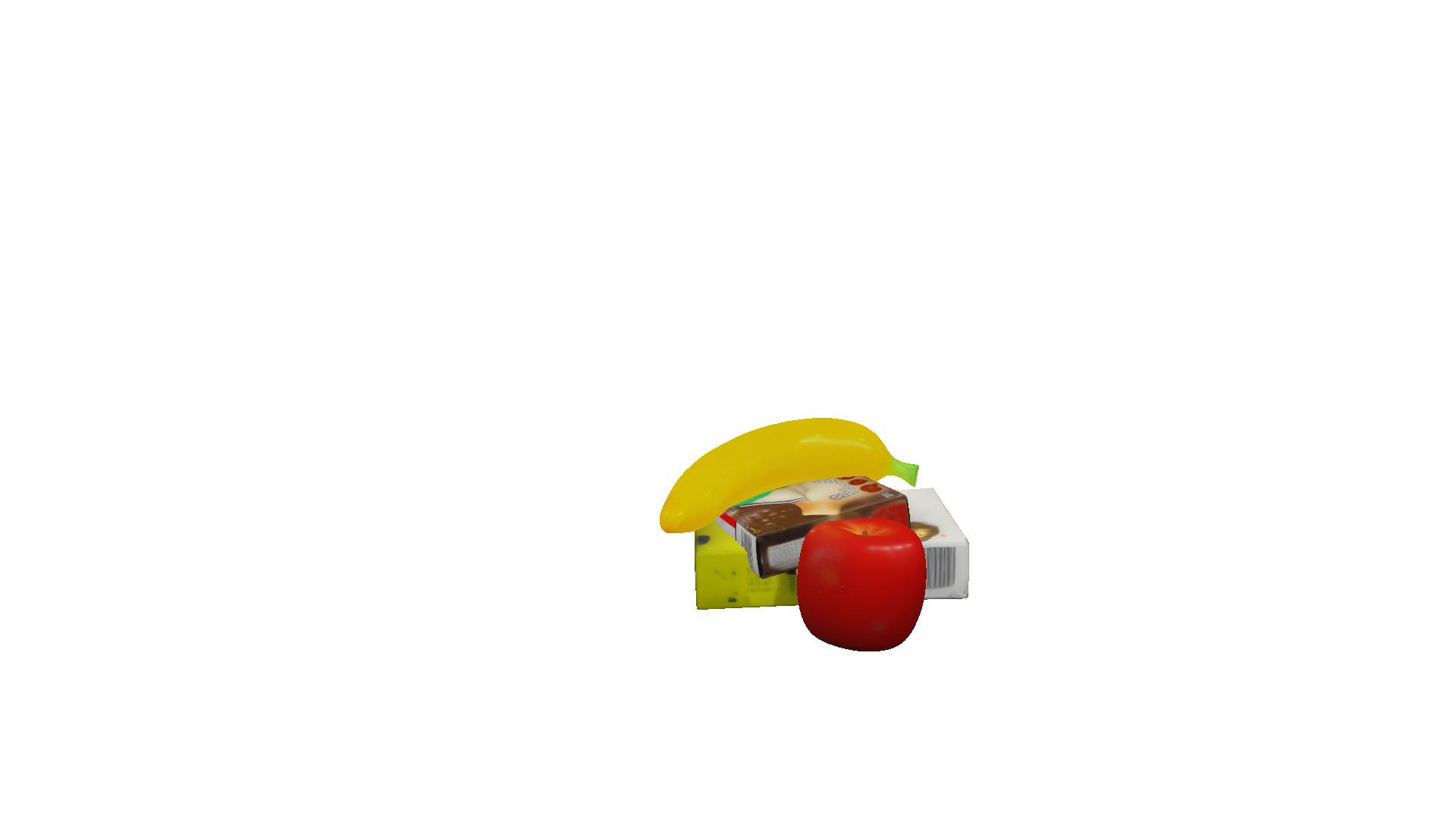} &
\includegraphics[width=0.165\textwidth]{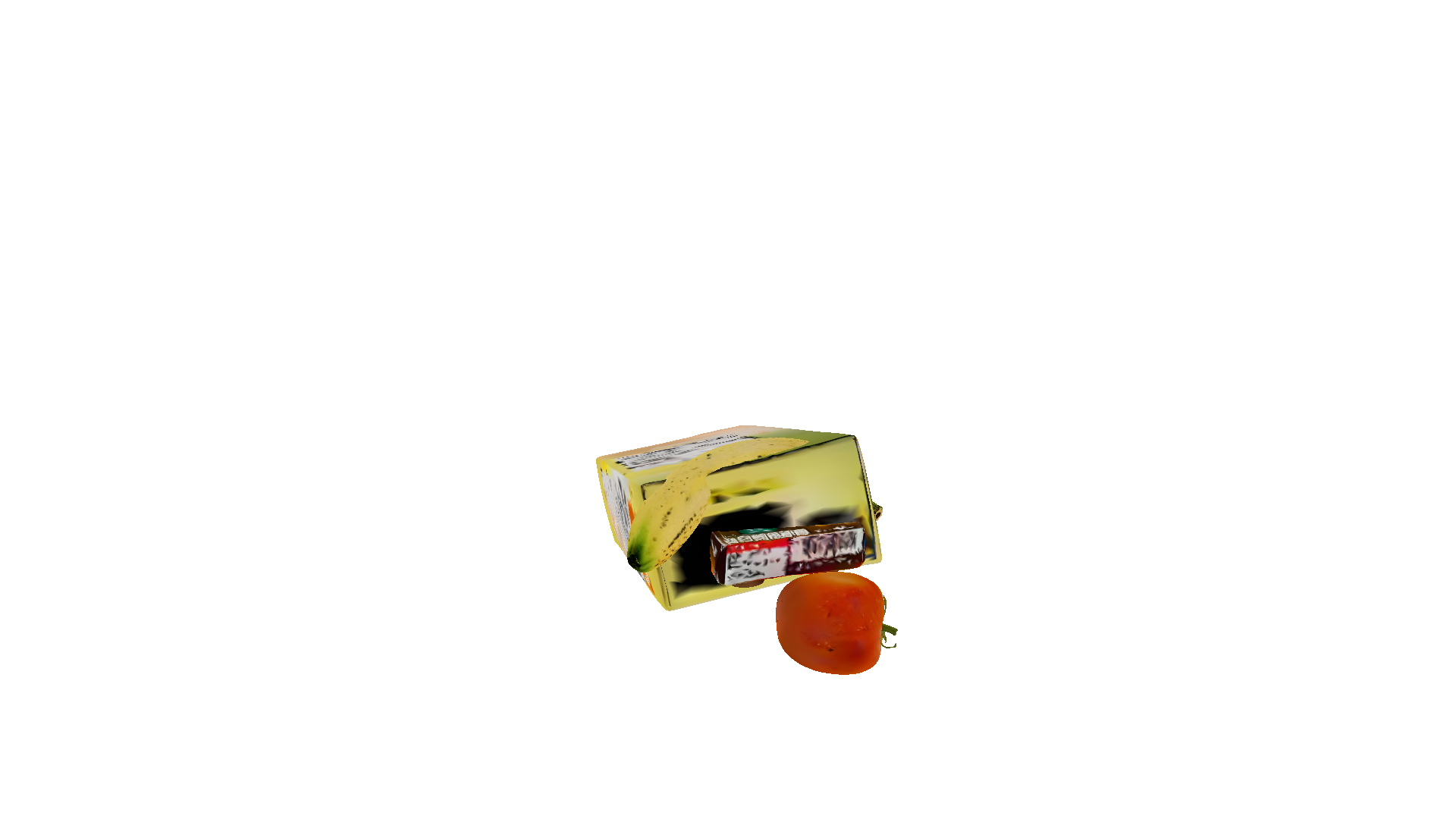} &
\includegraphics[width=0.165\textwidth]{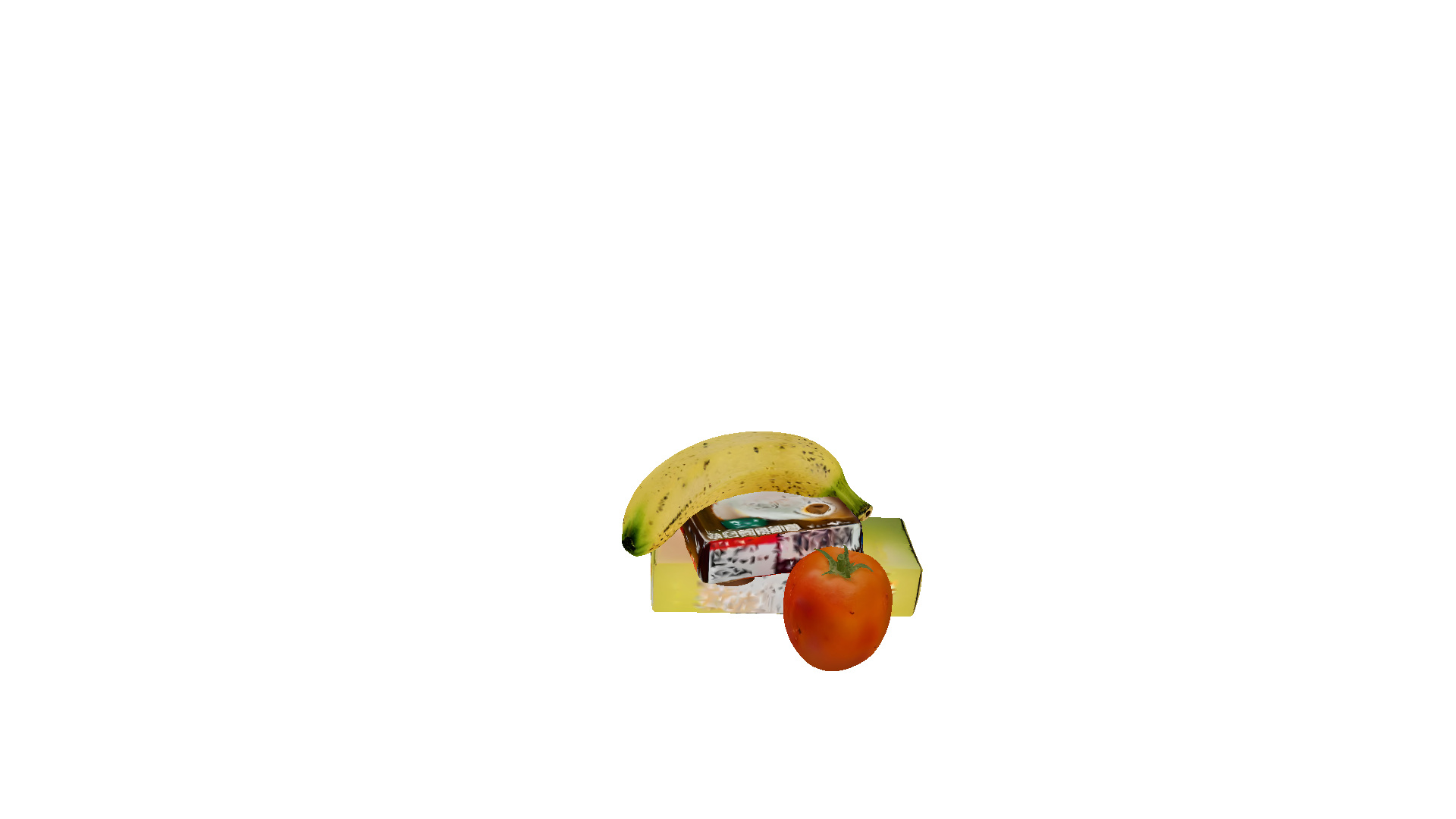} \\

\bottomrule
\end{tabular*}
\caption{\textbf{Scene Reconstruction Qualitative Results.} We compare \sysName against SAM3D across Easy, Medium, and Hard scenes.}
\label{tab:scene_reconstruction_qualitative}
\end{table*}

\begin{table}[t]
\centering
\setlength{\tabcolsep}{2pt}
\renewcommand{\arraystretch}{0.78}
\begin{tabular}{ccccc}
\toprule
\makecell{N\\Objects} &
\makecell{Time\\(min)} &
Original &
\makecell{Recon.\\Twin} &
\makecell{Cousins\\Image} \\
\midrule
\raisebox{0.04\textwidth}{2} & \raisebox{0.04\textwidth}{19.98} &
\includegraphics[width=0.20\textwidth,height=0.12\textwidth,keepaspectratio]{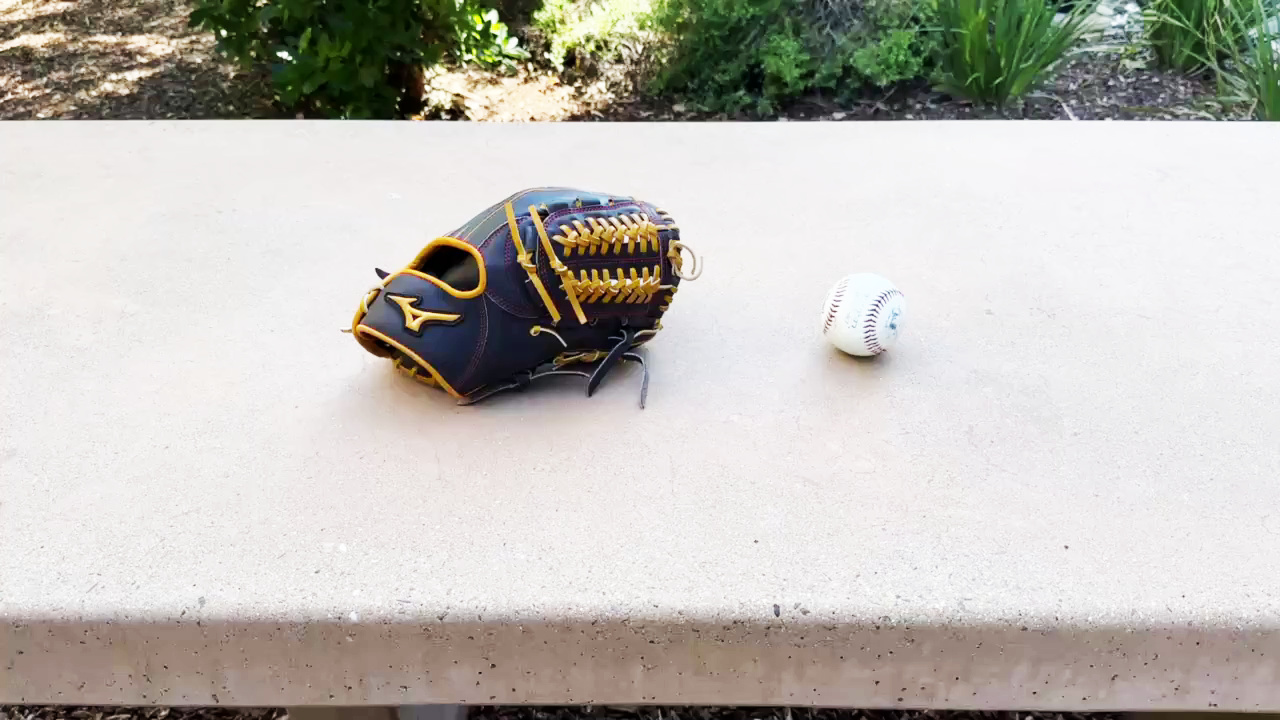} &
\includegraphics[width=0.20\textwidth,height=0.12\textwidth,keepaspectratio]{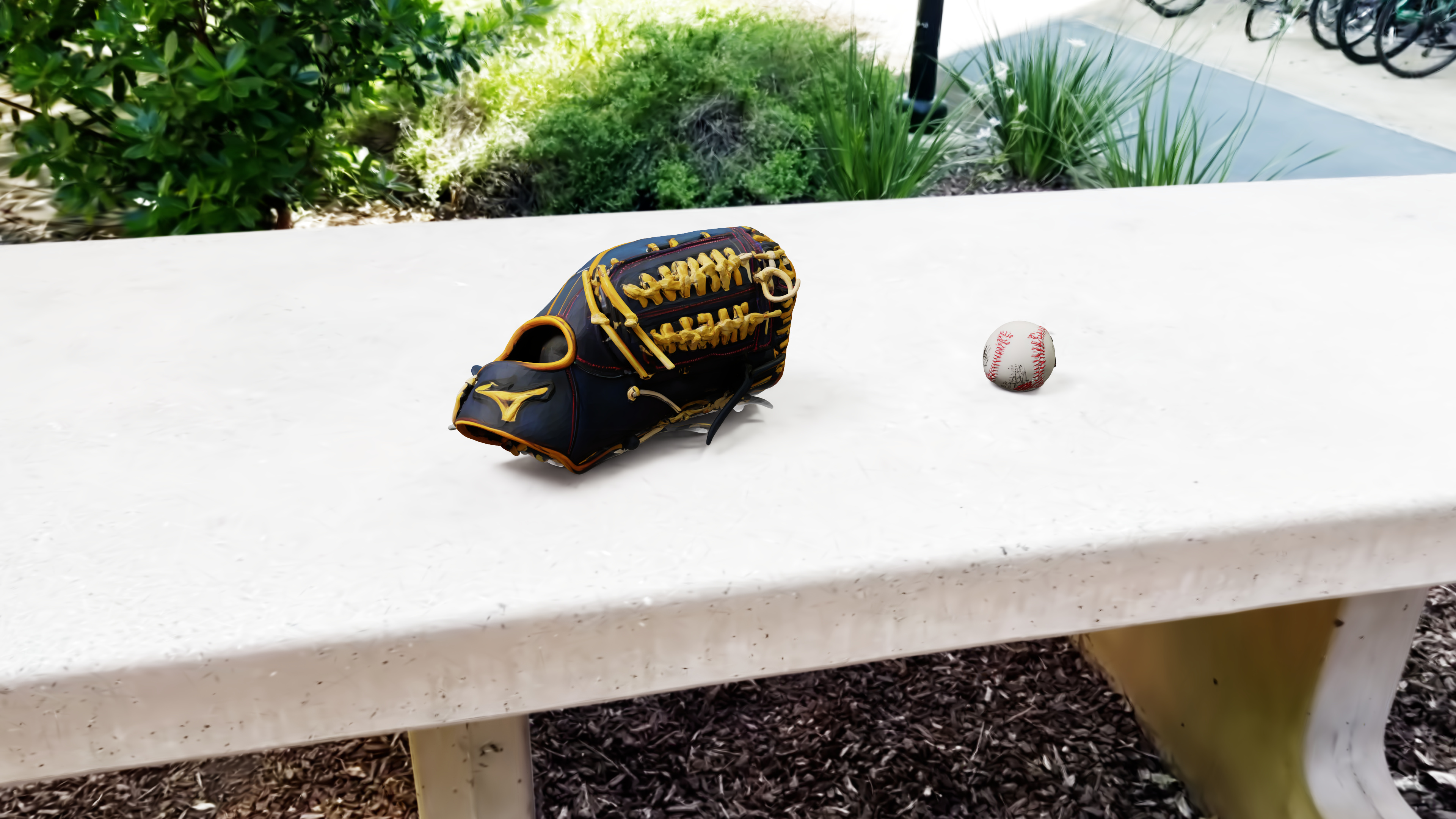} &
\includegraphics[width=0.20\textwidth,height=0.12\textwidth,keepaspectratio]{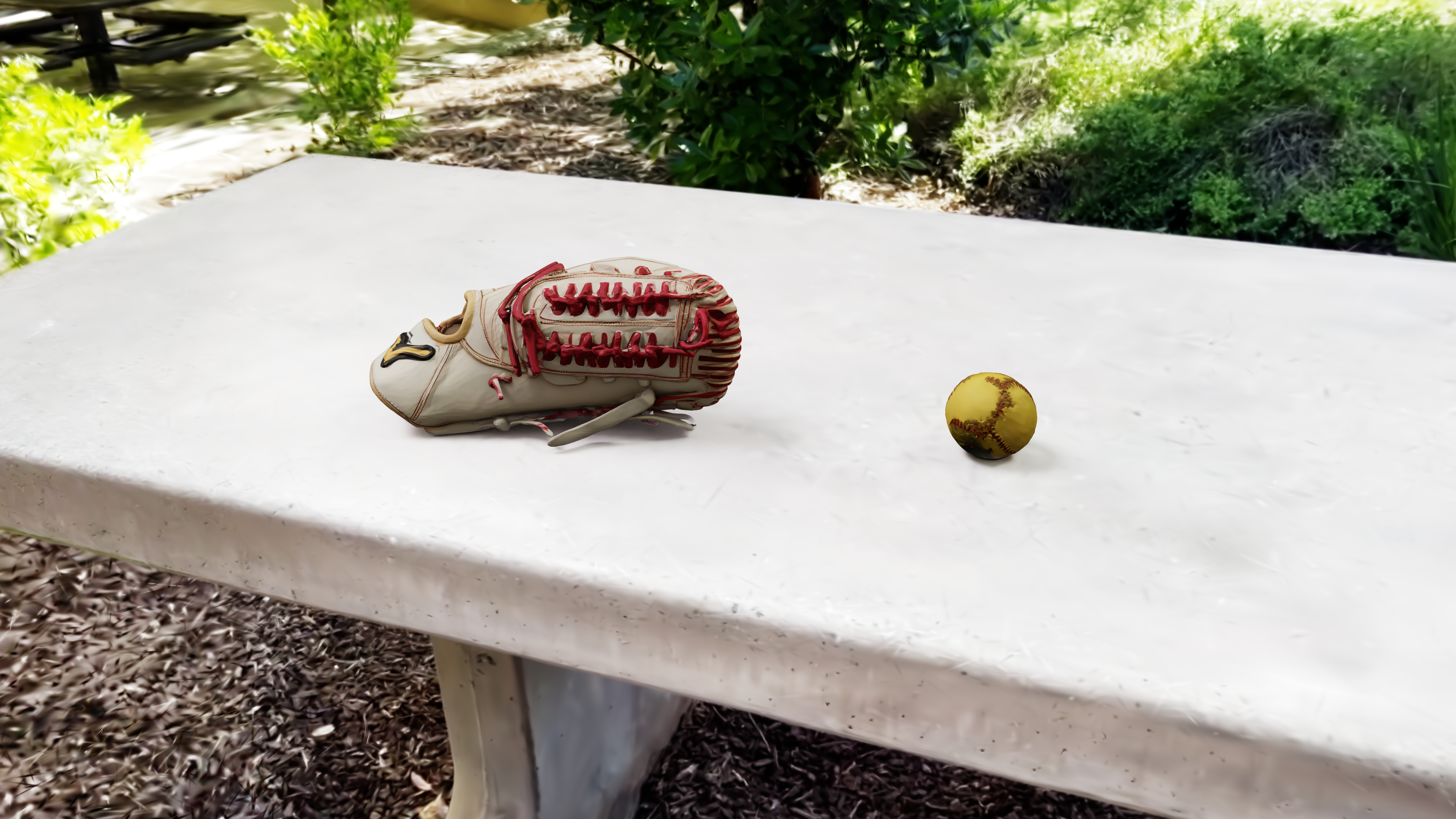} \\[1pt]

\raisebox{0.04\textwidth}{4} & \raisebox{0.04\textwidth}{34.07} &
\includegraphics[width=0.20\textwidth,height=0.12\textwidth,keepaspectratio]{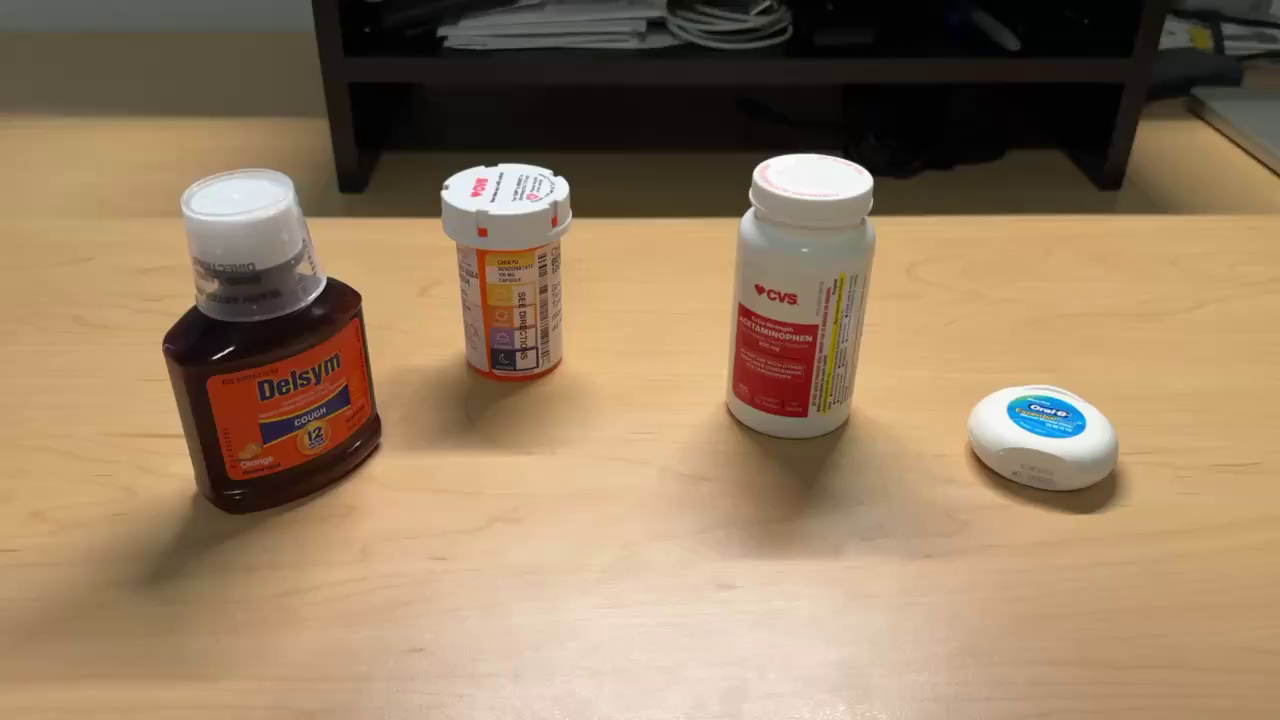} &
\includegraphics[width=0.20\textwidth,height=0.12\textwidth,keepaspectratio]{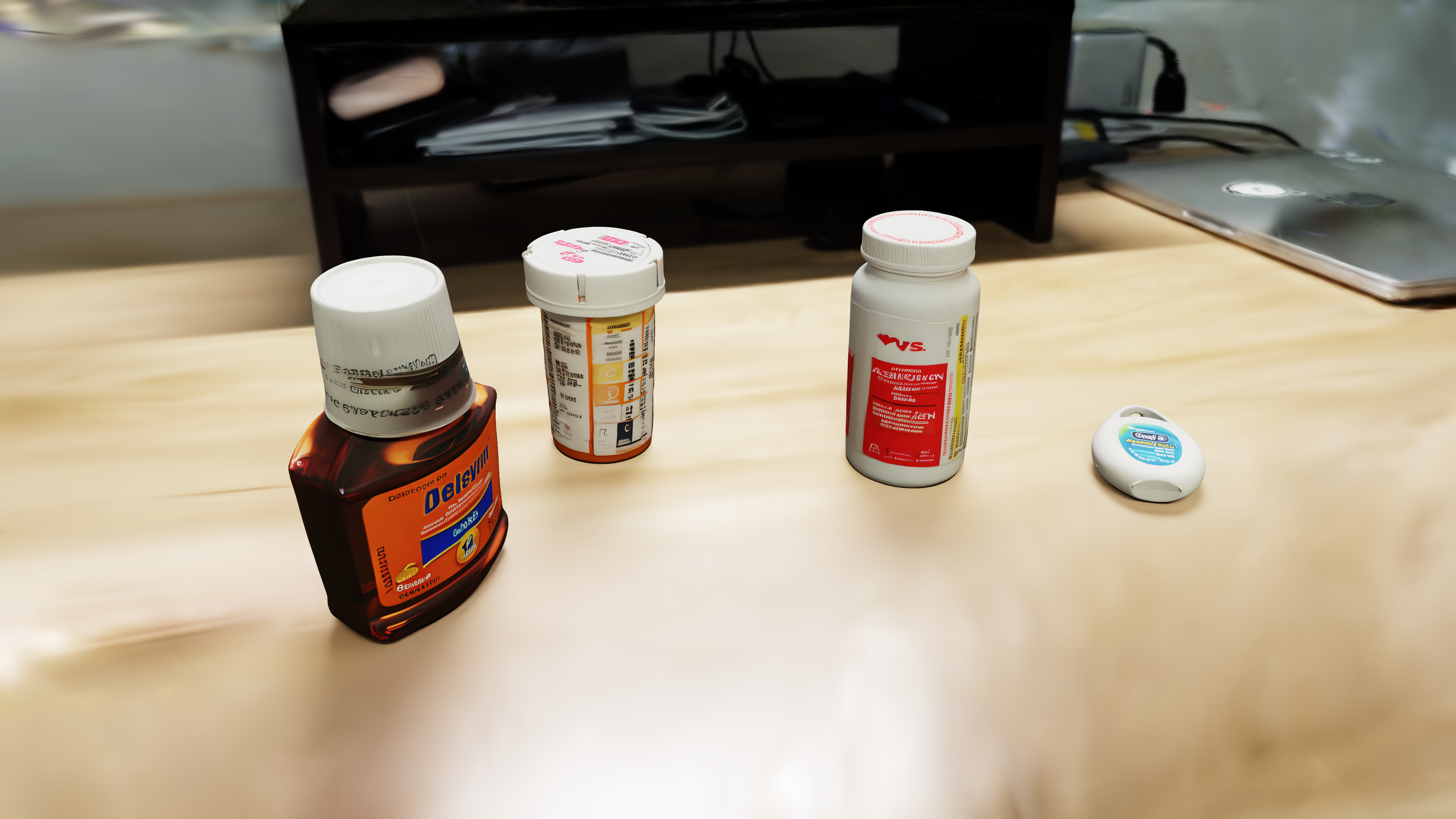} &
\includegraphics[width=0.20\textwidth,height=0.12\textwidth,keepaspectratio]{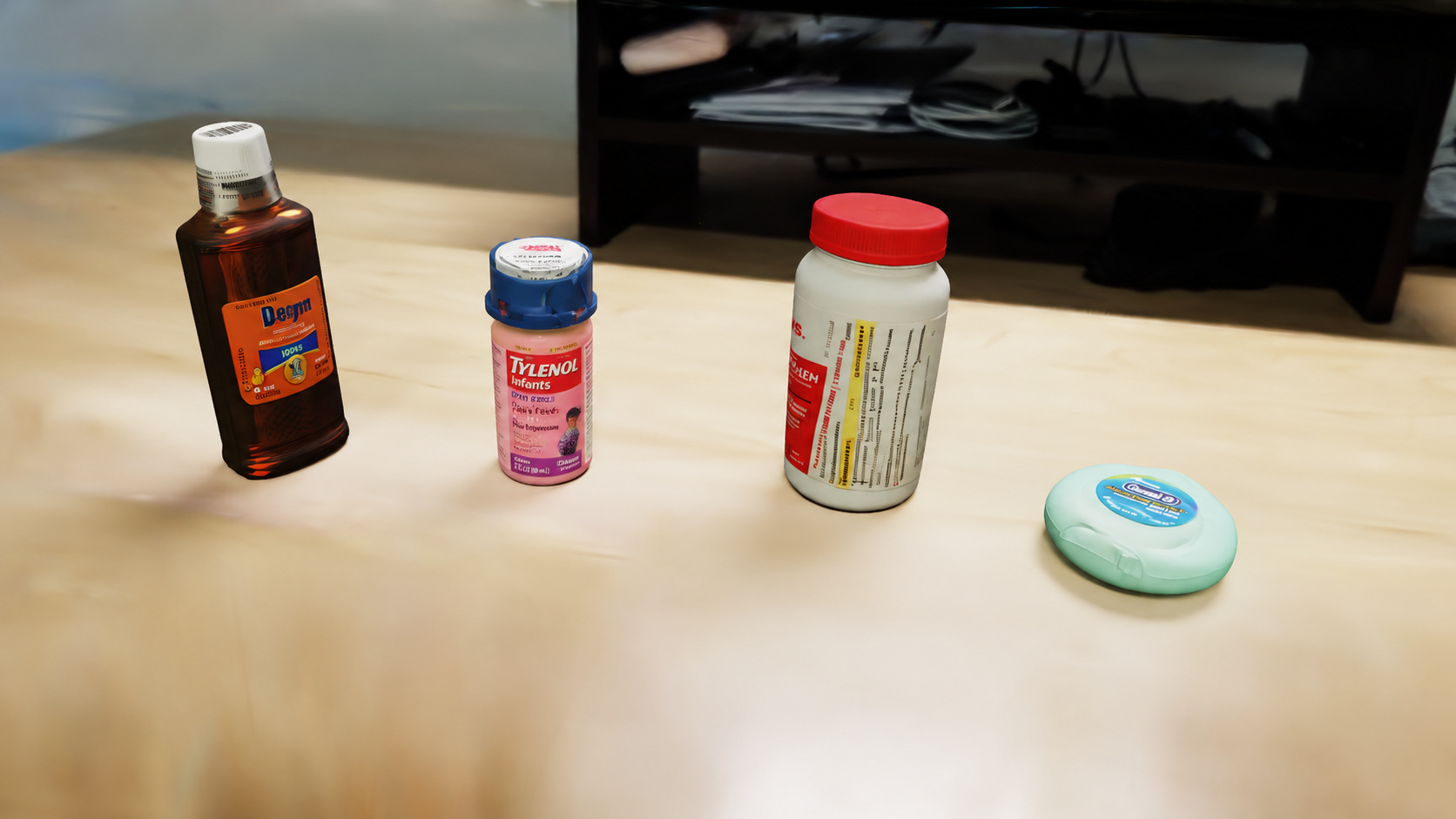} \\[1pt]

\raisebox{0.04\textwidth}{9} & \raisebox{0.04\textwidth}{41.36} &
\includegraphics[width=0.20\textwidth,height=0.12\textwidth,keepaspectratio]{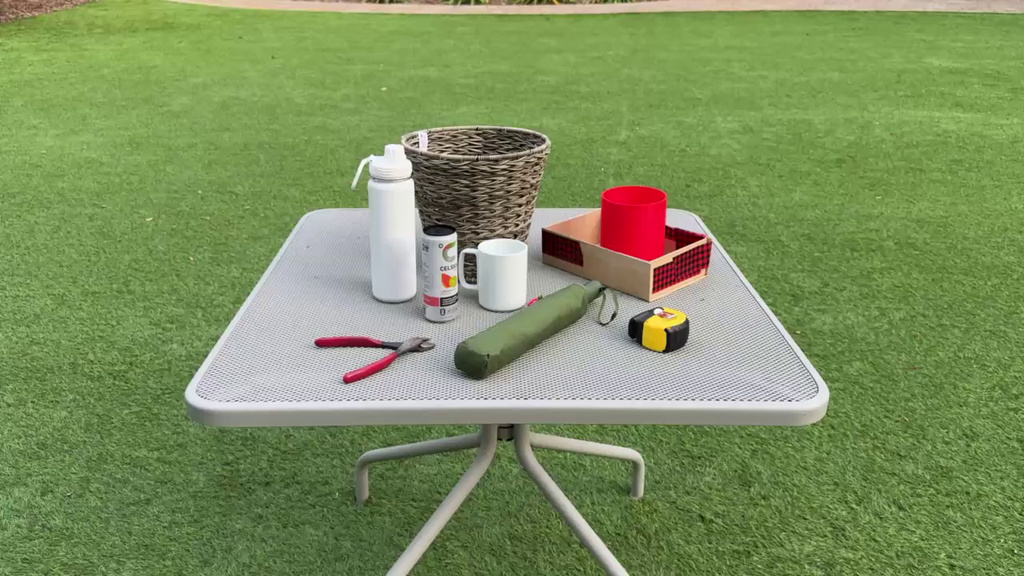} &
\includegraphics[width=0.20\textwidth,height=0.12\textwidth,keepaspectratio]{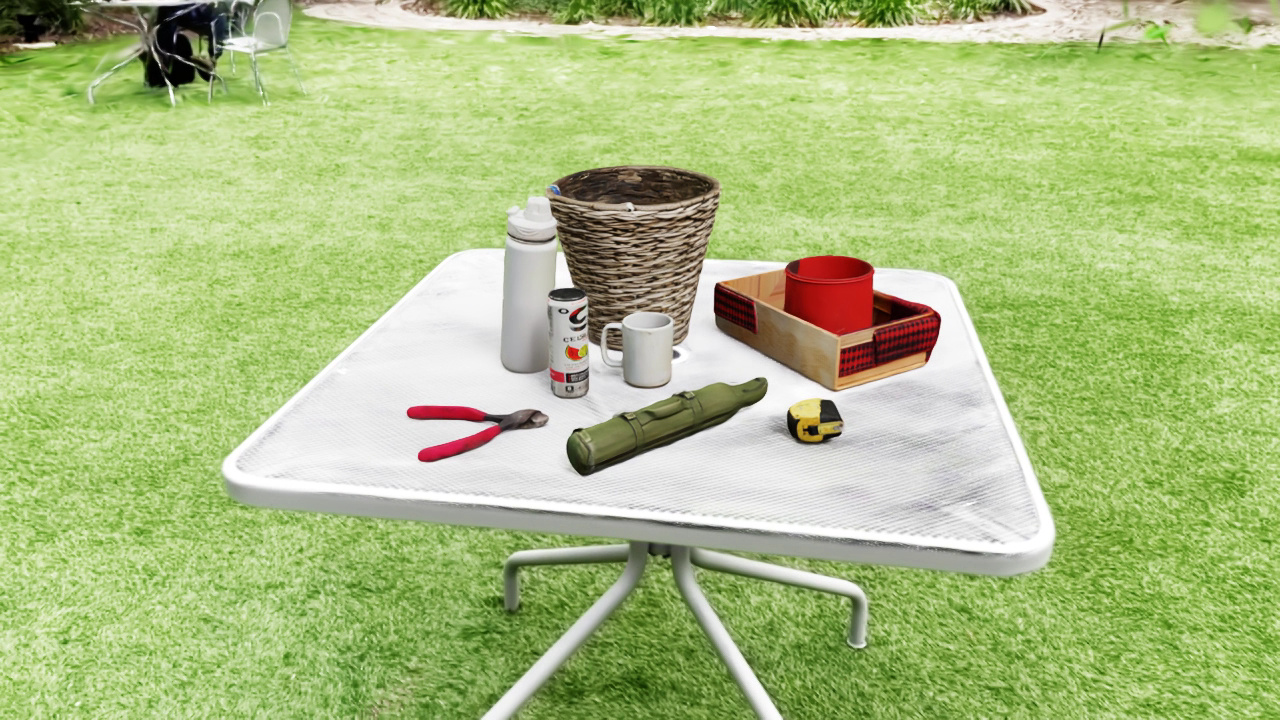} &
\includegraphics[width=0.20\textwidth,height=0.12\textwidth,keepaspectratio]{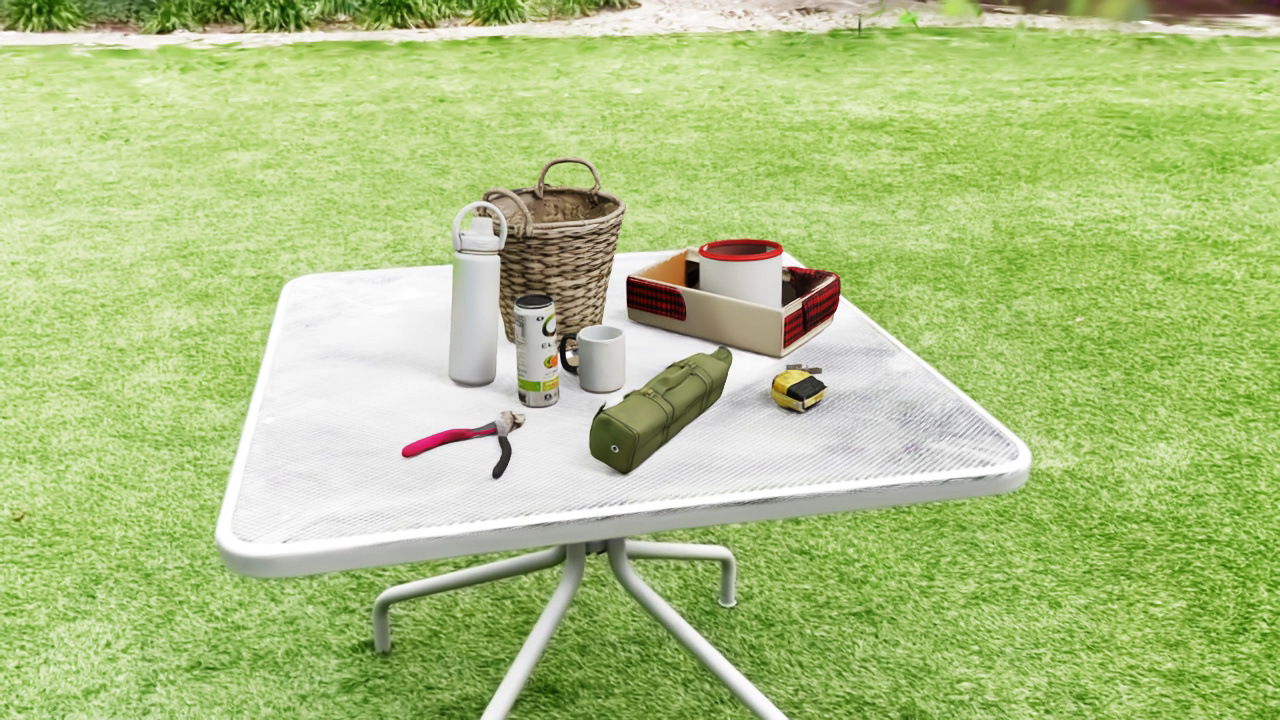} \\[1pt]

\raisebox{0.04\textwidth}{10} & \raisebox{0.04\textwidth}{42.76} &
\includegraphics[width=0.20\textwidth,height=0.12\textwidth,keepaspectratio]{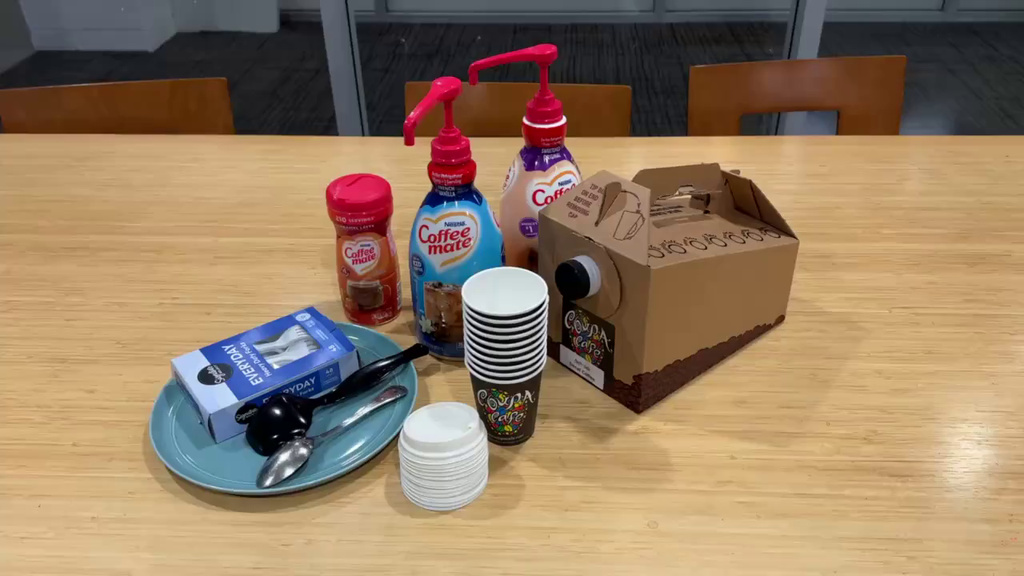} &
\includegraphics[width=0.20\textwidth,height=0.12\textwidth,keepaspectratio]{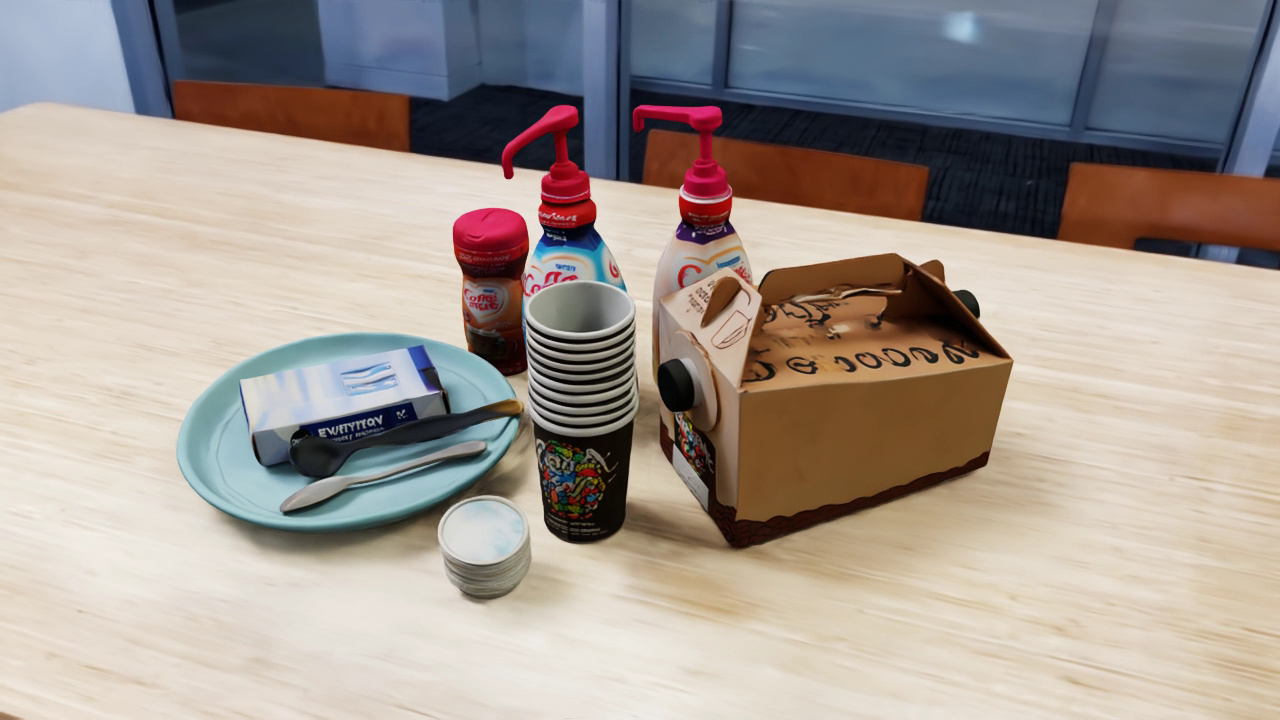} &
\includegraphics[width=0.20\textwidth,height=0.12\textwidth,keepaspectratio]{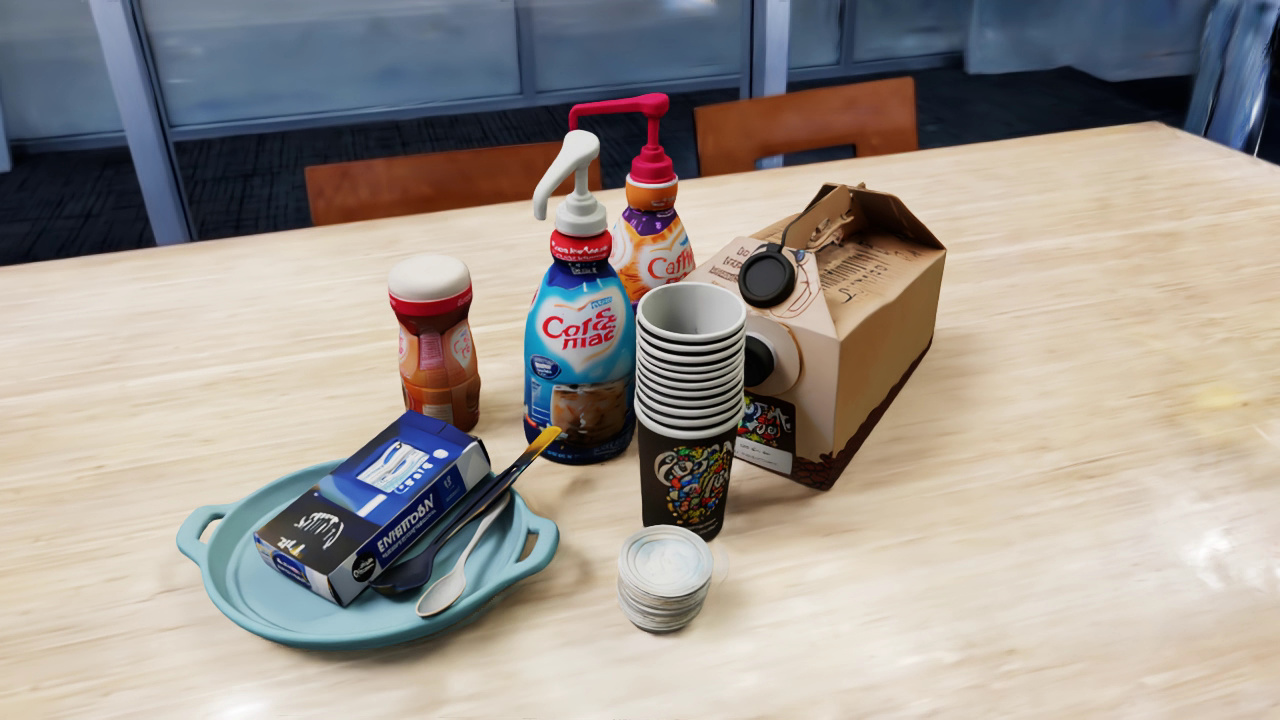} \\[1pt]

\raisebox{0.04\textwidth}{10} & \raisebox{0.04\textwidth}{49.02} &
\includegraphics[width=0.20\textwidth,height=0.12\textwidth,keepaspectratio]{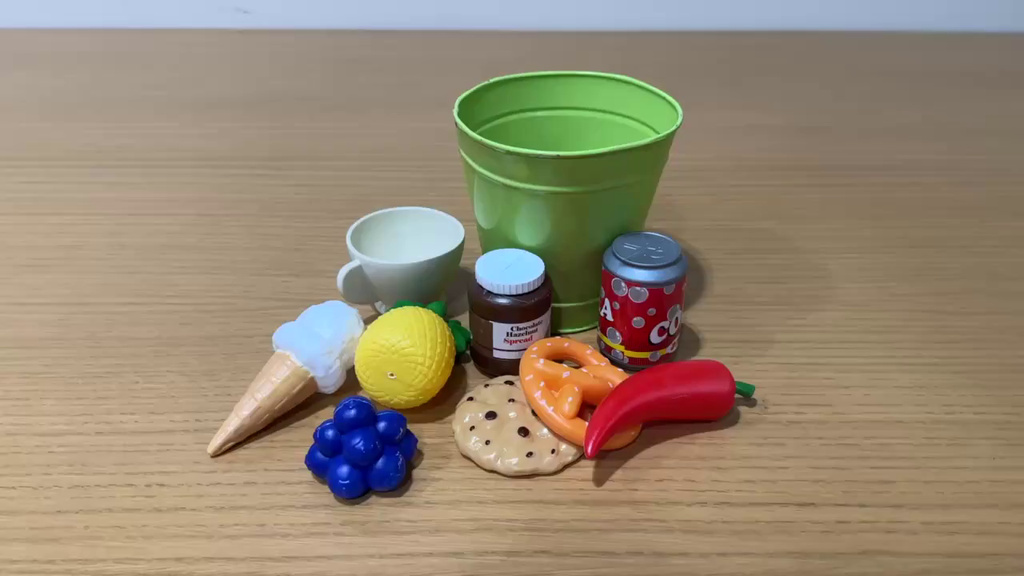} &
\includegraphics[width=0.20\textwidth,height=0.12\textwidth,keepaspectratio]{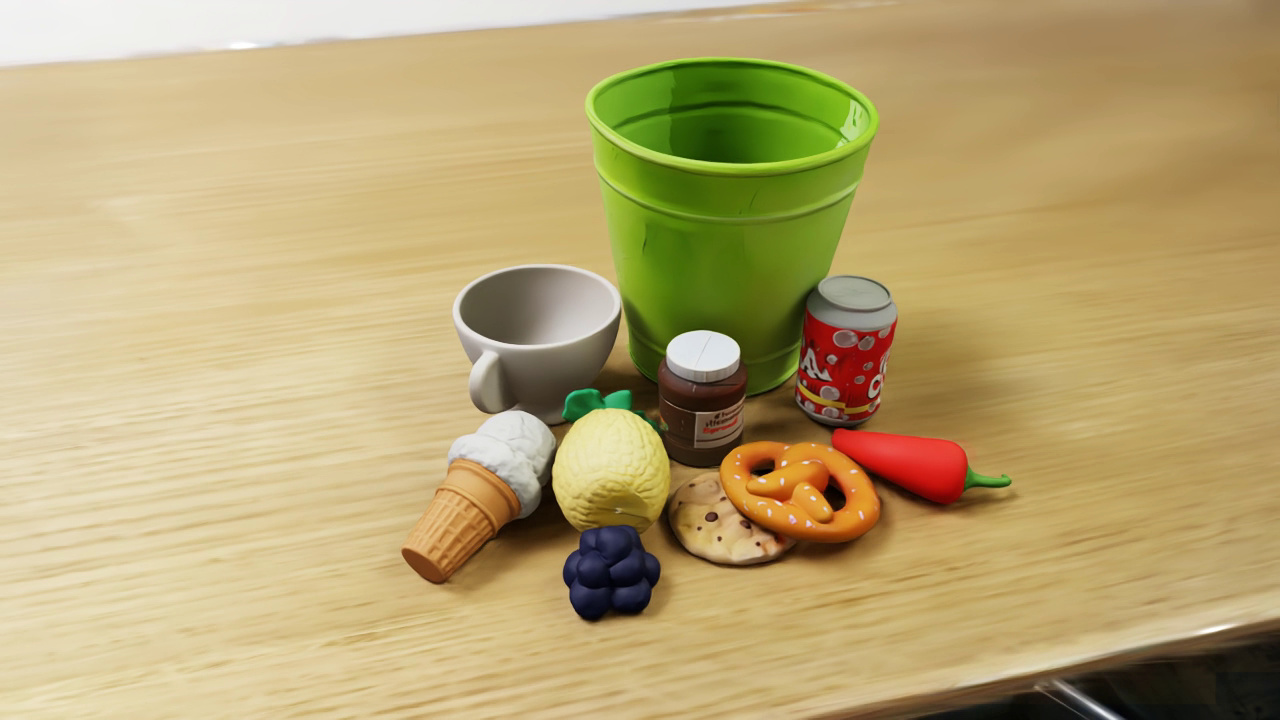} &
\includegraphics[width=0.20\textwidth,height=0.12\textwidth,keepaspectratio]{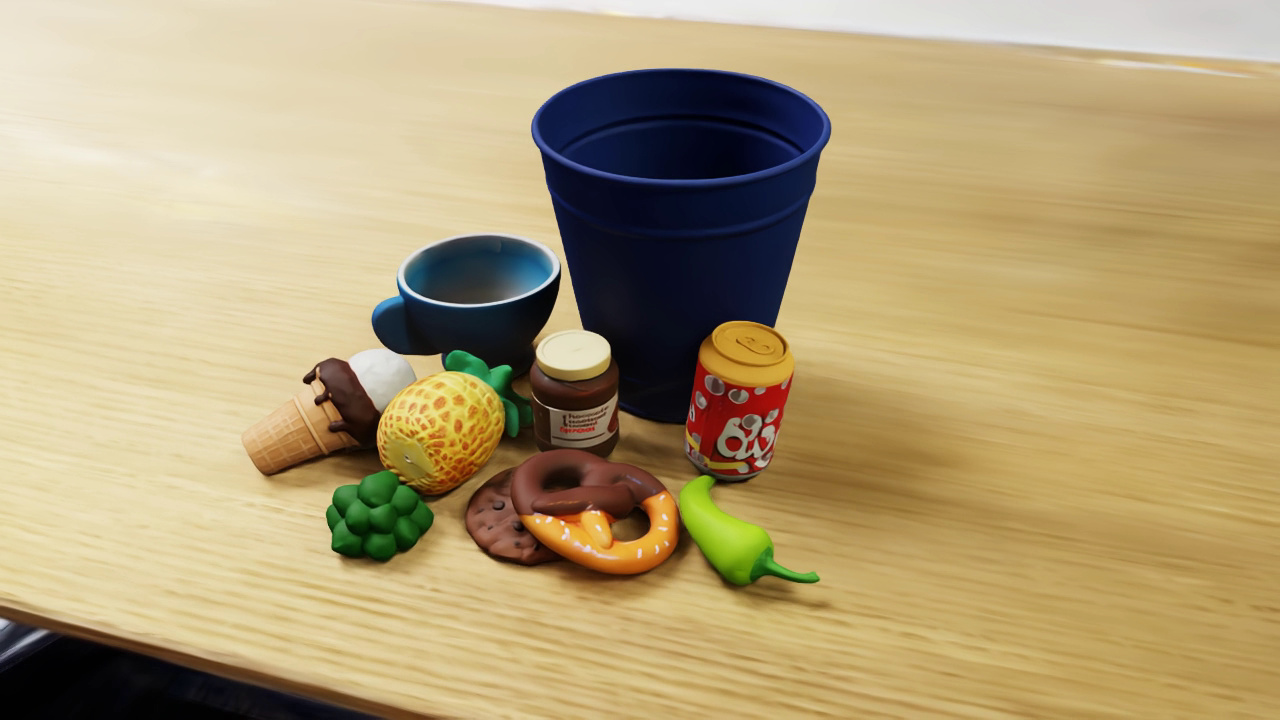} \\[1pt]

\raisebox{0.04\textwidth}{10} & \raisebox{0.04\textwidth}{51.01} &
\includegraphics[width=0.20\textwidth,height=0.12\textwidth,keepaspectratio]{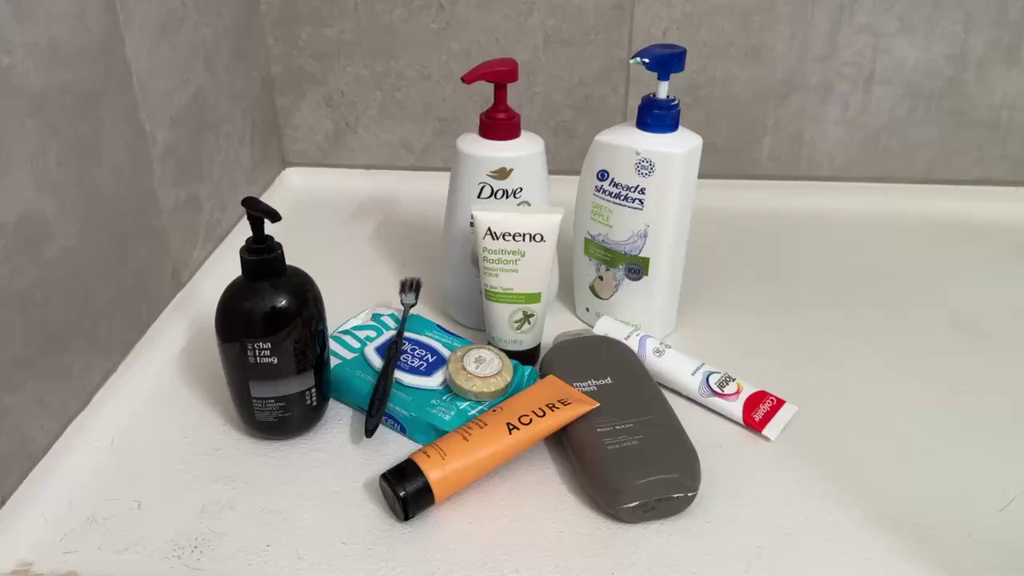} &
\includegraphics[width=0.20\textwidth,height=0.12\textwidth,keepaspectratio]{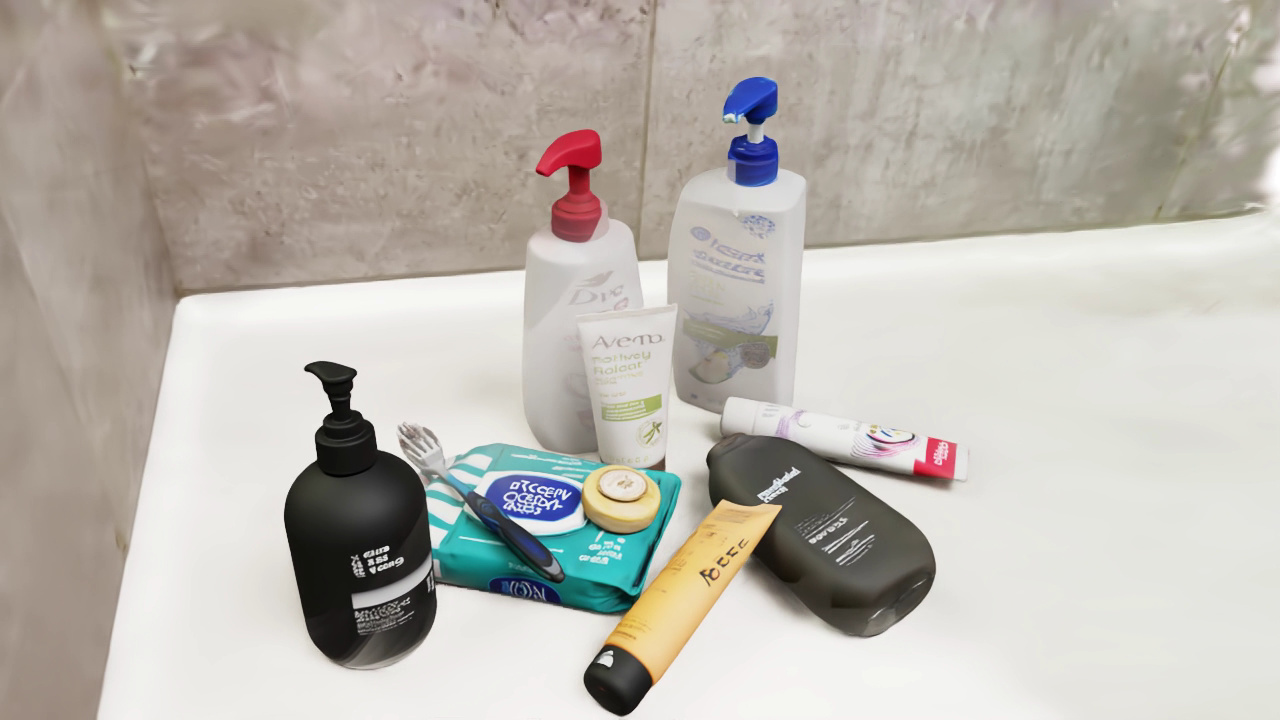} &
\includegraphics[width=0.20\textwidth,height=0.12\textwidth,keepaspectratio]{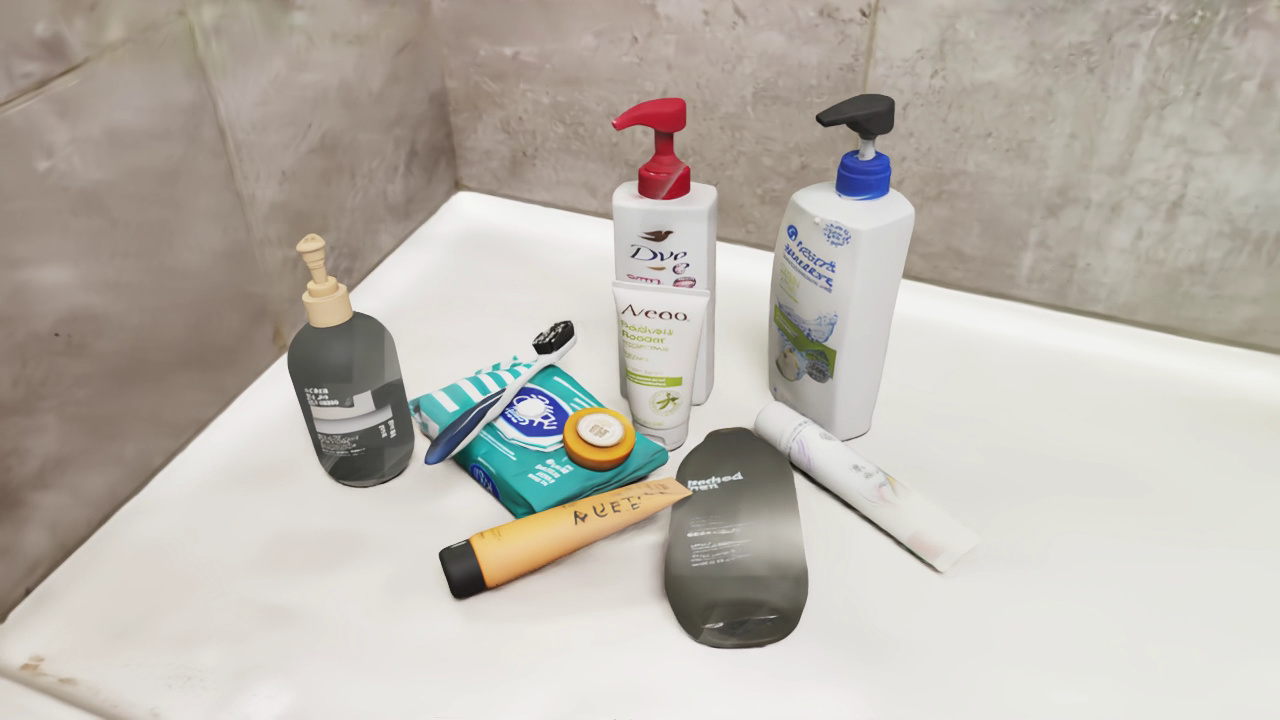} \\[1pt]

\raisebox{0.04\textwidth}{10} & \raisebox{0.04\textwidth}{51.39} &
\includegraphics[width=0.20\textwidth,height=0.12\textwidth,keepaspectratio]{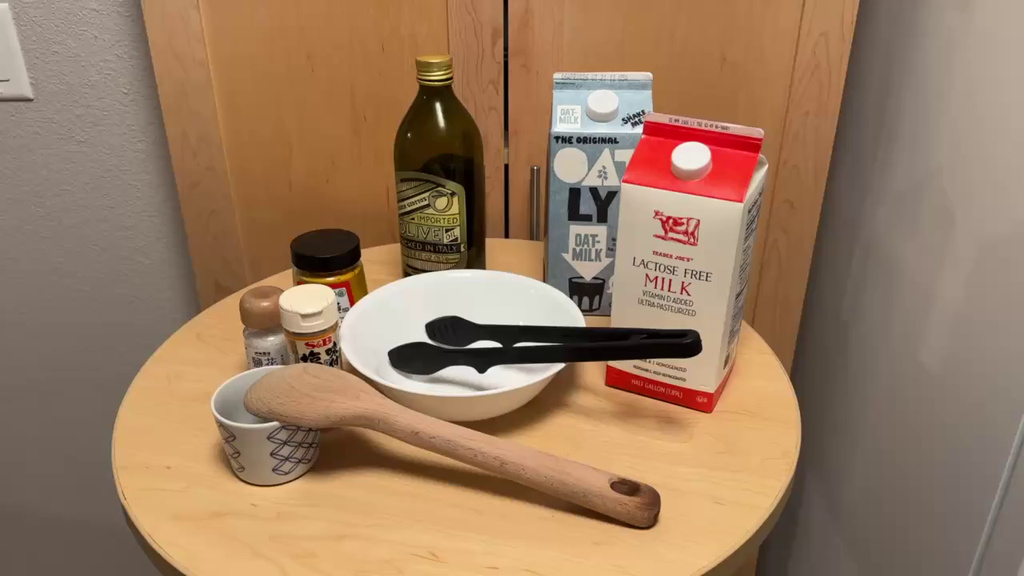} &
\includegraphics[width=0.20\textwidth,height=0.12\textwidth,keepaspectratio]{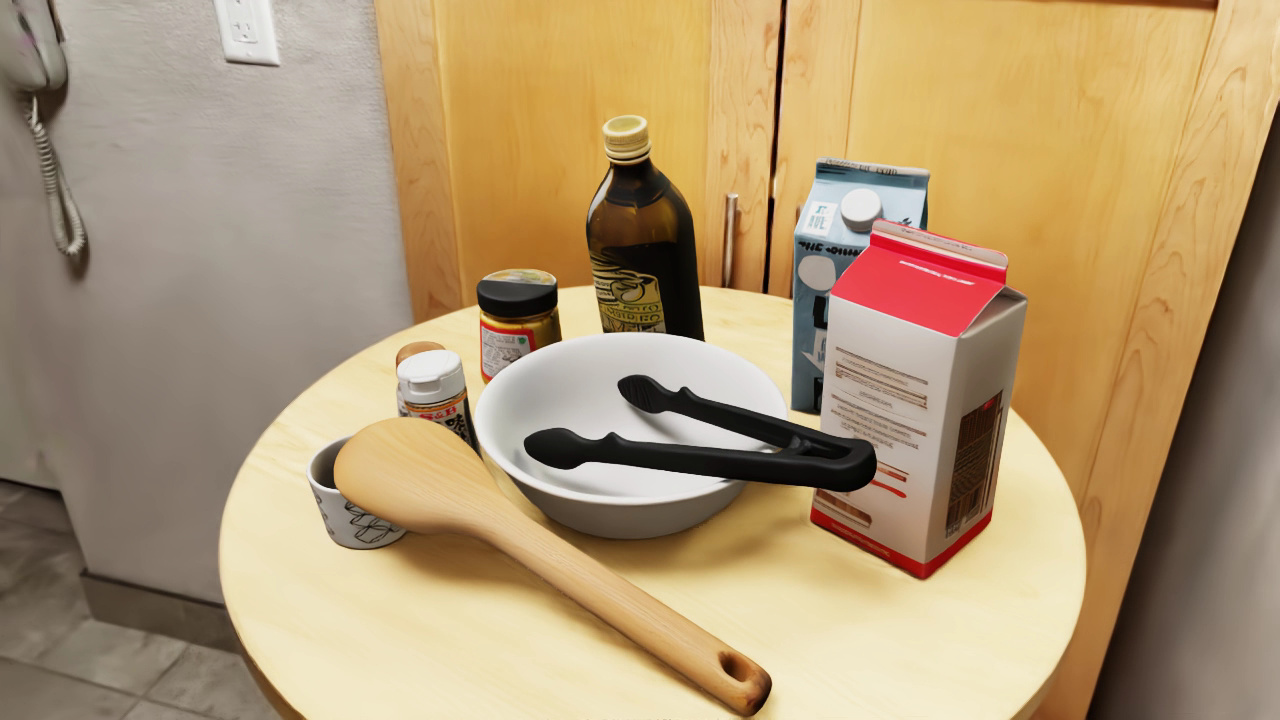} &
\includegraphics[width=0.20\textwidth,height=0.12\textwidth,keepaspectratio]{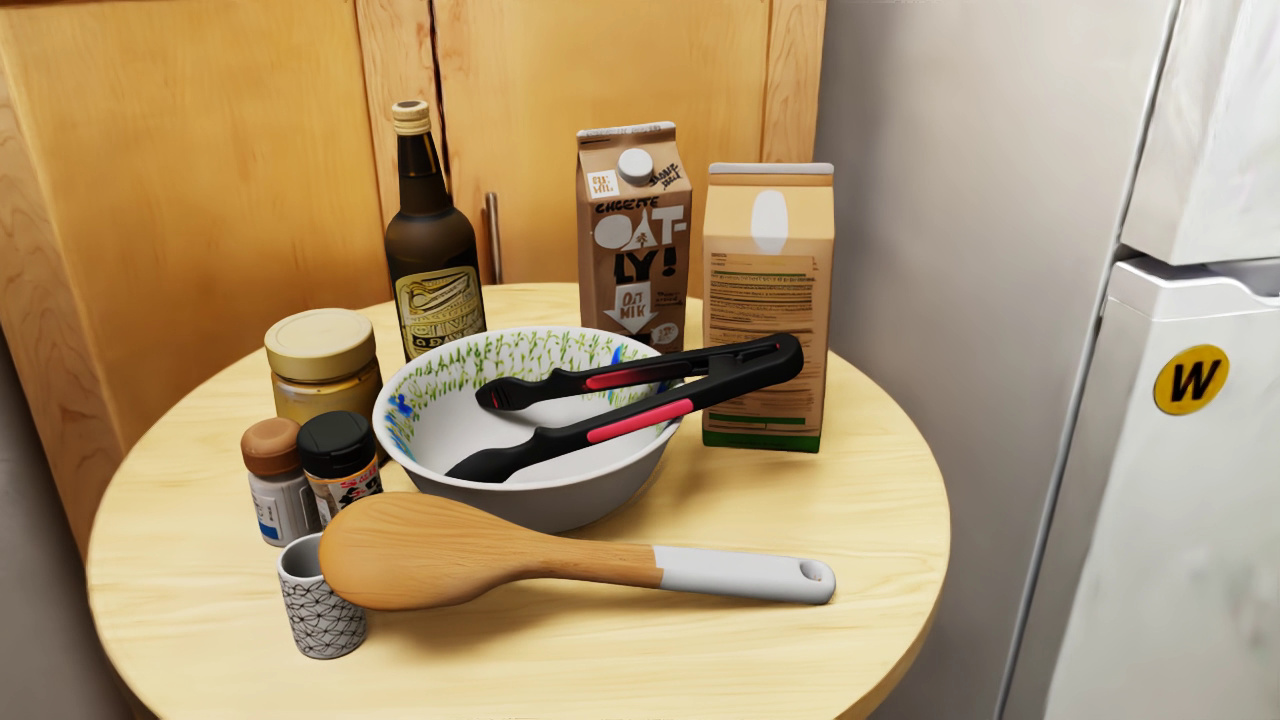} \\[1pt]
\raisebox{0.04\textwidth}{10} & \raisebox{0.04\textwidth}{58.71} &
\includegraphics[width=0.20\textwidth,height=0.12\textwidth,keepaspectratio]{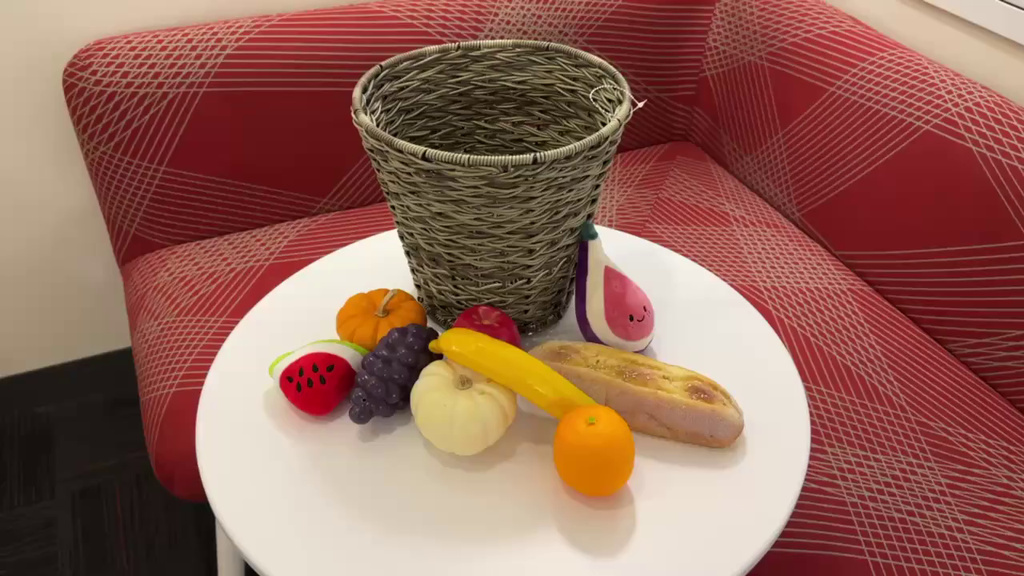} &
\includegraphics[width=0.20\textwidth,height=0.12\textwidth,keepaspectratio]{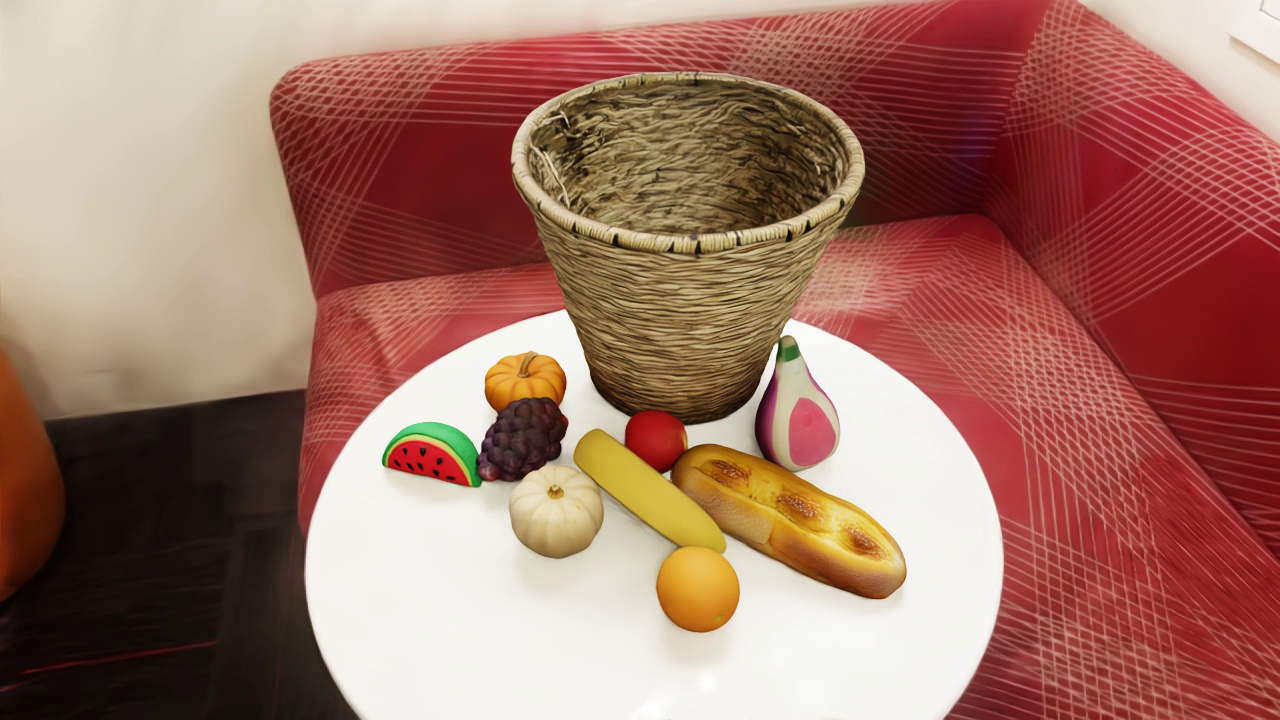} &
\includegraphics[width=0.20\textwidth,height=0.12\textwidth,keepaspectratio]{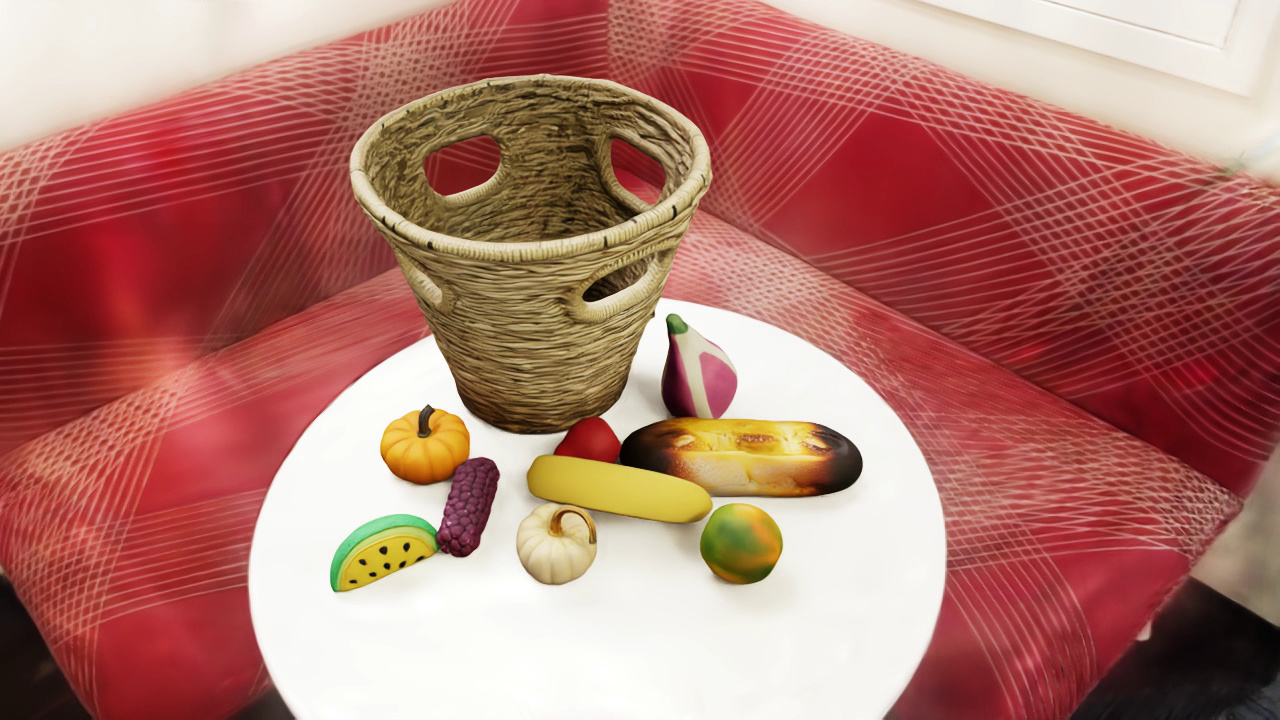} \\[1pt]
\raisebox{0.04\textwidth}{11} & \raisebox{0.04\textwidth}{67.13} &
\includegraphics[width=0.20\textwidth,height=0.12\textwidth,keepaspectratio]{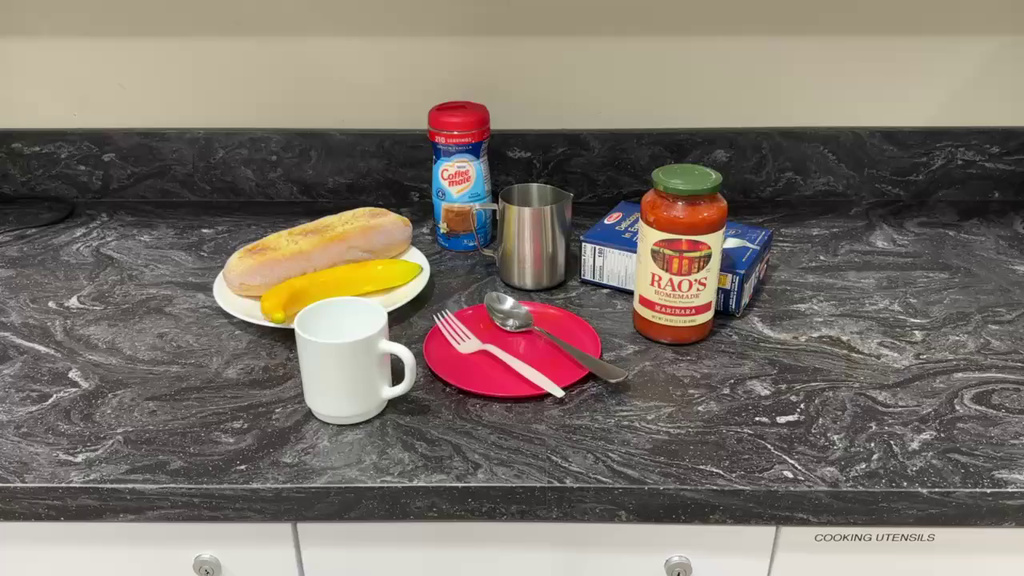} &
\includegraphics[width=0.20\textwidth,height=0.12\textwidth,keepaspectratio]{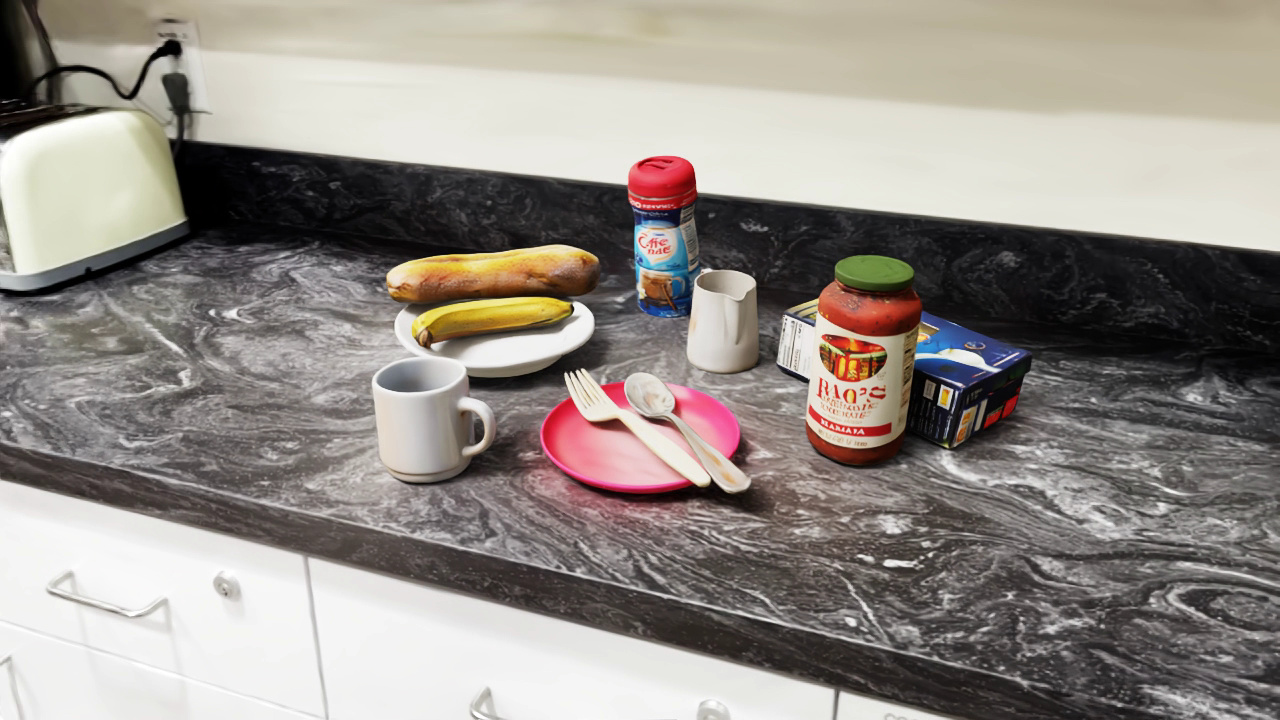} &
\includegraphics[width=0.20\textwidth,height=0.12\textwidth,keepaspectratio]{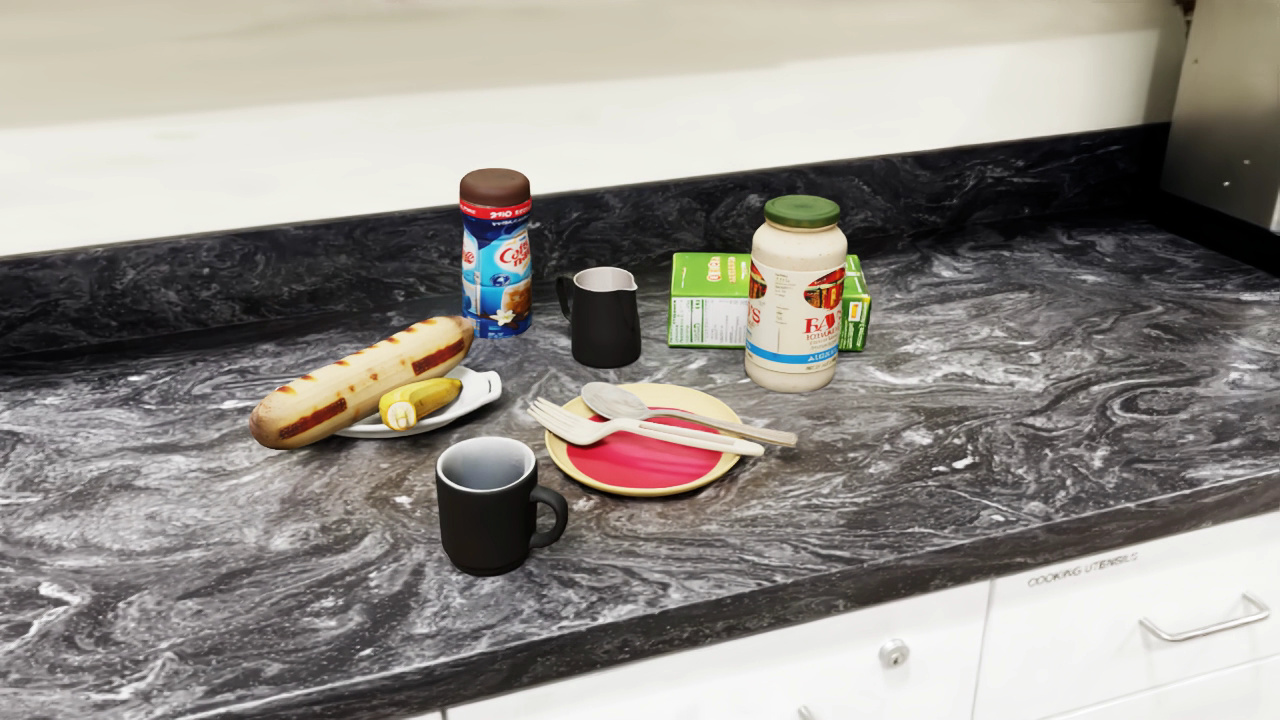} \\
\bottomrule
\end{tabular}
\caption{\textbf{Additional real-to-sim reconstruction results.} We show additional real-world input images, the corresponding reconstructed digital twins generated by \sysFull, and sampled digital cousin scene variations. The first two columns report the number of reconstructed objects and the total wallclock time to reconstruct the scene. All reconstructions are run on a machine with an NVIDIA GeForce RTX 3090 GPU with 24GB VRAM.}
\label{tab:additional_real2sim_reconstruction}
\end{table}

\subsection{Comparison between Manual and Automatic Background Pipeline}
\label{app:3dgs-background-analysis}
In this section, we compare the quantitative and qualitative reconstruction results between the manual background pipeline and automatic background pipeline mentioned in Appendix~\ref{sec:bg_details}.

\subsubsection{Qualitative Reconstruction Results}
We record five in-the-wild scenes and run both background reconstruction pipelines. The side-by-side visual results are shown in Table~\ref{tab:bg_qualitative}. The automatic background pipeline can produce floating artifacts around the support surface that partially occlude foreground objects, which may occur when the object-removal model hallucinates content during image inpainting.
\begin{table}[t]
\centering
\setlength{\tabcolsep}{2pt}
\renewcommand{\arraystretch}{0.78}
\begin{tabular}{cccc}
\toprule
Scene &
\makecell{Real\\Scene} &
\makecell{Manual\\Background Result} &
\makecell{Automatic\\Background Result} \\
\midrule
\raisebox{0.04\textwidth}{dorm\_1} &
\includegraphics[width=0.20\textwidth,height=0.12\textwidth,keepaspectratio]{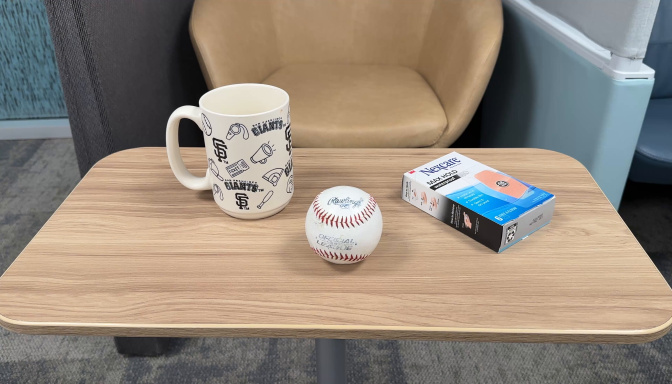} &
\includegraphics[width=0.20\textwidth,height=0.12\textwidth,keepaspectratio]{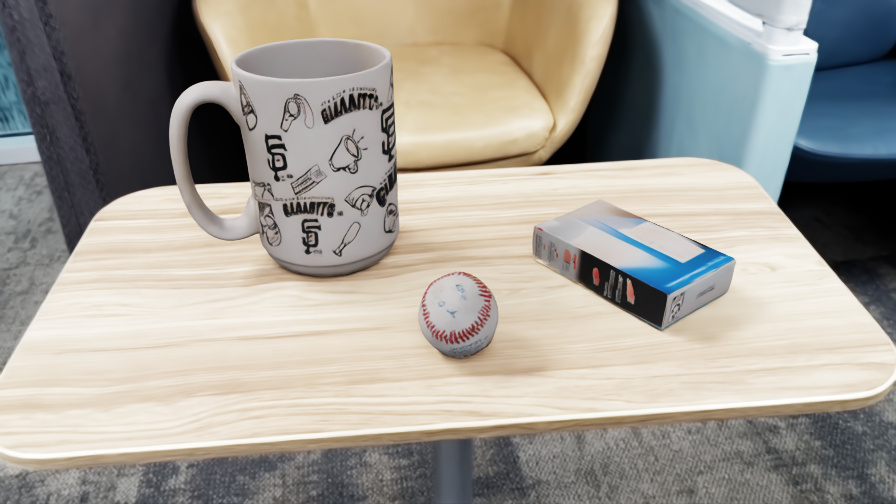} &
\includegraphics[width=0.20\textwidth,height=0.12\textwidth,keepaspectratio]{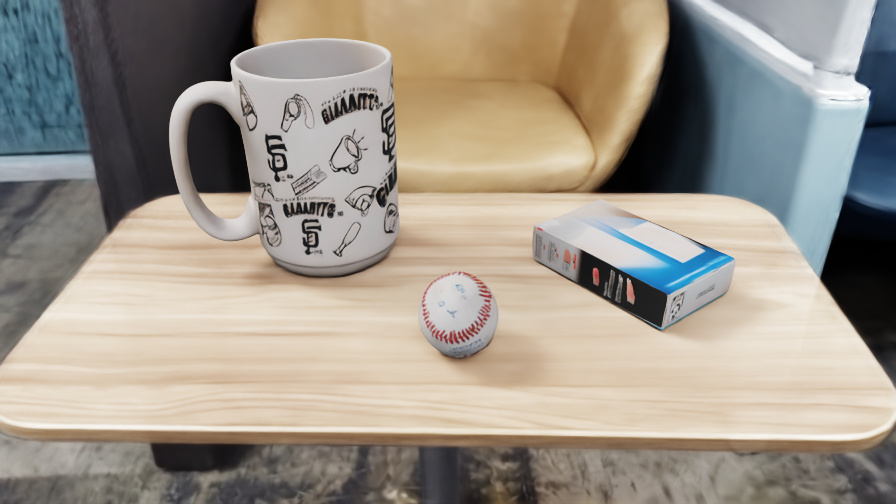} \\[1pt]

\raisebox{0.04\textwidth}{dorm\_2} &
\includegraphics[width=0.20\textwidth,height=0.12\textwidth,keepaspectratio]{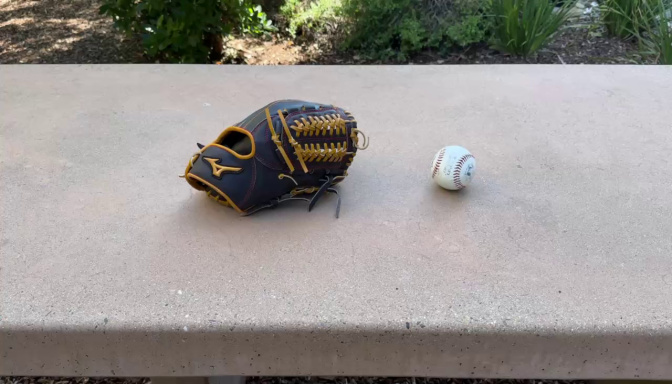} &
\includegraphics[width=0.20\textwidth,height=0.12\textwidth,keepaspectratio]{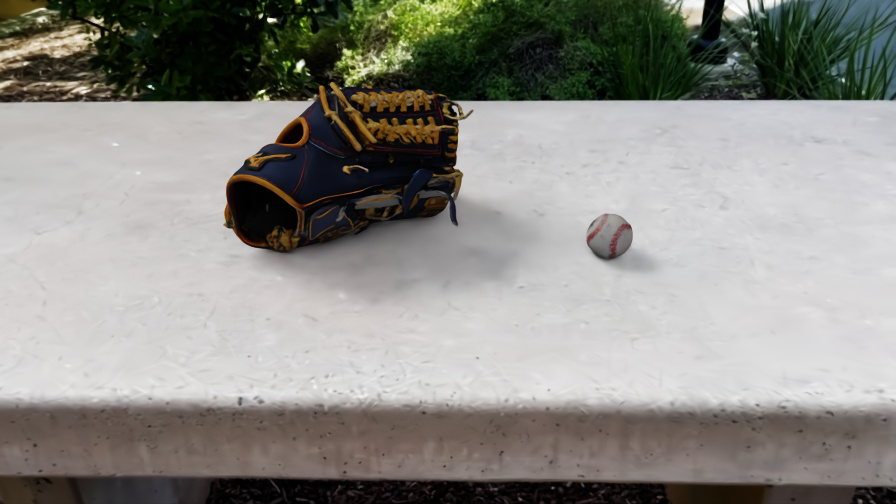} &
\includegraphics[width=0.20\textwidth,height=0.12\textwidth,keepaspectratio]{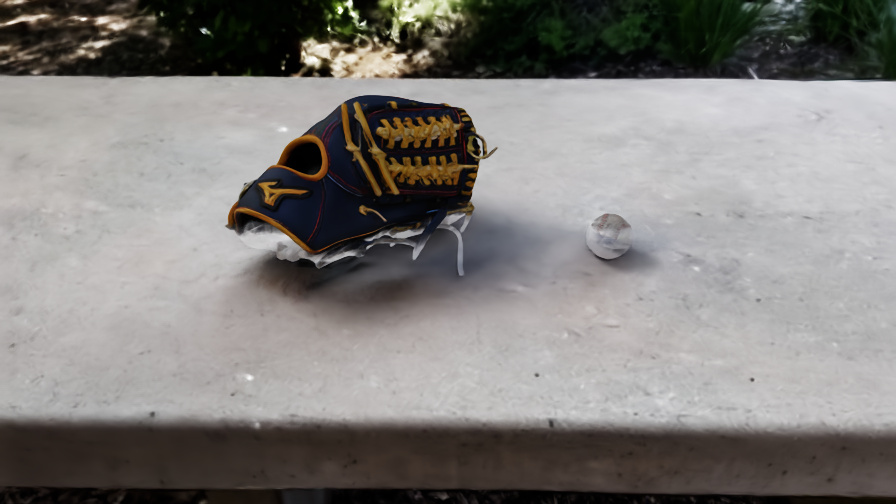} \\[1pt]

\raisebox{0.04\textwidth}{dorm\_3} &
\includegraphics[width=0.20\textwidth,height=0.12\textwidth,keepaspectratio]{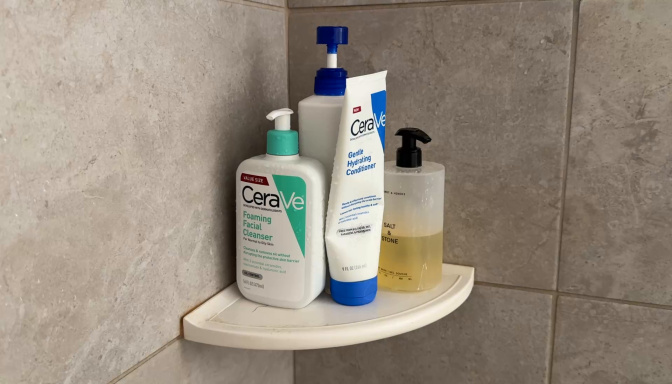} &
\includegraphics[width=0.20\textwidth,height=0.12\textwidth,keepaspectratio]{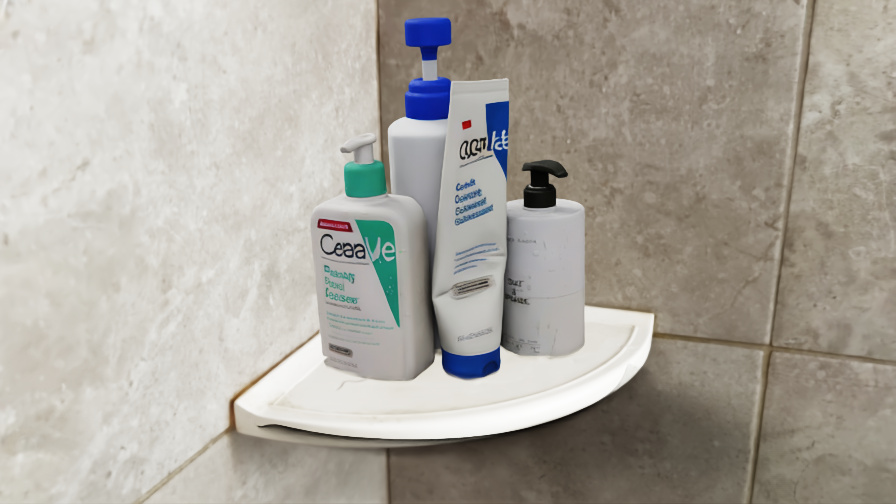} &
\includegraphics[width=0.20\textwidth,height=0.12\textwidth,keepaspectratio]{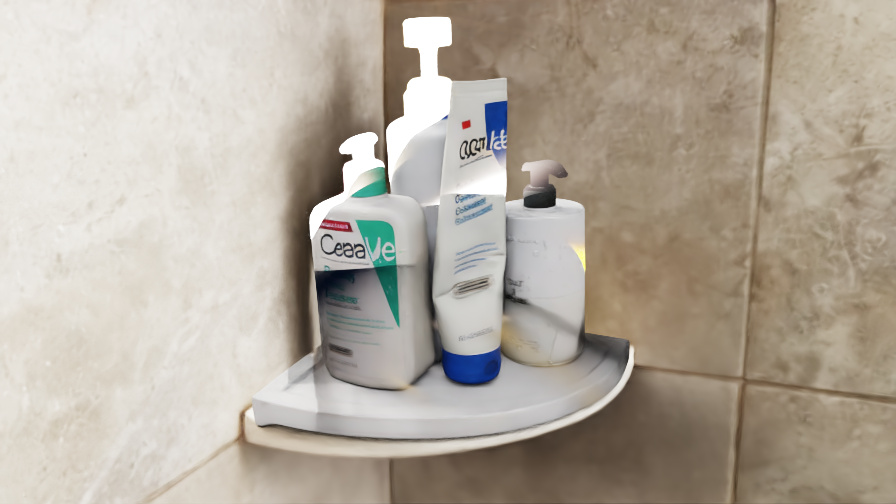} \\[1pt]

\raisebox{0.04\textwidth}{dorm\_4} &
\includegraphics[width=0.20\textwidth,height=0.12\textwidth,keepaspectratio]{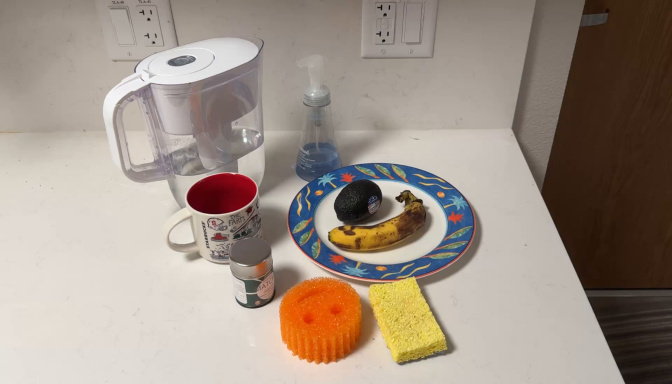} &
\includegraphics[width=0.20\textwidth,height=0.12\textwidth,keepaspectratio]{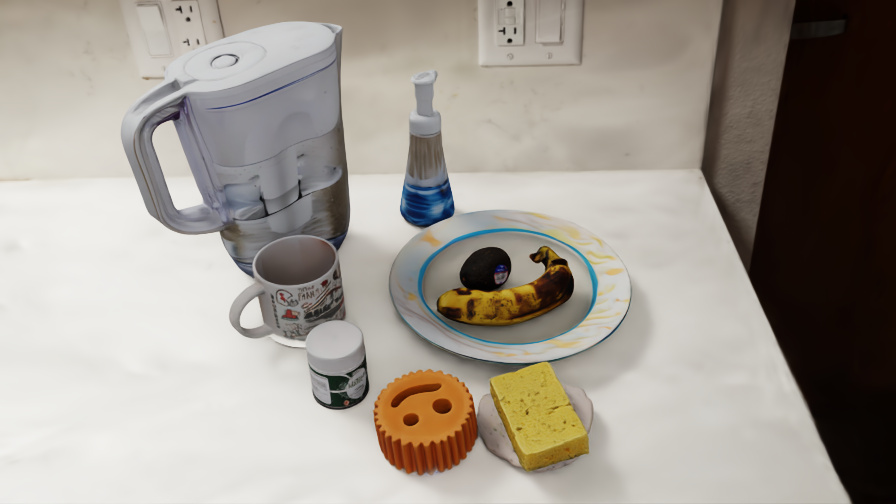} &
\includegraphics[width=0.20\textwidth,height=0.12\textwidth,keepaspectratio]{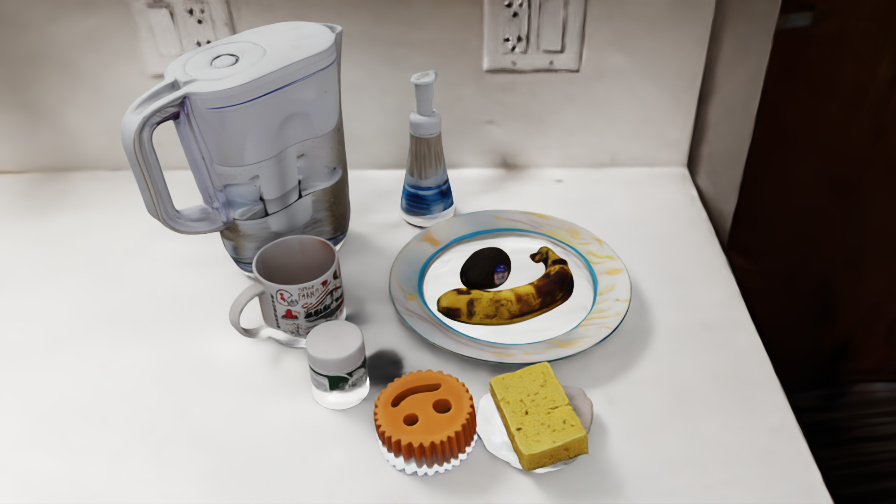} \\[1pt]

\raisebox{0.04\textwidth}{dorm\_5} &
\includegraphics[width=0.20\textwidth,height=0.12\textwidth,keepaspectratio]{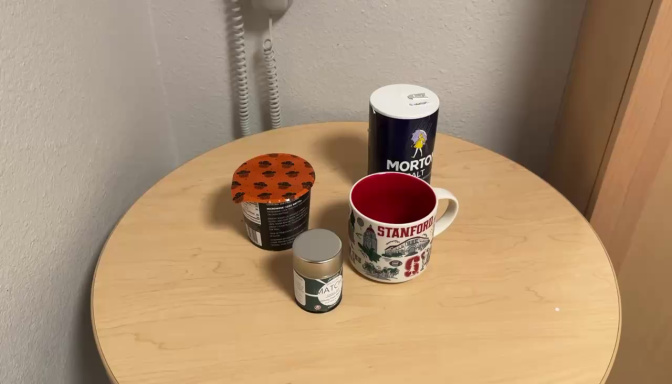} &
\includegraphics[width=0.20\textwidth,height=0.12\textwidth,keepaspectratio]{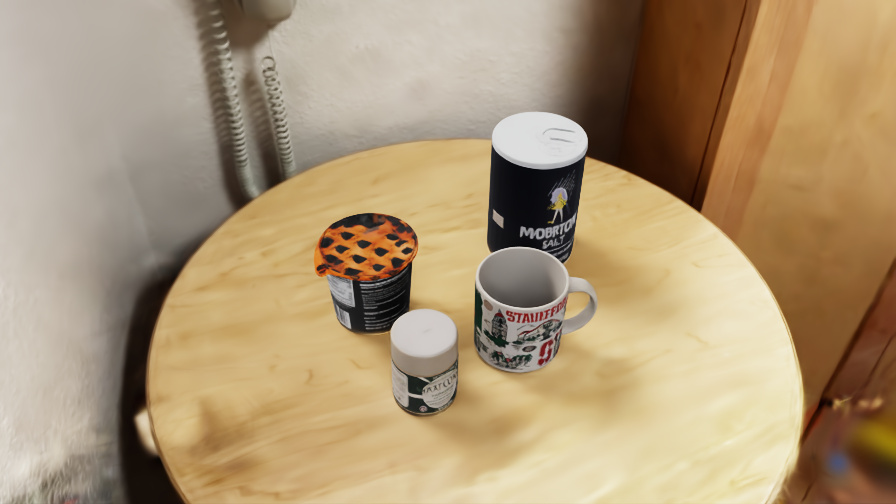} &
\includegraphics[width=0.20\textwidth,height=0.12\textwidth,keepaspectratio]{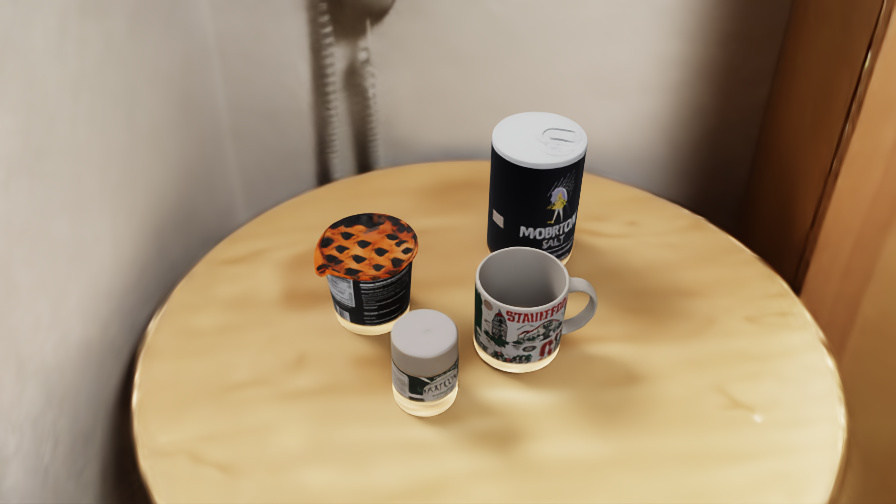} \\
\bottomrule
\end{tabular}
\caption{\textbf{Background reconstruction comparison.} The results for the real scene alongside results from the manual and automated background pipelines for five dorm scenes.}
\label{tab:bg_qualitative}
\end{table}

\subsubsection{Quantitative Reconstruction Results}
We quantify the agreement between each rendered scene shown in Table~\ref{tab:bg_qualitative} and the corresponding real video frame using 7 complementary metrics as follows:

\begin{itemize}
  \item \textbf{PSNR} (Peak Signal-to-Noise Ratio, dB, $\uparrow$) is defined as $10\log_{10}(1/\text{MSE})$ on intensities normalized to $[0,1]$ and measures global pixel-wise fidelity; because it derives from the mean squared error, it is dominated by a small number of large deviations and is sensitive to global misalignment.

  \item \textbf{SSIM} (Structural Similarity Index, $[-1,1]$, $\uparrow$) compares local luminance, contrast, and structure over sliding windows, capturing perceived structural agreement while remaining comparatively tolerant of uniform intensity shifts.

  \item \textbf{MAE} (Mean Absolute Error, $\downarrow$) is the mean of $|I_{\text{render}}-I_{\text{GT}}|$ over all pixels and channels, reporting the average per-pixel intensity deviation.

  \item \textbf{RMSE} (Root Mean Squared Error, $\downarrow$) is the square root of the mean squared error and penalizes large per-pixel errors more heavily than MAE.

  \item \textbf{NCC} (Normalized Cross-Correlation, $[-1,1]$, $\uparrow$) computes the correlation between the zero-mean, unit-variance render and ground-truth images; by factoring out global brightness and contrast it isolates spatial/structural alignment, making it the most direct indicator of whether the background geometry is registered to the real footage (values near zero indicate near-random alignment).

  \item \textbf{$\Delta E$} (CIE76 color difference, $\Delta E_{ab}^*$, $\downarrow$) is the mean Euclidean distance between the images in the perceptually-uniform CIELAB color space, quantifying color/appearance error in units calibrated to human color perception.

  \item \textbf{EdgeMAE} (Sobel edge MAE, $\downarrow$) is the mean absolute difference between the Sobel gradient-magnitude maps of the two images, measuring how well structural edges and contours coincide independently of absolute color.
\end{itemize}

Together these metrics span pixel fidelity (PSNR, MAE, RMSE), structural and perceptual similarity (SSIM, EdgeMAE), color accuracy ($\Delta E$), and geometric alignment (NCC), providing a comprehensive comparison.  We evaluate each metric on 50 uniformly sampled frames per scene across five scenes, and report the per-scene means together with their cross-scene average in Table~\ref{tab:bg_comparison}.
\begin{table}[ht]
  \centering
  \caption{\textbf{Render-vs-real Background Reconstruction Metrics.} Render-vs-real agreement averaged over five scenes (dorm\_1, dorm\_2, dorm\_3, dorm\_4, dorm\_5).}
  \label{tab:bg_comparison}
  \begin{tabular}{lcccccccc}
    \toprule
    Variant & PSNR$\uparrow$ & SSIM$\uparrow$ & MAE$\downarrow$ & RMSE$\downarrow$ & NCC$\uparrow$ & $\Delta E\downarrow$ & EdgeMAE$\downarrow$ \\
    \midrule
    Manual Pipeline & 12.91 & 0.497 & 0.1712 & 0.2294 & 0.549 & 20.13 & 0.0283 \\
    Automatic Pipeline  & \textbf{15.29} & \textbf{0.605} & \textbf{0.1275} & \textbf{0.1758} & \textbf{0.749} & \textbf{15.23} & \textbf{0.0248} \\
    \bottomrule
  \end{tabular}
\end{table}

The automated pipeline outperforms the manual, hand-curated background reconstruction across all reported metrics. One potential interpretation of this result is that the automatic pipeline does not estimate the background alignment but derives it: the background-to-world transform is composed analytically
  from the same camera poses that define the evaluation viewpoints, so the reconstruction is registered to the ground-truth frame by construction and stays consistent across views. Manual
  alignment instead approximates this six-degree-of-freedom transform by human eye, and since reprojection error scales with rotation error and scene depth, small imperceptible offsets compound into
  large pixel misalignment. This is clearest in NCC, which isolates spatial alignment; the PSNR, RMSE, and $\Delta E$ gains follow from the same tighter registration.

\clearpage

\end{document}